\ifdefined\pdfinfoomitdate\pdfinfoomitdate=1 \pdftrailerid{} \pdfsuppressptexinfo=-1 \fi
\documentclass
[
    twoside,                 %
    openright,               %
    cleardoublepage = empty, %
    fontsize = 12 pt,        %
    american,                %
    captions = tableheading, %
    numbers = noenddot,      %
    footheight = 35 pt,      %
]
{scrbook}

\newif\ifprintVersion   %
\newif\ifprofessionalPrint %
\newif\iffancyTheorems  %
\newif\ifboldNumberSets %

\printVersionfalse
\professionalPrintfalse
\fancyTheoremstrue
\boldNumberSetstrue

\newcommand*{\printTitle}{}
\newcommand*{\myTitle}[1]{\renewcommand*{\printTitle}{#1}}

\newcommand*{\printAuthor}{}
\newcommand*{\myName}[1]{\renewcommand*{\printAuthor}{#1}}

\newcommand*{\printSubject}{}
\newcommand*{\mySubject}[1]{\renewcommand*{\printSubject}{#1}}

\newcommand*{\printKeywords}{}
\newcommand*{\myKeywords}[1]{\renewcommand*{\printKeywords}{#1}}

\newlength{\extraborderlength}

\newlength{\mybindingcorrection}

    \myTitle{PriMa-Causa \\ Causal Prescriptive Maintenance}

    \myName{Felix Saretzky}

    \mySubject{Doctoral dissertation: PriMa-Causa, a causal reasoning layer and causal foundation model for prescriptive maintenance in manufacturing.}

    \myKeywords{prescriptive maintenance | causal inference | causal AI | do-calculus | structural causal models | causal process graphs | causal foundation models | prior-data fitted networks | synthetic data generation | root cause analysis | design-science research | overall equipment effectiveness}

\usepackage[utf8]{inputenc} %
\usepackage[T1]{fontenc}    %
\usepackage
[
    ngerman,         %
    main = american, %
]
{babel}                     %

\ifdefined\pdfgentounicode
  \input glyphtounicode
\fi

\usepackage{calc} %

\newlength{\myparindent}
\newlength{\myparskip}
\setfootnoterule{0 cm}

\deffootnote[1.2 em]{1.2 em}{0 em}{\makebox[1.4 em][l]{\textbf{\thefootnotemark}}}

\makeatletter%
\@removefromreset{footnote}{chapter}%
\makeatother

\usepackage[dvipsnames]{xcolor} %

\definecolor{stroke1}{HTML}{000000} %

\colorlet{captionlabel}{black}
\colorlet{footerpagenr}{black}
\colorlet{footerchapter}{stroke1}
\colorlet{footerchaptername}{black}
\colorlet{footersection}{stroke1}
\colorlet{footersectionname}{black}
\colorlet{chapternumber}{stroke1}

\newlength{\mypaperwidth}
\newlength{\mypaperheight}
\newlength{\mybodywidth}
\newlength{\mybodyheight}
\newlength{\myoutermargin}
\ifprintVersion
    \ifprofessionalPrint
    \else
    \fi
\else
\fi

\newlength{\mytopmargin}
\ifprintVersion
    \ifprofessionalPrint
    \fi
\fi

\newlength{\myinnermargin}
\ifprintVersion
    \ifprofessionalPrint
    \fi
\fi

\newlength{\mybottommargin}
\ifprintVersion
    \ifprofessionalPrint
    \fi
\fi

\newcommand{\goldenratio}{1.618}

\newlength{\myheadsep} %
\ifprintVersion
    \ifprofessionalPrint
    \fi
\fi

\newlength{\myfootskip} %
\ifprintVersion
    \ifprofessionalPrint
    \fi
\fi

\newlength{\mymargininnersep} %
\newlength{\mymarginoutersep} %
\ifprintVersion
    \ifprofessionalPrint
    \fi
\fi

\newlength{\mymarginwidth} %
\newlength{\mymarginwidthwithinnersep} %
\usepackage
[
    \ifprintVersion
        \ifprofessionalPrint
            paperwidth = \mypaperwidth + 2\extraborderlength + \mybindingcorrection,
            paperheight = \mypaperheight + 2\extraborderlength,
        \else
            paperwidth = \mypaperwidth,
            paperheight = \mypaperheight,
        \fi
    \else
        paperwidth = \mypaperwidth,
        paperheight = \mypaperheight,
    \fi
    textwidth = \mybodywidth,
    textheight = \mybodyheight,
    outer = \myoutermargin,
    top = \mytopmargin,
    headsep = \myheadsep,
    footskip = \myfootskip,
    marginparsep = \mymargininnersep,
    marginparwidth = \mymarginwidth,
]
{geometry} %

\usepackage
[
]
{scrlayer-scrpage} %

\clearpairofpagestyles

\KOMAoptions
{%
    headwidth = \textwidth + \mymarginwidthwithinnersep,%
    footwidth = \myoutermargin : \textwidth,%
}

\automark[chapter]{chapter}
\automark*[section]{}
\newmarks\sectionnumbermark
\AddToHook{cmd/sectionmark/before}{\marks\sectionnumbermark{\thesection}}

\lehead%
{%
    \begin{minipage}[b]{\mymarginwidth}%
        \small\raggedleft\normalfont\textsf{\textbf{\color{footerchapter}\chaptername\ \thechapter}}
    \end{minipage}
}
\cehead{\hspace*{\mymarginwidthwithinnersep}\parbox{\textwidth}{\raggedright\leftmark}}

\rohead%
{%
    \Ifstr{\rightmark}{\leftmark}%
    {%
        \begin{minipage}[b]{\mymarginwidth}%
            \small\raggedright\normalfont\textsf{\textbf{\color{footersection}Chapter\ \thechapter}}%
        \end{minipage}%
    }%
    {%
        \begin{minipage}[b]{\mymarginwidth}%
            \small\raggedright\normalfont\textsf{\textbf{\color{footersection}Section\ \firstmarks\sectionnumbermark}}%
        \end{minipage}%
    }%
}
\cohead{\hspace*{-\mymarginwidthwithinnersep}\parbox{\textwidth}{\raggedleft\rightmark}}

\lefoot*%
{%
    \vspace*{1 ex}%
    {\color{stroke1}\rule{\myoutermargin - \mymargininnersep}{0.5 mm}}\\
    \begin{minipage}[b]{\myoutermargin - \mymargininnersep}%
        \raggedleft\normalfont\color{footerpagenr}\textbf{\thepage}%
    \end{minipage}%
}
\rofoot*%
{%
    \vspace*{1 ex}%
    {\color{stroke1}\rule{\myoutermargin - \mymargininnersep}{0.5 mm}}\\
    \begin{minipage}[b]{\myoutermargin - \mymargininnersep}%
        \raggedright\normalfont\color{footerpagenr}\textbf{\thepage}%
    \end{minipage}%
}

\usepackage{caption}
\usepackage{tikz} %

\newlength{\mytmpa}
\newlength{\mytmpb}

\renewcommand*{\partlineswithprefixformat}[3]%
{%
    #2
    \thispagestyle{empty}
    \ifprintVersion
        \setlength{\mytmpa}{0.618\mypaperwidth + \mybindingcorrection + \extraborderlength}%
        \setlength{\mytmpb}{0.382\mypaperheight + \extraborderlength}%
    \else
        \setlength{\mytmpa}{0.618\mypaperwidth}%
        \setlength{\mytmpb}{0.382\mypaperheight}%
    \fi
    \begin{tikzpicture}[overlay, remember picture]%
        \node [inner sep = 0, outer sep = 0, anchor = north] at (current page.north west)%
        {%
            \begin{tikzpicture}[overlay, remember picture]%
                \draw[color = stroke1, line width = 0.7 mm] (\mytmpa, 0) -- (\mytmpa, -\mytmpb);%
            \end{tikzpicture}%
        };%
        \node (align) [align = right, below = \mytmpb - 2 ex, inner sep = 0, outer sep = 0, anchor = north west] at (current page.north west)%
        {%
            \hspace{\mytmpa}\hspace{0.5 em}\partname\ \thepart\\[1 ex]
            \color{stroke1}#3%
        };%
    \end{tikzpicture}%
}
\RedeclareSectionCommand%
[%
    font = \normalfont\Huge\sffamily,
    prefixfont = \normalfont\Huge\sffamily,
]
{part}

\usepackage{etoolbox}
\usepackage[strict]{changepage}
\usetikzlibrary{calc}

\newbool{chapterHasANumber}
\newbool{chapterHasAStar}
\newcommand*{\thischapterlabel}{}
\renewcommand*{\chapterlinesformat}[3]%
{%
    \Ifnumbered{#1}{\setbool{chapterHasANumber}{true}}{\setbool{chapterHasANumber}{false}}%
    \Ifstr{#2}{}{\setbool{chapterHasAStar}{true}}{\setbool{chapterHasAStar}{false}}%
    \ifboolexpr{bool{chapterHasANumber} and not bool{chapterHasAStar}}%
    {%
        \renewcommand*{\thischapterlabel}{\color{chapternumber}\thechapter}
    }%
    {%
        \renewcommand*{\thischapterlabel}{\vphantom{0}}
    }%
    \checkoddpage
    \begin{tikzpicture}[overlay, remember picture]%
        \ifoddpage%
            \coordinate (anchor) at ($(current page.north west) + (\myinnermargin, -\mytopmargin)$);
            \node [inner sep = 0, outer sep = 0, anchor = north west] (numbernode) at (anchor)%
            {%
                \sffamily\fontsize{60}{60}\selectfont\thischapterlabel
            };%
            \draw[color = stroke1, line width = 0.7 mm] ($(numbernode.south west)-(0, 1 ex)$) -- ++(\paperwidth, 0);%
            \ifboolexpr{bool{chapterHasANumber} and not bool{chapterHasAStar}}%
            {%
                \node (align) [text width = \textwidth - 2 cm, align = right, right = \mybodywidth, inner sep = 0, outer sep = 0, anchor = east] at (numbernode.west) {#3};
            }%
            {%
                \node (align) [align = left, inner sep = 0, outer sep = 0, anchor = west] at (numbernode.west) {#3};
            }%
        \else%
            \coordinate (anchor) at ($(current page.north east) - (\myinnermargin, \mytopmargin)$);
            \node [inner sep = 0, outer sep = 0, anchor = north east] (numbernode) at (anchor)%
            {%
                \sffamily\fontsize{60}{60}\selectfont\thischapterlabel
            };%
            \draw[color = stroke1, line width = 0.7 mm] ($(numbernode.south east)-(0, 1 ex)$) -- ++(-\paperwidth, 0);%
            \ifboolexpr{bool{chapterHasANumber} and not bool{chapterHasAStar}}%
            {%
                \node (align) [text width = \textwidth - 2 cm, align = left, left = \mybodywidth, inner sep = 0, outer sep = 0, anchor = west] at (numbernode.east) {#3};
            }%
            {%
                \node (align) [align = right, inner sep = 0, outer sep = 0, anchor = east] at (numbernode.east) {#3};
            }%
        \fi%
    \end{tikzpicture}%
}
\RedeclareSectionCommand%
[%
    font = \color{stroke1}\normalfont\huge\sffamily,
    afterskip = 20 pt,
]
{chapter}

\BeforeStartingTOC[toc]{\pagestyle{plain}}
\AfterStartingTOC{\thispagestyle{plain}}

\usepackage
[
    sortcites,              %
    style = numeric,     %
    sorting = none,         %
    defernumbers,           %
    safeinputenc,           %
    backref = true,         %
    backrefstyle = three,   %
    hyperref = true,        %
    maxbibnames = 99,       %
    maxcitenames = 2,       %
]
{biblatex} %

\renewbibmacro{in:}%
{%
    \ifentrytype{article}{}{\printtext{\bibstring{in}\intitlepunct}}%
}

\renewbibmacro*{volume+number+eid}%
{%
    \printfield{volume}%
    \iffieldundef{number}{}{\addcolon}%
    \printfield{number}%
    \setunit*{\addcomma\space}%
    \printfield{eid}%
}

\DefineBibliographyStrings{english}%
{%
    backrefpage  = {\lowercase{s}ee page}, %
    backrefpages = {\lowercase{s}ee pages} %
}

\DeclareFieldFormat[article]{title}{\textbf{\color{stroke1}#1}}
\DeclareFieldFormat[inproceedings]{title}{\textbf{\color{stroke1}#1}}
\DeclareFieldFormat[thesis]{title}{\textbf{\color{stroke1}#1}}
\DeclareFieldFormat[book]{title}{\textbf{\color{stroke1}#1}}
\DeclareFieldFormat[unpublished]{title}{\textbf{\color{stroke1}#1}}
\DeclareFieldFormat[report]{title}{\textbf{\color{stroke1}#1}}
\DeclareFieldFormat[inbook]{chapter}{\textbf{\color{stroke1}#1}}
\DeclareFieldFormat[inbook]{title}{#1}
\DeclareFieldFormat{title}{\textbf{\color{stroke1}#1}}
\DeclareFieldFormat{pages}{#1}

\newtoggle{authorend}
\togglefalse{authorend}

\DeclareBibliographyDriver{article}%
{%
  \usebibmacro{bibindex}%
  \usebibmacro{begentry}%
  \iftoggle{authorend}{}{\usebibmacro{author/translator+others}}%
  \setunit{\labelnamepunct}\newblock
  \usebibmacro{title}%
  \newunit
  \printlist{language}%
  \newunit\newblock
  \usebibmacro{byauthor}%
  \newunit\newblock
  \usebibmacro{bytranslator+others}%
  \newunit\newblock
  \printfield{version}%
  \newunit\newblock
  \usebibmacro{in:}%
  \usebibmacro{journal+issuetitle}%
  \newunit
  \usebibmacro{byeditor+others}%
  \newunit
  \usebibmacro{note+pages}%
  \newunit\newblock
  \iftoggle{bbx:isbn}
  {\printfield{issn}}
  {}%
  \newunit\newblock
  \usebibmacro{doi+eprint+url}%
  \newunit\newblock
  \usebibmacro{addendum+pubstate}%
  \setunit{\bibpagerefpunct}\newblock
  \usebibmacro{pageref}%
  \newunit\newblock
  \iftoggle{bbx:related}
  {\usebibmacro{related:init}%
    \usebibmacro{related}}
  {}%
  \usebibmacro{finentry}%
  \iftoggle{authorend}{\usebibmacro{author/translator+others}}{}%
}

\DeclareBibliographyDriver{inbook}%
{%
  \usebibmacro{bibindex}%
  \usebibmacro{begentry}%
  \iftoggle{authorend}{}{\usebibmacro{author/translator+others}}%
  \setunit{\labelnamepunct}\newblock
  \usebibmacro{chapter+pages}%
  \newunit
  \printlist{language}%
  \newunit\newblock
  \usebibmacro{byauthor}%
  \newunit\newblock
  \usebibmacro{in:}%
  \usebibmacro{bybookauthor}%
  \newunit\newblock
  \usebibmacro{maintitle+booktitle}%
  \newunit\newblock
  \usebibmacro{byeditor+others}%
  \newunit\newblock
  \printfield{edition}%
  \newunit
  \iffieldundef{maintitle}
  {\printfield{volume}%
    \printfield{part}}
  {}%
  \newunit
  \printfield{volumes}%
  \newunit\newblock
  \usebibmacro{series+number}%
  \newunit\newblock
  \printfield{note}%
  \newunit\newblock
  \usebibmacro{publisher+location+date}%
  \newunit\newblock
  \newunit\newblock
  \iftoggle{bbx:isbn}
  {\printfield{isbn}}
  {}%
  \newunit\newblock
  \usebibmacro{doi+eprint+url}%
  \newunit\newblock
  \usebibmacro{addendum+pubstate}%
  \setunit{\bibpagerefpunct}\newblock
  \usebibmacro{pageref}%
  \newunit\newblock
  \iftoggle{bbx:related}
  {\usebibmacro{related:init}%
    \usebibmacro{related}}
  {}%
  \usebibmacro{finentry}%
  \iftoggle{authorend}{\usebibmacro{author/translator+others}}{}%
}

\DeclareBibliographyDriver{inproceedings}%
{%
  \usebibmacro{bibindex}%
  \usebibmacro{begentry}%
  \iftoggle{authorend}{}{\usebibmacro{author/translator+others}}%
  \setunit{\labelnamepunct}\newblock
  \usebibmacro{title}%
  \newunit
  \printlist{language}%
  \newunit\newblock
  \usebibmacro{byauthor}%
  \newunit\newblock
  \usebibmacro{in:}%
  \usebibmacro{maintitle+booktitle}%
  \newunit\newblock
  \usebibmacro{event+venue+date}%
  \newunit\newblock
  \usebibmacro{byeditor+others}%
  \newunit\newblock
  \iffieldundef{maintitle}
  {\printfield{volume}%
    \printfield{part}}
  {}%
  \newunit
  \printfield{volumes}%
  \newunit\newblock
  \usebibmacro{series+number}%
  \newunit\newblock
  \printfield{note}%
  \newunit\newblock
  \printlist{organization}%
  \newunit
  \usebibmacro{publisher+location+date}%
  \newunit\newblock
  \usebibmacro{chapter+pages}%
  \newunit\newblock
  \iftoggle{bbx:isbn}
  {\printfield{isbn}}
  {}%
  \newunit\newblock
  \usebibmacro{doi+eprint+url}%
  \newunit\newblock
  \usebibmacro{addendum+pubstate}%
  \setunit{\bibpagerefpunct}\newblock
  \usebibmacro{pageref}%
  \newunit\newblock
  \iftoggle{bbx:related}
  {\usebibmacro{related:init}%
    \usebibmacro{related}}
  {}%
  \usebibmacro{finentry}%
  \iftoggle{authorend}{\usebibmacro{author/translator+others}}{}%
}

\DeclareBibliographyDriver{thesis}%
{%
  \usebibmacro{bibindex}%
  \usebibmacro{begentry}%
  \iftoggle{authorend}{}{\usebibmacro{author}}%
  \setunit{\labelnamepunct}\newblock
  \usebibmacro{title}%
  \newunit
  \printlist{language}%
  \newunit\newblock
  \usebibmacro{byauthor}%
  \newunit\newblock
  \printfield{note}%
  \newunit\newblock
  \printfield{type}%
  \newunit
  \usebibmacro{institution+location+date}%
  \newunit\newblock
  \usebibmacro{chapter+pages}%
  \newunit
  \printfield{pagetotal}%
  \newunit\newblock
  \iftoggle{bbx:isbn}
  {\printfield{isbn}}
  {}%
  \newunit\newblock
  \usebibmacro{doi+eprint+url}%
  \newunit\newblock
  \usebibmacro{addendum+pubstate}%
  \setunit{\bibpagerefpunct}\newblock
  \usebibmacro{pageref}%
  \newunit\newblock
  \iftoggle{bbx:related}
  {\usebibmacro{related:init}%
    \usebibmacro{related}}
  {}%
  \usebibmacro{finentry}%
  \iftoggle{authorend}{\usebibmacro{author}}{}%
}

\usepackage
[
    babel = true, %
]
{microtype}           %
\usepackage{csquotes} %

\usepackage{amsmath}
\usepackage{amssymb}
\usepackage{mathtools}
\usepackage{braceMnSymbol} %
\usepackage[nohyperlinks]{acronym}

\usepackage
[
    ttscale = 0.85, %
]
{libertine} %
\usepackage
[
    libertine,    %
    slantedGreek, %
    vvarbb,       %
    libaltvw,     %
]
{newtxmath} %
\usepackage{url} %

\usepackage{graphicx} %
\usepackage
[
    subrefformat = simple, %
    labelformat = simple,  %
]
{subcaption}         %

\columnsep = \mymargininnersep

\usepackage{array}     %
\usepackage{booktabs}  %
\usepackage{tabularx}  %
\usepackage{longtable} %

\newcolumntype{Y}{>{\raggedright\arraybackslash}X}

\usepackage{enumitem} %

\usepackage
[
    ruled,         %
    vlined,        %
    linesnumbered, %
]
{algorithm2e} %

\usepackage{setspace} %

\usepackage{footnote}  %

\usepackage[most]{tcolorbox}

\newtcolorbox{quotebox}[1][]{%
    enhanced, breakable,
    colback=black!5, colframe=black!60,
    boxrule=0pt, frame hidden,
    borderline west={1.2pt}{0pt}{black!60},
    sharp corners,
    left=10pt, right=8pt, top=6pt, bottom=6pt,
    before skip=10pt, after skip=10pt,
    fonttitle=\bfseries, #1
}

\usepackage
[
    bookmarks = true,                 %
    bookmarksopen = false,            %
    bookmarksnumbered = true,         %
    pdfstartpage = 1,                 %
    pdftitle = {{\printTitle}},       %
    pdfauthor = {{\printAuthor}},     %
    pdfsubject = {{\printSubject}},   %
    pdfkeywords = {{\printKeywords}}, %
    pdfcreator = {},                  %
    pdfproducer = {},                 %
    breaklinks = true,                %
    colorlinks = true,            %
    allcolors = stroke1,          %
]
{hyperref} %

\input{packages_and_commands/general_commands}      %

\DeclareMathOperator{\dop}{do}

\makeatletter
\renewcommand*\AC@placelabel@part[1]{}
\renewcommand*\AC@placelabel[1]{%
  \expandafter\ifx\csname AC@\AC@prefix#1\endcsname\AC@used
  \else
    \ifAC@starred\else
      \global\expandafter\let\csname AC@\AC@prefix#1\endcsname\AC@used
    \fi
    \AC@addtoAC@clearlist{#1}%
  \fi
}
\makeatother

\usetikzlibrary{arrows.meta, calc, positioning}
\usepackage{pdfpages} %

\begin{document}

    \pagenumbering{Alph}
    \includepdf[pages=-]{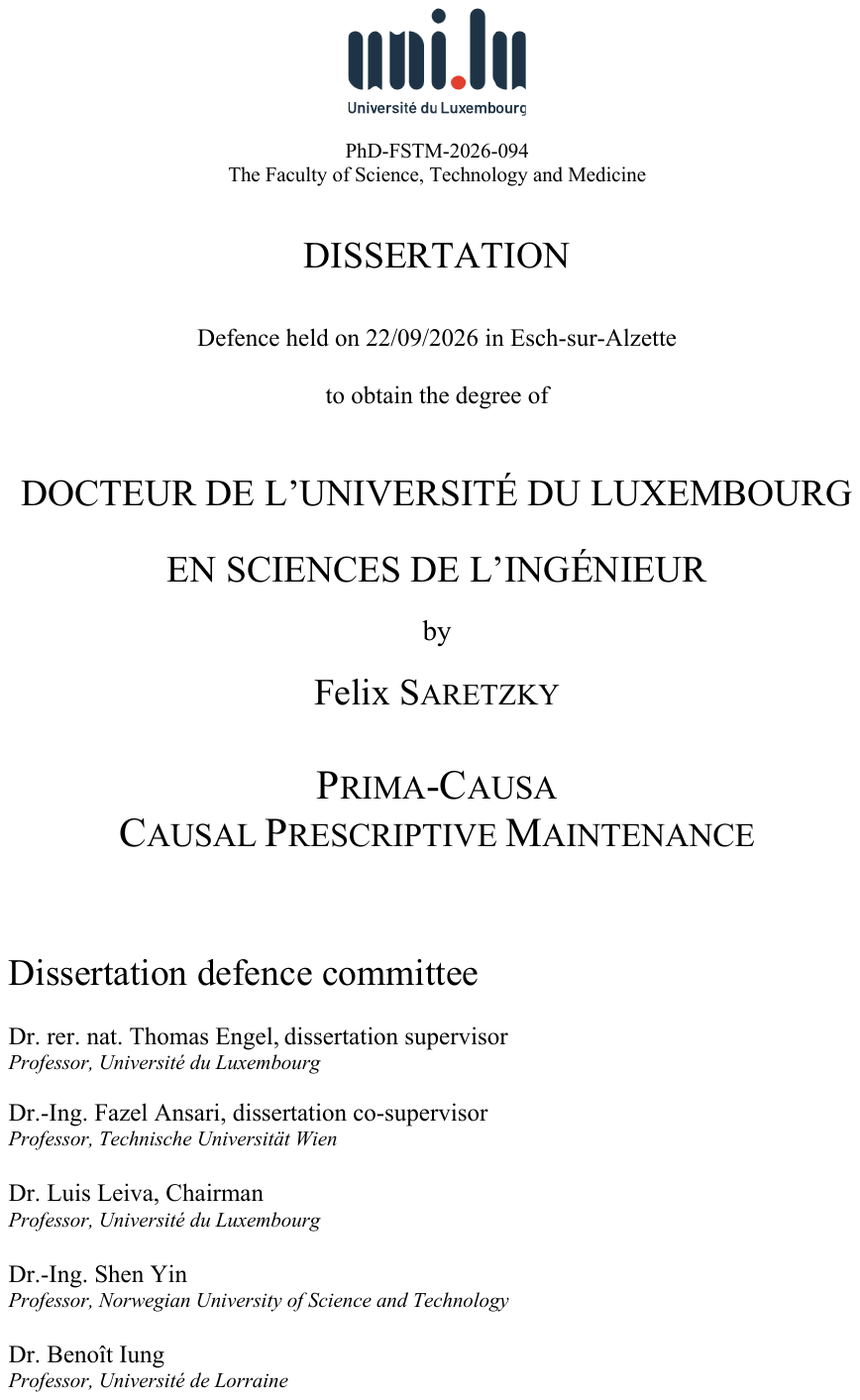}

    \frontmatter

    \cleardoublepage
\thispagestyle{empty}
\vspace*{\fill}
\begin{center}
\textit{In memory of Prof.\,Dr.-Ing. Peter Plapper}
\end{center}
\vspace*{\fill}
\cleardoublepage

    \pagestyle{plain}

    \addchap{Abstract}
    Automated manufacturing requires data-driven decision making under uncertainty, incomplete information, and sparse observational records. Unplanned downtime is a dominant cost of getting those decisions wrong. Prescriptive maintenance denotes the AI-driven recommendation of a maintenance action against an impending failure rather than its mere prediction. Under non-stationary production conditions, each decision point presents a different set of candidate actions and a different production-line state. Established reference architectures support these decisions by coupling predictive models, fitted to observational records, with recommendation logic that encodes curated expert knowledge in rules or optimisation objectives. Selecting a corrective action requires an interventional quantity: how the line behaviour would change if the action were set. Manufacturing data, however, is observational, generated under a condition-dependent policy rather than under experimentation, so a model fitted to predict the outcome estimates an associational quantity. It may describe observed outcomes accurately and still misdirect the choice of action, without the prediction error revealing it. Causal estimators can target the interventional quantity, but they require an expert-led causal study per case, for which the bounded decision window of operation leaves no room. This dissertation analyses how causal reasoning can be embedded into prescriptive maintenance so that actions are ranked by their estimated interventional effect rather than by a predicted outcome. This results in PriMa-Causa, built from three components. A causal reasoning layer integrates the estimate into the PriMa reference model. A parametric simulator supplies the interventional ground truth absent from manufacturing data. A graph-conditioned tabular causal foundation model, pre-trained once on this synthetic production-line prior, answers interventional queries in a single forward pass. The results show that PriMa-Causa attains a lower prescription error than the associational baselines, with the largest margin under weak overlap and sparse context, and that a partially disclosed causal graph reduces this error further. The causal inductive bias carried by the synthetic prior and the graph conditioning can thus replace per-case estimator fitting. The framework complements the estimator with the formal step of translating an operational maintenance problem into the matching causal query, from effect estimation to root-cause analysis. More generally, PriMa-Causa is a step toward automating causal inference on the path to autonomous and self-healing maintenance.

    \selectlanguage{ngerman}
    \addchap{Zusammenfassung}
    \enlargethispage{2\baselineskip}%
Automatisierte Fertigung erfordert datengetriebene Entscheidungen unter Unsicherheit, unvollständiger Information und spärlichen Beobachtungsdaten. Ungeplante Ausfallzeiten sind ein wesentlicher Kostenfaktor, der durch falsche Entscheidungen entsteht. Präskriptive Instandhaltung bezeichnet die KI-gestützte Empfehlung einer Maßnahme gegen einen bevorstehenden Ausfall statt dessen bloßer Vorhersage. Unter nichtstationären Bedingungen stellt sich die Maßnahmenwahl wiederholt, jeweils mit anderer Maßnahmenmenge und anderem Linienzustand. Etablierte Referenzarchitekturen koppeln dafür prädiktive, an Beobachtungsdaten angepasste Modelle mit einer Empfehlungslogik, die kuratiertes Expertenwissen in Regeln oder Optimierungszielen kodiert. Die Wahl einer Maßnahme erfordert eine interventionelle Größe: wie sich das Linienverhalten unter der gesetzten Maßnahme ändern würde. Fertigungsdaten entstehen jedoch unter einer zustandsabhängigen Betriebspolitik statt unter Experimenten, sodass ein auf Vorhersage angepasstes Modell eine assoziative Größe schätzt. Es kann beobachtete Ergebnisse genau beschreiben und dennoch die Maßnahmenwahl fehlleiten, ohne dass die Vorhersagegüte dies offenbart. Kausale Schätzverfahren erfassen die interventionelle Größe zwar, erfordern jedoch eine expertengeführte Kausalstudie je Fall, für die das Entscheidungsfenster der laufenden Produktion keinen Raum lässt. Die Dissertation untersucht, wie sich kausales Schließen in die präskriptive Instandhaltung einbetten lässt, sodass Maßnahmen nach geschätztem interventionellem Effekt statt nach vorhergesagtem Ergebnis priorisiert werden. Daraus resultiert PriMa-Causa: eine Schicht für kausales Schließen im PriMa-Referenzmodell, ein parametrischer Simulator für die in Fertigungsdaten fehlenden wahren interventionellen Größen und ein graph-konditioniertes tabellarisches kausales Foundation-Modell, das, einmalig auf dieser synthetischen A-priori-Verteilung vortrainiert, interventionelle Anfragen in einem einzigen Vorwärtsdurchlauf beantwortet. Die Ergebnisse zeigen, dass PriMa-Causa einen geringeren Präskriptionsfehler erreicht als die assoziativen Vergleichsverfahren, am deutlichsten bei schwacher Überlappung und spärlichem Kontext, und dass ein partiell verfügbarer Kausalgraph ihn weiter verringert. Damit kann der kausale induktive Bias aus A-priori-Verteilung und Graph-Konditionierung die fallweise Schätzeranpassung ersetzen. Der Ansatz ergänzt den Schätzer um die formale Übersetzung des operativen Problems in die passende kausale Anfrage, von Effektschätzung bis Ursachenanalyse. Allgemeiner ist PriMa-Causa ein Schritt zur Automatisierung kausaler Inferenz hin zu autonomer, selbstheilender Instandhaltung.

    \selectlanguage{american}

    \addchap{Acknowledgments}
    \begingroup
\setlength{\parindent}{0pt}
\setlength{\parskip}{0.8em}

This dissertation began under Prof.\,Dr.-Ing.~Peter Plapper. He recognised my potential, gave me the opportunity to start, and supported me with a trust I have not forgotten. I am grateful that this work could begin with him.

I thank my supervisor, Prof.\,Dr.\ rer.\ nat.~Thomas Engel, and Univ.-Prof.\ Dr.-Ing.\ habil.~Fazel Ansari for continuing this path with me, and for a support that went far beyond supervision.

I also thank Prof.\,Dr.~Luis Leiva for serving on the thesis supervisory committee and for chairing the defence. I am grateful to Prof.\,Dr.-Ing.~Shen Yin and Prof.\,Dr.~Benoît Iung for serving on the defence committee and for evaluating this dissertation.

I thank my colleagues at the University of Luxembourg for the discussions that accompanied this work, and in particular Dr.~Marko Orošnjak for the joint work on the systematic review of prescriptive maintenance. I also thank Luisa Braun of Saarland University of Applied Sciences (htw~saar) for the shared path through the doctoral programme.

This doctoral research was funded by Imatec. Its industrial application was developed within the research project PANIC, which was supported by the Ministry of the Economy of Luxembourg. I thank Ludovic Maton, Olivier Leroux, Alain Ruspeler, and Christian Trappen for opening their production line to this research, for their time in the many discussions about it, and for grounding this work in practice.

I also thank Dr.-Ing.~Meysam Minoufekr (Dropslab Technologies) for the open exchange of ideas and the many discussions throughout this work.

Finally, I thank my family and friends, and in particular my partner Isabelle, for their patience and support during my doctoral studies.

\endgroup

    \setuptoc{toc}{totoc}
    \tableofcontents
    
    \addchap{List of Acronyms}
    \begin{acronym}[Fast Moving Consumer Goods] %
    	\acro{CATE}{Conditional Average Treatment Effect}
\acro{CPG}{Causal Process Graph}
\acro{CRL}{Causal Reasoning Layer}
\acro{DAG}{Directed Acyclic Graph}
\acro{FMCG}{Fast Moving Consumer Goods}
\acro{MTBF}{Mean Time Between Failures}
\acro{MTTR}{Mean Time to Repair}
\acro{OEE}{Overall Equipment Effectiveness}
\acro{PdM}{Predictive Maintenance}
\acro{PEHE}{Precision in Estimation of Heterogeneous Effects}
\acro{PFN}{Prior-Data Fitted Network}
\acro{PPD}{Posterior Predictive Distribution}
\acro{PsM}{Prescriptive Maintenance}
\acro{RCA}{Root Cause Analysis}
\acro{SCM}{Structural Causal Model}
\acro{TRL}{Technology Readiness Level}
\acro{XAI}{Explainable Artificial Intelligence}

    \end{acronym}

    \addchap{List of Symbols}
    \label{ch:list-of-symbols}
    
\vspace{0.5em}

\begingroup
\renewcommand{\arraystretch}{1.1}
\setlength{\tabcolsep}{4pt}
\footnotesize

\begin{longtable}{@{}p{0.30\textwidth}p{0.64\textwidth}@{}}
\toprule
\textbf{Symbol} & \textbf{Meaning} \\
\midrule
\endfirsthead

\multicolumn{2}{@{}l}{\footnotesize\textit{List of Symbols (continued).}} \\
\toprule
\textbf{Symbol} & \textbf{Meaning} \\
\midrule
\endhead

\midrule
\multicolumn{2}{r@{}}{\footnotesize\textit{Continued on next page.}} \\
\endfoot

\bottomrule
\endlastfoot

\multicolumn{2}{@{}l}{\textbf{Process variables and sets}} \\*
\addlinespace[2pt]
$\mathbf{V}=\{V_1,\dots,V_n\}$
& Endogenous process variables of the production system. \\

$T\in\mathbf{V}$
& Intervention target (the variable set by $\dop(T{=}t)$). \\

$t$
& Intervention level assigned to $T$. \\

$Y\in\mathbf{V}$
& Decision-relevant outcome variable, typically a recovery- or quality-related key performance indicator such as \acs{MTTR}. \\

$\mathcal{Y}^{\star}\subseteq\mathrm{range}(Y)$
& Desired outcome regime of $Y$, a set of outcome values. \\

$\mathcal{Y}_{\mathrm{cand}}\subseteq\mathbf{V}\setminus\{T\}$
& Mode-2 candidate-outcome set, a set of process variables admissible as downstream outcomes of $T$. \\

$X\subseteq\mathbf{V}\setminus\{T,Y\}$
& Pre-intervention decision-time context. \\

$x$
& Observed realisation of $X$ at the decision time $t_0$. \\

$T_j,\;t_j$
& Candidate intervention target and candidate intervention level. \\

$a_j=(T_j,t_j)$
& Candidate action (variable-level pair). \\

$t_{\mathrm{ref}}$
& Mode-2 reference intervention level used for post-estimation comparison of candidate-outcome distributions. \\

$t_0$
& Decision time of the current case. \\

$\mathrm{MTTI}_{\mathrm{PC}}$
& Mean time to intervention by PriMa-Causa: elapsed time from predictive trigger to recommendation. \\

$\mathrm{MTTR}_{\mathrm{conv}}$
& Conventional \acs{MTTR} interval covering manual troubleshooting, intervention selection, execution, and process restoration. \\

\addlinespace[6pt]
\multicolumn{2}{@{}l}{\textbf{Causal operators and queries}} \\*
\addlinespace[2pt]
$\dop(T{=}t)$
& Pearl's interventional operator: set $T$ to value $t$. \\

$P(Y\mid X{=}x)$
& Observational outcome distribution under context $X{=}x$. \\

$P(Y\mid \dop(T{=}t),X{=}x)$
& Interventional outcome distribution under intervention $\dop(T{=}t)$ and pre-intervention context $X{=}x$. \\

$\mathcal{L}_1,\,\mathcal{L}_2,\,\mathcal{L}_3$
& Layers of the Pearl causal hierarchy: associational, interventional, and counterfactual. \\

$\mathbf{X}{=}\mathbf{x}$,\ $\mathbf{Y}$,\ $\mathbf{z}$
& Generic intervened variable set and its assigned value, generic outcome set, and generic pre-intervention context, used in the set-valued formulation of Section~\ref{sec:bg-causal}. From Chapter~\ref{chap:prima-causa} onwards the case-specific instances are used, with $\mathbf{X}{=}\mathbf{x}$ specialised to $T{=}t$, $\mathbf{Y}$ to $Y$, and $\mathbf{z}$ to $X{=}x$. \\

$\mathcal{M}_{\mathbf{x}}$
& Submodel of $\mathcal{M}$ under the intervention $\dop(\mathbf{x})$, in which the mechanisms of the intervened variables are replaced by constants. \\

$\tau(x)$
& Conditional total effect at context $x$, the interventional contrast $\mathbb{E}[Y\mid \dop(T{=}1),X{=}x]-\mathbb{E}[Y\mid \dop(T{=}0),X{=}x]$. \\

\addlinespace[6pt]
\multicolumn{2}{@{}l}{\textbf{Operationalised prior knowledge}} \\*
\addlinespace[2pt]
$\mathcal{K}$
& Operationalised prior knowledge of the production system. \\

$\mathcal{K}_{\mathrm{op}}\in\{\mathrm{none},\mathrm{ordered},\mathrm{complete}\}$
& Structural completeness of $\mathcal{K}$, an ordinal level $\in\{\mathrm{none},\mathrm{ordered},\mathrm{complete}\}$. \\

$\Phi_{\mathcal{K}}(x)$
& Constraint map that turns $\mathcal{K}$ into machine-usable constraints at context $x$. \\

$\mathcal{G}_{\mathcal{K}}$
& Admissible causal-graph search space induced by $\mathcal{K}$. \\

$\mathcal{A}_{\mathcal{K}}(x)=\{a_j\}_{j}$
& Candidate-action set at context $x$. \\

$\mathcal{R}_{\mathrm{role}}$
& Set of variable-role labels: intervention target, context, outcome, post-intervention descendant, or excluded. \\

$r_{\mathcal{K}}:\mathbf{V}\rightarrow\mathcal{R}_{\mathrm{role}}$
& Variable-role map. \\

$\mathcal{I}_{\mathcal{K}}$
& Knowledge-derived identification constraints (such as temporal ordering or exclusion restrictions). \\

\addlinespace[6pt]
\multicolumn{2}{@{}l}{\textbf{Causal Reasoning Layer pipeline}} \\*
\addlinespace[2pt]
$c$
& Case frame at the input of the \acs{CRL}. \\

$M\in\{\mathrm{M1},\mathrm{M2},\mathrm{M3},\mathrm{M4}\}$
& Decision mode of the case. \\

$T_c,\;Y_c\in\mathbf{V}\cup\{\bot\}$
& Case-specific intervention target and outcome ($\bot$ marks an unspecified slot, for instance $T_c=\bot$ if not yet fixed; not missing sensor data). \\

$d_c$
& Study-design object derived from $c$ and $\Phi_{\mathcal{K}}(x)$. \\

$Q_M$
& Mode-specific target query implied by $M$. \\

$\mathcal{D}_c^{(M)}$
& Planned digital-intervention tensor indexed by context, action, and outcome. \\

$\widehat{\mathcal{D}}_c^{(M)}$
& Estimated counterpart of $\mathcal{D}_c^{(M)}$. \\

$\ell,\,j,\,k$;\ $x_\ell$,\ $Y_k$
& Indices of the digital-intervention tensor: context scenarios, candidate actions, and outcomes. $x_\ell$ and $Y_k$ denote the indexed context scenario and outcome. \\

$S_j(x)$
& Mode-3 candidate-action score. \\

$s_c$
& Case-specific, robustness-qualified scores produced by the \acs{CRL}. \\

$\rho_c$
& Prescription recommendation object passed to the PriMa recommendation layer. \\

\addlinespace[6pt]
\multicolumn{2}{@{}l}{\textbf{Causal discovery}} \\*
\addlinespace[2pt]
$\widehat{G}=(\mathbf{V},\widehat{\mathbf{E}})$
& Estimated causal graph with vertex set $\mathbf{V}$ and edge set $\widehat{\mathbf{E}}$. \\

\addlinespace[6pt]
\multicolumn{2}{@{}l}{\textbf{Causal Process Graph}} \\*
\addlinespace[2pt]
$G=(\mathbf{V},\mathbf{E})$
& Causal Process Graph: directed acyclic graph over the sensor and process-variable set $\mathbf{V}$ with edge set $\mathbf{E}\subseteq\mathbf{V}\times\mathbf{V}$ encoding physically plausible, process-temporally preceding cause-effect relations. \\

$G_{\mathrm{CPG}}$
& Generic graph instance sampled from the \acs{CPG} sampler of Algorithm~\ref{alg:cpg-sample}. \\

$G_{\mathrm{ER}}$
& Graph instance sampled from the Erdős--Rényi \acs{DAG} baseline sampler. \\

$n,\,\eta,\,\mathrm{seed},\,K$
& Parameters of the \acs{CPG} sampler: node count $n$, density level $\eta\in[0,1]$ setting the within-station and cross-station edge rates, reproducibility seed, and station count $K$. \\

$w_b$
& Heavy-tailed in-attractiveness weight of a node $V_b$, drawn lognormally with unit mean. It multiplies the probability of $V_b$ being selected as an edge target, within and across stations. \\

\addlinespace[6pt]
\multicolumn{2}{@{}l}{\textbf{Structural causal model}} \\*
\addlinespace[2pt]
$\mathcal{M}=(\mathbf{U},\mathbf{V},\mathcal{F},P(\mathbf{U}))$
& Structural causal model over the \acs{CPG}: exogenous noise $\mathbf{U}$, endogenous variables $\mathbf{V}$, structural functions $\mathcal{F}$, and the noise distribution $P(\mathbf{U})$. Distinct from the decision mode $M$. \\

$\mathbf{U}=\{U_1,\dots,U_n\}$,\ $U_i$
& Exogenous noise variables with joint distribution $P(\mathbf{U})$. \\

$f_i\in\mathcal{F}$
& Structural function assigning $V_i$ from its parents $\mathrm{pa}(V_i)$ and noise $U_i$. \\

$m_i$
& Deterministic mechanism of the additive-noise family, $V_i=m_i(\mathrm{pa}(V_i))+U_i$. \\

$\mathrm{pa}(V_i)$
& Parent set of $V_i$ in the \acs{CPG}. \\

$z_i$
& Continuous latent score $m_i(\mathrm{pa}(V_i))+U_i$ thresholded to a binary signal. \\

$\theta_i$,\ $\mathrm{LTL}_i$,\ $\mathrm{UTL}_i$
& Switch threshold, and lower and upper tolerance limit of an in-spec flag. \\

$C$
& Number of categories of a nominal discrete variable. \\

$N$
& Number of sample rows (units) of the dataset under consideration. \\

$F$
& Number of covariate columns of the dataset under consideration. \\

\addlinespace[6pt]
\multicolumn{2}{@{}l}{\textbf{Semi-synthetic anchoring}} \\*
\addlinespace[2pt]
$X^{\mathrm{host}}\in\mathbb{R}^{N_{\mathrm{host}}\times F}$
& Covariate matrix of a host dataset used for semi-synthetic anchoring: $N_{\mathrm{host}}$ units (rows) and $F$ retained covariates (columns) of real numbers. \\

$x^{\mathrm{host}}$
& Observed covariate row of $X^{\mathrm{host}}$ for one unit. \\

$f_{\mathrm{trt}},\;f_{\mathrm{rsp}}$
& Treatment-assignment and response-surface functions of the semi-synthetic generator, drawn as generalised additive functions of $X^{\mathrm{host}}$. \\

$e(x^{\mathrm{host}})$
& Propensity score $\operatorname{expit}\bigl(f_{\mathrm{trt}}(x^{\mathrm{host}})\bigr)$ of the simulated treatment. \\

$\mu_0(x^{\mathrm{host}}),\;\mu_1(x^{\mathrm{host}})$
& Response surfaces: evaluations of $f_{\mathrm{rsp}}$ with the treatment indicator held at $0$ and at $1$. \\

$\tau_i$
& True per-unit effect $\mu_1(x^{\mathrm{host}}_i)-\mu_0(x^{\mathrm{host}}_i)$, the oracle quantity scored by the empirical evaluation. \\

\addlinespace[6pt]
\multicolumn{2}{@{}l}{\textbf{Pre-trained estimator and evaluation}} \\*
\addlinespace[2pt]
$q_\theta(\cdot\mid x,D)$
& Predictive distribution of the prior-data fitted network with parameters $\theta$, given query input $x$ and labelled context $D$, approximating the posterior predictive distribution (Section~\ref{sec:bg-pfn}). \\

$\theta$
& Overlap parameter of the empirical evaluation. Distinct from the network parameters in $q_\theta$ and the threshold $\theta_i$. \\

$N_c$
& Number of observational rows in the context of the pre-trained estimator in the empirical evaluation, equal to the training-set size of the fitted baselines. \\

$N_{\mathrm{SCM}}$
& Number of sampled \acs{SCM} instances over which a result of the empirical evaluation is aggregated. \\

\end{longtable}
\endgroup

    \clearpage

    \pagestyle{headings}
    \mainmatter

    \chapter{Introduction}
    \label{chap:introduction}
    
\begin{jointwork}
This chapter sets out the motivation, scope and structure of the dissertation.
The economic and technological context of high-speed production of \acf{FMCG}, as well as
the limitations of correlative decision support, are explained in
Section~\ref{sec:intro-motivation}. The conceptual differentiation on which the
dissertation is based, namely between predicting a fault and recommending a corrective action, is presented in Section~\ref{sec:intro-gentle-intro}. The
challenges arising from this differentiation are examined in
Section~\ref{sec:intro-challenges-rqs}, where the central hypothesis of the
dissertation and the three research questions are derived. The corresponding contributions are
summarised in Section~\ref{sec:intro-contributions}, and the structure of the remainder
of the dissertation is outlined in Section~\ref{sec:intro-outline}. Figure~\ref{fig:ch1_structure} outlines the structure of this chapter.
\end{jointwork}

\vspace{\baselineskip}

\begin{center}
    \includegraphics[width=\textwidth]{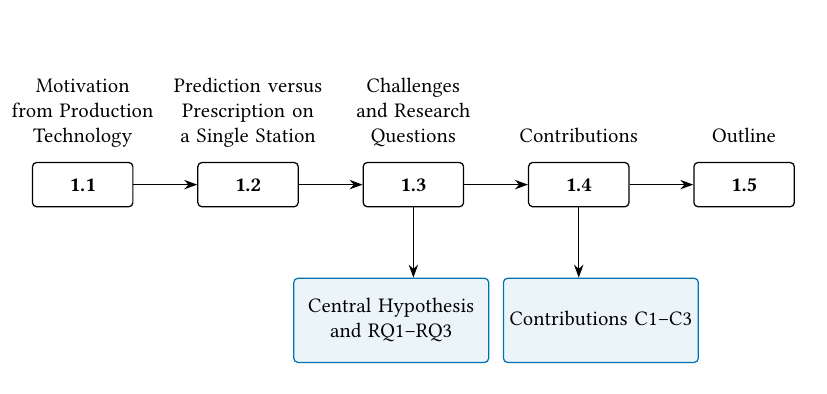}%
    \captionof{figure}{Structure of Chapter~\ref{chap:introduction}.}
    \label{fig:ch1_structure}
\end{center}

\newpage

\section{Motivation from Production Technology}
\label{sec:intro-motivation}

In high-volume, high-precision production lines, such as high-speed lines in the \acs{FMCG} sector,
process steps are closely interlinked, so that even a minor deviation in a single step
can lead to costly quality issues or unplanned downtime~\cite{saretzkyNetworks}.
The economic burden of such downtime is considerable. Unplanned downtime costs
the world’s 500 largest companies an estimated \$1.4~trillion annually, which
corresponds to approximately $11\%$ of their turnover and the gross domestic product of a
major industrialised nation~\cite{siemens}. Specifically for the \acs{FMCG} sector, one hour of
downtime is reported to cost an average of around \$36{,}000, and the sector has recorded the
sharpest rise in total annual downtime costs per plant since 2019~\cite{siemens}. Operationally, these losses are reflected in the \acf{OEE}, whose availability component
is determined by the \acf{MTTR} and the \acf{MTBF}.
Among these key performance indicators, the \acs{MTTR} can be directly reduced through effective fault troubleshooting. The duration of an unplanned shutdown depends on the speed and accuracy of the \acf{RCA}, followed by the selection of a corrective action.

\begin{figure}[!htbp]
	\centering
	\includegraphics[width=\textwidth]{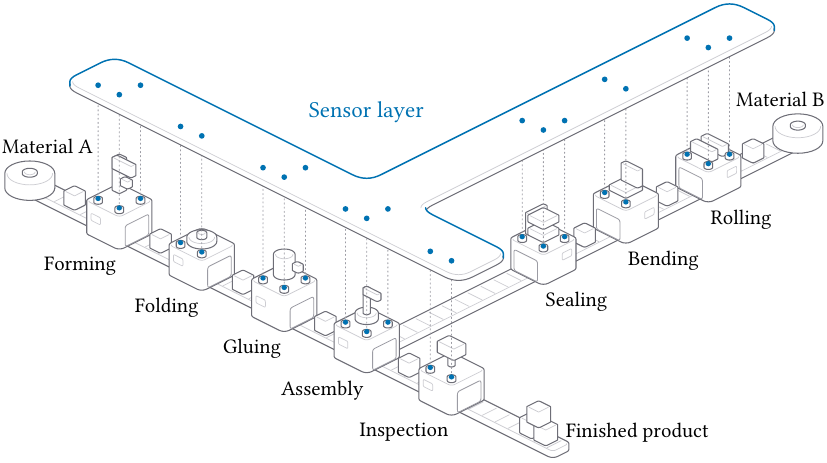}
	\caption{Isometric view of a multi-station high-speed \acs{FMCG} production line and its sensor layer.}
	\label{fig:intro-line}
\end{figure}

A high-speed \acs{FMCG} production line consists of several stations which, under synchronised operating conditions, carry out various
process steps, as shown in
Figure~\ref{fig:intro-line}. The process states of every station are recorded by sensors, which together form the observable sensor layer of the line. Despite a high degree of automation, fault diagnosis and resolution continue to
depend heavily on human operators. Corrective hypotheses are typically derived from
implicit expert knowledge, which is difficult to scale when expertise is scarce or
unevenly distributed across shifts and production sites. This limitation is further
compounded by demographic change and severe labor shortages~\cite{EuropeanCommission2024shortages}. Experienced personnel
leave the workforce, causing a loss of domain knowledge, while qualified
replacements remain difficult to recruit. This consequence is reflected in the recovery statistics. Around $70\%$ of machine downtimes are reported to be resolved within two minutes~\cite{rahm2018}, a recovery time that suggests a rapid
elimination of symptoms rather than the resolution of the underlying cause. If only the
symptom is addressed, the same downtime tends to recur and troubleshooting becomes an iterative process, so the resulting downtime is not permanently reduced.

Established troubleshooting methods offer only limited scope for counteracting this pattern.
Manual \acs{RCA} approaches such as Ishikawa diagrams or the Five Whys method are
subjective and time-consuming, and are difficult to scale as process complexity increases.
Although data-driven methods such as \acf{PdM} and \acf{XAI} can predict \textit{when} a failure
is likely to occur or \textit{which} fault might arise, most do not address \textit{why} it
occurred or \textit{which} action would rectify it. The fundamental difficulty arises because industrial maintenance data is sparse, noisy, heterogeneous, and context-dependent. Consequently, robust decision support requires more than pattern recognition. It must incorporate structural assumptions, including causal ones, alongside domain expertise~\cite{finkPHMPartI,finkPHMPartII}. Furthermore,
purely correlational models are vulnerable to confounding factors and to conditioning on
process variables that lie on the causal path, which means that recommendations tend
to focus on symptoms rather than on causes that can be influenced.

Recent advances in the fields of causal discovery and causal inference have established a methodological
foundation for evaluating such cause-and-effect relationships using observational data from manufacturing~\cite{hagedorn2022,VANDERSCHUEREN2023108798, doubleML}. However, design knowledge on how intervention-oriented maintenance methods should be
structured and evaluated is missing, as existing research is fragmented across \acs{RCA}, predictive maintenance,
prescriptive maintenance and causal machine learning. This dissertation addresses this
gap by developing PriMa-Causa, a layer for causal reasoning that inserts interventional
reasoning between the predictive analysis stage and the maintenance recommendation. The
individual station in Section~\ref{sec:intro-gentle-intro} serves as a reference for the production line as a
whole, which is represented by this layer as a Causal Process Graph. The conceptual framework,
the instantiation as a trained model, the industrial application and the empirical evaluation under production-line data conditions
are developed in the core chapters of this dissertation and outlined in
Section~\ref{sec:intro-outline}.

\begin{figure}[!htbp]
	\centering
	\includegraphics[width=\textwidth]{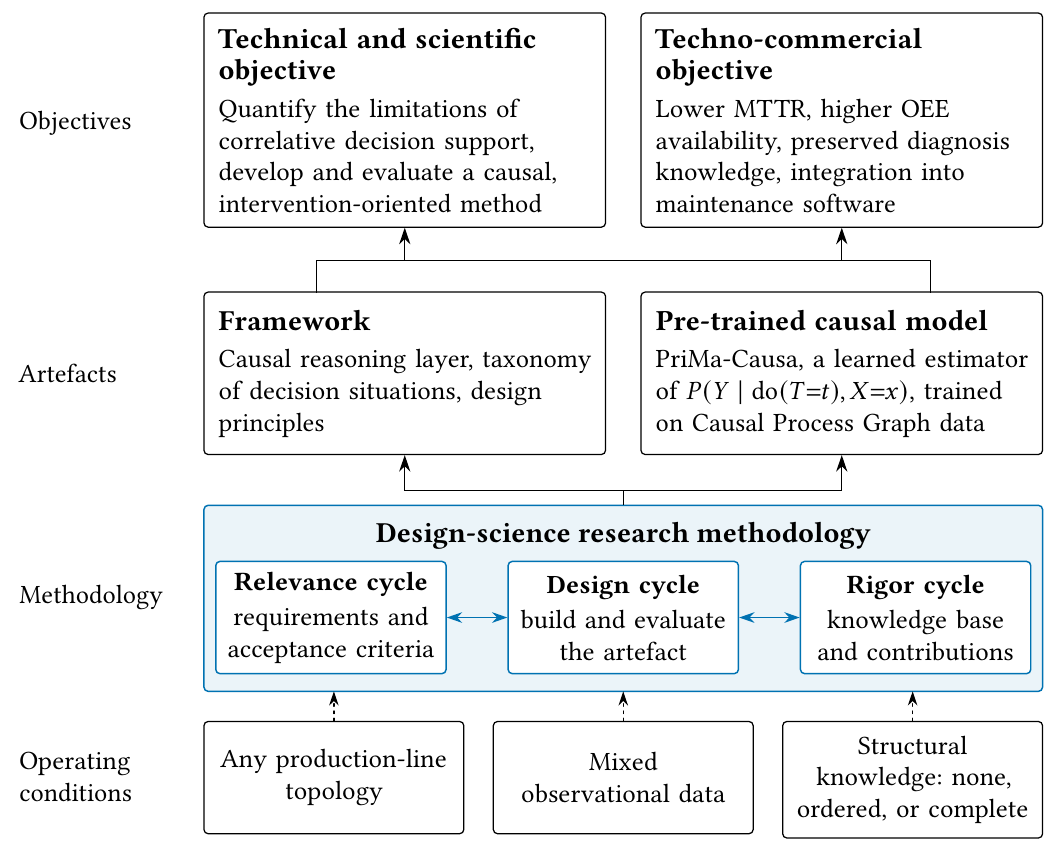}
	\caption{Techno-commercial and technical-scientific objectives of the dissertation.}
	\label{fig:intro-goals}
\end{figure}

This work is guided by two complementary objectives that share a common methodology and are constrained by the operating conditions of \acs{FMCG} production lines, as summarised in
Figure~\ref{fig:intro-goals}. In the figure, solid connectors denote the production relation between
the tiers and dashed connectors denote a constraint. The techno-commercial objective is a sustainable reduction in
\acs{MTTR} and the resulting increase in the availability component of \acs{OEE}, combined
with the preservation of scarce specialist knowledge on fault diagnosis and its integration into
existing maintenance software. The technical and scientific objective is to
quantify the limitations of correlative decision support and to develop and evaluate a causal,
intervention-oriented method that overcomes these limitations. The two objectives are linked by the
design-science research methodology~\cite{hevner2004,peffers2007}, which is organised into a relevance,
a rigor, and a design cycle~\cite{hevner2007}. The method is designed, instantiated, and evaluated
within the design cycle (Section~\ref{sec:intro-outline}). The technical and scientific work is implemented in the form of two artefacts: a framework and a pre-trained causal model, through which the techno-commercial objective
is pursued. The method must function under three operating conditions, namely any
production-line topology, mixed observational data, and structural knowledge across the none, ordered, and complete levels. These objectives are addressed by the three research questions
mentioned in Section~\ref{sec:intro-challenges-rqs}.

\section{Prediction versus Prescription on a Single Station}
\label{sec:intro-gentle-intro}

\begin{figure}[!htbp]
	\centering
	\includegraphics[width=0.4\textwidth]{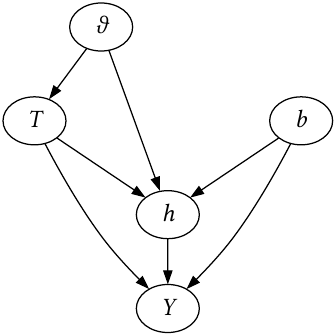}
	\caption[Simplified causal graph of one station.]{Simplified causal graph of one station, with context $\{\vartheta,b\}$, corrective action $T$, in-line
	mediator $h$, and seal-quality outcome $Y$.}
	\label{fig:intro-dag}
\end{figure}
\begin{figure}[!htbp]
	\centering
	\includegraphics[width=\textwidth]{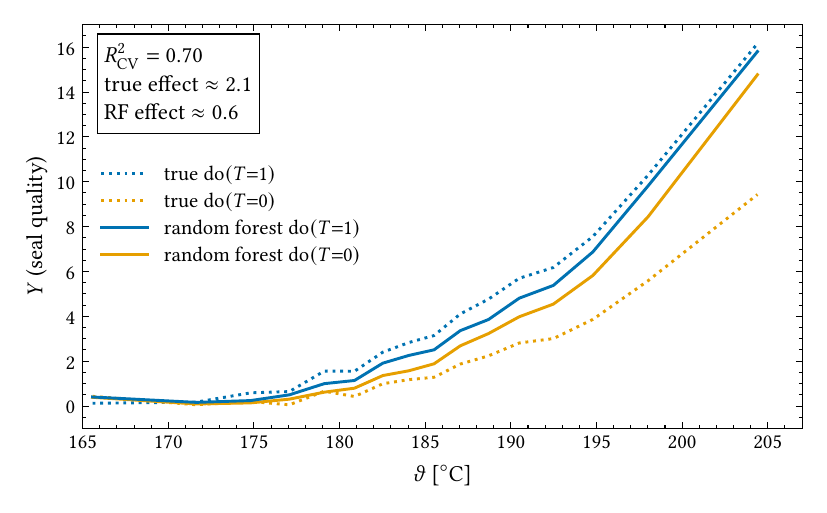}
	\caption[True and estimated interventional responses.]{True responses under $\dop(T{=}0)$ and $\dop(T{=}1)$ (dotted) and their random-forest estimates (solid), on data simulated from the structure of Figure~\ref{fig:intro-dag}.}
	\label{fig:intro-predict-prescribe}
\end{figure}

The vulnerability of correlative decision support can be precisely determined at a single
station on the production line, without yet introducing the formal methods of causal inference
developed in Chapter~\ref{chap: Background}. Figure~\ref{fig:intro-dag} shows a simplified causal graph of an adhesive application and sealing station. In this graph, the process temperature $\vartheta$ and a binary machine status signal $b$ define the context. A corrective action $T$ represents the intervention, while the adhesive height $h$ serves as a measured intermediate signal. The sealing quality $Y$ constitutes the decision-relevant outcome. The structure involves two features.
The temperature $\vartheta$ influences both the choice of intervention and the outcome, thereby
confounding the relationship between $T$ and $Y$. The adhesive height $h$ lies on the path from
the intervention to the outcome. It is therefore a mediator that is itself influenced by $T$. The signal $h$ is a strong predictor of $Y$, since it is a direct cause of $Y$ that carries much of the influence of $\vartheta$, $b$, and $T$, and is therefore likely to be included in a predictive model. 
A model designed to predict $Y$ from all the measured data uses both the confounding factor $\vartheta$ and the mediator $h$ for the prediction. Conditioning on $\vartheta$ removes the confounding bias, while conditioning on the mediator $h$ eliminates part of the effect of the intervention, meaning that good prediction and valid recommendations can diverge.
Figure~\ref{fig:intro-predict-prescribe} illustrates this distinction using data simulated on the basis of the structure shown in Figure~\ref{fig:intro-dag}. The data are generated by the structural equations
\begin{equation}
  \begin{aligned}
    \vartheta &\sim \mathcal{N}(185,\,10^2), \qquad b \sim \mathrm{Bernoulli}(0.5),\\
    T &\sim \mathrm{Bernoulli}\bigl(\operatorname{expit}(0.13\,\Delta\vartheta + 0.003\,\Delta\vartheta^2 + U_T)\bigr),\\
    h &:= 0.12\,\Delta\vartheta + 0.4\,b + 0.6\,T + 0.02\,\Delta\vartheta\,T + 0.001\,\Delta\vartheta^2 + U_h,\\
    Y &:= 1.3\,h + 0.5\,b + 0.4\,T + 0.4\,h\,T + 0.5\,h^2 + U_Y,
  \end{aligned}
  \label{eq:intro-station-scm}
\end{equation}
where $\Delta\vartheta=\vartheta-185$ is the deviation of the process temperature from its set point in degrees Celsius, and the exogenous terms $U_T\sim\mathcal{N}(0,1.4^2)$, $U_h\sim\mathcal{N}(0,1.2^2)$, and $U_Y\sim\mathcal{N}(0,3.5^2)$ are mutually independent.\footnote{The structural equations are adapted, with renamed and rescaled variables, from a tutorial example by J.~Cooiman, \emph{Why Better Predictive Accuracy Does Not Guarantee Better Causal Inference}, CausAI, October 2025 (unpublished course material).} Each equation assigns a variable from its parents in Figure~\ref{fig:intro-dag}, which is the defining property of a structural causal model (Section~\ref{sec:bg-causal}). On a sample of $10\,000$ independent units, a random forest trained on all the measured data captures the seal quality well, with a coefficient of determination
of approximately $0.70$ under three-fold cross-validation, close to the value of about $0.71$ that the true structural equation attains. The same model cannot capture the effect of the
corrective action. Its estimated response under $\dop(T{=}1)$ remains close to its response
under $\dop(T{=}0)$, while the actual intervention responses differ from one another, meaning that the effect of the action is largely ignored. The estimated intervention contrast of approximately $0.6$ is well below the actual contrast of approximately $2.1$. The key driver of this behavior is the mediator $h$, which lies directly on the causal path from the intervention to the outcome. Because $h$ serves as a dominant predictor, conditioning on it blocks the indirect path $T\rightarrow h\rightarrow Y$, so that the estimated contrast retains only the direct part of the effect and misses the larger part transmitted through $h$. The bias could be avoided by excluding $h$ from the model. This remedy, however, presupposes that $h$ is recognised as a mediator, which is structural knowledge. Under predictive model selection, no criterion suggests the omission, since the inclusion of $h$ raises the predictive accuracy. A model selected on the basis of its
predictive accuracy is therefore not necessarily a model that supports decision-making~\cite{fromPredictionToPrescription,DoutreligneHowToSelect}.

In principle, this example illustrates a change in the formulation of the problem rather than a shortcoming in the quality of the model (data problem $\neq$ structural problem).
A prediction estimates what is likely to be observed, summarised by the association quantity
$P(Y\mid X{=}x)$. A maintenance decision, on the other hand, requires the effect of an intervention,
summarised by the intervention quantity $P(Y\mid \dop(T{=}t),X{=}x)$. The operationally
relevant question is therefore not which variable is most strongly associated with a
fault, but which corrective action would reduce the mean time to repair the most.
\textit{What would the seal quality be like if the corrective action were taken, and what
would it be like if no action were taken?} Such a question concerns the impact of a measure in a
context in which that measure may never have been recorded. Therefore, a prescription is most valuable precisely where
the records provide no data, and answering such a question requires extrapolation beyond the
observed operating range, which cannot be justified by predictive accuracy alone. A sustained reduction
in \acs{MTTR} can reduce unplanned downtime and increase the availability component of \acs{OEE}.
In this way, the \acs{MTTR} can act as a bridge between prescriptive maintenance and the assessment
of the effects of interventions. 
Since a model with high predictive accuracy can nevertheless
prescribe an ineffective measure, the gap between prediction and recommendation is a central obstacle to data-driven
decision-making in high-risk areas such as high-speed manufacturing.
In this setting, the gap carries two operational consequences. First, a permanent reduction in downtime depends on interventional estimates that purely predictive models may not be able to provide. Second, executing an incorrect corrective action at line speed can itself incur substantial operational costs. This separation between
association and intervention is
formally justified for prescriptive maintenance in Chapter~\ref{chap: literature review}
(Section~\ref{sec: The Prediction and Prescription Gap}) and incorporated into the
reference architecture for prescriptive maintenance in Chapter~\ref{chap:prima-causa}
(Section~\ref{sec:ch4-bridge-problem}).

\section{Challenges and Research Questions}
\label{sec:intro-challenges-rqs}

Chapter~\ref{chap: literature review} shows that reference models for prescriptive maintenance
are often based on the assumption that an association model of the form
$P(Y\mid X{=}x)$ is sufficient for selecting corrective actions.
In Section~\ref{sec:intro-gentle-intro}, it was shown that this assumption fails in the case of a single station where the
action is distorted by the context and its effect is transmitted via a measured intermediate signal.
The systematic review summarises the resulting shortcomings in four research gaps, RG1 to
RG4 (Section~\ref{sec: Synthesis and Deduction of Research}), which are operationalised by the design objectives DO1 to DO4 of Chapter~\ref{chap:prima-causa} (Subsection~\ref{subsec:ch4-objectives}).
This section formulates the central hypothesis of the dissertation and derives the three research questions from it.

\begin{quotebox}
Under the data conditions of \acs{FMCG} production lines, an interventional estimator of $P(Y\mid \dop(T{=}t),X{=}x)$ equipped with causal inductive bias yields more accurate intervention-effect estimates and a higher expected maintenance efficiency than an associational estimator of $P(Y\mid X{=}x)$.
\end{quotebox}

\noindent
The hypothesis is tested in the simulation studies of Chapter~\ref{chap:empirical-evaluation}, which answer RQ3, while RQ1 and RQ2 establish the design and the instantiation on which this test rests. The three questions correspond to the
build-and-evaluate activities of the design cycle~\cite{hevner2007}, in which the artefact is
designed at reference level (RQ1), instantiated as a learned model (RQ2), and evaluated against
controllable ground truth (RQ3). Their mapping to the research gaps, the design objectives, and the
chapters that address them is summarised in Table~\ref{tab:intro-crosswalk}.

\medskip
\noindent\textbf{RQ1 (design).}\enspace Under which structural conditions can causal inductive
bias be embedded into a prescriptive-maintenance reference model so that recommendations follow
from the interventional quantity $P(Y\mid \dop(T{=}t),X{=}x)$ rather than from the associational
quantity $P(Y\mid X{=}x)$?
\emph{Significance.}\enspace A reference model estimates the associational quantity in its
predictive layer, whereas the selection of a corrective action requires the interventional one.
The reviewed literature specifies neither how causal structure should be inserted to bridge the
two (RG1) nor how causal diagnosis should be connected to action selection (RG4). The answer is
the conceptual core of Chapter~\ref{chap:prima-causa}: a taxonomy of causal-prescriptive decision
situations, a causal reasoning layer, and a set of design principles.

\medskip
\noindent\textbf{RQ2 (instantiation).}\enspace How can design knowledge be
formalised as a trainable model and made analysable under the data conditions of \acs{FMCG} production lines,
where structural knowledge is incomplete, the data is of mixed types, fault events occur rarely and
no interventional reference data is available?
\emph{Significance.}\enspace The interventional outcomes can be identified from observational data
only under explicit conditions and are never directly recorded, meaning that validation, rather than
estimation, constitutes the binding constraint (RG2, RG3). Answering RQ2 therefore requires both a data set
with verifiable ground truth and a model that operates under conditions of incomplete structural
knowledge, as developed in Chapters~\ref{chap:synthetic-data} and~\ref{chap:prima-causa-model}.

\medskip
\noindent\textbf{RQ3 (evaluation).}\enspace To what extent does the resulting causal approach
improve the accuracy of the estimation of intervention effects and the expected maintenance efficiency
compared with associational estimators under controlled data conditions of \acs{FMCG} production lines?
\emph{Significance.}\enspace A causal method is justified only if it identifies intervention effects
that associational models overlook, as illustrated in Section~\ref{sec:intro-gentle-intro},
and only if this advantage is reflected in an operational measure such as the availability component
of the \acs{OEE} (RG2, RG3). The evaluation is carried out against controllable ground truth
and associational baselines in Chapter~\ref{chap:empirical-evaluation}.

\begin{table}[!htbp]
    \centering
    \caption[Mapping of research gaps, design objectives, and research questions.]{Mapping of the
    research gaps of Chapter~\ref{chap: literature review} and the design objectives of
    Chapter~\ref{chap:prima-causa} to the three research questions and the chapters that address
    them.}
    \label{tab:intro-crosswalk}
    \begin{tabular}{@{}lllll@{}}
        \toprule
        \textbf{RQ} & \textbf{Focus} & \textbf{Research gaps} & \textbf{Design objectives} & \textbf{Addressed in} \\
        \midrule
        RQ1 & design        & RG1, RG4 & DO1, DO4 & Ch.~\ref{chap:prima-causa} \\
        RQ2 & instantiation & RG2, RG3 & DO2, DO3 & Ch.~\ref{chap:synthetic-data}, Ch.~\ref{chap:prima-causa-model} \\
        RQ3 & evaluation    & RG2, RG3 & DO2, DO3 & Ch.~\ref{chap:usecase}, Ch.~\ref{chap:empirical-evaluation} \\
        \bottomrule
    \end{tabular}
\end{table}

\begin{figure}[!htbp]
    \centering
    \includegraphics[width=\textwidth]{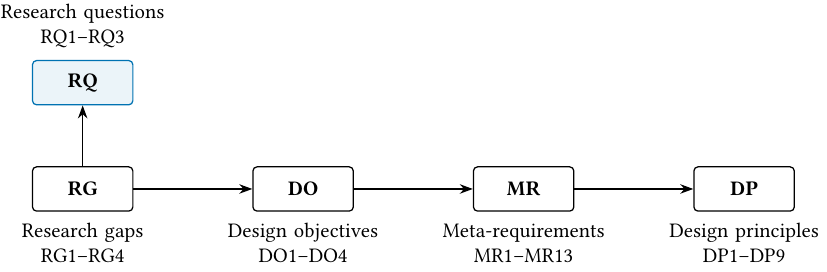}
    \caption[From research gaps to design principles.]{Derivation from research gaps to design principles, with the research questions posed from the gaps.}
    \label{fig:intro-design-ladder}
\end{figure}

\noindent
Research gap RG2 couples an end-to-end methodology with intervention-oriented evaluation, so it is
addressed across RQ2 and RQ3, and research gap RG3 constrains both the instantiation and its
evaluation. The contributions that answer the three questions are summarised in
Section~\ref{sec:intro-contributions}. Figure~\ref{fig:intro-design-ladder} summarises this derivation from research gaps to the meta-require\-ments and design principles of Chapter~\ref{chap:prima-causa}.

\section{Contributions}
\label{sec:intro-contributions}

\begin{figure}[!htbp]
	\centering
	\includegraphics[width=\textwidth]{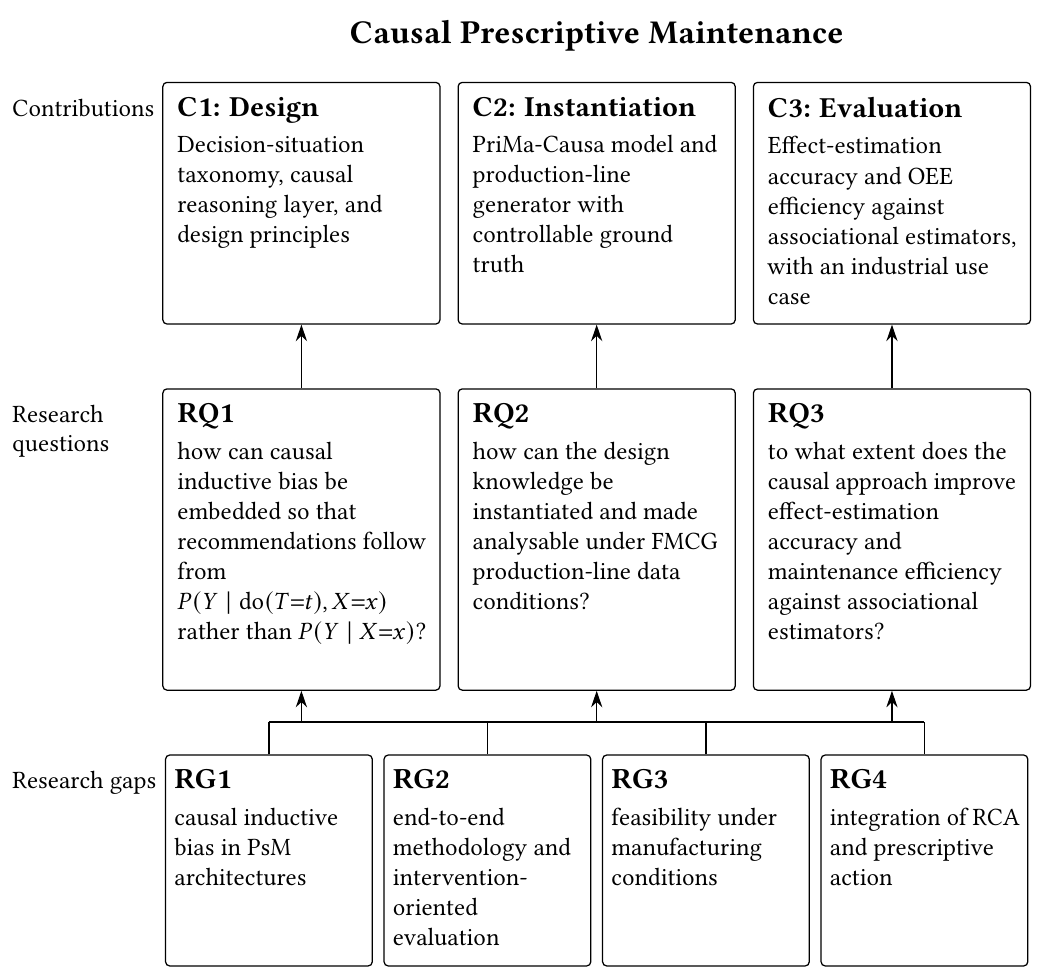}
	\caption[Contributions in relation to research questions and gaps.]{Relation of
	the three contributions to the research questions and to the four research gaps
	that jointly motivate them.}
	\label{fig:intro-contribution-pyramid}
\end{figure}

The three contributions, labelled C1 to C3, rest on the peer-reviewed
publications \cite{saretzkyNetworks, markoReview, saretzky2026integratingcausalfoundationmodel}.
Their relationship to the research questions and research gaps is summarised in
Figure~\ref{fig:intro-contribution-pyramid}. The figure shows the four gaps as a
joint basis for the three questions. The mapping of individual gaps to individual
questions is given in Table~\ref{tab:intro-crosswalk}.

\begin{itemize}[leftmargin=*]
    \item[] \noindent
    \textbf{Saretzky, F.}, Andersen, L., Engel, T., \& Ansari, F. (2026). Integrating a Causal Foundation Model into a Prescriptive Maintenance Framework for Optimising Production-Line OEE. \textit{Procedia CIRP}, \textit{146}, 78--83. \href{https://doi.org/10.1016/j.procir.2026.03.207}{doi.org/\allowbreak 10.1016/\allowbreak j.procir.2026.03.207}

    \item[] \noindent
    \textbf{Saretzky, F.}, Engel, T., \& Ansari, F. (2025). Network-based Root Cause Identification to Improve OEE in High-Precision Manufacturing. \textit{IFAC-PapersOnLine}, \textit{59}(10), 2415--2420. \href{https://doi.org/10.1016/j.ifacol.2025.09.406}{doi.org/10.1016/j.ifacol.2025.09.406}

    \item[] \noindent
    Oro\v{s}njak, M., \textbf{Saretzky, F.}, \& Kedziora, S. (2025). Prescriptive Maintenance: A Systematic Literature Review and Exploratory Meta-Synthesis. \textit{Applied Sciences}, \textit{15}(15), 8507. \href{https://doi.org/10.3390/app15158507}{doi.org/10.3390/app15158507}
\end{itemize}

\medskip
\noindent\textbf{C1 (design).}\enspace Answering RQ1, C1 is a reference-level design artefact that
comprises a taxonomy of causal-prescriptive decision situations, a causal reasoning layer that
inserts interventional estimation between the predictive-analytics step and the recommendation
step, and a set of design principles consolidated from meta-requirements~\cite{walls1992}
(Chapter~\ref{chap:prima-causa}). Its diagnostic foundation, a network-based
root-cause-identification method on the Causal Process Graph, was published
in~\cite{saretzkyNetworks}. The systematic review of the
prescriptive-maintenance landscape from which the architectural gap is derived
was published in~\cite{markoReview}.

\medskip
\noindent\textbf{C2 (instantiation).}\enspace Answering RQ2, C2 instantiates the artefact as the
PriMa-Causa model, a learned causal estimator of $P(Y\mid \dop(T{=}t),X{=}x)$
(Chapter~\ref{chap:prima-causa-model}), together with its data foundation, a parametric
production-line simulator that samples Causal Process Graphs and draws observational and
interventional data with controllable ground truth (Chapter~\ref{chap:synthetic-data}). The
simulator is released as open-source software. The integration of a pre-trained
causal model into the prescriptive-maintenance pipeline was published
in~\cite{saretzky2026integratingcausalfoundationmodel}.

\medskip
\noindent\textbf{C3 (evaluation).}\enspace Answering RQ3, C3 is an empirical evaluation of the
instantiated model against controllable ground truth and associational baselines, measuring the
accuracy of interventional-effect estimation and an expected, \acs{OEE}-based
maintenance-efficiency criterion (Chapter~\ref{chap:empirical-evaluation}), together with a
demonstration on a real production-line use case (Chapter~\ref{chap:usecase}). The
\acs{OEE}-oriented evaluation was reported
in~\cite{saretzky2026integratingcausalfoundationmodel}.

\section{Outline}
\label{sec:intro-outline}

The three cycles introduced in Section~\ref{sec:intro-motivation} organise the dissertation as follows.

\begin{itemize}
  \item \textbf{Relevance cycle.} The application domain of \acs{FMCG} maintenance supplies the
  requirements and the acceptance criteria. The research gaps RG1 to RG4 are derived from the state
  of the art (Chapter~\ref{chap: literature review}) and operationalised as the design objectives
  DO1 to DO4 (Chapter~\ref{chap:prima-causa}). The artefact is demonstrated on a real
  production-line use case (Chapter~\ref{chap:usecase}).
  \item \textbf{Rigor cycle.} The knowledge base of causal inference, machine learning, and
  prescriptive maintenance provides the scientific grounding (Chapters~\ref{chap: Background}
  and~\ref{chap: literature review}). The resulting design knowledge is returned to it through the
  peer-reviewed publications underlying this dissertation.
  \item \textbf{Design cycle.} The artefact is built and evaluated iteratively. Building consolidates
  meta-requirements into design principles at reference level (Chapter~\ref{chap:prima-causa}) and
  instantiates them as a learned model with its data foundation (Chapters~\ref{chap:synthetic-data}
  and~\ref{chap:prima-causa-model}). Evaluation tests the instantiation against controllable ground
  truth (Chapter~\ref{chap:empirical-evaluation}).
\end{itemize}

The methodology is referred to at three levels of granularity throughout the dissertation. The three
cycles describe the research process. Within the design cycle, the build-and-evaluate activities are
divided into design, instantiation, and evaluation, which is the division underlying research
questions RQ1 to RQ3 and contributions C1 to C3. The design knowledge produced in the design cycle is
structured as design objectives, meta-requirements~\cite{walls1992}, and design
principles~\cite{gregor_anatomy} in Chapter~\ref{chap:prima-causa}. The remainder of the dissertation is organised in two parts, related to the research questions and
contributions in Figure~\ref{fig:intro-contribution-pyramid}:

\begin{itemize}
  \item \textbf{Part~I, Foundations and research context:} background of causal inference and
  machine learning (Chapter~\ref{chap: Background}) and the state-of-the-art review that deduces the
  research gaps (Chapter~\ref{chap: literature review}).
  \item \textbf{Part~II, Causal prescriptive maintenance:} the framework, decision-situation
  taxonomy, and design principles (Chapter~\ref{chap:prima-causa}: RQ1, C1), the production-line
  simulator and the PriMa-Causa model (Chapters~\ref{chap:synthetic-data}
  and~\ref{chap:prima-causa-model}: RQ2, C2), the industrial application of the framework
  (Chapter~\ref{chap:usecase}), and the empirical evaluation
  (Chapter~\ref{chap:empirical-evaluation}: RQ3, C3). The dissertation closes with the
  conclusions and outlook of Chapter~\ref{chap:conclusions}.
\end{itemize}

	\part{Foundations and Research Context}

		\chapter{Brief Theoretical Background}
	    \label{chap: Background}
		\begin{jointwork}
This chapter sets out the mathematical and methodological foundations on which the prescriptive methods used in this dissertation are based. Figure~\ref{fig:ch2_structure} provides an overview of the structure of this chapter, in which two complementary foundations are presented for the prescriptive methods discussed in Chapters~\ref{chap:prima-causa}–\ref{chap:prima-causa-model}. Section~\ref{sec:bg-causal} develops the framework for causal inference, which formalises interventions and the interventional estimand $P(Y\mid \dop(T{=}t),X{=}x)$. Section~\ref{sec:bg-pfn} introduces the paradigm of the prior-data fitted network, the amortised in-context estimator, which is specifically tailored to this estimand in Part~II. The structural causal model defined in Section~\ref{sec:bg-causal} is the common object of both approaches, as it reappears in Section~\ref{sec:bg-pfn} as a model prior.
\end{jointwork}

\vspace{\baselineskip}

\begin{center}
    \includegraphics[width=\textwidth]{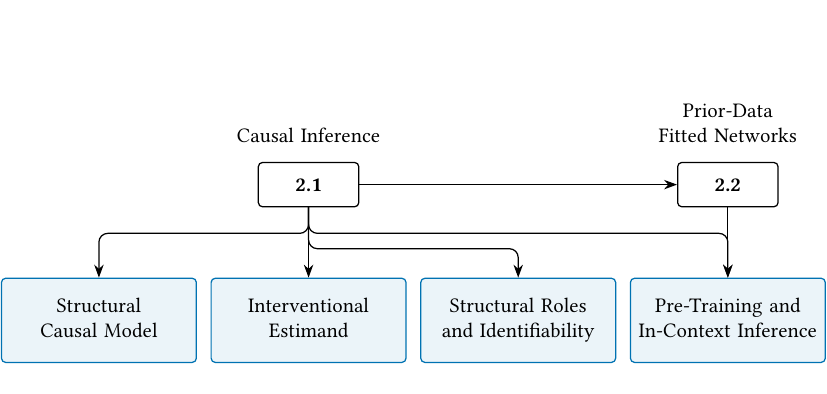}%
    \captionof{figure}{Structure of Chapter~\ref{chap: Background}.}
    \label{fig:ch2_structure}
\end{center}

\newpage

\section{Causal Inference}
\label{sec:bg-causal}

This section introduces the fundamentals of causal inference. The theory of structural causal models
and the underlying do-calculus originate from Pearl~\cite{pearl2009} and
are presented in the formulation by Bareinboim~\cite{bareinboim2025causal}. The
definitions and equations listed below follow these two sources.

\paragraph{Causal inductive bias.}
Learning from limited and noisy production data requires assumptions that the data
alone do not provide. In machine learning, these assumptions constitute the \emph{inductive bias} of a model, the structural prior that constrains its hypothesis space and enables generalisation beyond the observed datasets.
The injection of prior knowledge into models for prognosis and condition monitoring has
recently been organised into architectural inductive biases, training-level learning biases,
and data-level observational biases~\cite{finkPHMPartI,finkPHMPartII}. The architectural inductive biases
alone span relational, spectral, temporal, operator, interpretability, and causal
priors~\cite{finkPHMPartI}. The bias adopted in this dissertation is the causal inductive bias, understood in a broad sense.

\begin{figure}[!htbp]
	\centering
	\begin{minipage}{0.48\textwidth}
		\centering
		\includegraphics[width=0.9\linewidth]{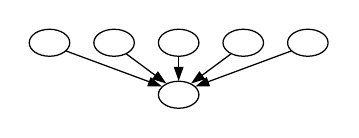}
	\end{minipage}%
	\hfill
	\begin{minipage}{0.48\textwidth}
		\centering
		\includegraphics[width=0.4\linewidth]{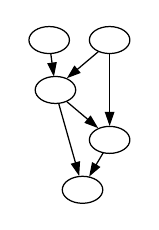}
	\end{minipage}
	\caption[Machine learning versus structural causal modelling.]{Illustration of standard machine learning (left) and structural causal modelling (right) as two representations of a data-generating process.}
	\label{fig:ml-vs-scm}
\end{figure}

A \emph{causal inductive bias} is the totality of causal assumptions that are not contained in the data and that transform an associational learning problem into a well-posed intervention problem. It operates on two levels. At the
level of the model, it ties the estimator to the cause-and-effect structure of the
process, to a structural causal model and its causal graph, rather than to the
correlations between the observed signals~\cite{finkPHMPartI,pearl2009}. At the level
of the problem, it determines how a maintenance situation is formulated as an interventional
query, namely the choice of the intervention target, the outcome, the conditioning context, the
role of each remaining variable, and the permissible interventions. Both levels are
necessary, as an interventional quantity cannot be reconstructed from observational data
without such assumptions~\cite{bareinboim2025causal}.

Figure~\ref{fig:ml-vs-scm} compares the two forms of representation at the model level, the correlative and the structural. Their differences for a single manufacturing station were highlighted in Chapter~\ref{chap:introduction}. The problem level is addressed systematically by the framework set out in Chapter~\ref{chap:prima-causa}, which guides the
translation of a maintenance case into an intervention query and the encoding of prior knowledge as machine-readable constraints.

\paragraph{Structural causal models.}
A structural causal model formalises the mechanisms underlying the production
system. It is a tuple $\mathcal{M}=(\mathbf{U},\mathbf{V},\mathcal{F},P(\mathbf{U}))$,
where $\mathbf{U}$ is a set of exogenous variables determined outside the model,
$\mathbf{V}=\{V_1,\dots,V_n\}$ a set of endogenous variables, $\mathcal{F}=\{f_1,\dots,f_n\}$
a set of structural functions, and $P(\mathbf{U})$ a probability function over the
domain of $\mathbf{U}$. Each endogenous variable is assigned by its mechanism from its
parents and an exogenous term,
\begin{equation}
  V_i := f_i\bigl(\mathrm{pa}(V_i),\,U_i\bigr), \qquad i=1,\dots,n .
  \label{eq:bg-scm-assignment}
\end{equation}
In the production-line
setting the endogenous set $\mathbf{V}$ collects the sensor and process variables of
the line, for instance station temperatures, conveyor speeds, in-line measurements,
and OK and not-OK quality flags. The exogenous set $\mathbf{U}$ collects the variation
left outside the model, such as ambient conditions, raw-material fluctuation, and
wear. Each mechanism $f_i$ encodes the physical dependence of a process variable on
its direct causes. The model is assumed acyclic, so a topological order exists and each draw of the
exogenous variables uniquely fixes the endogenous variables. Together with $P(\mathbf{U})$
this induces a unique observational distribution $P(\mathbf{V})$. The production line is
represented in this way as a structural causal model over the Causal Process Graph,
defined in Section~\ref{sec:synth-cpg}.

\paragraph{The Pearl causal hierarchy.}
A structural causal model induces a three-layer hierarchy of questions, summarised
in Table~\ref{tab:causal-hierarchy}~\cite{pearl2009, bareinboim2025causal}. The associational
layer $\mathcal{L}_1$ concerns what is observed, the interventional layer
$\mathcal{L}_2$ concerns the effect of acting, and the counterfactual layer
$\mathcal{L}_3$ concerns what would have happened under an alternative to a realised
action. The methods of this dissertation operate at the interventional layer, since
a prescriptive decision asks for the effect of a corrective action rather than for
an association.

\begin{table}[!htbp]
  \centering
  \caption[Pearl's causal hierarchy.]{The three layers of the Pearl causal
  hierarchy~\cite{pearl2009}, adapted from~\cite{bareinboim2025causal}.}
  \label{tab:causal-hierarchy}
  \footnotesize
  \setlength{\tabcolsep}{4pt}
  \renewcommand{\arraystretch}{1.1}
  \begin{tabularx}{\textwidth}{@{} c l l X X @{}}
    \toprule
    \textbf{\#} &
    \textbf{Layer (Formalism)} &
    \textbf{Activity} &
    \textbf{Typical Question} &
    \textbf{PsM Example} \\
    \midrule
    1 &
    \shortstack[l]{Association \\ $P(y \mid x)$} &
    Seeing &
    \emph{What is?}\newline How does $X$ change the belief in $Y$? &
    Does elevated process temperature predict a higher seal-defect rate? \\
    \addlinespace[3pt]
    2 &
    \shortstack[l]{Intervention \\ $P(y \mid \dop(x), z)$} &
    Doing &
    \emph{What if?}\newline What happens if $X$ is set? &
    Will recalibrating the sealing station reduce the seal-defect rate? \\
    \addlinespace[3pt]
    3 &
    \shortstack[l]{Counterfactual \\ $P(y_{x} \mid x', y')$} &
    Imagining &
    \emph{Why?}\newline Was it $X$ that caused $Y$? &
    Had the sealing station been recalibrated, would this stop still have occurred? \\
    \bottomrule
  \end{tabularx}
\end{table}

\paragraph{Interventions and the $\dop$-operator.}

The $\dop$-operator is a mathematical device that formally separates active manipulation
``doing'' from passive observation ``seeing''. The conditional probability
$P(\mathbf{Y}\mid\mathbf{X}{=}\mathbf{x})$ describes how the belief about $\mathbf{Y}$ is updated
once $\mathbf{X}$ is observed to be $\mathbf{x}$, whereas the interventional probability
$P(\mathbf{Y}\mid \dop(\mathbf{x}))$ describes the distribution of $\mathbf{Y}$ when an external
force sets $\mathbf{X}$ to $\mathbf{x}$, altering the mechanism of the intervened variable while
leaving the remaining mechanisms unchanged. In industrial practice the functional mechanisms
$f_i$ and the exogenous distributions $P(\mathbf{U})$ are rarely known. The $\dop$-operator nevertheless retains practical value, as the following identification argument shows. Through \emph{identification}, an interventional quantity is mathematically reduced to a functional of the observational distribution, provided the effect is identifiable from the assumed causal graph. This allows the effect to be estimated from passive data, without performing the physical intervention and without knowing the exact mechanisms. Such estimates represent a \emph{provisional causality}, conditional on the set of assumptions encoded by the causal graph~\cite{pearl2018bookOfWhy}.

Formally, an intervention is therefore represented by an operation on the model rather than by conditioning on the observed distribution. Intervening on a set $\mathbf{X}$ and setting it to a value $\mathbf{x}$ replaces the mechanism of each intervened variable by that constant and retains every other mechanism, which yields the submodel

\begin{equation}
  \mathcal{M}_{\mathbf{x}} = (\mathbf{U},\mathbf{V},\mathcal{F}_{\mathbf{x}},P(\mathbf{U})),
  \qquad
  \mathcal{F}_{\mathbf{x}} = \{f_i : V_i \notin \mathbf{X}\} \cup \{\mathbf{X} \leftarrow \mathbf{x}\}.
  \label{eq:bg-submodel}
\end{equation}
This replacement, the $\dop$-operator~\cite{pearl2009,bareinboim2025causal}, severs the
intervened variables from their parents, so that any resulting change in the outcome
follows from the external setting alone, which is what renders the effect causal. On
the production line it corresponds to forcing a controllable lever to a chosen value, for
instance setting the temperature setpoint at a station regardless of the recipe
that would otherwise determine it. The submodel induces the interventional
distribution as the law of $\mathbf{Y}$ when the exogenous noise is drawn from
$P(\mathbf{U})$ and the mechanisms of $\mathcal{M}_{\mathbf{x}}$ are evaluated.
Following the Layer-2 valuation of~\cite{bareinboim2025causal}, for an
outcome set $\mathbf{Y}\subseteq\mathbf{V}$ and a set $A$ of its values,
\begin{equation}
  P\bigl(\mathbf{Y}\in A\mid \dop(\mathbf{x})\bigr)
  = \mathbb{E}_{\mathbf{U}}\bigl[\mathbf{1}\{\mathbf{Y}_{\mathcal{M}_{\mathbf{x}}}(\mathbf{U})\in A\}\bigr],
  \label{eq:bg-do-valuation}
\end{equation}
where $\mathbf{Y}_{\mathcal{M}_{\mathbf{x}}}(\mathbf{U})$ is the value of $\mathbf{Y}$
obtained by solving the submodel $\mathcal{M}_{\mathbf{x}}$ at the exogenous draw and
the expectation is taken over $\mathbf{U}\sim P(\mathbf{U})$.
Conditioning on a pre-intervention context $\mathbf{z}$ is written
$P(\mathbf{Y}\mid \dop(\mathbf{x}),\mathbf{z})$.
Bold symbols denote variable sets and their value assignments. From
Chapter~\ref{chap:prima-causa} onwards the intervened set is the single target $T$ at
level $t$, the outcome set is the single variable $Y$, and the pre-intervention context
is written $X{=}x$.

\paragraph{The interventional estimand.}
\label{subsec: prediction vs prescription}
The transition from prediction to prescription is a change in the type of question
being asked. A predictive model learns the observational conditional expectation
$\mathbb{E}[Y\mid X{=}x]$, an associational quantity on layer $\mathcal{L}_1$. A
prescriptive decision instead requires the corresponding interventional quantity,
on layer $\mathcal{L}_2$. When the goal is to choose between two actions, the decision is
governed by a contrast of two interventional means. For a binary action the contrast
at context $x$ is the conditional total effect
\begin{equation}
	\tau(x) := \mathbb{E}[Y \mid \dop(T{=}1), X{=}x]
	- \mathbb{E}[Y \mid \dop(T{=}0), X{=}x] ,
	\label{eq:cate}
\end{equation}
following the $z$-specific total effect~\cite{bareinboim2025causal}, expressed here as a contrast of interventional means at the line state $x$. In the
production-line setting $\tau(x)$ quantifies how much a corrective action would
change the mean recovery time at line state $x$. 

\begin{figure}[!htbp]
  \centering
  \includegraphics[width=\textwidth]{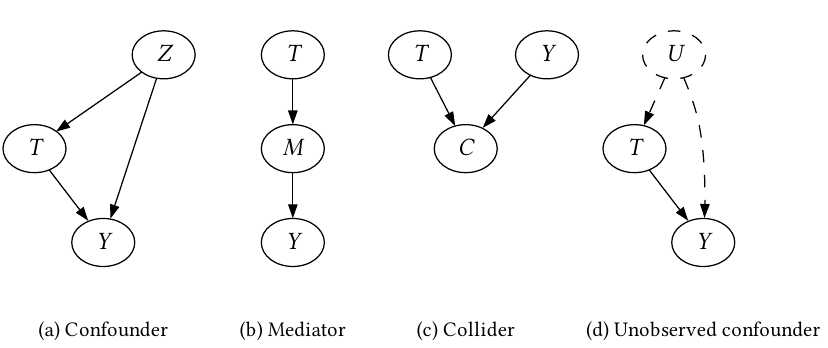}
  \caption[Structural roles of a process variable.]{Roles of a process variable
  relative to an action $T$ and an outcome $Y$: (a)~confounder, (b)~mediator,
  (c)~collider, and (d)~unobserved confounder.}
  \label{fig:bg-structures}
\end{figure}

\paragraph{Confounders, mediators, and colliders.}
The role a process variable plays on the paths between an action and an outcome
determines how it enters an interventional analysis~\cite{pearl2009,bareinboim2025causal}.
A \emph{confounder} is a common cause of the action and the outcome, for instance the
machine state, which drives both the maintenance decision and the recovery time, and it
must be adjusted for. A \emph{mediator} lies on a directed path from the action to the
outcome, for instance an in-line measurement that the action changes and that in turn
affects the quality flag, and conditioning on it blocks the part of the effect
transmitted through it. A \emph{collider} is a common effect of two variables, and
conditioning on it opens a path that was otherwise closed, which is the structural origin
of selection bias, for instance when only parts that reach the final inspection are
recorded. These roles, together with the case of an \emph{unobserved} confounder, represented as a shared exogenous term in $\mathbf{U}$, for instance unlogged tool wear or material batch,
are shown in Figure~\ref{fig:bg-structures}, where the unobserved common cause is drawn
with dashed edges. 

\paragraph{Identifiability and estimation.}
Before causal effects can be estimated from purely observational data, it must be determined whether they are \emph{identifiable}. In causal inference, a causal query, such as an interventional distribution $P(Y \mid \dop(T{=}t))$, is considered identifiable if it can be uniquely computed from any positive joint distribution of the observed variables. The qualitative assumptions encoded in the causal graph determine whether this computation is admissible~\cite{pearl2009}. Identification is the translation of a causal expression containing a $\dop$-operator into an equivalent, $\dop$-free expression consisting solely of observational conditional probabilities. 

This results in a two-step process: First, provided such a translation is possible, the effect is identified. Second, in a subsequent \emph{estimation step}, machine learning models or statistical estimators can be used to calculate these $\dop$-free conditional probabilities from observational data. If no such translation exists, the effect is non-identifiable, and no statistical or machine learning estimator can recover it from observational data alone, however large the sample, in the absence of further assumptions.

Whether an interventional effect can be recovered from observational data is commonly established by
graphical criteria. The \emph{back-door criterion}~\cite{pearl2009}
holds for a set $Z$ that contains no descendant of $T$ and blocks every path between $T$ and
$Y$ that carries an arrow into $T$. Such a set identifies the effect through the back-door
adjustment~\cite{pearl2009},
\begin{equation}
  P\bigl(Y \mid \dop(T{=}t)\bigr) = \sum_{z} P\bigl(Y \mid T{=}t,\, Z{=}z\bigr)\,P(Z{=}z).
  \label{eq:bg-backdoor}
\end{equation}
The \emph{front-door criterion}~\cite{pearl2009} holds if no valid $Z$
is observed, but a set $W$ intersects every directed path from $T$ to $Y$, has no open
back-door path from $T$, and all of its back-door paths to $Y$ are blocked by $T$. Provided
in addition that $P(t,w)>0$, the effect is then recovered from the two stages $T\to W$ and
$W\to Y$ by the front-door adjustment~\cite{pearl2009},
\begin{equation}
  P\bigl(Y \mid \dop(T{=}t)\bigr)
  = \sum_{w} P\bigl(W{=}w \mid T{=}t\bigr)\sum_{t'} P\bigl(Y \mid T{=}t',\, W{=}w\bigr)\,P(T{=}t').
  \label{eq:bg-frontdoor}
\end{equation}
When neither set is available,
an \emph{instrumental variable}~\cite{pearl2009}, a variable that influences
$T$, shares no common cause with $Y$, and affects $Y$ only through $T$, identifies or bounds
the effect under unobserved confounding. The manufacturing-specific form of these conditions
is developed in Chapter~\ref{chap:prima-causa} (Subsection~\ref{subsec:ch4-crl-assumptions}).

\paragraph{Why causal assumptions are unavoidable.}
These identification results are based on structural assumptions that cannot be derived from observational data alone. According to the Causal Hierarchy Theorem (CHT), the hierarchy almost never collapses, which means that questions at a higher layer (such as the effect of an intervention at layer $\mathcal{L}_2$) cannot be answered exclusively on the basis of data from a lower layer (such as passive observations at layer $\mathcal{L}_1$)~\cite{bareinboim2025causal}. Therefore, the causal effect of a corrective maintenance measure cannot, in principle, be inferred from observation logs without resorting to the qualitative structural assumptions encoded in a causal model.

\section{Prior-Data Fitted Networks}
\label{sec:bg-pfn}

\begin{jointwork}
The notation of this section is local to it and follows the cited prior-data fitted
network literature.
\end{jointwork} 

\paragraph{Prior-data fitted networks.}
A \acf{PFN} is a parameterised conditional distribution
$q_\theta(\cdot \mid x, D)$ over a query label $y$, given a query input $x$ and a
labelled context $D=\{(x_i,y_i)\}_{i=1}^{n}$, realised by a single transformer trained
once across many synthetic datasets sampled from a prior
$p(D)$~\cite{muller2022transformers}. Positional encodings are omitted, so the prediction
is invariant to permutations of the context $D$. The object $q_\theta$ approximates is the
Bayesian posterior predictive distribution (PPD)
$p(y\mid x,D)=\int p(y\mid x,\psi)\,p(\psi\mid D)\,\mathrm{d}\psi$, which marginalises a
latent task $\psi$ given the observed context~\cite{muller2022transformers}. This
integral admits a closed form only in special cases and is otherwise intractable. The only
requirement of the construction is the ability to sample whole datasets from the
prior~\cite{muller2022transformers}.

\paragraph{Pre-training.}
Pre-training fits $\theta$ once, offline. Each step draws a dataset together with one
further labelled pair from the prior, masks the held-out label, and fits the network to
predict it from the remaining context. With $q_\theta$ the returned distribution, the
objective is the prior-data negative log-likelihood
\begin{equation}
\ell_\theta
  = \mathbb{E}_{\,D\cup\{x,y\}\sim p(D)}
    \bigl[-\log q_\theta\bigl(y \mid x,D\bigr)\bigr],
\label{eq:bg-pfn-loss}
\end{equation}
minimised by stochastic gradient descent over datasets sampled afresh at each
step~\cite{muller2022transformers}. This objective equals the expected
cross-entropy, and hence the expected Kullback-Leibler divergence up to a
$\theta$-independent constant, between the exact PPD and
$q_\theta$~\cite{muller2022transformers}. Provided the PPD is representable
within the family $\{q_\theta\}$, its global minimiser therefore satisfies
$q_{\theta^\ast}(\cdot\mid x,D)=p(\cdot\mid x,D)$~\cite{muller2022transformers},
so a forward pass performs amortised Bayesian inference rather than per-task optimisation.

\paragraph{In-context inference.}
At deployment the network performs in-context learning, adapting to a task through its
inputs with $\theta$ held fixed. A labelled context $D$ and a query $x$ are read together,
and a full predictive distribution over the query label is returned in a single forward
pass, with no gradient step or parameter update at decision time. The task-specific
computation that a classical estimator performs by fitting is thus amortised into $\theta$
during pre-training. This in-context evaluation, in which the context is consumed as input
while the weights remain fixed, is the property relied upon in the later chapters, where no
estimator is fitted at decision time.

\paragraph{The prior and its structural-causal specialisation.}
\looseness=-1
The prior $p(D)$ is the primary inductive bias of the estimator and is a free, replaceable
design choice, constrained only by the requirement that whole datasets can be sampled from
it~\cite{muller2022transformers}. In the tabular foundation model TabPFN this prior is
constructed from sampled structural causal models, each drawing a mechanism graph and
propagating exogenous noise through it to emit a labelled table~\cite{hollmann2023tabpfn}.
The prior therefore reuses the structural-causal-model object of
Section~\ref{sec:bg-causal} (Equation~\ref{eq:bg-scm-assignment}) as its generative core.

		\chapter{State of the Art and Research Gap}
    \label{chap: literature review}
	
\label{chap:sota}%
\begin{jointwork}
This chapter incorporates my prior peer-reviewed publications~\cite{markoReview, saretzkyNetworks}. Figure~\ref{fig:ch3_structure} outlines the structure of this chapter, which reviews the state of the art from three complementary perspectives that jointly motivate the methodological contributions of this dissertation. These comprise a systematic review of the \acs{PsM} landscape (Section~\ref{sec: Systematic Literature Review on Prescriptive Maintenance}), an evaluation of root cause analysis as the diagnostic step preceding intervention selection (Section~\ref{sec: Review of Root Cause Analysis in Industrial Maintenance}), and an analysis of the associational and interventional inference underlying prescription (Section~\ref{sec: The Prediction and Prescription Gap}), from which the research gaps addressed in this dissertation are derived (Section~\ref{sec: Synthesis and Deduction of Research}).
\end{jointwork}

\vspace{\baselineskip}

\begin{center}
    \includegraphics[width=\textwidth]{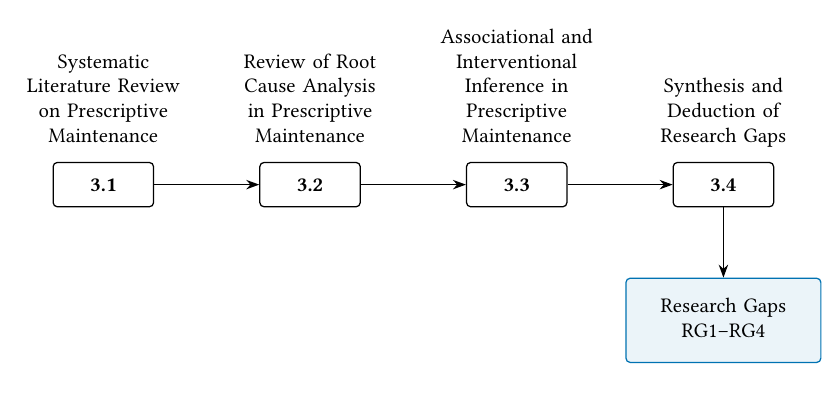}%
    \captionof{figure}{Structure of Chapter~\ref{chap:sota}.}
    \label{fig:ch3_structure}
\end{center}

\newpage

\section{Systematic Literature Review on Prescriptive Maintenance}
\label{sec: Systematic Literature Review on Prescriptive Maintenance}

This section follows the systematic literature review in the full paper \cite{markoReview}.

Although \acf{PdM} has become a well-established paradigm for forecasting equipment failures, it falls short in determining \textit{what} actions should be taken, \textit{when} to take them, and \textit{how} to optimise decisions under limited resources and environmental uncertainty \cite{primaDesign}. In response, \acf{PsM} extends \acs{PdM} by additionally prescribing and optimising the maintenance intervention itself \cite{Prima, Psm_nuclear, Psm_monitoring, ai_dirven_belt}. \acs{PsM} builds on Industry 4.0 technologies, such as the Internet of Things, Cyber-Physical Systems \cite{fazelCPS}, and Digital Twins \cite{digitalTwin}, which supply the required data and integration infrastructure. Industry 5.0 \cite{Industrie50_eu} further directs decision support towards human-centricity, resilience, and sustainability \cite{I50_sustainable, Peruzzini01112024}.
Despite this convergence of technology and policy, the industrial implementation of \acs{PsM} remains fragmented. Existing reviews on the topic are either domain-specific, such as the biopharmaceutical review by Carvalho et al.~\cite{Biopharmaceutical} or the wind turbine studies by Santiago et al.~\cite{windturbinesPSM} and Fox et al.\ \cite{en15020504}, or they address \acs{PsM} only as a peripheral extension of \acs{PdM} \cite{MALLIORIS202480, SHADI2025115668}. While Giacotto et al.\ \cite{Psm_review_twente25} provided the first comprehensive \acs{PsM}-specific review with a taxonomy of prescriptive outputs and methods, their work focused primarily on methodological classification without sufficiently examining implementation maturity or industry-specific applications. Collectively, none of these reviews have offered a fully integrated synthesis of prescriptive actions and enabling technologies, nor have they clustered \acs{PsM} knowledge into thematic research areas grounded in practical deployment. To address this gap, this section presents the findings of an application-oriented systematic literature review of \acs{PsM} studies published between 2013 and 2024, following the methodology and results reported in \cite{markoReview}. Where relevant, the discussion is complemented by more recent developments published after the review period. The review is guided by the following four literature-review questions (LRQ1--LRQ4):

\begin{description}[font=\normalfont\bfseries, leftmargin=1.6cm, labelwidth=1.3cm, labelindent=0pt]
	\item[LRQ1:] What applications, methods, and tools are utilised within the \acs{PsM} body of knowledge?
	\item[LRQ2:] What activities characterise the prescriptive part of \acs{PsM}?
	\item[LRQ3:] What clusters of research can be identified within the existing \acs{PsM} literature?
	\item[LRQ4:] What are the current challenges and future research directions for \acs{PsM}?
\end{description}

\subsection{Methodology and Literature Demographics}
\label{subsec: Methodology and Literature Demographics}

The review followed a six-stage protocol comprising systematic study retrieval via the PRISMA procedure \cite{prisma}, an objective review strategy, structured data extraction, bibliometric metadata analysis, content analysis through multiple correspondence analysis with hierarchical clustering on principal components, and a synthesis addressing the research questions. The search string \textit{``prescriptive maintenance''} was applied to titles, abstracts, and keywords across the Scopus, Web of Science, Google Scholar, OpenAlex, and Dimensions index bases and restricted to primary studies published between 2013 and 2024. After duplicate removal and a two-stage eligibility screening reaching a final interrater agreement of Fleiss' $\kappa = 0.83$ \cite{fleissKappa}, a final corpus of $n = 61$ studies was retained. From each study, metadata, thirteen content-based categorical variables, and qualitative evidence were extracted. Latent research clusters were identified by the correspondence analysis with hierarchical clustering on the categorical codings, with the first three principal components retained (approximately 40\% of the total inertia). The complete search strings, per-database counts, screening flow, coding scheme, and the formal correspondence-analysis and clustering derivation are reported in \cite{markoReview}. Bibliometrically, \acs{PsM} is a comparatively young field with accelerating output. Two primary studies were mentioned before 2017 \cite{before17_1, before17_2}, after which the publication volume grew by at least 50\% ($R^2 = 0.69$). The corpus concentrates in manufacturing (32\%), energy (23\%), and aerospace (13\%), which jointly account for approximately 69\% of the studies. Most studies rely on single case studies (57\%) or simulation (54\%), which indicates that scalable real-world \acs{PsM} deployments remain at an early maturity stage. Optimisation objectives are dominated by time, maintenance cost, and operational cost, whereas social and environmental criteria remain underrepresented and appear mainly in studies published after 2020. The detailed demographic distributions across sectors, methods, and sustainability dimensions are reported in \cite{markoReview}.

\subsection{Technological Enablers and Prescriptive Actions}
\label{subsec: Technological Enablers and Prescriptive Actions}
This subsection analyses the core methods and technologies that enable \acs{PsM} and characterises the spectrum of prescriptive actions identified across industrial sectors. The discussion first addresses the machine learning and deep-learning methods that underpin failure prediction and decision support, before examining how these predictive capabilities are translated into actionable maintenance recommendations.
The \acs{PsM} landscape is methodologically dominated by supervised machine learning models for the predictive layer. The most frequently employed algorithms across the reviewed studies are catalogued in~\cite{markoReview}. Random Forest (RF) is especially prevalent, since its classification accuracy and inherent feature-importance ranking are reported to make it well-suited for root cause analysis and anomaly detection~\cite{markoReview, GIBEY2024234864, sun2023ai}. Support Vector Machines (SVMs) and Artificial Neural Networks (ANNs) are similarly widespread, typically trained on historical failure data or sensor signals to forecast remaining useful life (RUL) or classify failure types~\cite{BIEBL202064}. Unsupervised methods such as $k$-Means clustering and Principal Component Analysis complement these approaches by supporting exploratory analysis and dimensionality reduction.

Complementing classical machine learning, deep-learning models have expanded the scope of \acs{PsM}. Several functional categories emerge from the reviewed studies. \textit{Temporal modelling and prognosis} is primarily addressed by Long Short-Term Memory (LSTM) Networks, Recurrent Neural Networks, and Echo State Networks, which are well-suited for capturing long-term dependencies in time-series sensor data. LSTM-based architectures are identified in the reviewed literature as a prominent approach to remaining useful life estimation in complex assets such as turbofan engines and batteries~\cite{2024101405}, while Echo State Networks offer adaptive modelling capabilities in noisy, real-time environments~\cite{GIBEY2024234864}. \textit{Spatial-temporal feature extraction} is addressed by Convolutional Neural Networks and Deep Neural Networks, which are applied for early fault detection from vibration and thermal data~\cite{Vermesan}. \textit{Generative learning and anomaly detection} using Generative Adversarial Networks enable the simulation of rare failure events and the generation of synthetic training data for imbalanced datasets. Such imbalance is a common challenge in industrial settings, where failure samples are scarce~\cite{jmse12030493}.
Unlike the supervised and unsupervised methods discussed above, which primarily serve the predictive stage, deep reinforcement learning directly addresses the decision-making problem by learning optimal maintenance policies through interaction with an environment. These agents recommend maintenance actions by balancing cost, operational constraints, and degradation state, and have been deployed for component-level repair-versus-replace decisions~\cite{deepreinforcement}, fleet-level scheduling~\cite{GOBY2023109165}, and sustainability-aware maintenance optimisation~\cite{sun2023ai}.
Despite the growing adoption of deep reinforcement learning, the interpretability of 
\acs{PsM} models remains a recognised challenge, particularly in 
safety-critical industrial applications. Only isolated studies address this through \acf{XAI}: Eider et al.~\cite{EIDER2023100845} proposed context-aware natural language explanations for battery management, Petroutsatou and Ladopoulos~\cite{Petroutsatou_2022} linked \acs{XAI} to repair-versus-replace decisions, and 
Madathil et al.~\cite{xaiManu} highlighted that interpretable 
machine learning methods can improve operator understanding of recommended 
actions. The epistemological limitations of applying post-hoc 
explanation methods to associational models are examined in 
Subsection~\ref{subsec: structural limitations}.
More fundamentally, the review indicates that causal inference methods are largely absent from the \acs{PsM} corpus. Among the 61 reviewed studies, only Vanderschueren et al.~\cite{VANDERSCHUEREN2023108798} explicitly leveraged causal machine learning to estimate individualised treatment effects for maintenance scheduling. The implications of this observation for the prescriptive objective are analysed in Section~\ref{sec: The Prediction and Prescription Gap}.
Beyond the technological enablers, the prescriptive dimension of \acs{PsM} is characterised by the ability to convert predictive outputs, such as failure diagnoses or remaining useful life estimates, into recommendations that specify \textit{what} activities should be taken, \textit{when}, \textit{by whom}, and under \textit{what constraints}. The spectrum of prescriptive activities identified in the review ranges from localised component-level decisions to systemic fleet-level interventions and is catalogued in~\cite{markoReview}.

At the most granular level, component-level repair-versus-replace decisions rely on degradation assessments and resource constraints~\cite{GIBEY2024234864, DUTTA2022107825, gordon2021datadriven, Mao_2023, ESPOSITO20221974}. These are complemented by parts-planning activities that identify components for repair or replacement in advance based on sensor-based diagnosis~\cite{Belov}. At a higher level, prioritisation and intelligent scheduling aim to reduce unplanned downtime at optimal cost~\cite{choubey2020holistic}, while strategic decision support encompasses the selection among alternative maintenance strategies such as preventive versus corrective or outsourcing versus in-house repair~\cite{ELBASHEER20222605, Vermesan, markoReview}. A distinctive feature of \acs{PsM} is the co-optimisation of maintenance and production activities, synchronising schedules to minimise disruptions and balance operational load~\cite{GIBEY2024234864, WESENDRUP20241210, Vermesan}. The most advanced prescriptive capability is automated execution and self-regulation, where systems autonomously adjust operating parameters, such as reducing charging rates when thermal thresholds are reached, without human intervention~\cite{GIBEY2024234864, Vermesan}. At the broadest scope, fleet- and mission-level prescriptions coordinate maintenance across multiple assets, considering mission requirements, availability windows, and component fatigue~\cite{bond, choubey2020holistic, GOBY2023109165}.

The operationalisation of these prescriptive activities is highly dependent on the industrial domain. In \textit{manufacturing}, actions are oriented towards preventing disruptions in production flows by minimising downtime and aligning maintenance with operational schedules~\cite{wessendrup22, ELBASHEER20222605}. Here, horizontal and vertical integration (H\&VI) connects maintenance with planning systems for workforce assignment, spare parts forecasting, and production reconfiguration. In the \textit{energy sector}, \acs{PsM} exhibits a higher degree of autonomy, with self-optimising systems that rely on real-time data and machine learning algorithms to trigger early warnings and adapt operations. Prescriptive actions are embedded within broader strategic goals such as grid stability, load balancing, and emission reduction. In \textit{aerospace}, \acs{PsM} is driven by mission-critical priorities including flight delay minimisation~\cite{Meissner1, MEISSNER2021107812} and safety-aware recommendations, integrating data from onboard diagnostics, flight history, and sensor logs for real-time adaptive decisions. The \textit{railway sector} prioritises user experience, infrastructure safety, and service reliability, with prescriptions determined by spatial targeting of high-risk infrastructure segments and strategic asset prioritisation.
Across all sectors, a common trajectory emerges:
\begin{quotebox}
\acs{PsM} moves industrial maintenance from discrete decision events towards continuous prescriptive control. This transition requires the integration capabilities of Industry 4.0 technologies, particularly horizontal and vertical integration, and mature predictive and causal models.
\end{quotebox}

\subsection{Thematic Research Clusters}
\label{subsec: Thematic Research Clusters}

\begin{figure}[!htbp]
	\centering
	\includegraphics[width=0.8\textwidth]{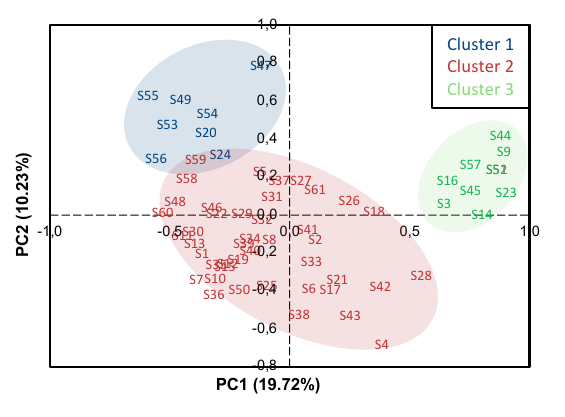}
	\caption[Cluster plot of \acs{PsM} research studies.]{Cluster plot of \acs{PsM} research studies in the PC1--PC2 space, coloured by cluster assignment resulting from the hierarchical clustering on the first three principal components. Adapted from \cite{markoReview}.}
	\label{fig: MCA HCPC}
\end{figure}

Applying multiple correspondence analysis with hierarchical clustering on principal components to the categorical dimensions of Application Scope, Study Focus, Systemic and Sustainability Aspects, and Hierarchical Decision-Making Level resulted in three distinguishable research clusters (Figure~\ref{fig: MCA HCPC}). Their characteristics are described below, with the full description matrix reported in \cite{markoReview}.

\textbf{Cluster~1} focuses on technical integration and real-time diagnostics at the machine- and (multi)component level \cite{sun2023ai, gordon2021datadriven, Vermesan, deepreinforcement}. Studies in this cluster are predominantly situated in the manufacturing and energy sectors, where the prescriptive layer relies on deep reinforcement learning and Internet-of-Things deployments to identify and prevent failures in sensor-rich operational settings. While these applications demonstrate the potential of artificial intelligence at the shop-floor level, enterprise-scale integration and generalisability are limited.

\textbf{Cluster~2} is the largest cluster and extends the scope to tactical and strategic decision-making for fleets and process-wide systems \cite{Prima, primaDesign, Psm_nuclear, ELBASHEER20222605, GIBEY2024234864, bond, choubey2020holistic, EIDER2023100845, 2024101405, GOBY2023109165, Mao_2023, Meissner1, MEISSNER2021107812, primaX}. It encompasses enterprise-level decision-support systems employing probabilistic modelling, fuzzy logic, Monte Carlo simulations, and optimisation frameworks. Unlike Cluster~1, these studies explicitly incorporate human factors and address planning under uncertainty. The PriMa reference models \cite{Prima, primaX} are representative examples, spanning multiple functional dimensions across hierarchical decision-making levels.
 
\textbf{Cluster~3} diverges from the other two by adopting a strategic and conceptual perspective \cite{wessendrup22, Pinar, Petroutsatou_2022, franzen}. Rather than proposing algorithmic solutions, these studies frame \acs{PsM} within broader discussions of digital transformation, organisational readiness, and workforce dynamics. This top-down orientation treats \acs{PsM} as a management-level concern rather than a purely operational function.
 
The three clusters are complementary rather than hierarchical. The analytics of Cluster~1 become effective only with the organisational alignment of Cluster~2 and the socio-technical support structures of Cluster~3. Three observations from this analysis are particularly relevant for the remainder of this chapter:
 
\begin{enumerate}
	\item \textbf{Prescriptive validation is underrepresented.} While prediction is well-established, few studies demonstrate through empirical validation that recommended actions improve outcomes.
	
	\item \textbf{Cross-level integration is limited.} The integration of \acs{PsM} across decision-making levels depends critically on horizontal and vertical integration technologies, yet many studies operate in isolation at a single hierarchical level.
	
	\item \textbf{Causal reasoning is rare.} All three clusters rely on associative models that detect correlations between symptoms and failures but do not model the underlying causal mechanisms. This limitation is fundamental, since the prescriptive objective, which answers \textit{what should be done} and \textit{what actions are required to achieve a desired outcome}, is inherently interventional.
\end{enumerate}

The third observation raises the question whether the diagnostic step that typically precedes intervention selection can compensate for the limitations of the predictive layer. In manufacturing practice, \acs{RCA} occupies this intermediate position between fault detection and the choice of corrective action. The following section examines the extent to which current \acs{RCA} methods, whether knowledge-driven, data-driven, or causal, provide the transition from diagnosis to prescription.

\section{Review of Root Cause Analysis in Prescriptive Maintenance}
\label{sec: Review of Root Cause Analysis in Industrial Maintenance}
 
Root cause analysis (\acs{RCA}) is a systematic approach to identifying the underlying causes of problems in production processes, which facilitates the implementation of corrective actions~\cite{Pietsch}. In the context of industrial maintenance, \acs{RCA} occupies a central position between fault detection and the selection of corrective actions: While predictive models can indicate \textit{that} a failure is imminent, \acs{RCA} aims to determine \textit{why} the failure occurs, thereby narrowing down the range of possible corrective actions. This diagnostic function makes \acs{RCA} an obvious, yet largely unexploited component of \acs{PsM}, since effective measures require knowledge of the causal mechanisms that cause the failure. Traditionally, \acs{RCA} in manufacturing has relied on knowledge-driven methods grounded in expert judgement. Established quality management tools such as Ishikawa diagrams, the Five Whys technique, Pareto analysis, and Failure Mode and Effect Analysis remain widely used for tracing problems back to their sources~\cite{Pietsch}. These approaches are effective in well-understood processes with experienced operators but are inherently limited by their dependence on individual expertise, the difficulty of formalising tacit knowledge, and their inability to scale to complex, high-dimensional production environments~\cite{Pietsch}. As manufacturing systems increase in complexity through the integration of Internet-of-Things sensors, cyber-physical systems, and multi-stage process chains, the volume of potentially relevant signals exceeds the capacity of purely manual inspection~\cite{saretzkyNetworks}. In response, there has been a growing body of work on data-driven approaches to \acs{RCA}. A recent scoping review by Pietsch et al.~\cite{Pietsch} surveyed both knowledge-driven and data-driven \acs{RCA} methods in manufacturing and identified an accelerating shift towards the latter. Among data-driven methods, \acs{XAI} techniques, in particular Shapley Additive Explanations and Local Interpretable Model-agnostic Explanations, are the most prevalent approach~\cite{Pietsch}. These methods interpret a trained classification or regression model post hoc by attributing its predictions to individual input features, which are then presented to process experts as candidate causes~\cite{Pietsch, Lin2025CLARiCCL}. Tree-based models are similarly used, exploiting their built-in feature importance rankings for fault attribution~\cite{saiRCA}. Despite their practical utility, purely data-driven \acs{RCA} methods have a fundamental limitation that arises from the models they interpret: they may capture associational patterns rather than causal mechanisms. An \acs{XAI} explanation identifies features that are statistically relevant for a model's prediction, but these features need not be causal drivers of the fault. They may instead be consequences of the true root cause, correlated symptoms, or artefacts of confounding~\cite{Carlonicausality}. As the Fraunhofer IAIS white paper on AI-based \acs{RCA} emphasises, interpretable models can indicate \textit{which} variables are relevant for a given prediction task, but this does not automatically establish the causal chain from root cause to failure event~\cite{kobialka2021kibasierte}. This distinction is critical for maintenance decision-making: recommending an intervention on a correlated but non-causal variable may not resolve the underlying problem. Recognising this limitation, a growing body of work has begun to integrate causal reasoning into \acs{RCA}. These causal approaches can be broadly organised along Pearl's causal hierarchy~\cite{pearl2009}. At the interventional layer $\mathcal{L}_2$, several methods model faults as unexpected interventions on a causal graph and identify root causes by detecting which causal mechanisms have shifted. Budhathoki et al.~\cite{budhathoki21aWhy} formalised the question \textit{``Why did the distribution change?''} by decomposing changes in the joint distribution into contributions of individual causal mechanisms using KL divergence, attributing distributional shifts to the nodes whose conditional distributions have changed. The CIRCA algorithm~\cite{CIRCAalgo} and the RCD method~\cite{RCD} operationalise a similar idea by testing which variables' causal conditionals have shifted between normal and abnormal periods, with the RCD method specifically doing so by introducing a binary intervention indicator. At the counterfactual layer $\mathcal{L}_3$, Budhathoki et al.~\cite{budhathoki22a} proposed a formal method for attributing outlier events to their root causes using \acs{SCM}s. Their approach asks the counterfactual question \textit{``Would the target event not have been an outlier had the causal mechanism at a given node behaved normally?''} and quantifies contributions using Shapley values. Related counterfactual \acs{RCA} approaches combine anomaly detection with causal graphs to produce causally justified explanations~\cite{Rehak2023}.
Despite these advances, no \acs{PsM} reference model has yet fully integrated the output of causal \acs{RCA} into action selection. Most \acs{RCA} methods, whether knowledge-driven, data-driven, or causal, are designed as reactive diagnostic tools: they are applied \textit{after} a fault has occurred to explain its origin. The transition from diagnosis to prescription, from identifying \textit{why} a fault occurred to determining \textit{what action} should be taken to prevent its recurrence, requires an additional inferential step that current \acs{RCA} frameworks do not systematically provide. Causal \acs{RCA} methods are the closest precursor, as the interventional reasoning they employ is, in principle, the same reasoning required for causal effect estimation and intervention selection. The prescriptive implications of causal root cause identification, such as ranking candidate interventions by their expected causal effect on the target outcome, have not yet been examined in the manufacturing context.

\section{Associational and Interventional Inference in Prescriptive Maintenance}
\label{sec: The Prediction and Prescription Gap}

As established in Subsection~\ref{subsec: Thematic Research Clusters}, all three identified research clusters within the \acs{PsM} literature are predominantly built on associational models that learn through correlations between symptoms and faults, with explicit modelling of the underlying causal mechanisms remaining a rare exception. This section examines the analytical scope of existing \acs{PsM} reference models, identifies structural properties of observational manufacturing data that limit the prescriptive validity of associational methods, and surveys emerging causal approaches that address these limitations. The chapter concludes with a synthesis of the resulting research gaps.
\subsection{Prescriptive Maintenance Reference Models and Analytical Scope}
\label{subsec: from pdm to psm}
The distinction between \acs{PdM} and \acs{PsM} has been characterised across the literature along three dimensions: objective, method, and outcome. \acs{PdM} mainly aims to forecast \textit{when} equipment failure might occur, using predictive analytics such as statistical models and sensor-based condition monitoring to schedule maintenance at the most opportune time~\cite{Psm_review_twente25}. \acs{PsM} extends this objective by additionally prescribing \textit{what} specific actions should be taken to prevent or mitigate failures, leveraging prescriptive analytics, optimisation algorithms, and decision support systems~\cite{Prima, primaX, Psm_review_twente25}. While \acs{PdM} produces a \textit{failure prediction} as its primary output, \acs{PsM} aims to deliver a \textit{course of action prescription}, a set of recommended tasks that answers the question ``What should be done?'' to optimise maintenance ahead of time~\cite{Psm_review_twente25}. The PriMa reference model, introduced by Ansari et al.~\cite{Prima}, operationalises this transition through a four-layer architecture: (1)~a data management layer for processing multimodal and heterogeneous data from multidimensional sources, (2)~a predictive data analytics toolbox comprising supervised, unsupervised, and semi-supervised learning methods, (3)~a recommender and decision support dashboard, and (4)~an overarching layer for semantic-based learning and reasoning. The model explicitly combines predictive analytics with domain-specific knowledge representation and semantic reasoning to generate maintenance recommendations that are integrated with production planning and control~\cite{Prima, primaDesign}. The related PriMa-X extension~\cite{primaX} introduced a maturity assessment framework, recognising that the pathway from data acquisition to actionable prescription involves multiple stages of increasing analytical complexity. Across both formulations, the prescriptive capability relies on three interacting components: predictive data analytics, structured domain knowledge, and rule- or simulation-based reasoning to derive action plans from predicted states. In the review, Giacotto et al.~\cite{Psm_review_twente25} consolidate the defining characteristics of \acs{PsM} into six attributes: it is the most mature maintenance strategy, it is knowledge-based and data-driven, it is predictive, holistic (integrating maintenance with production and operations), self-learning (incorporating feedback loops for continuous improvement), and adaptive to different industries and asset types. The review identifies six enablers (prescriptive analytics, the IoT, failure prognosis systems, data analysis, decision support systems, and ontology) and six primary outputs, of which the most prominent is the course of action prescription, reported in 53 of 76 reviewed studies. However, most of the prescriptive analytics approaches reported in these reviews are associational. The systematic review presented in Section~\ref{sec: Systematic Literature Review on Prescriptive Maintenance} confirms that machine-learning, deep-learning, and deep-reinforcement-learning models dominate the predictive and prescriptive layers, while causal inference methods appear in only isolated instances~\cite{VANDERSCHUEREN2023108798}. Similarly, Giacotto et al.~\cite{Psm_review_twente25} list causal inference and causal machine learning as prescriptive analytics methods but identify only two studies employing them. This observation is directly relevant to the PriMa architecture. As formalised in Section~\ref{subsec: prediction vs prescription}, the predictive analytics toolbox (Layer~2) of PriMa estimates the observational conditional $\mathbb{E}[Y \mid X{=}x]$, whereas the recommender (Layer~3) must evaluate the interventional quantity $P(Y \mid \dop(T{=}t), X{=}x)$. This transition constitutes a shift from an $\mathcal{L}_1$ associational query to an $\mathcal{L}_2$ interventional query in Pearl's causal hierarchy (Table~\ref{tab:causal-hierarchy}). If the recommender layer relies directly on the associational output of the predictive layer without adjusting for the causal structure of the data-generating process, the resulting prescriptions may be confounded. 

\subsection{Confounding, Selection Bias, and Evaluation Criteria in Observational Maintenance Data}
\label{subsec: structural limitations}
When purely associational models are used for intervention planning, prescriptive validity is limited on two levels. The first concerns the observational data itself, specifically through three structural characteristics: confounding, selection bias, and the distribution shift caused by the intervention. The second concerns standard predictive practice based on these data: the metric used to select a model and the scope of the post-hoc methods employed to explain it.
  
\paragraph{Confounding.}
In observational data from manufacturing, the choice of intervention is not random. Maintenance teams replace components when they detect signs of wear, adjust parameters when quality deviates, and escalate interventions when initial corrective actions fail. As a result, the choice of action~$T$ and the outcome~$Y$ (such as \acs{MTTR}) share common causes (machine condition, product format, operator experience, or environmental conditions) that act as confounding factors. A model trained to predict $Y$ from $(T, X)$ may absorb these confounded relationships and, for example, indicate that replacing sensors is associated with long downtime. However, the causal effect could have the opposite sign: sensors are only replaced once the machine is already severely impaired, and the replacement itself shortens the recovery time compared to the counterfactual scenario of inaction. This phenomenon, referred to in the literature as \textit{confounding bias}, has a direct parallel in the maintenance context~\cite{fromPredictionToPrescription,Hernan2024-HERCIW,VANDERSCHUEREN2023108798}.

\paragraph{Selection bias.}
Industrial process data exhibit pronounced selection bias. Production systems are optimised for high yield, and the majority of recorded observations correspond to nominal operating conditions~\cite{saretzkyNetworks,saretzky2026integratingcausalfoundationmodel}. Failures are rare, and when they do occur, they are often resolved before detailed diagnostic data can be collected. Furthermore, the actions recorded in maintenance logs tend to reflect past decision-making guidelines (typically conservative, experience-based heuristics) rather than a systematic exploration of the action space. This leads to a situation in which the observational data cover only a narrow range of the covariate-treatment space, calling into question the assumption of \textit{overlap} required for valid causal inference~\cite{fromPredictionToPrescription}. 

\paragraph{Distribution shift under intervention.}
Predictive models are reliable as long as the data-generating distribution remains stationary. However, the purpose of a prescription is to change the behaviour of the system. As soon as a recommended measure is implemented, the dynamics of the production process shift: failure modes may be eliminated, deterioration trends may change, and the statistical relationships that the model has learned may no longer hold. In the literature on causal inference, this is formalised as the distinction between observational and interventional distributions: $P(Y \mid T) \neq P(Y \mid \dop(T))$, typically when confounding factors are present~\cite{pearl2009,bareinboim2025causal, pearl2018bookOfWhy}. For \acs{PsM} frameworks based on supervised learning models trained on historical data, this means that the prediction level can become unreliable precisely when the prescriptive level influences its output.
 
\paragraph{Model selection criteria.}
A key methodological challenge in prescriptive interventions is that conventional model selection based on the minimum prediction error (known as the $\mu$-risk or MSE) does not necessarily lead to valid causal estimators. Although highly flexible machine learning models provide excellent approximations of the majority population, they extrapolate incorrectly in covariate regions with weak overlap, which systematically compromises the validity of interventional estimates. Paradoxically, simpler models with more restrictive inductive bias often generate more robust causal effect estimates in these sparse regions, even with a nominally lower $R^2$~\cite{DoutreligneHowToSelect}. To overcome this validation deficit, Doutreligne and Varoquaux~\cite{DoutreligneHowToSelect} identify the \textit{R-risk} as an appropriate evaluation metric. By incorporating nuisance parameters for the conditional mean and the propensity score to correct for confounding, this criterion empirically exhibits the highest congruence with the true causal estimation error (the oracle $\tau$-risk).

\paragraph{Scope of post-hoc explanation methods.}
A widely used strategy for interpreting predictive models is to supplement them with post-hoc explanation methods such as SHAP~\cite{shapLundberg} and LIME~\cite{LimeXAI}. However, as Carloni et al.~\cite{Carlonicausality} emphasise in their systematic review of causality in explainable AI, these methods are based on associative models and therefore quantify the contribution of each feature to the model’s prediction, but not its causal influence on the outcome. In the field of manufacturing, Madathil et al.~\cite{xaiManu} note that SHAP-based analyses are limited to feature selection and dimensionality reduction, as the underlying process physics cannot be effectively identified. Fink et al.~\cite{finkPHMPartI,finkPHMPartII} note that feature attribution methods are sensitive to input noise and are often expressed in terms of low-level features that do not align with domain concepts. To counteract this, causal extensions such as counterfactual Shapley values~\cite{budhathoki22a} have been proposed. However, these methods rely on access to a complete structural causal model. In complex manufacturing environments, it is difficult to derive and validate complete causal graphs on a large scale, as the underlying physical interactions are often only partially known~\cite{finkPHMPartI,finkPHMPartII,saretzky2026integratingcausalfoundationmodel}.

\subsection{Causal Approaches in Manufacturing and Open Research Needs}
\label{subsec: emerging causal}
 
The limitations discussed in the previous subsection have inspired an emerging field of research that incorporates causal reasoning into decision support for maintenance and manufacturing. These approaches can be broadly categorised based on two dimensions: the availability of a causal graph and the type of causal query they are designed to address.
 
\paragraph{Graph-based approaches.}
Structural causal models (SCMs), causal Bayesian networks (CBNs), and their dynamic extensions provide a formal framework for encoding cause-effect relationships and drawing interventional conclusions. In the context of \acs{PsM}, Ansari et al.~\cite{fazelCPS} proposed integrating dynamic Bayesian networks (DBNs) with multimodal data for prescriptive maintenance to enable predictive decision-making and optimal maintenance planning through probabilistic causal inference. To expand this line of research beyond purely economic KPIs, Madreiter and Ansari~\cite{OSEE} introduced the OSEE (Overall Sustainable Equipment Effectiveness) framework, which extends the classical \acs{OEE} by incorporating indicators of environmental and social sustainability. To identify the causal relationships between these indicators, the authors developed a DBN-based model that enables scenario-based “what-if” simulations at the manufacturing level, for example, to assess how a revised maintenance strategy aimed at improving availability and employee training affects the dimensions of operational, environmental, and social performance over time. The OSEE-DBN thus represents an application of $\mathcal{L}_2$ causal inference to prescriptive decision-making in sustainable manufacturing. Saretzky et al.~\cite{saretzkyNetworks} demonstrated a network-based approach for identifying root causes to improve \acs{OEE} in high-precision manufacturing. Hennebold et al.~\cite{fraunhoferSyntheticData} proposed the use of causal models to generate synthetic data in injection moulding processes in order to address the problem of small sample sizes and the high costs associated with generating additional data. However, graph-based approaches require either expert-defined or data-driven causal graphs. In manufacturing environments, creating such graphs is challenging due to complexity, noisy data, sensor signals with varying sampling rates, heterogeneous components, maintenance interventions, and the general lack of reference causal structures. Causal discovery algorithms can partially automate graph creation. Nonparametric tests for conditional independence have been applied in specific manufacturing use cases, such as automotive assembly lines~\cite{huegleDiss, hagedorn2022}. However, many standard algorithms struggle with mixed discrete-continuous data and unobserved confounding factors, so their reliability in complex, high-dimensional industrial environments has not yet been established~\cite{hagedorn2022, causalAssembly}. Furthermore, the need to consult experts for graph specification, as is done in the OSEE-DBN, can be time-consuming and lead to subjectivity, especially when there are a large number of variables and interrelationships.

\paragraph{Graph-free causal machine learning.}
An alternative line of research addresses causal effect estimation without requiring an explicit causal graph at inference time. Vanderschueren et al.~\cite{VANDERSCHUEREN2023108798} applied causal machine learning to estimate individualised treatment effects for optimising preventive maintenance frequency, using counterfactual generative models to account for selection bias in observational service contract data. Their approach demonstrated that causal effect estimation enables cost-effective, machine-specific maintenance scheduling that outperforms policies derived from standard predictive models. Schwarz et al.~\cite{doubleML} developed a framework based on double machine learning to estimate the causal effect of rework decisions in semiconductor manufacturing, deriving profit-increasing decision guidelines from observational production data. These approaches share a common principle: they estimate interventional quantities rather than associational ones, thereby enabling the what-if reasoning that prescriptive maintenance requires without the overhead of specifying a full causal graph.

\section{Synthesis and Deduction of Research Gaps}
\label{sec: Synthesis and Deduction of Research}
The analysis in this chapter highlights a structural gap in the current literature on \acs{PsM} and established reference models. Existing frameworks such as PriMa~\cite{Prima} and related extensions~\cite{primaX, primaDesign} define \acs{PsM} as a process ranging from data collection through prediction to recommendations for action. The comprehensive review by Giacotto et al.~\cite{Psm_review_twente25} and the systematic review presented in Section~\ref{sec: Systematic Literature Review on Prescriptive Maintenance} confirm that the prescriptive analysis methods used in this field are predominantly associative. The prescriptive goal of determining \textit{which measures should be taken to achieve a desired outcome} requires the estimation of interventional quantities, which associational models generally cannot identify without additional assumptions~\cite{pearl2009, fromPredictionToPrescription}. The following research gaps can be derived from this analysis:

\begin{figure}[!htbp]
    \centering
    \includegraphics[width=0.75\textwidth]{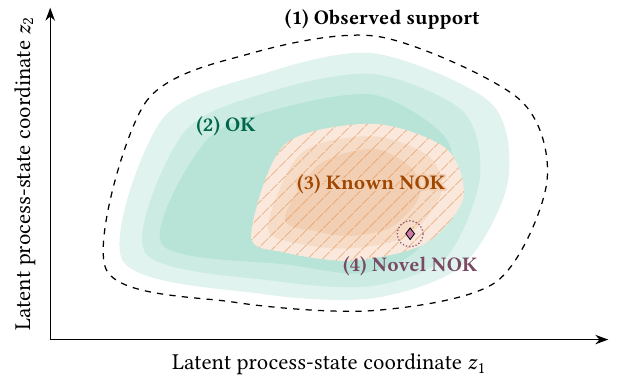}
    \caption{Schematic data-support gap under manufacturing production-line conditions.}
    \label{fig:sota-data-support-gap}
\end{figure}

\begin{enumerate}

    \item \textbf{RG1: Causal inductive bias in \acs{PsM} 
    architectures.}
    Current \acs{PsM} reference models estimate the 
    observational conditional $\mathbb{E}[Y \mid X{=}x]$ rather 
    than the interventional quantity 
    $P(Y \mid \dop(T{=}t), X{=}x)$ required for prescriptive decision support. These architectures do not include systematic mechanisms for encoding causal structures or causal inductive biases in the analytical layers. The literature reviewed did not address how such biases can be formalised and integrated into existing reference models.

    \item \textbf{RG2: End-to-end causal methodology and 
    intervention-oriented evaluation.}
    The reviewed literature provides isolated components 
    (predictive models, rule-based recommenders, post-hoc 
    explanations) but no integrated methodology connecting 
    causal effect estimation to intervention ranking and 
    prescriptive action selection. This includes the absence 
    of causal model selection criteria and semi-synthetic 
    benchmarks for evaluating prescriptive recommendations
    against interventional ground truth in the manufacturing domain. The development 
    of a validated end-to-end framework with 
    intervention-oriented evaluation constitutes an open 
    research need.

    \item \textbf{RG3: Feasibility under manufacturing and production line conditions.}
    Causal inference in real-world manufacturing deviates from standard benchmark assumptions. Observational production logs suffer from severe confounding bias, practical overlap violations, and cyclic feedback induced by closed-loop control systems. Consequently, recorded process adjustments conflate operator actions, controller responses, and disturbance reactions, precluding an unambiguous mapping of the $\dop(T{=}t)$ operator. Furthermore, non-stationarity (such as tooling wear or recipe changes) violates the IID assumption and limits model transferability, while the data itself is complicated by asynchronous multimodal streams, state-dependent measurement errors, non-random data gaps, and the frequent absence of reference causal graphs. Due to the extreme class imbalance, events with critical failure (NOK) typically occur in sparsely covered parts of the observed support (see Figure~\ref{fig:sota-data-support-gap}), where associative extrapolation fails. Since counterfactual ground truth is fundamentally unobservable, validation is constrained to (semi-)synthetic protocols. To address the lack of a unified prescriptive maintenance framework satisfying these stringent constraints, this dissertation takes up this gap under the operating conditions of \acs{FMCG} production lines (Chapter~\ref{chap:introduction}).

    \item \textbf{RG4: Integration of root cause analysis and prescriptive action.}
    Existing \acs{RCA} approaches in manufacturing, whether 
    knowledge-driven, XAI-based, or causal, remain 
    predominantly reactive and diagnostically oriented. Although causal 
    \acs{RCA} methods use the same interventional inferences required for prescription, 
    they have not yet been integrated into \acs{PsM} reference models. The systematic transition from causal diagnosis to causal prescription represents an open research question.
    
\end{enumerate}

 	\part{Causal Prescriptive Maintenance}

\chapter{A Causal Reasoning Layer for Prescriptive Maintenance}
\label{chap:prima-causa}

\begin{jointwork}
This chapter develops the PriMa-Causa artefact that addresses the research gaps derived in Chapter~\ref{chap: literature review}. It incorporates my prior peer-reviewed publication~\cite{saretzky2026integratingcausalfoundationmodel}. Figure~\ref{fig:ch4_structure} outlines the structure of this chapter, which specifies this artefact in five steps. The chapter objective and artefact scope are stated first (Section~\ref{sec:ch4-objective}), followed by the positioning of the causal reasoning layer within the canonical prescriptive-maintenance skeleton (Section~\ref{sec:ch4-bridge-problem}). A taxonomy of causal-prescriptive decision situations is then derived (Section~\ref{sec: The Prescriptive-Question Quadrant}), from which the design theory and design principles are consolidated (Section~\ref{sec:ch4-design-principles}) and operationalised as the Causal Reasoning Layer (Section~\ref{sec:ch4-causal-reasoning-layer}).
\end{jointwork}

\vspace{\baselineskip}

\begin{center}
    \includegraphics[width=\textwidth]{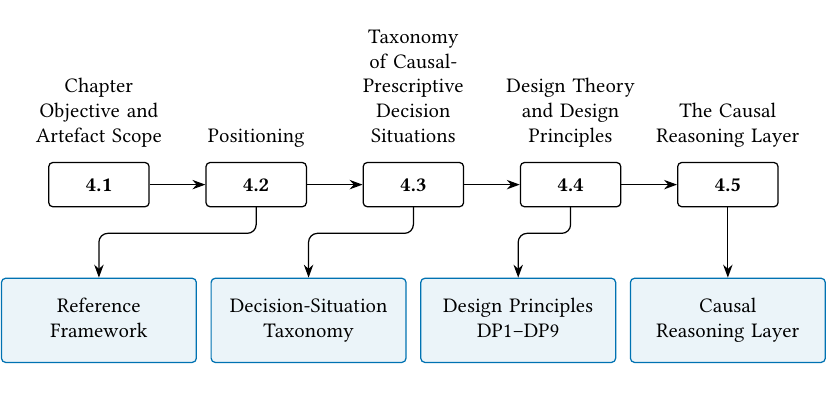}%
    \captionof{figure}{Structure of Chapter~\ref{chap:prima-causa}.}
    \label{fig:ch4_structure}
\end{center}

\newpage

\section{Chapter Objective and Artefact Scope}
\label{sec:ch4-objective}

The objective of this chapter is to define the conceptual core of PriMa-Causa. The chapter does not yet evaluate one specific estimator or one specific implementation. Instead, it specifies the theory artefact that later chapters instantiate and evaluate. This artefact comprises three design-scientific outputs at the reference level: a reference framework (model), a decision-situation taxonomy (constructs) together with a set of design principles (prescriptive statements of the design theory), and an operational pipeline (method) for turning observed key performance indicator deviations into intervention recommendations. The literature reviewed in Chapter~\ref{chap: literature review} sets out considerable architectural ambition for prescriptive maintenance, yet the transition from diagnosis or prediction to justified action remains weakly specified across many of the surveyed frameworks~\cite{Prima,Psm_review_twente25}. PriMa-Causa therefore addresses a more specific question:

\begin{quotebox}
\enquote{How can observed operational deviations in cyber-physical production systems be formalised into interventional queries to systematically derive causally justified prescriptive recommendations?}
\end{quotebox}

This question operationalises the design research question RQ1 introduced in Chapter~\ref{chap:introduction} at artefact level. The artefact developed in this chapter answers RQ1 and constitutes contribution~C1. The chapter establishes the full design theory, comprising design objectives DO1--DO4, meta-requirements MR1--MR13, and design principles DP1--DP9. Of these, contribution~C1 addresses research gaps RG1 and RG4 through design objectives DO1 and DO4. Research gaps RG2 and RG3 are taken up by the instantiation and its evaluation as contributions~C2 and~C3 through design objectives DO2 and DO3. The design principles derived from these objectives are specified in this chapter and, where testable, tested empirically in the evaluation. Table~\ref{tab:ch4-notation} summarises the notation used in this chapter. The complete list of symbols recurring across the dissertation is given in the \hyperref[ch:list-of-symbols]{List of Symbols}.

\begin{table}[!htbp]
    \centering
    \caption{Notation used in Chapter~\ref{chap:prima-causa}.}
    \label{tab:ch4-notation}
    \footnotesize
    \setlength{\tabcolsep}{4pt}
    \renewcommand{\arraystretch}{1.1}
    \begin{tabularx}{\textwidth}{@{}p{0.34\textwidth}Y@{}}
        \toprule
        \textbf{Symbol} & \textbf{Meaning} \\
        \midrule
        \multicolumn{2}{@{}l}{\textit{Variables and queries}} \\
        \addlinespace[2pt]
        $\mathbf{V}=\{V_1,\dots,V_n\}$ & Endogenous process variables. \\
        $T\in\mathbf{V}$,\ $t$ & Intervention target and intervention level. \\
        $Y\in\mathbf{V}$,\ $\mathcal{Y}^{\star}\subseteq\mathrm{range}(Y)$ & Outcome variable (such as \acs{MTTR}) and desired outcome regime. \\
        $\mathcal{Y}_{\mathrm{cand}}\subseteq\mathbf{V}\setminus\{T\}$ & Mode-2 candidate-outcome set. \\
        $X\subseteq\mathbf{V}\setminus\{T,Y\}$,\ $x$ & Pre-intervention decision-time context and its observed realisation. \\
        $T_j\in\mathbf{V}$,\ $t_j$,\ $a_j=(T_j,t_j)$ & Candidate intervention target, candidate level, and candidate action. \\
        $t_{\mathrm{ref}}$ & Mode-2 reference intervention level used for post-estimation comparison. \\
        $t_0$ & Decision time of the current case. \\
        $\mathrm{MTTI}_{\mathrm{PC}}$,\ $\mathrm{MTTR}_{\mathrm{conv}}$ & Time from predictive trigger to PriMa-Causa recommendation, and conventional recovery interval (Subsection~\ref{subsec:decision-time-knowledge-horizon}). \\
        $\dop(T{=}t)$ & Pearl's interventional operator. \\
        $P(Y\mid X{=}x)$ & Observational outcome distribution. \\
        $P(Y\mid \dop(T{=}t),X{=}x)$ & Interventional outcome distribution. \\
        \addlinespace[6pt]
        \multicolumn{2}{@{}l}{\textit{Operationalised prior knowledge}} \\
        \addlinespace[2pt]
        $\mathcal{K}$,\ $\mathcal{K}_{\mathrm{op}}\in$\newline$\{\mathrm{none},\mathrm{ordered},\mathrm{complete}\}$ & Operationalised prior knowledge and its structural completeness. \\
        $\Phi_{\mathcal{K}}(x)$ & Constraint map that turns $\mathcal{K}$ into machine-usable constraints at context $x$. \\
        $\mathcal{G}_{\mathcal{K}}$,\ $\mathcal{A}_{\mathcal{K}}(x)$,\ $r_{\mathcal{K}}$,\ $\mathcal{I}_{\mathcal{K}}$ & Components of $\Phi_{\mathcal{K}}(x)$: admissible causal-graph search space, candidate-action set, variable-role map, and identification constraints. \\
        $\mathcal{R}_{\mathrm{role}}$ & Role labels (intervention target, context, outcome, post-intervention descendant, excluded). \\
        \addlinespace[6pt]
        \multicolumn{2}{@{}l}{\textit{Causal Reasoning Layer pipeline}} \\
        \addlinespace[2pt]
        $c$,\ $M\in\{\mathrm{M1},\mathrm{M2},\mathrm{M3},\mathrm{M4}\}$ & Case frame and decision mode. \\
        $T_c,Y_c\in\mathbf{V}\cup\{\bot\}$ & Case-specific intervention target and outcome ($\bot$ marks an unspecified slot, for instance $T_c = \bot$ if not yet fixed; not missing sensor data). \\
        $d_c$,\ $Q_M$ & Study-design object and mode-specific target query. \\
        $\mathcal{D}_c^{(M)}$,\ $\widehat{\mathcal{D}}_c^{(M)}$ & Planned and estimated digital-intervention tensors. \\
        $\ell,\,j,\,k$;\ $x_\ell$,\ $Y_k$ & Digital-intervention-tensor indices: context scenarios, candidate actions, and outcomes. $x_\ell$ and $Y_k$ denote the indexed context scenario and outcome. \\
        $S_j(x)$,\ $s_c$,\ $\rho_c$ & Mode-3 candidate-action score, qualified scores, and prescription object. \\
        $\widehat{G}=(\mathbf{V},\widehat{\mathbf{E}})$ & Estimated causal graph (Mode~4 output). \\
        \bottomrule
    \end{tabularx}
\end{table}

\clearpage

\section{Positioning}
\label{sec:ch4-bridge-problem}

\begin{figure}[!htbp]
    \centering
    \makebox[\textwidth][c]{%
        \includegraphics[width=\textwidth]{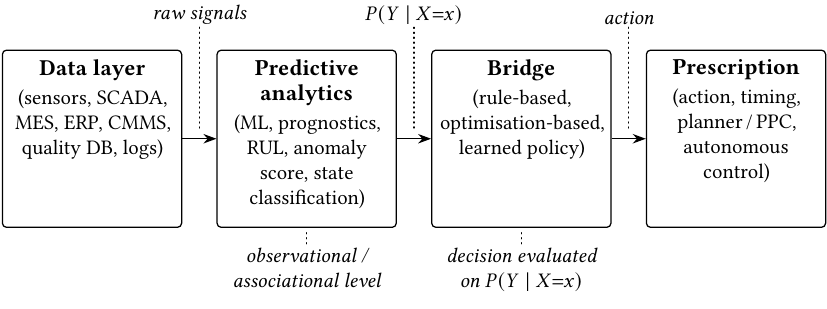}%
    }
    \caption{Operational PsM structure according to the body of knowledge.}
    \label{fig:ch4-traditional-psm}
\end{figure}

Chapter~\ref{chap: literature review} established the current state of prescriptive maintenance and derived the research gaps that the dissertation addresses. Existing \acs{PsM} reference models share a common operational skeleton: raw data are processed into predictive quantities, and those predictions are converted into maintenance actions through an intermediate decision object, here generically labelled the \emph{bridge}, that appears as the \emph{reaction model} in~\cite{Prima}, as the \emph{decision-support method} in the prognostic decision-support literature or as the \emph{prescriptive engine} in the broader prescriptive-analytics literature. Figure~\ref{fig:ch4-traditional-psm} summarises this canonical structure consisting of data, prediction, linkage, and prescription.

\begin{figure}[!htbp]
    \centering
    \includegraphics[width=\textwidth]{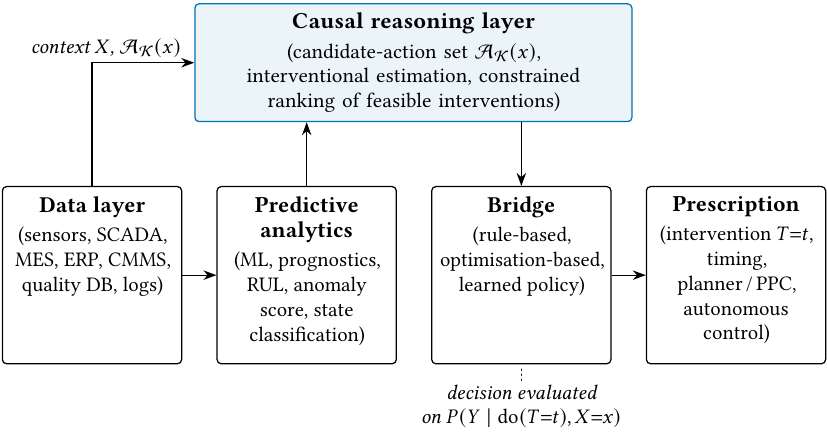}
    \caption{Operational PriMa-Causa structure with the inserted causal reasoning layer.}
    \label{fig:ch4-prima-causa-psm}
\end{figure}

In the literature, the bridge is presented as a set of rules, an optimisation model, or a learned strategy, and evaluates possible actions based on observable predictive distributions of the form $P(Y\mid X{=}x)$. However, selecting among the possible actions requires the interventional distribution $P(Y \mid \dop(T{=}t), X{=}x)$. Prediction and action recommendations answer different questions, and conflating the two leaves maintenance recommendations structurally unjustified as soon as the regime under which data are observed differs from the regime under which interventions are carried out~\cite{fromPredictionToPrescription}. A prescriptive recommendation must compare possible measures in the current context that have not yet been taken, so their outcomes are unobserved at the decision point. Industrial process data additionally cover only a narrow region of the relevant state-action space, which compounds this extrapolation. An associative model offers weaker justification for this extrapolation, whereas a causal model encodes the mechanism invariances under intervention on which such a conclusion rests~\cite{pearl2009,elementsOfCausalInference}. This discrepancy forms the structural core of the architectural gap derived in Chapter~\ref{chap: literature review}. PriMa-Causa therefore introduces a \emph{causal reasoning layer} between the level of predictive analytics and the bridge, as illustrated in Figure~\ref{fig:ch4-prima-causa-psm}. 

The level of predictive analytics remains as an event trigger and context provider, signalling when a case requires a decision. The causal reasoning layer extracts the decision-time context \(X{=}x\) and the candidate-action set \(\mathcal{A}_{\mathcal{K}}(x)\) from the data level, estimates the intervention outcomes under \(\dop(T_j{=}t_j)\) for each candidate action \(a_j=(T_j,t_j)\), and forwards the intervention outcomes to the bridge. The bridge then operates on \(P(Y \mid \dop(T_j{=}t_j), X{=}x)\) rather than on \(P(Y \mid X{=}x)\). The integration is confined to the interface between the prediction and the bridge. The data, predictive-analytics, bridge, and recommendation sections of the canonical \acs{PsM} framework remain structurally unchanged. Within PriMa’s four-layer architecture, the new layer is integrated into the existing data, analytics, and recommendation layers and draws on the semantic-based learning and reasoning layer for operationalised prior knowledge, without adding a fifth layer.

\subsection{Formalising the Operational Problem: The Causal Translation}
\label{sec:ch4-causal-translation}

\begin{figure}[!htbp]
    \centering
    \includegraphics[width=\textwidth]{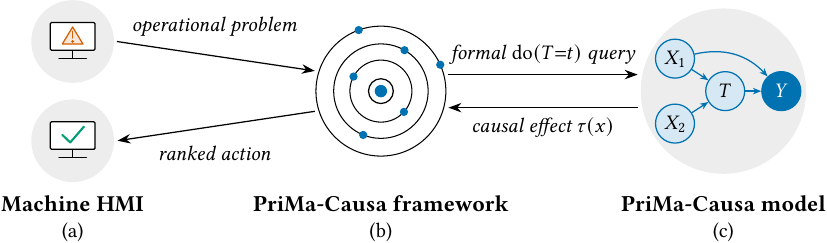}
    \caption[From operational challenge to operational action.]{The translation of a maintenance-related operational challenge into a mathematical formalisation of actions, and the resulting retranslation into a comprehensible operational action.}
    \label{fig:ch4-causal-translation}
\end{figure}

In order to respond to the question posed in Section~\ref{sec:ch4-objective}, it is first necessary to define the term \textit{prescription} in this dissertation. The term \textit{prescription} carries several established meanings across the analytics and maintenance literature. A \textit{prescription} is defined here as a decision recommendation that names a concrete action under stated operational constraints. It differs from a prediction in that it selects an intervention instead of describing an expected outcome. Figure~\ref{fig:ch4-causal-translation} illustrates how the framework proposed in this chapter translates an operational problem at the machine level (a) into a mathematical problem (b), solves it (c), and then converts the result back into an action recommendation that is readable by an operator or engineer (return path from (b) to (a)). This concept is formally outlined in Section~\ref{sec:ch4-causal-reasoning-layer}.

\section{Taxonomy of Causal-Prescriptive Decision Situations}
\label{sec: The Prescriptive-Question Quadrant}

The literature reviewed in Chapter \ref{chap: literature review} suggests that prescriptive-maintenance decisions are heterogeneous with respect to identification difficulty, available knowledge, and operational scope. A taxonomy is therefore introduced to distinguish between the different causal questions being posed at a given decision point and the subsequent analytical tasks that follow from these questions. The taxonomy developed here is problem-oriented. The classification object is not the maintenance system in its entirety, but rather the decision situation confronted by the artefact.

\subsection{Dimensions of Prescriptive Decisions}
\label{subsec: Two dimensions of prescriptive decisions}

Prescriptive-maintenance reference models are guided by action-oriented questions, for instance how a specific event should be made to happen~\cite{Prima}, when equipment will fail and what can be done about it~\cite{Psm_review_twente25}, or, in general prescriptive analytics, what should be done~\cite{Wissuchek}. Each formulation, despite the differences in phrasing, requires the specification of two entities: an intervention target $T$ and an outcome variable $Y$. Both are drawn from the endogenous process variable set $\mathbf{V}$ of the production system, following the notation of structural causal models~\cite{bareinboim2025causal}. The intervention target is defined as the variable, component, or process lever whose value may be modified. The outcome denotes the key performance indicator whose expected value the intervention is meant to improve, such as \acs{MTTR}.

\subsection{Decision Modes}
\label{subsec: Four quadrants}

In accordance with the previously outlined logic, four distinct decision modes are derived. The classification is conceptually informed by the distinction between forward and reverse causal questions~\cite{why_ask_why} and by recent work that organises decision-support questions by intervention and outcome specificity~\cite{taylor2024causaldiscovery,taylor2024bringingcausality}. The purpose of the modes is not merely descriptive. Each mode identifies the analytical task that a prescriptive-maintenance artefact is intended to solve before a recommendation can be justified. The objective is, where feasible, to reduce a Mode~2 or Mode~3 problem to a Mode~1 problem, as Mode~1 is the standard estimation case and common algorithms and methodologies apply there. Mode~4 is instead a structural prerequisite of the other modes. 

In \textbf{Mode~1}, both the intervention $T$ and the outcome of interest $Y$ are defined at the time of decision-making. In manufacturing, this corresponds to cases in which the affected key performance indicator is known and a specific corrective intervention is being evaluated, validated, or optimised. The analytical task is to estimate the conditional interventional outcome distribution under the current line state. The hard intervention $\dop(T{=}t)$ that induces this distribution is treated as a primitive operation of causal analysis~\cite{bareinboim2025causal}. Task-specific quantities such as expected values, threshold probabilities, quantiles or utility values are derived accordingly from the decision objective. Depending on the available data and intervention regime, Mode~1 can be implemented by controlled experiments, quasi-experimental designs, or model-based estimators. In all these implementations, the intervention-outcome pair is specified before the analysis.

\begin{tcolorbox}[
    title={Mode 1: What is the effect of this intervention on this outcome?},
    colback=white,
    colframe=black,
    boxrule=0.4pt,
    arc=2mm,
    top=4pt, bottom=4pt,
    before skip=10pt, after skip=10pt
]
\noindent
\begin{minipage}[c]{0.26\textwidth}
    \centering
    \includegraphics[scale=0.58]{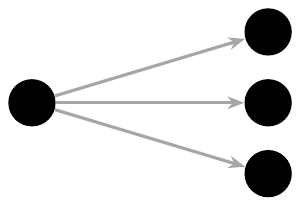}
\end{minipage}
\hfill
\begin{minipage}[c]{0.70\textwidth}
    \begin{equation}
        P\!\bigl(Y \mid \dop(T{=}t),\, X{=}x\bigr).
    \end{equation}
\end{minipage}
\end{tcolorbox}
\vspace{0.5cm}

In \textbf{Mode~2}, the intervention $T$ is defined, while the outcome of interest $Y$ is not specified in advance. The analysis involves the search for a valid candidate-outcome set $\mathcal{Y}_{\mathrm{cand}} \subseteq \mathbf{V}\setminus\{T\}$. This set contains observable process variables that can meaningfully serve as downstream outcomes under the current schema and process ontology. In the context of manufacturing, this corresponds to cases in which a specific process modification is being considered, yet the downstream implications on quality, availability, energy, or sustainability indicators remain uncertain. The analytical task is exploratory. It applies the same Layer-2 query used in Mode~1 to each candidate outcome and identifies which candidate outcomes show decision-relevant distributional changes under the decision-time context $X{=}x$. Only after estimation is each candidate distribution \(P(Y_k\mid \dop(T{=}t),X{=}x)\) compared with its baseline \(P(Y_k\mid \dop(T{=}t_{\mathrm{ref}}),X{=}x)\) at the reference level \(t_{\mathrm{ref}}\), which yields a decision-relevant effect measure for each candidate outcome. 

\vspace{0.5cm}
\begin{tcolorbox}[
    title={Mode 2: Which process outcomes are affected by this intervention?},
    colback=white,
    colframe=black,
    boxrule=0.4pt,
    arc=2mm,
    top=4pt, bottom=4pt,
    before skip=10pt, after skip=10pt
]
\noindent
\begin{minipage}[c]{0.26\textwidth}
    \centering
    \includegraphics[scale=0.58]{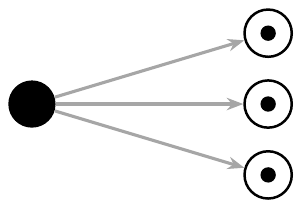}
\end{minipage}
\hfill
\begin{minipage}[c]{0.70\textwidth}
    \begin{equation}
    \left\{
    P\!\bigl(Y_k \mid \dop(T{=}t), X{=}x\bigr)
    \right\}_{Y_k \in \mathcal{Y}_{\mathrm{cand}}}
    \end{equation}
\end{minipage}
\end{tcolorbox}

\vspace{0.5cm}
In \textbf{Mode~3}, the outcome of interest $Y$ is defined by an observed deviation, whereas the intervention target $T$ is not yet known. In a manufacturing context, this is the scenario that arises following a decline in quality, a failure occurrence or a loss of process capability, where the upstream cause has yet to be identified and is the target of a root cause analysis or troubleshooting process. The case is initiated with the reverse causal question of why the deviation in $Y$ occurred. Questions of this form, which are reverse causal in nature, rarely permit a single, clearly defined answer~\cite{why_ask_why}. Mode~3 thus formulates the reverse question as a finite set of forward what-if interventions~\cite{saretzky2026integratingcausalfoundationmodel}, evaluated over the candidate-action set \(\mathcal{A}_{\mathcal{K}}(x)\) obtained from upstream diagnosis such as causal-graph-based backward traversal~\cite{saretzkyNetworks} or rule-based troubleshooting. Each candidate action is evaluated by the probability that setting \(T_j\) to a candidate corrective value \(t_j\) places the fixed outcome \(Y\) in the desired outcome regime \(\mathcal{Y}^{\star}\subseteq\mathrm{range}(Y)\) under context \(X{=}x\). Candidates with higher scores provide stronger fix-condition evidence and are therefore prioritised as actionable root-cause candidates. The score conditions on the pre-intervention context \(x\), not on the observed deviation. It is an interventional quantity on layer \(\mathcal{L}_2\) and not a counterfactual attribution on layer \(\mathcal{L}_3\).

\vspace{0.5cm}
\begin{tcolorbox}[
    title={Mode 3: What are the potential root causes of the observed deviation?},
    colback=white,
    colframe=black,
    boxrule=0.4pt,
    arc=2mm,
    top=4pt, bottom=4pt,
    before skip=10pt, after skip=10pt
]
\noindent
\begin{minipage}[c]{0.26\textwidth}
    \centering
    \includegraphics[scale=0.58]{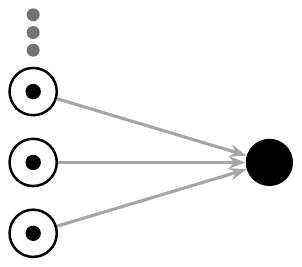}
\end{minipage}
\hfill
\begin{minipage}[c]{0.70\textwidth}
    \begin{equation}
    \begin{aligned}
        S_j(x)
        &=
        P\!\bigl(Y \in \mathcal{Y}^{\star}
        \mid \dop(T_j{=}t_j),\, X{=}x\bigr).
    \end{aligned}
    \end{equation}
\end{minipage}
\end{tcolorbox}
\vspace{0.5cm}

In \textbf{Mode~4}, neither the intervention target $T$ nor the outcome of interest $Y$ is defined at the time of decision-making. This setting arises, for example, at the commissioning of a production line, at the transfer of production to a previously unmodelled line, or at a major process redesign. In such cases, the absence of a causal graph hinders the support of Modes~1--3. The analytical task is therefore structural rather than intervention-specific. Accordingly, causal discovery methods are used to estimate a causal graph, denoted $\widehat{G}$, over the endogenous process variables $\mathbf{V}$, from observational data and available process knowledge~\cite{elementsOfCausalInference,causalAssembly}. Operationalised prior knowledge $\mathcal{K}$, such as temporal ordering,
station layering, physical asymmetries, or known impossibility
constraints, restricts the admissible causal-graph search space
\(\mathcal{G}_{\mathcal{K}}\). The resulting structure should be interpreted
as an estimated causal graph rather than as a guaranteed recovery of the
unique true graph. Depending on the available data and assumptions, it
may correspond to a single estimated graph, a partially oriented
structure, or an equivalence class of admissible graphs. Mode~4 is thus a
structural setup task rather than the dominant operational mode of the
artefact. Its output provides the structural basis on which the
interventional queries in Modes~1--3 can later be specified and assessed.
Causal discovery is treated here as the preparatory complement to the
prescriptive modes rather than as a contribution of this dissertation. The
artefact consumes the available structure through the structural completeness
$\mathcal{K}_{\mathrm{op}}$ (Subsection~\ref{subsec:knowledge-availability})
instead of recovering the graph, so Mode~4 lies outside the scope of the
instantiation and empirical evaluation that follow.

\vspace{0.5cm}
\begin{tcolorbox}[
    title={Mode 4: What is the causal structure of the production process?},
    colback=white,
    colframe=black,
    boxrule=0.4pt,
    arc=2mm,
    top=4pt, bottom=4pt,
    before skip=10pt, after skip=10pt
]
\noindent
\begin{minipage}[c]{0.26\textwidth}
    \centering
    \includegraphics[scale=0.58]{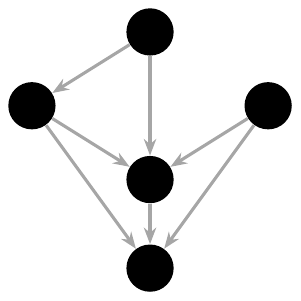}
\end{minipage}
\hfill
\begin{minipage}[c]{0.70\textwidth}
    \begin{equation}
        \widehat{G}=(\mathbf{V},\widehat{\mathbf{E}}),
        \qquad
        \widehat{G}\in\mathcal{G}_{\mathcal{K}}.
    \end{equation}
\end{minipage}
\end{tcolorbox}

\subsection{Availability of Prior Knowledge}
\label{subsec:knowledge-availability}

The taxonomy outlined above differentiates between decision-making scenarios based on the availability of intervention targets and outcomes. Moreover, a secondary dimension can be defined: the availability and operationalisation of prior knowledge. In this dissertation, the term \enquote{prior knowledge} is employed to denote any information regarding the physical, structural, temporal, causal or operational characteristics of a production system that is not derived exclusively from the observational data set utilised for the present estimation problem. This includes physical laws, process topology, station sequence, intervention constraints, assumptions regarding wear and tear, measurement metadata, simulation models, ontological descriptions and expert-maintained troubleshooting rules. Neuer~\cite{neuer2024maschinelles} describes such knowledge as transformations of input data, functional dependencies, differential equations, and more general lawful regularities, and discusses its digital representation through tables, variable metadata, enriched data objects, and ontologies. Fink et al.~\cite{finkPHMPartI,finkPHMPartII} distinguish the point at which such knowledge enters machine learning as inductive, observational, or learning bias.

For PriMa-Causa, the status of \(\mathcal{K}\) is described along four conceptual axes: its \emph{content} (physical laws, causal structure, process topology, degradation behaviour, measurement semantics, admissible interventions), its \emph{form} (expert judgement, table, metadata, ontology, simulation model, equation, rule base, or graph), its \emph{point of entry} in the analytical pipeline (whether $\mathcal{K}$ constrains the causal graph, the candidate-action set, variable-role assignment, feature construction, estimator selection, training, or decision ranking), and its \emph{structural completeness}. Only the structural completeness is carried forward as a formal variable, denoted \(\mathcal{K}_{\mathrm{op}}\). It separates the mere existence of knowledge from its machine-usable role in causal-prescriptive reasoning. A physical relation may be known to an engineer, but it constrains PriMa-Causa only once it is encoded in a form that can restrict causal search, define admissible interventions, specify variable roles, constrain estimation, or support interventional queries.

Operationally, usable knowledge induces a constraint map. For a given set of prior knowledge \(\mathcal{K}\) and a decision-time context \(x\), this map is written as

\begin{equation}
\Phi_{\mathcal{K}}(x)
=
\bigl(
\mathcal{G}_{\mathcal{K}},\,
\mathcal{A}_{\mathcal{K}}(x),\,
r_{\mathcal{K}},\,
\mathcal{I}_{\mathcal{K}}
\bigr).
\end{equation}

The constraint map \(\Phi_{\mathcal{K}}(x)\) is neither an estimator nor an interventional distribution. The mechanism through which causal inductive bias enters the PriMa-Causa pipeline is as follows. The map $\Phi_{\mathcal{K}}(x)$ restricts the admissible causal-graph search space ($\mathcal{G}_{\mathcal{K}}$), defines the candidate actions ($\mathcal{A}_{\mathcal{K}}(x)$), assigns variable roles ($r_{\mathcal{K}}$), and encodes the identification constraints ($\mathcal{I}_{\mathcal{K}}$). The map comprises only these four components. Prior knowledge can also constrain estimation or training, or inform the ranking of actions by cost, safety, feasibility, and latency. These additional effects are not represented as further components of $\Phi_{\mathcal{K}}(x)$. They enter the estimation and recommendation blocks of the Causal Reasoning Layer (Section~\ref{sec:ch4-causal-reasoning-layer}).

The candidate action set is formally instantiated as follows
\begin{equation}
\mathcal{A}_{\mathcal{K}}(x)=\{a_j\}_{j},\qquad a_j=(T_j,t_j).
\end{equation}

\(T_j\in\mathbf{V}\) denotes an intervenable target variable or operational action lever, and \(t_j\in\mathrm{range}(T_j)\) denotes the candidate intervention level, setting, protocol, or regime for that variable. The set \(\mathcal{A}_{\mathcal{K}}(x)\) is finite. The pair \((T_j,t_j)\) is required to be operationally realisable, that is, \(T_j\) must correspond to a controllable degree of freedom on the line and \(t_j\) must lie within its admissible operating range. The pair \(a_j=(T_j,t_j)\) may correspond to a hard (atomic) intervention that fixes \(T_j\) to a constant level, to a stochastic intervention that draws \(t_j\) from a prescribed distribution~\cite{pearl2009,elementsOfCausalInference,ness2025causalai}, or to a conditional intervention that sets \(t_j\) as a function of the observed context, written \(\dop(T_j{=}g_j(x))\)~\cite{pearl2009} (Table~\ref{tab:ch4-intervention-types}). The hard case is recovered when \(g_j\) is constant. Across these three types, the intervention is required to retain a stable and reproducible technical meaning on the line.

\begin{table}[!htbp]
    \centering
    \caption[Intervention types represented by candidate actions.]{Intervention types represented by candidate actions \(a_j=(T_j,t_j)\).}
    \label{tab:ch4-intervention-types}
    \footnotesize
    \setlength{\tabcolsep}{4pt}
    \renewcommand{\arraystretch}{1.1}
    \begin{tabular}{@{}p{0.18\textwidth}p{0.40\textwidth}p{0.34\textwidth}@{}}
        \toprule
        \textbf{Type} & \textbf{Description} & \textbf{Manufacturing example} \\
        \midrule
        Hard (atomic)
        & \(T_j\) is set to a fixed value.
        & Fix process temperature to \(85.0\,^{\circ}\mathrm{C}\). \\
        Stochastic
        & \(T_j\) is drawn from a fixed distribution.
        & Sample process temperature from \(\mathcal{N}(85.0,\,0.5^2)\,^{\circ}\mathrm{C}\). \\
        Conditional
        & \(T_j\) is set as a function of the observed context.
        & Set process temperature to \(85.0+0.05\,v\;^{\circ}\mathrm{C}\), where \(v\) is line speed. \\
        \bottomrule
    \end{tabular}
\end{table}

The role map $r_{\mathcal{K}}$ assigns each variable in $\mathbf{V}$ to one of the labels in $\mathcal{R}_{\mathrm{role}}$ (Table~\ref{tab:ch4-notation}). The assignment is task-specific: the same process variable may be an outcome in one query and a context variable in another. The role map prevents intervention-target variables, pre-intervention context variables, outcomes, and post-intervention descendant variables from being conflated. Variables labelled as excluded are neither admissible intervention targets nor part of the current query. Interventional queries $P(Y\mid \dop(T{=}t), X{=}x)$ are interpreted with $X$ as pre-intervention decision-time context. Post-intervention descendants of $T$ are not admissible adjustment variables unless the estimand is explicitly defined to condition on them. Knowledge availability is not binary. In this dissertation, $\mathcal{K}_{\mathrm{op}}$ is treated as an ordinal measure of structural completeness with three levels: \emph{none}, \emph{ordered} and \emph{complete}. The level is assessed over the variable set relevant to the current case. If the completeness differs across subsystems, the lowest applicable level is used. At the \emph{none} level, no machine-readable graph is provided. Informal expert opinions, historical experience in fault-finding or general process intuition may well exist, but they are not encoded as a graph and therefore do not justify causal identification. At the \emph{ordered} level, the process sequence of the production line is known, so that the temporal order of the process steps is available as a causal chain along the line, in the sense of the causal flow through a technical process chain, as described by Neuer~\cite{neuer2024maschinelles}. This sequence, together with the station topology, prohibited edges or permissible interventions, constrains the graph to a partially specified structure, which reduces the search space and supports the assignment of variable roles. Explicit identification checks are nevertheless still required. At the \emph{complete} level, the complete causal graph is available as validated, machine-readable knowledge, which is sufficient to define the graph constraints, permissible intervention variables and levels, variable roles, and exclusion rules for conditioning on the descendants of $T$. The complete graph represents the idealised upper limit and is rarely achievable in practice. The three levels do not describe the correctness of the knowledge per se. They describe its operational status within the PriMa-Causa framework, and even a complete graph remains subject to empirical validation, expert review and revision when new operating conditions, products, materials or failure modes are observed.

\subsection{Operational Knowledge States and Outcome Horizon}
\label{subsec:decision-time-knowledge-horizon}

In the preceding subsections, the definition of causal decision-making modes and the operational status of prior knowledge were provided. A further distinction is required at implementation time. It concerns whether the current fault is already covered by fault histories, diagnostic codes, and records of actions taken and their outcomes. This distinction is operational rather than formal and is not identical to the taxonomy of known and unknown \(T\) and \(Y\) in Section~\ref{sec: The Prescriptive-Question Quadrant}. The taxonomy determines the causal question that is asked. The present distinction is determined by the question of whether the current case on the production line is sufficiently covered by data and encoded knowledge to support a direct prescription. It is likewise independent of the structural completeness $\mathcal{K}_{\mathrm{op}}$. A complete causal graph does not guarantee historical coverage of a specific fault, and sparse structural knowledge does not preclude rich case-history coverage.

In the context of a \textbf{data-intensive} operational scenario, the warning pattern, the failure progression, the diagnosis and the recommended action are considered to be sufficiently supported by historical data, rules or validated expert knowledge (see the OK and NOK regions in Figure~\ref{fig:sota-data-support-gap}). In this context, a \acs{PdM} signal such as a remaining-useful-life estimate, an anomaly score or a key performance indicator deviation can be linked to an established \acs{PsM} action path. If there is sufficient lead time, the action can be planned before the predicted downtime occurs. This is the best-case scenario assumed by many prescriptive maintenance workflows. It requires not only a prediction of when a failure might occur, but also a justified mapping from the predicted state to a specific action. \acs{PsM} can then rely on an established rule, an optimisation model or a learned policy that maps the predicted state to an action.

As illustrated by the novel NOK region in Figure~\ref{fig:sota-data-support-gap}, an operationally \textbf{data-sparse} case differs from the data-intensive one. While the system may still detect an anomalous trend, the corresponding failure mechanism, the path to the root cause, or the history of interventions and their outcomes is not sufficiently documented or available to enable a directly supervised recommendation for action. This is common in production data, as historical logs consist predominantly of nominal operating conditions, while flagged NOK events, fault-resolution logs and the results of interventions carried out are scarce or influenced by regime changes, as was deduced in RG3. PriMa-Causa addresses the latter situation without being limited to extremely rare events. The relevance of this distinction arises where the available data are sufficient to describe the current context and narrow the permissible interventions, but insufficient to justify a directly supervised recommendation based solely on past NOK histories.

The outcome horizon of PriMa-Causa is the recovery-oriented variable \(Y\), with \acs{MTTR} as the primary horizon. Of the three \acs{OEE} factors, availability, performance, and quality, only availability depends on recovery time, so a shorter \acs{MTTR} raises availability and through it, the \acs{OEE}~\cite{iso22400}. These relations are made precise where they are used, in the budget-constrained intervention-prioritisation experiment of Subsection~\ref{sec:eval-mttr}. For quality-related disturbances, \(Y\) may instead be a return-to-OK indicator, a quality-loss variable, or a joint recovery criterion. The intervention target \(T\) is restricted to technically actionable variables, and PriMa-Causa scores each candidate action \(a_j=(T_j,t_j)\in\mathcal{A}_{\mathcal{K}}(x)\) by the interventional quantity \(P(Y\mid \dop(T_j{=}t_j), X{=}x)\), where \(X{=}x\) is the decision-time context from the data and predictive layers.

\begin{figure}[!htbp]
    \centering
    \includegraphics[width=\textwidth]{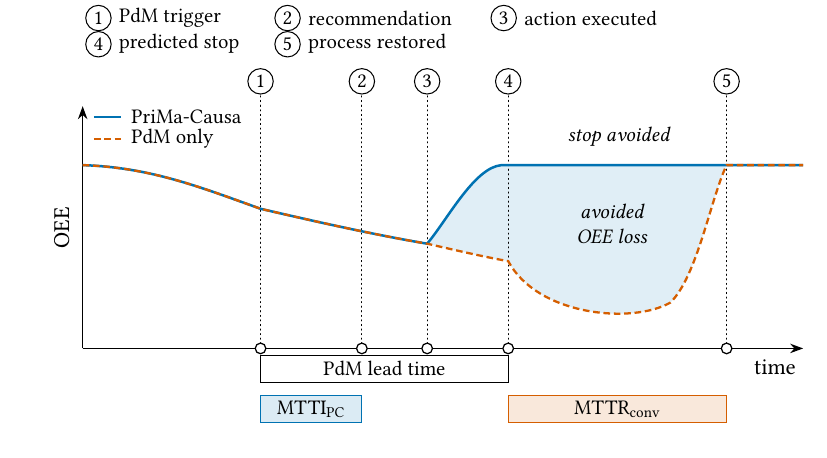}
    \caption[Operational horizon for MTTR reduction.]{Operational horizon for MTTR reduction through PriMa-Causa.}
    \label{fig:operational-knowledge-mttr}
\end{figure}

Figure~\ref{fig:operational-knowledge-mttr} contrasts this horizon for the data-sparse case. When the predictive trigger is not supported by an encoded action, the \acs{PdM}-only path reaches the predicted stop and incurs the manual recovery interval \(\mathrm{MTTR}_{\mathrm{conv}}\), which contains troubleshooting, intervention selection, execution, and restoration. PriMa-Causa instead evaluates candidate actions in silico within the \acs{PdM} lead time, over the interval \(\mathrm{MTTI}_{\mathrm{PC}}\) from trigger to recommendation, so the recovery can be shortened or the stop avoided when the recommendation arrives in time. The shaded region is the resulting avoided \acs{OEE} loss. The \acs{PdM} trigger, the recommendation, and the executed action are marked in sequence on the timeline, and the depicted PriMa-Causa trajectory corresponds to the stop-avoided branch.

The reduction is not improved failure prediction, which remains in the predictive layer, but the removal of the troubleshooting and intervention-selection latency that dominates \(\mathrm{MTTR}_{\mathrm{conv}}\). Using operationalised prior knowledge \(\mathcal{K}\) and the context \(x\), PriMa-Causa computes action-outcome expectations with models prepared before the incident, since a case-specific estimator cannot be fitted at decision time and labelled intervention outcomes are not yet available. The recommendation is prescriptive inside the \acs{PdM} lead time and acts as recovery support once the stop has occurred.

\section{Design Theory and Design Principles}
\label{sec:ch4-design-principles}

The taxonomy of Section~\ref{sec: The Prescriptive-Question Quadrant} specifies which decision situations the artefact must cover. Following the design-science research methodology~\cite{hevner2004,peffers2007}, and within the design cycle introduced in Section~\ref{sec:intro-outline}, this section consolidates the design knowledge of PriMa-Causa in three layers: design objectives derived from the research gaps of Chapter~\ref{chap: literature review}, meta-requirements~\cite{walls1992} that specify the conditions any admissible instantiation must satisfy, and design principles that consolidate the meta-requirements into prescriptive claims at reference-architecture level. The principles follow the specification schema of Gregor et al.~\cite{gregor_anatomy}, which is used to specify each principle by its \emph{aim}, \emph{context}, \emph{mechanism}, and \emph{rationale}.

\subsection{Design Objectives}
\label{subsec:ch4-objectives}

The four research gaps derived in Chapter~\ref{chap: literature review} translate into four design objectives (Table~\ref{tab:ch4-design-objectives}) that bound the PriMa-Causa scope.

\begin{table}[!htbp]
    \centering
    \caption[Design objectives of PriMa-Causa.]{Design objectives of PriMa-Causa, derived from the research gaps of Chapter~\ref{chap: literature review}.}
    \label{tab:ch4-design-objectives}
    \footnotesize
    \setlength{\tabcolsep}{4pt}
    \renewcommand{\arraystretch}{1.15}
    \begin{tabularx}{\textwidth}{@{}p{1.1cm}p{1.7cm}Y@{}}
        \toprule
        \textbf{ID} & \textbf{Source} & \textbf{Design objective} \\
        \midrule
        DO1 & RG1 & Introduce causal inductive bias as a structural component of the prescriptive-maintenance framework rather than as an after-the-fact correction on a predictive output. \\
        DO2 & RG2 & Provide an end-to-end methodology that connects predictive triggers, causal reasoning, and validated interventional recommendations under explicit identification conditions and across multiple evaluation regimes. \\
        DO3 & RG3 & Operate under the data and knowledge regimes characteristic of \acs{FMCG} manufacturing. \\
        DO4 & RG4 & Integrate root-cause evidence with action selection while preserving the analytical separation between causal diagnosis, candidate-action generation, and prescriptive recommendation. \\
        \bottomrule
    \end{tabularx}
\end{table}

\subsection{Meta-Requirements}
\label{subsec:ch4-metareq}

The four design objectives translate into a set of meta-requirements (Table~\ref{tab:ch4-meta-requirements}). The requirements specify a class of admissible PriMa-Causa artefacts rather than one estimator implementation. Several requirements serve more than one design objective. The grouping below names each block's primary dimension, and the Source column of the table gives the binding per-requirement attribution. MR1--MR5 address the architectural and causal-inductive-bias dimension (DO1). MR6--MR8 unpack the manufacturing-feasibility dimension (DO3) into data conditions, identification, and estimator diagnostics. MR9--MR10 address the \acs{RCA}-integration dimension (DO4). MR11--MR13 translate the end-to-end methodology and evaluation dimension (DO2) into decision ranking, operational optimisation, and validation.

\begin{table}[!htbp]
    \centering
    \caption{Meta-requirements for PriMa-Causa, derived from the design objectives of Subsection~\ref{subsec:ch4-objectives}.}
    \label{tab:ch4-meta-requirements}
    \footnotesize
    \setlength{\tabcolsep}{4pt}
    \renewcommand{\arraystretch}{1.15}
    \begin{tabularx}{\textwidth}{@{}p{1.1cm}p{1.7cm}Y@{}}
        \toprule
        \textbf{ID} & \textbf{Source} & \textbf{Meta-requirement} \\
        \midrule
        MR1 & DO1 &
        Insert a causal reasoning layer between the predictive-analytics layer and the bridge (Figure~\ref{fig:ch4-prima-causa-psm}) such that the four PriMa layers remain structurally intact. \\
        MR2 & DO1, DO2 &
        Classify each case into a decision mode $M\in\{\mathrm{M1},\mathrm{M2},\mathrm{M3},\mathrm{M4}\}$ based on the descriptors $T_c, Y_c \in \mathbf{V}\cup\{\bot\}$, carrying the structural completeness $\mathcal{K}_{\mathrm{op}}$ forward in the case frame (Subsection~\ref{subsec: Four quadrants}). \\
        MR3 & DO1, DO2 &
        Specify each prescriptive query through the tuple $(T, t, X, Y)$ and a decision objective, with $T\in\mathbf{V}$ as the intervention target, $t$ as the intervention level, $X$ as pre-intervention context, and $Y$ as the recovery- or quality-oriented outcome. \\
        MR4 & DO1 &
        Justify each recommendation through an interventional quantity of the form $P(Y\mid \dop(T{=}t), X{=}x)$ derived from the mode-specific target $Q_M$, rather than through an associational predictive score. \\
        MR5 & DO1, DO3 &
        Operationalise prior knowledge $\mathcal{K}$ through the constraint map $\Phi_{\mathcal{K}}(x) = (\mathcal{G}_{\mathcal{K}}, \mathcal{A}_{\mathcal{K}}(x), r_{\mathcal{K}}, \mathcal{I}_{\mathcal{K}})$, with the structural completeness $\mathcal{K}_{\mathrm{op}}\in\{\mathrm{none},\mathrm{ordered},\mathrm{complete}\}$ explicit at decision time. \\
        MR6 & DO3 &
        Remain usable under data-sparse manufacturing conditions, including missingness, measurement error, non-stationarity, closed-loop control, and partially observed causal structure (Subsection~\ref{subsec:decision-time-knowledge-horizon}). \\
        MR7 & DO2, DO3 &
        Make identification assumptions explicit and assess identifiability, overlap (common support), confounding, selection bias, and risks from conditioning on descendants of $T$ before ranking actions. \\
        MR8 & DO2, DO3 &
        Support estimator selection, sensitivity analysis, refutation tests, and model diagnostics appropriate to the available data and to the structural completeness $\mathcal{K}_{\mathrm{op}}$. \\
        MR9 & DO4 &
        Transform diagnostic (\acs{RCA}) evidence into the candidate-action set $\mathcal{A}_{\mathcal{K}}(x)$ while keeping root-cause identification analytically distinct from action selection. \\
        MR10 & DO3, DO4 &
        Restrict $\mathcal{A}_{\mathcal{K}}(x)$ to intervenable, safe, and operationally feasible variables through the role map $r_{\mathcal{K}}$, and exclude variables that $r_{\mathcal{K}}$ labels as outcomes or post-intervention descendants from the candidate intervention-target set. \\
        MR11 & DO2, DO3 &
        Rank candidate actions through uncertainty-aware decision logic, including abstention, escalation, or multiple alternatives when the evidence carried by $\widehat{\mathcal{D}}_c^{(M)}$ is insufficient. \\
        MR12 & DO2, DO3 &
        Optimise recommendations under operational constraints including cost, safety, downtime budget, implementation latency, production planning, and expected \acs{MTTR}/\acs{OEE} effect. \\
        MR13 & DO2 &
        Provide auditable causal recommendation evidence through the digital-intervention tensor $\widehat{\mathcal{D}}_c^{(M)}$, and support staged validation across synthetic, semi-synthetic, simulated, and real-world regimes. \\
        \bottomrule
    \end{tabularx}
\end{table}
\clearpage

\subsection{Design Principles}
\label{subsec:ch4-design-principles-sub}

The meta-requirements consolidate into nine design principles. Each principle is specified through the aim, context, mechanism, and rationale components of the schema of Gregor et al.~\cite{gregor_anatomy}, together with the meta-requirements it operationalises. Table~\ref{tab:ch4-design-principles} lists the principles. The empirical consequences of the testable principles are formulated as evaluation hypotheses and tested in Chapter~\ref{chap:empirical-evaluation}.

\renewcommand{\arraystretch}{1.25}
\setlength{\tabcolsep}{5pt}
\footnotesize
\begin{longtable}{@{}>{\raggedright\arraybackslash}p{2.6cm}>{\raggedright\arraybackslash}p{\dimexpr\linewidth-2.6cm-2\tabcolsep\relax}@{}}
    \caption{Design principles of PriMa-Causa, specified in the schema \cite{gregor_anatomy}.}
    \label{tab:ch4-design-principles}\\
    \toprule
    \textbf{Principle} & \textbf{Specification} \\
    \midrule
    \endfirsthead

    \multicolumn{2}{@{}l}{\footnotesize\textit{Table~\ref{tab:ch4-design-principles} (continued).}}\\
    \toprule
    \textbf{Principle} & \textbf{Specification} \\
    \midrule
    \endhead

    \midrule
    \multicolumn{2}{r@{}}{\footnotesize\textit{Continued on next page.}}\\
    \endfoot

    \bottomrule
    \endlastfoot

    \textbf{DP1.}\par\emph{Separate diagnosis from prescription}
    & \textbf{\textit{Aim.}}~Preserve the analytical separation between root-cause identification and action selection.\par\smallskip
    \textbf{\textit{Context.}}~Mode-3 cases that begin with an observed deviation in $Y$ and an unknown intervention target.\par\smallskip
    \textbf{\textit{Mechanism.}}~Diagnostic evidence enters only as a restriction on $\mathcal{A}_{\mathcal{K}}(x)$. The prescription $\rho_c$ is derived from $P(Y\mid \dop(T_j{=}t_j), X{=}x)$, not from the diagnostic score directly.\par\smallskip
    \textbf{\textit{Rationale.}}~Conflating diagnosis and prescription leaves recommendations structurally unjustified (DO4).\par\smallskip
    \textbf{\textit{Sources.}}~MR4, MR9, MR10. \\
    \midrule

    \textbf{DP2.}\par\emph{Restrict intervention targets to intervenable variables}
    & \textbf{\textit{Aim.}}~Avoid prescribing on variables that cannot be changed through maintenance, control, or operator action.\par\smallskip
    \textbf{\textit{Context.}}~All modes involving action selection (Modes 1, 2, and 3).\par\smallskip
    \textbf{\textit{Mechanism.}}~The role map $r_{\mathcal{K}}: \mathbf{V}\to\mathcal{R}_{\mathrm{role}}$ assigns each variable an intervention-target, context, outcome, post-intervention-descendant, or excluded label. Only variables labelled as intervention target enter $\mathcal{A}_{\mathcal{K}}(x)$.\par\smallskip
    \textbf{\textit{Rationale.}}~The operator $\dop(T_j{=}t_j)$ is formally defined irrespective of physical manipulability~\cite{pearl2009}. A prescriptive recommendation, however, is executable only if $T_j$ corresponds to a controllable degree of freedom. A variable that cannot be set through maintenance, control, or operator action yields a recommendation that cannot be carried out, so non-actuatable variables and observed outcomes are excluded from the candidate target set.\par\smallskip
    \textbf{\textit{Sources.}}~MR3, MR5, MR10. \\
    \midrule

    \textbf{DP3.}\par\emph{Resolve reverse questions through forward intervention logic}
    & \textbf{\textit{Aim.}}~Convert reverse causal questions into actionable forward queries.\par\smallskip
    \textbf{\textit{Context.}}~Mode 3, where the intervention target is unspecified at decision time.\par\smallskip
    \textbf{\textit{Mechanism.}}~The reverse question is reframed as a finite set of forward queries $P(Y\in\mathcal{Y}^{\star}\mid \dop(T_j{=}t_j), X{=}x)$ over $\mathcal{A}_{\mathcal{K}}(x)$, with $X$ restricted to pre-intervention decision context.\par\smallskip
    \textbf{\textit{Rationale.}}~Reverse causal questions rarely admit a unique answer~\cite{why_ask_why}. Forward $\dop$-queries yield comparable, rank-able fix-condition evidence, trading explanatory completeness about the origin of the deviation for decision-relevant comparability among the available candidate actions.\par\smallskip
    \textbf{\textit{Sources.}}~MR2, MR3, MR4. \\
    \midrule

    \textbf{DP4.}\par\emph{Operationalise prior knowledge}
    & \textbf{\textit{Aim.}}~Move manufacturing knowledge from informal background to machine-usable inductive bias.\par\smallskip
    \textbf{\textit{Context.}}~All decision modes and all \acs{CRL} blocks, parametrised by the structural completeness $\mathcal{K}_{\mathrm{op}}$.\par\smallskip
    \textbf{\textit{Mechanism.}}~The constraint map $\Phi_{\mathcal{K}}(x)$ binds graph search, candidate actions, variable-role assignment, and identification constraints into a single object available at each \acs{CRL} block.\par\smallskip
    \textbf{\textit{Rationale.}}~Knowledge constrains PriMa-Causa only once it is encoded in machine-usable form. Informal expertise alone does not justify causal identification.\par\smallskip
    \textbf{\textit{Sources.}}~MR5. \\
    \midrule

    \textbf{DP5.}\par\emph{Make uncertainty decision-relevant}
    & \textbf{\textit{Aim.}}~Couple recommendation behaviour to the uncertainty of the underlying evidence.\par\smallskip
    \textbf{\textit{Context.}}~The robust recommendation synthesis block of the \acs{CRL} (Section~\ref{sec:ch4-causal-reasoning-layer}), regardless of mode.\par\smallskip
    \textbf{\textit{Mechanism.}}~The recommendation object $\rho_c$ carries a recommendation status (\emph{recommend}, \emph{return alternatives}, \emph{abstain}, \emph{escalate}) jointly with the qualified scores $s_c$ and their uncertainty diagnostics.\par\smallskip
    \textbf{\textit{Rationale.}}~Uncertainty reported only as an auxiliary statistic does not change the recommended action and therefore does not lower the decision risk. Coupling the recommendation status to the uncertainty of the estimate makes the diagnostic operationally consequential.\par\smallskip
    \textbf{\textit{Sources.}}~MR11. \\
    \midrule

    \textbf{DP6.}\par\emph{Optimise under operational constraints}
    & \textbf{\textit{Aim.}}~Align causal effectiveness with operational feasibility.\par\smallskip
    \textbf{\textit{Context.}}~Ranking within the robust recommendation synthesis block, whenever candidate actions differ in operational cost, feasibility, or risk, including, but not limited to, ties in interventional outcome.\par\smallskip
    \textbf{\textit{Mechanism.}}~The qualified scores $s_c$ attached to each candidate action $a_j\in\mathcal{A}_{\mathcal{K}}(x)$ combine the interventional estimate $P(Y\mid \dop(T_j{=}t_j), X{=}x)$ with cost, downtime budget, safety, production-planning, and implementation-latency constraints.\par\smallskip
    \textbf{\textit{Rationale.}}~A causally effective action is not automatically the preferred action under plant constraints.\par\smallskip
    \textbf{\textit{Sources.}}~MR12. \\
    \midrule

    \textbf{DP7.}\par\emph{Support iterative learning and multi-regime validation}
    & \textbf{\textit{Aim.}}~Treat the artefact lifecycle as iterative across data regimes and time.\par\smallskip
    \textbf{\textit{Context.}}~The semantic-based learning and reasoning layer (Layer~4) of PriMa (Subsection~\ref{subsec: from pdm to psm}), across synthetic, semi-synthetic, simulated, and real-world regimes.\par\smallskip
    \textbf{\textit{Mechanism.}}~Execution feedback, expert overrides, and unsuccessful recommendations are recorded and used to re-characterise $\mathcal{K}$, so that $\Phi_{\mathcal{K}}(x)$ tracks process drift and newly observed structure in subsequent cases. Before deployment, candidate recommendations are validated across these regimes through the auditable evidence object $\widehat{\mathcal{D}}_c^{(M)}$.\par\smallskip
    \textbf{\textit{Rationale.}}~Production systems change through tooling wear, product variants, and recipe updates. Data-sparse cases transition to data-dense ones through use.\par\smallskip
    \textbf{\textit{Sources.}}~MR5, MR6, MR13. \\
    \midrule

    \textbf{DP8.}\par\emph{Guard recommendations by identifiability and support}
    & \textbf{\textit{Aim.}}~Prevent unsupported interventional claims.\par\smallskip
    \textbf{\textit{Context.}}~Estimation steps for any selected cell of $\mathcal{D}_c^{(M)}$.\par\smallskip
    \textbf{\textit{Mechanism.}}~Each selected cell is checked for identifiability against the causal-graph assumptions encoded in $\mathcal{G}_{\mathcal{K}}$ under the constraints $\mathcal{I}_{\mathcal{K}}$, together with back-door admissibility of $X$ and common support (Subsection~\ref{subsec:ch4-crl-assumptions}), before it is estimated. Insufficient support, overlap, or assumption strength returns bounds, sensitivity results, or escalation rather than a point estimate.\par\smallskip
    \textbf{\textit{Rationale.}}~Causal effects estimated outside the empirical support are extrapolations, not interventional evidence.\par\smallskip
    \textbf{\textit{Sources.}}~MR4, MR7, MR8, MR13. \\
    \midrule

    \textbf{DP9.}\par\emph{Preserve PriMa-compatible modularity}

    & \textbf{\textit{Aim.}}~Integrate without restructuring the PriMa reference architecture.\par\smallskip
    \textbf{\textit{Context.}}~Architectural insertion of the \acs{CRL} (Section~\ref{sec:ch4-bridge-problem}).\par\smallskip
    \textbf{\textit{Mechanism.}}~The \acs{CRL} consumes predictive-analytics-layer triggers and decision-time context, draws on inputs from the data layer and the semantic-based learning and reasoning layer (Layer~4), and forwards the interventional outcomes $P(Y\mid \dop(T_j{=}t_j), X{=}x)$ to the bridge. The qualified scores $s_c$ and the recommendation object $\rho_c$ are emitted to the recommendation layer, and the four PriMa layers remain structurally intact.\par\smallskip
    \textbf{\textit{Rationale.}}~Architectural reuse keeps PriMa-Causa adoptable in existing \acs{PsM} deployments.\par\smallskip
    \textbf{\textit{Sources.}}~MR1, MR5. \\
\end{longtable}
\normalsize
\renewcommand{\arraystretch}{1.0}

\section{The Causal Reasoning Layer}
\label{sec:ch4-causal-reasoning-layer}

The Causal Reasoning Layer (\acs{CRL}) operationalises the design objectives and principles of Section~\ref{sec:ch4-design-principles} as a sequence of analytical blocks. To remain compatible with quality and maintenance practice on \acs{FMCG} production lines, the \acs{CRL} operationalises the localisation, diagnosis, and prescription steps of the standard-oriented troubleshooting process summarised in Table~\ref{tab:quality-troubleshooting-standards}. Detection remains with the predictive-analytics layer, containment and execution remain with line operations and the bridge, and standardisation is carried by the semantic-based learning and reasoning layer (DP7).

\begin{table}[!htbp]
    \centering
    \caption{Standard-oriented troubleshooting process for quality problems in production lines.}
    \label{tab:quality-troubleshooting-standards}
    \footnotesize
    \setlength{\tabcolsep}{4pt}
    \renewcommand{\arraystretch}{1.15}
    \begin{tabularx}{\textwidth}{@{}p{1.9cm}Y>{\raggedright\arraybackslash}p{3.8cm}@{}}
        \toprule
        \textbf{Step} & \textbf{Purpose} & \textbf{Standards and guidelines} \\
        \midrule
        Detect &
        Detect deviations in quality-related key performance indicators (such as scrap rate, rework rate, first-pass yield, or \acs{OEE}). &
        ISO~22400-2~\cite{iso22400}; ISO~11462~\cite{iso11462}; ISO~7870~\cite{iso7870-1,iso7870-2} \\

        Contain &
        Isolate affected products, batches, or process segments to limit immediate impact. &
        ISO~9001:2026, cl.~8.7~\cite{iso9001} \\

        Localise &
        Identify the affected station, machine, product variant, shift, material batch, or process window. &
        ISO~11462~\cite{iso11462}; ISO~7870~\cite{iso7870-1,iso7870-2}; ISA~95 / IEC~62264~\cite{iec62264} \\

        Diagnose &
        Identify the failure mode and most plausible root cause of the quality deviation. &
        IEC~62740~\cite{iec62740}; IEC~60812~\cite{iec60812}; IEC~61025~\cite{iec61025} \\

        Decide / Prescribe &
        Select a corrective intervention based on the diagnosed cause and expected effect on the key performance indicator. &
        ISO~9001:2026, cl.~10.2~\cite{iso9001} \\

        Fix and Verify &
        Implement the corrective action and verify recovery of the quality-related key performance indicator. &
        DIN~31051~\cite{din31051}; DIN~EN~13306~\cite{din13306}; ISO~9001:2026, cl.~10.2~\cite{iso9001} \\

        Learn and Standardise &
        Update work instructions, maintenance plans, FMEA, control plans, and knowledge bases. &
        ISO~9001:2026, cl.~7.1.6, 10.2~\cite{iso9001} \\
        \bottomrule
    \end{tabularx}
\end{table}

The \acs{CRL} abstracts this sequence into two recurring reasoning patterns. The first is \emph{systematic search-space reduction}. Before acting, a competent engineer narrows the set of plausible causes by exploiting process structure, physical asymmetries, and prior incident history. The second is \emph{experimentation}. Given a narrowed candidate set, the engineer hypothesises which lever to pull, observes the effect, and updates the diagnosis. The two patterns motivate a sequence of \acs{CRL} blocks rather than one predictive step. Search-space reduction appears first as prescriptive case framing. Experimentation appears later as in-silico evaluation of interventional outcomes.

\begin{figure}[!htbp]
	\centering
	\includegraphics[width=\textwidth]{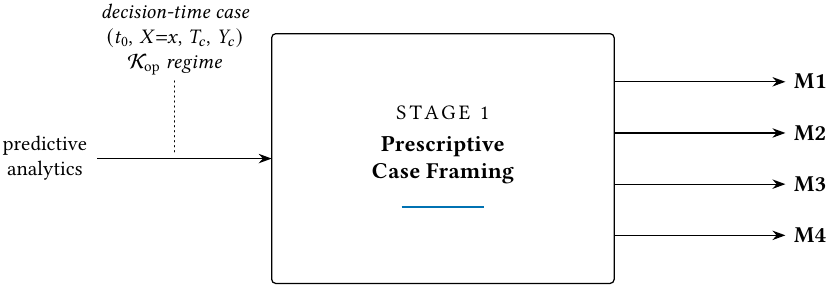}
		\caption{Prescriptive case framing.}
		\label{fig:ch4-prescriptive-case-framing}
\end{figure}

Prescriptive case framing is the input-normalisation and mode-classification block of the \acs{CRL} (Figure~\ref{fig:ch4-prescriptive-case-framing}). Its function is not to estimate an effect, identify an adjustment set, or rank actions. It receives a predictive trigger at decision time \(t_0\), the pre-intervention context \(X{=}x\), the case-specific intervention target \(T_c\), the case-specific outcome \(Y_c\), and the structural completeness \(\mathcal{K}_{\mathrm{op}}\). The descriptors \(T_c\) and \(Y_c\) either name a process variable in \(\mathbf{V}\) or remain unspecified, marked by $\bot$, which records that the corresponding slot has not yet been fixed for the decision situation rather than denoting missing sensor data. Mode selection can be performed by an engineer, by a rule-based workflow, or by an implementation-specific user interface. The dissertation therefore treats the result as the case frame $c=(t_0, X{=}x, T_c, Y_c, \mathcal{K}_{\mathrm{op}}, M)$ rather than as the output of a required automated classifier. The mode $M\in\{\mathrm{M1},\mathrm{M2},\mathrm{M3},\mathrm{M4}\}$ is selected by the decision-mode taxonomy of Subsection~\ref{subsec: Four quadrants}:
\begin{equation}
M =
\begin{cases}
\mathrm{M1}, & T_c\neq\bot,\quad Y_c\neq\bot,\\
\mathrm{M2}, & T_c\neq\bot,\quad Y_c=\bot,\\
\mathrm{M3}, & T_c=\bot,\quad Y_c\neq\bot,\\
\mathrm{M4}, & T_c=\bot,\quad Y_c=\bot.
\end{cases}
\end{equation}
This classification fixes the causal question class of the case. It does not yet specify the estimator, the adjustment set, or the final prescription. The structural completeness \(\mathcal{K}_{\mathrm{op}}\) is carried forward because it determines which components of the constraint map \(\Phi_{\mathcal{K}}(x)\) can be used in the subsequent study-design, estimation, and ranking steps.

\begin{figure}[!htbp]
    \centering
    \includegraphics[width=\textwidth]{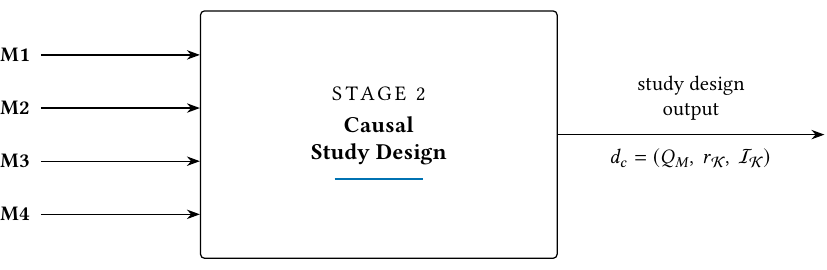}
    \caption{Causal study design.}
    \label{fig:ch4-causal-study-design}
\end{figure}

Causal study design is the second block of the \acs{CRL}. It translates the case frame \(c\) and the constraint map \(\Phi_{\mathcal{K}}(x)\) into a case-specific study-design object \(d_c\). The object \(d_c\) specifies the mode-specific target \(Q_M\), the variable-role map \(r_{\mathcal{K}}\), and the identification-relevant knowledge constraints \(\mathcal{I}_{\mathcal{K}}\):
\begin{equation}
d_c
=
\left(
Q_M,\ r_{\mathcal{K}},\ \mathcal{I}_{\mathcal{K}}
\right).
\end{equation}
Here, \(Q_M\) denotes the target query or structural target implied by mode \(M\). The role map \(r_{\mathcal{K}}\) was defined in Subsection~\ref{subsec:knowledge-availability}. It assigns variables in \(\mathbf{V}\) to roles such as intervention target, context, outcome, post-intervention descendant, or excluded variable. The set \(\mathcal{I}_{\mathcal{K}}\) denotes the knowledge-derived identification constraints carried by \(\Phi_{\mathcal{K}}(x)\), for example temporal ordering, exclusion restrictions, constraints on the admissible adjustment set, or mechanism assumptions. The study-design block still does not estimate \(Q_M\). It defines what has to be estimated and which role and identification constraints must be respected by the subsequent estimation block. The translation from case frame to study design is mode-conditional, see Figure~\ref{fig:ch4-causal-study-design}.

Each of the four modes defined in Subsection~\ref{subsec: Four quadrants} expresses a different causal question and therefore requires a different study design.

\begin{itemize}
    \item Under \textbf{Mode~1}, \(T_c=T\) and \(Y_c=Y\) are specified. The target is the conditional interventional outcome distribution
    \(P(Y \mid \dop(T{=}t), X{=}x)\). Candidate intervention levels must be drawn from the candidate-action set \(\mathcal{A}_{\mathcal{K}}(x)\) defined in Subsection~\ref{subsec:knowledge-availability}. Decision-specific summaries are derived from this distribution.
    \item Under \textbf{Mode~2}, the target is the outcome-indexed
    family of conditional interventional outcome distributions
    \(\{P(Y_k \mid \dop(T{=}t), X{=}x) : Y_k \in \mathcal{Y}_{\mathrm{cand}}\}\).
    Decision-relevant changes are obtained by comparing summaries of these distributions
    with the corresponding reference distributions under \(\dop(T{=}t_{\mathrm{ref}})\).
    \item Under \textbf{Mode~3}, the target is a score over the candidate-action set
    \(\mathcal{A}_{\mathcal{K}}(x)\) that ranks diagnosis-restricted candidate actions by the probability that
    intervening on $T_j$ at level $t_j$ places the fixed outcome $Y$ in the desired regime
    \(\mathcal{Y}^{\star}\), as defined in the Mode~3 box of
    Subsection~\ref{subsec: Four quadrants}.
    \item Under \textbf{Mode~4}, the target is a causal graph \(\widehat{G}\)
    over $\mathbf{V}$ recovered from observational data
    under the process-knowledge constraints \(\mathcal{G}_{\mathcal{K}}\). Without
    functional assumptions, observational data alone identify the graph at most
    up to its Markov equivalence class, and \(\mathcal{G}_{\mathcal{K}}\) can orient
    further edges~\cite{elementsOfCausalInference}.
\end{itemize}

Experimentation is rarely admissible on a running \acs{FMCG} production line. Randomised evaluation of maintenance actions is generally costly and often infeasible in industrial maintenance settings~\cite{VANDERSCHUEREN2023108798}. On \acs{FMCG} lines it is further constrained by safety requirements, stoppage costs, batch-level quality outcomes, and the regulatory framing of changes to qualified processes. Digital Intervention Experiments therefore form the third block of the \acs{CRL}. The block receives \(d_c\), the constraint map \(\Phi_{\mathcal{K}}(x)\), and the decision-time context \(x\). For each candidate action \(a_j=(T_j,t_j) \in \mathcal{A}_{\mathcal{K}}(x)\), it defines the interventional query \(P(Y \mid \dop(T_j{=}t_j), X{=}x)\) to be evaluated under explicit identification assumptions. The resulting digital-intervention tensor \(\mathcal{D}_c^{(M)}\) arranges these queries across contexts, candidate actions, and decision-relevant outcomes, following the logic of contextual and complex experimentation~\cite{cinelli2025challengesstatisticsdozenchallenges}. This substitution preserves the engineer's reasoning pattern while keeping the plant inside its certified operating envelope.

\begin{figure}[!htbp]
    \centering
    \includegraphics[width=\textwidth]{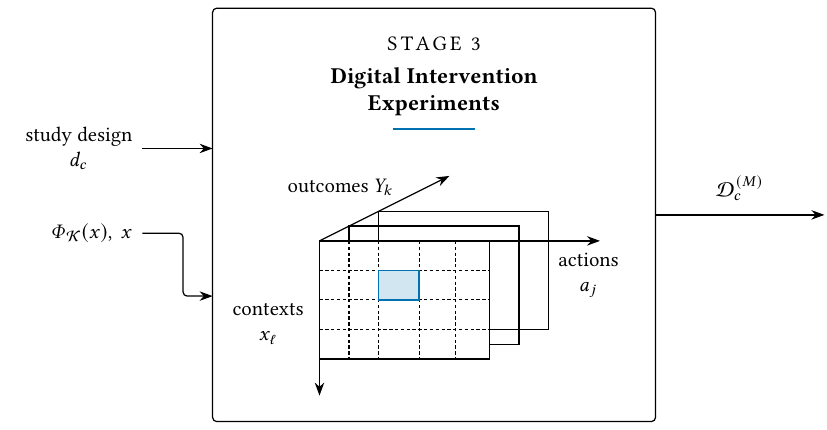}
    \caption{Digital Intervention Experiments.}
    \label{fig:ch4-digital-intervention-experiments}
\end{figure}

The tensor uses three fixed index roles. Context scenarios are indexed by \(\ell\), candidate actions by \(j\), and decision-relevant outcomes by \(k\). Thus, \(x_\ell\) denotes the context scenario with index \(\ell\), \(a_j=(T_j,t_j)\) the candidate action with index \(j\), and \(Y_k\) the outcome with index \(k\) for the selected mode. For a single decision-time case, the context axis may contain only the observed context \(x\). Additional \(x_\ell\) values represent engineer-defined or implementation-defined scenario contexts. The highlighted cell in Figure~\ref{fig:ch4-digital-intervention-experiments} represents one planned digital intervention query:
\begin{equation}
\mathcal{D}_c^{(M)}[\ell,j,k]
=
P(Y_k \mid \dop(T_j{=}t_j), X{=}x_\ell).
\end{equation}
The tensor does not imply that all possible combinations are evaluated automatically. It provides a structured workspace for the engineer or implementation to organise the case-specific queries derived from \(d_c\). A cell is considered only when the corresponding action is admissible, the variable roles in \(r_{\mathcal{K}}\) are coherent, and the identification constraints in \(\mathcal{I}_{\mathcal{K}}\) are not violated. During estimation, selected cells may be assigned estimates, uncertainty information, empirical-support diagnostics, or sensitivity results. The tensor therefore records the planned in-silico intervention study before it becomes an evidence object for downstream recommendation. For Mode~3, a selected cell holds the interventional distribution of the fixed outcome, from which the probability of reaching the desired regime is derived in robust recommendation synthesis. Mode~4 does not populate the tensor, as its target is the estimated causal graph rather than an interventional distribution.

\begin{figure}[!htbp]
    \centering
    \includegraphics[width=\textwidth]{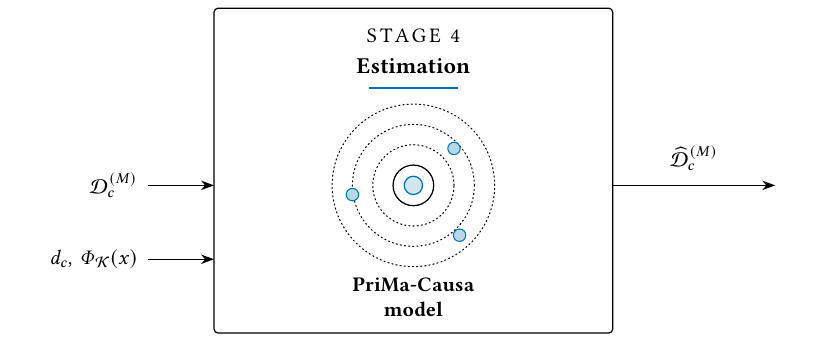}
    \caption{Estimation.}
    \label{fig:ch4-estimation}
\end{figure}

Estimation is the fourth block of the \acs{CRL}. The block receives the tensor \(\mathcal{D}_c^{(M)}\), the study-design object \(d_c\), the constraint map \(\Phi_{\mathcal{K}}(x)\), and the available production history. Its purpose is to estimate the selected interventional queries represented by \(\mathcal{D}_c^{(M)}\). The estimation model is therefore constrained by the role map \(r_{\mathcal{K}}\), the identification constraints \(\mathcal{I}_{\mathcal{K}}\), and the empirical support available for the selected context-action-outcome cell. The choice of the estimator class is instantiated in Chapter~\ref{chap:prima-causa-model}. This block specifies the constraints any admissible estimator must respect. Figure~\ref{fig:ch4-estimation} visualises this block as the PriMa-Causa model operating on the digital-intervention tensor.

\begin{figure}[!htbp]
    \centering
    \includegraphics[width=\textwidth]{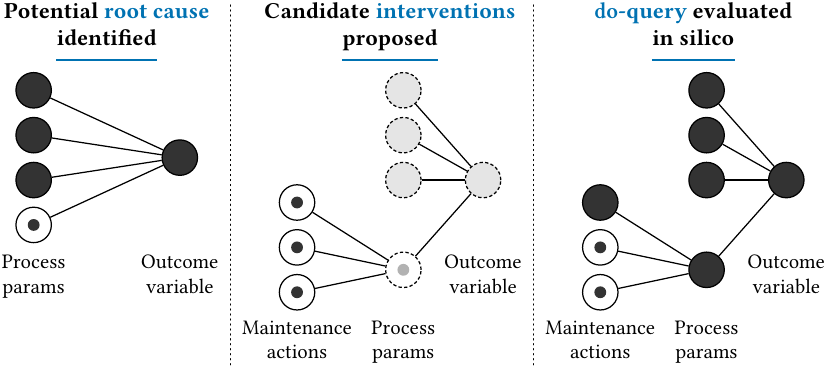}
    \caption{Intervention hypotheses and in-silico evaluation, adapted from~\cite{taylor2024causaldiscovery,taylor2024bringingcausality}.}
    \label{fig:ch4-intervention-experiment-progression}
\end{figure}

For each selected tensor cell, the estimation block returns an estimated interventional distribution:
\begin{equation}
\widehat{\mathcal{D}}_c^{(M)}[\ell,j,k] = \widehat{P}(Y_k \mid \dop(T_j{=}t_j), X{=}x_\ell).
\end{equation}
The hat marks an estimate, not a new target quantity. Depending on the available data and assumptions, a selected cell can remain unestimated, be reported as a point estimate, or be accompanied by uncertainty and empirical-support diagnostics. These diagnostics are passed to robust recommendation synthesis rather than treated as final recommendation evidence. The tensor representation also supports traceability. A later prescription recommendation can be traced back to the evaluated \(\dop\)-queries (Figure~\ref{fig:ch4-intervention-experiment-progression}), their estimated outcomes, and the uncertainty and support diagnostics carried as auxiliary metadata alongside the estimated tensor. This addresses the validated end-to-end methodology required by RG2 and operationalises MR4 and MR13. It also supports DP3, DP5, and DP8 because the recommendation evidence remains linked to forward intervention queries, diagnostics, and identification conditions.

\begin{figure}[!htbp]
    \centering
    \includegraphics[width=\textwidth]{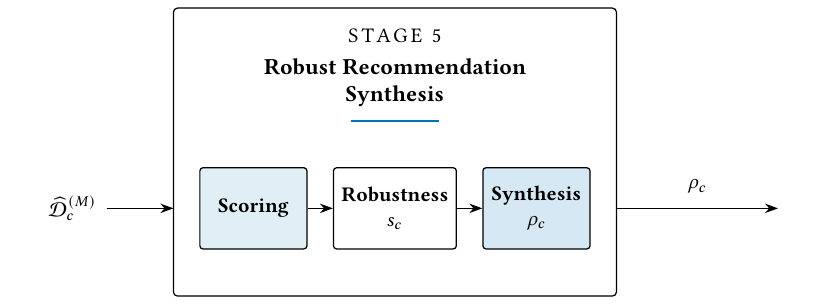}
    \caption{Robust recommendation synthesis.}
    \label{fig:ch4-robust-recommendation-synthesis}
\end{figure}

Robust recommendation synthesis is the fifth block of the \acs{CRL}. At framework level, the block denotes the post-estimation reasoning step that prepares estimated interventional quantities for the PriMa recommendation layer. Its input is the evidence object produced by the fourth block, represented by the estimated tensor \(\widehat{\mathcal{D}}_c^{(M)}\) and its associated uncertainty and empirical-support diagnostics. The block does not imply a fully automated prescription system. The operations can be implemented by software, encoded workflow rules, or operator judgement. The block produces two outputs: the case-specific qualified scores \(s_c\), and the prescription recommendation object \(\rho_c\) prepared for the PriMa recommendation layer. First, mode-specific scoring turns estimated interventional distributions into comparable recommendation quantities. Depending on the selected mode, these quantities can be probabilities of reaching the desired regime $\mathcal{Y}^{\star}$, expected outcome changes, side-effect scores, or operationally cost-adjusted utilities. Second, robustness assessment qualifies these scores by uncertainty, empirical support, sensitivity to identification assumptions, and stability under admissible estimator or specification choices. The qualified scores are reported as $s_c$ and assembled into $\rho_c$, see Figure~\ref{fig:ch4-robust-recommendation-synthesis}.

\begin{figure}[p]
    \centering
    \makebox[\linewidth][c]{%
        \includegraphics[angle=90, width=\linewidth]{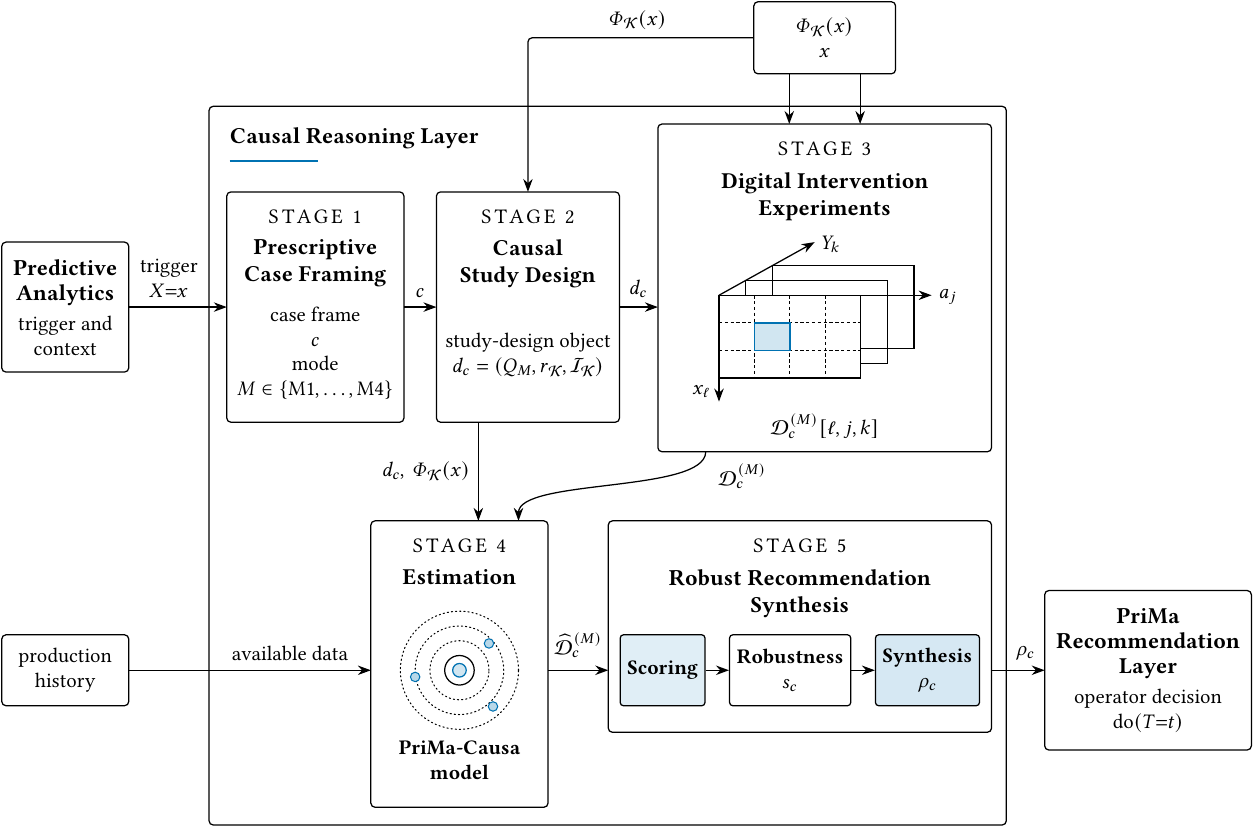}
    }
    \caption{Connected pipeline of the Causal Reasoning Layer.}
    \label{fig:ch4-crl-full-pipeline}
\end{figure}

Operationally, the block compares selected entries of \(\widehat{\mathcal{D}}_c^{(M)}\), qualifies them with robustness information, and prepares \(\rho_c\) for the PriMa recommendation layer. The object \(\rho_c\) denotes a prescription recommendation object, not a final autonomous decision. It can contain a recommended candidate action or action set, qualified scores, uncertainty and support diagnostics, and a recommendation status such as recommend, return alternatives, abstain, or escalate. The final intervention decision remains with the responsible operator or engineer. The complete \acs{CRL} pipeline is summarised in Figure~\ref{fig:ch4-crl-full-pipeline}. It connects the five blocks defined above into one framework sequence.

This sequence preserves the separation between prediction, causal reasoning, and prescription. Predictive analytics triggers the case and provides the decision-time context. The \acs{CRL} reformulates the case as an interventional question, evaluates admissible digital interventions under explicit assumptions, and prepares qualified recommendation evidence. The PriMa recommendation layer, which contains the bridge, then remains responsible for the operational decision process and for human approval of any intervention.

\subsection{Identifiability Assumptions in Manufacturing}
\label{subsec:ch4-crl-assumptions}

The back-door and front-door criteria and instrumental variables are introduced in Chapter~\ref{chap: Background} (Section~\ref{sec:bg-causal}). This subsection states the manufacturing-specific form of the resulting identifiability conditions.

Identification of $P(Y\mid \dop(T_j{=}t_j), X{=}x)$ from observational data rests on the structural conditions of the assumed causal graph, drawn from the admissible search space $\mathcal{G}_{\mathcal{K}}$, namely the Markov condition~\cite{pearl2009,bareinboim2025causal} and causal sufficiency~\cite{elementsOfCausalInference}. Given these, four operational conditions are verified for each query before it is estimated and acted upon: a well-defined intervention $\dop(T_j{=}t_j)$, the back-door admissibility of the context $X$, common support of the considered intervention levels, and the absence of selection bias in the records.

\paragraph{Well-defined intervention.}
Each candidate action $a_j=(T_j,t_j)$ shall designate a single, technically reproducible operating condition rather than a bundle of overlapping interventions, so that the submodel induced by $\dop(T_j{=}t_j)$ is unambiguous~\cite{pearl2009,bareinboim2025causal}. The interventional query $P(Y\mid \dop(T_j{=}t_j),X{=}x)$ then refers to one defined operating condition, obtained by fixing a single controllable degree of freedom on the line.

\paragraph{Back-door admissibility.}
The context $X$ shall block all back-door paths from $T$ to $Y$ and contain no descendant of $T$, the back-door criterion~\cite{pearl2009}. Causal sufficiency (no unobserved common cause of two or more variables in $\mathbf{V}$) is assumed at the graph level and can fail in practice. In \acs{FMCG} lines, unmeasured confounders such as operator behaviour, ambient conditions, or recipe history, as well as closed-loop controller feedback, can break the back-door requirement. They shall be addressed through explicit encoding in $\mathcal{G}_{\mathcal{K}}$, sensitivity analysis, or escalation.

\paragraph{Overlap (common support).}
Each intervention level under consideration shall be supported by the observational regime, that is, the relevant context-action region $(X{=}x,\,T_j{=}t_j)$ shall be represented in the data on which the estimate relies~\cite{Hernan2024-HERCIW}. Sparse NOK regimes regularly leave some intervention levels without support. Sensor drift, quantisation, and missing channels can further restrict the supported domain and shall be reflected in the empirical-support diagnostics attached to $\widehat{\mathcal{D}}_c^{(M)}$.

\paragraph{Selection bias.}
The recorded population shall not be restricted by a common effect of the candidate action and the outcome, nor by any other selection that opens a path between them. In \acs{FMCG} lines, inspection gating and scrap removal before logging restrict the records to units that reached a downstream station, which can open a collider path between action and outcome (Section~\ref{sec:bg-causal}). This risk shall be assessed for each candidate action before ranking.

When one or more of these conditions cannot be defended for a given case, MR7 mandates explicit reporting, and DP8 enforces an identification-aware approximation, that is, bounds, sensitivity intervals under stated departures, or escalation rather than a definitive prescription.

\subsection{Scope and Limitations}
\label{subsec:ch4-crl-abgrenzung}

\begin{figure}[!htbp]
    \centering
    \includegraphics[width=0.9\textwidth]{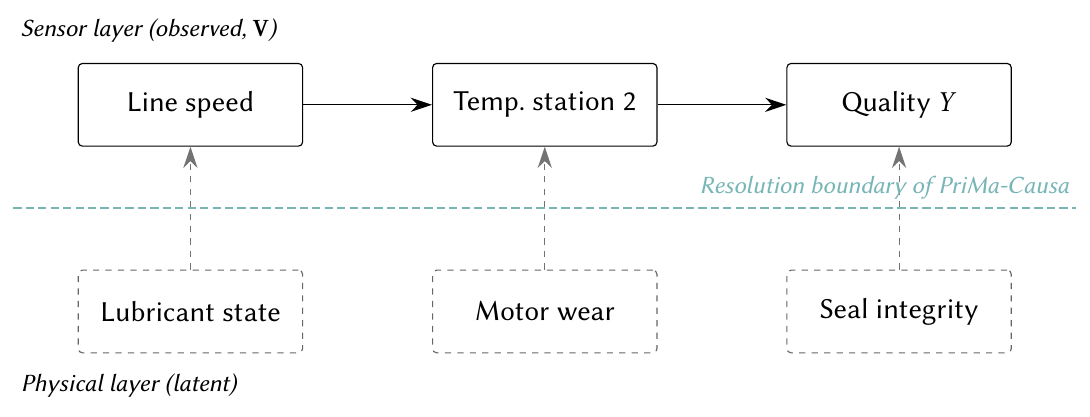}
    \caption{Scope and resolution boundary of PriMa-Causa.}
    \label{fig:ch4-scope-boundary}
\end{figure}

PriMa-Causa is scoped to causal reasoning on the observable sensor layer of the production system. Its estimates and intervention recommendations are confined to relations expressible in the observed variable set $\mathbf{V}$. The framework does not claim to identify physical root causes that lie below the sensor abstraction (Figure~\ref{fig:ch4-scope-boundary}). In the figure, solid edges denote the cause-effect relations on the sensor layer over which the framework reasons, and dashed edges denote latent physical mechanisms that manifest in the sensor layer but are not directly recovered.

Consider a quality deviation in $Y$ at a packaging station that is statistically and interventionally linked to elevated temperature at station~2. A PriMa-Causa instantiation would identify the temperature setpoint as an actionable lever for reducing the deviation. The underlying physical mechanism, however, may lie below the sensor layer, for instance a worn drive motor whose increased friction raises the local temperature. The resulting sensor-layer recommendation reduces \acs{MTTR} for the current incident, but it does not eliminate the physical root cause, and the issue may resurface.

This positions the framework between purely associational machine learning and physics-based digital twins. Where a digital twin reconstructs the latent physical state from first principles, PriMa-Causa approximates intervenable cause-effect relations from sensor data under the available process knowledge $\mathcal{K}$. The resulting recommendations are engineering-pragmatic, bounded by sensor expressiveness rather than by physical first principles, and confined to search-space narrowing rather than autonomous execution. As a decision-support layer, the framework supports \acs{MTTR} reduction and complements, rather than replaces, physics-based modelling and human expert validation.

		    \chapter{Production-Line Simulator}
\label{chap:synthetic-data}

\begin{jointwork}
This chapter incorporates my prior peer-reviewed publications~\cite{saretzkyNetworks,saretzky2026integratingcausalfoundationmodel}. Figure~\ref{fig:ch5_structure} outlines the structure of this chapter, which develops the data foundation for the causal-prescriptive methods of this dissertation. Four sections build from the chapter objective through the production-line topological prior and the synthetic data pipeline to the semi-synthetic anchoring procedure. The chapter yields two artefacts: pre-training datasets drawn from the synthetic pipeline, and semi-synthetic datasets anchored on real covariates.
\end{jointwork}

\vspace{\baselineskip}

\begin{center}
    \includegraphics[width=\textwidth]{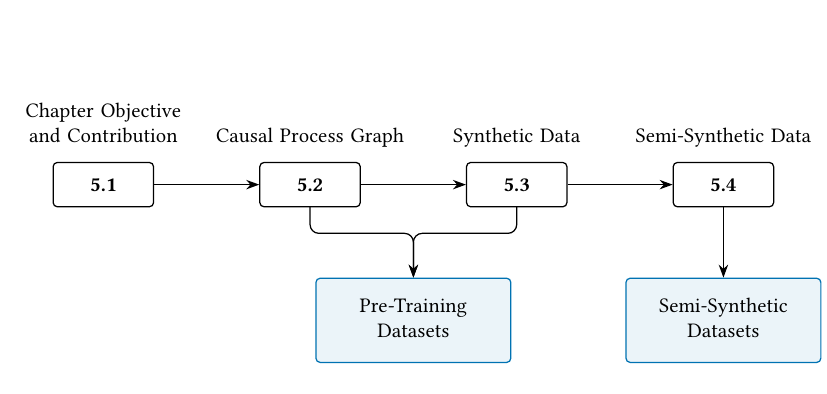}%
    \captionof{figure}{Structure of Chapter~\ref{chap:synthetic-data}.}
    \label{fig:ch5_structure}
\end{center}

\newpage

The Causal Reasoning Layer of Chapter~\ref{chap:prima-causa} evaluates interventional queries, whereas the recorded production data are observational. They cover a narrow operating range and contain no interventional ground truth. Validation, rather than estimation, is therefore the bottleneck of causal-prescriptive maintenance. An interventional estimate can be produced from observational records under assumptions, but the field data contain nothing against which the estimate could be checked. This chapter develops the data foundation that addresses this bottleneck: synthetic datasets drawn from structural causal models with production-line topology, and semi-synthetic datasets anchored on real covariates. In both, the interventional answers are available by construction. Section~\ref{sec:synth-objective} derives this objective from the data conditions of manufacturing and states the contributions of the chapter. Sections~\ref{sec:synth-cpg} to~\ref{sec:synth-semi-fmcg} develop the Causal Process Graph sampler as a topological prior, the synthetic data pipeline that produces observational and interventional data from it, and the semi-synthetic anchoring procedure that retains real covariates while simulating treatment and outcome. The resulting datasets feed the pre-training and evaluation of the causal foundation model in Chapter~\ref{chap:prima-causa-model}.

\section{Chapter Objective and Contribution}
\label{sec:synth-objective}

The \acs{CRL} of Chapter~\ref{chap:prima-causa} is built to evaluate the interventional estimand $P(Y\mid \dop(T{=}t),X{=}x)$. Under the identification assumptions of Subsection~\ref{subsec:ch4-crl-assumptions}, this estimand can be estimated from observational production records. It is, however, never directly recorded. Recorded production data contain only the outcome under the action actually taken. The outcome under any alternative action is never observed, so no measurement exists against which an interventional estimate could be validated (Research Gaps RG2 and RG3, Section~\ref{sec: Synthesis and Deduction of Research}). This quantity can, however, be constructed under a known causal mechanism. In the absence of experimental access, such a controlled construction is the only basis on which methods with an unobservable target can be validated. The same conclusion is reached in causal machine learning more broadly, where synthetic experiments are argued to be necessary rather than merely convenient for assessing methods whose target quantities are unobservable~\cite{poinsot2025synthetic}. For the maintenance domain, a systematic review identifies the scarcity of run-to-failure data as the dominant data challenge and reports it as the limiting factor for data-driven maintenance models across eighty-six studies~\cite{nieminen2026synthetic}.

\begin{figure}[!htbp]
    \centering
    \includegraphics[width=\textwidth]{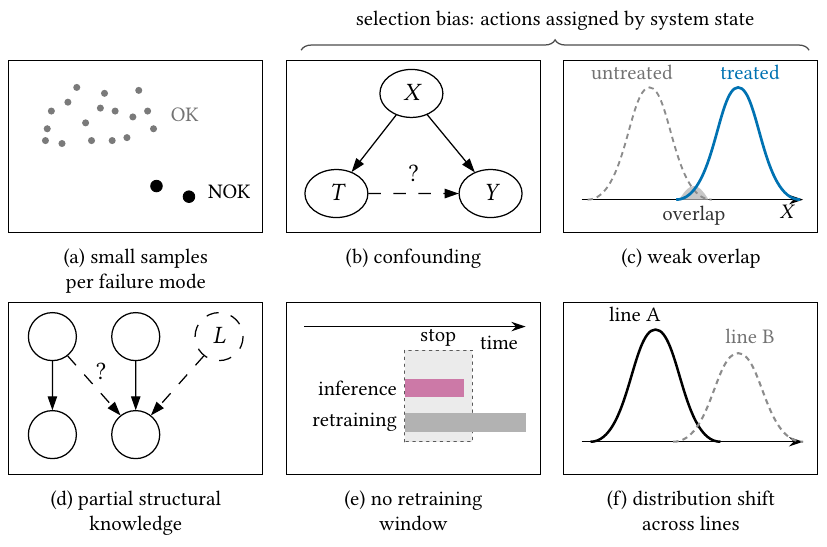}
    \caption[Operating conditions of observational manufacturing data.]{Operating conditions (a) to (f) of observational manufacturing data that constrain causal estimation and its validation.}
    \label{fig:synth-data-conditions}
\end{figure}

Validation is not the only obstacle. Estimation itself must operate under the data conditions of running production, which were derived in RG3 and are summarised in Figure~\ref{fig:synth-data-conditions}. Failure events are rare relative to nominal operation, so the effective sample size per failure mode is small. Prognostics and health management data are also sparse, noisy, and strongly context-dependent~\cite{finkPHMPartI,finkPHMPartII} (panel~a).

Under closed-loop control and historical maintenance policies, the recorded action additionally tracks the observed process state, which creates two distinct problems. The first is confounding. Maintenance is performed more often on already degraded equipment, so the process state acts as a common cause of action and outcome. The raw association between action and outcome is therefore a biased estimate of the interventional effect, even where both action levels are observed (Subsection~\ref{subsec: structural limitations}, panel~b). Adjusting for the process state through the back-door criterion removes this bias under the conditions of Equation~\ref{eq:bg-backdoor}. The second is weak overlap. Causal estimation requires sufficient overlap: every admissible action must occur with positive probability in every relevant context (Subsection~\ref{subsec:ch4-crl-assumptions}). A deterministic policy collapses the action onto a function of the process state. For instance, if an intensive maintenance protocol is applied only to machines beyond a certain age, the records contain neither old machines without the protocol nor young machines with it, so the back-door adjustment has no support at those contexts and the protocol effect is not identifiable there without extrapolation (panel~c). Near-violations of overlap soften the problem but do not resolve it, since actions recorded only rarely in a context destabilise model selection and inflate the variance of estimators that reweight by the conditional action probability $P(T\mid X)$~\cite{DoutreligneHowToSelect}. Confounding and weak overlap arise from a common origin: past actions were not assigned at random, so the records cover only a restricted region of the possible state-action combinations~\cite{VANDERSCHUEREN2023108798}. A further selection acts on the records, since parts that fail early or block the machine may never reach the downstream quality measurements at all (Subsection~\ref{subsec: structural limitations}). Structural knowledge of the line is typically incomplete. Causal graphs are absent or only partially known, which corresponds to the \emph{none} and \emph{ordered} knowledge regimes of Subsection~\ref{subsec:knowledge-availability} (panel~d). A recommendation must further be produced within the operational window of a production stop, so fitting or retraining a case-specific estimator at decision time is not feasible. Estimation must rely on models prepared before the incident (Subsection~\ref{subsec:decision-time-knowledge-horizon}, panel~e). The data-generating process finally drifts within a line through tooling wear, raw-material variation, and recipe changes. It also differs across lines, products, and plants. The covariate distribution thus shifts between training and deployment. Across lines the mechanisms themselves may differ. A model fitted on the history of one line can therefore not be assumed to transfer unchanged~\cite{finkPHMPartII} (panel~f).

A prescriptive query asks for the effect of an action in a context where that action was never recorded. Answering it therefore requires extrapolation into regions of the state-action space that the recorded operation never visited, for example degraded equipment under corrective actions that were never tried (panel~c). This extrapolation is not an edge case but the substance of the prescriptive question, because a recommendation is most valuable exactly where the records are sparse. A model fitted on the records must extrapolate into these regions. No ground truth exists against which its predictions could be checked. Synthetic and semi-synthetic data address both obstacles at once. Data drawn from a known \acs{SCM} carry interventional answers by construction. Sample size, the disclosed structural knowledge, the assignment mechanism, and the distribution shift across lines can be set directly. Confounding and overlap follow from the sampled structure and the assignment mechanism, and can be quantified exactly on the generated data. Estimators are thereby evaluated under controlled and measurable versions of the conditions of Figure~\ref{fig:synth-data-conditions}. This yields the ground truth needed for robust evaluation and the diverse structures needed for pre-training the learned causal estimators of Chapter~\ref{chap:prima-causa-model}.

Existing causal data generators for manufacturing do not close this gap at the level of process structure. The \texttt{causalAssembly} generator fits a single assembly line, deriving a partial ground-truth graph from process physics and estimating the attached conditionals by distributional random forests~\cite{causalAssembly}. A comparable generator fits one injection-moulding machine through a graphical causal model~\cite{fraunhoferSyntheticData}. A physics-based simulator models one production line~\cite{causalMan2025}. Parametric benchmark generators sample structural causal models from an abstract space without a manufacturing topology~\cite{poinsot2026causalProfiler}. What remains absent is a parametric topological prior over production-line structure, that is, a distribution over production-line graphs from which new line topologies can be sampled, generalising across lines rather than refitting one dataset.

This chapter contributes such a prior together with the data pipeline built on it. The first contribution is a parametric sampler of \acs{CPG} instances $G_{\mathrm{CPG}}$, provided as a software artefact, that serves as a topological source for production-line synthetic data, independent of any one reference line (Section~\ref{sec:synth-cpg}). The second adapts open-source \acs{SCM} generation frameworks~\cite{sauter2024causalplayground,qu2025tabicl,pfefferle2025tfmplayground} to the manufacturing setting by replacing their structureless graph source with the $G_{\mathrm{CPG}}$ prior and adding manufacturing mixed-signal mechanisms. From the resulting \acs{SCM} instances, observational and interventional data are drawn (Section~\ref{sec:synth-pipeline}). The composition of mechanism families into \acs{SCM} instances and the $\dop$-operator are reused from these frameworks. The third is a semi-synthetic anchoring procedure that retains real measurement scales while keeping the causal mechanism known (Section~\ref{sec:synth-semi-fmcg}). Together these operationalise the synthetic and semi-synthetic stages of meta-requirement MR13~(Table~\ref{tab:ch4-meta-requirements}). The structural plausibility of the sampler is established in Section~\ref{sec:synth-cpg} through its distinguishability from a structureless random-graph null. Distinguishability is a necessary but not a sufficient condition for utility.

The \acs{CPG} and the data drawn from it remain on the observable sensor layer to which the \acs{CRL} is scoped (Subsection~\ref{subsec:ch4-crl-abgrenzung}). Within one decision case, the graph is treated as acyclic and time-invariant over the operational horizon of interest. This is a modelling assumption rather than a property of the line, since closed-loop control induces feedback over time (RG3). The assumption is stated with the edge semantics of the \acs{CPG} in Section~\ref{sec:synth-cpg} and is not claimed beyond that horizon.

\section{Causal Process Graph}
\label{sec:synth-cpg}

A \acf{CPG} is a directed acyclic graph $G=(\mathbf{V},\mathbf{E})$ over the sensor and process-variable set $\mathbf{V}$ of the production system, where $\mathbf{V}$ is the endogenous variable set of the structural causal model introduced in Chapter~\ref{chap: Background} and summarised in Table~\ref{tab:ch4-notation}. An edge $(V_a,V_b)\in\mathbf{E}$ denotes a physically plausible, process-temporally preceding cause-effect relationship between a process state or variable $V_a$ and a downstream process state, variable, or quality event $V_b$~\cite{neuer2024maschinelles}. The edge thus carries two requirements simultaneously: it must be admissible under the physics of the process, and the cause must precede the effect along the operational timeline. \acsp{CPG} of this form have served as offline reference structures for network-based root cause identification~\cite{saretzkyNetworks} and as the topological backbone of graph-based synthetic data generators in manufacturing~\cite{saretzky2026integratingcausalfoundationmodel,fraunhoferSyntheticData,causalAssembly}.

A single \acs{CPG} describes one production line at one resolution. The contribution of this section is a parametric sampling procedure that produces \acs{CPG} instances $G_{\mathrm{CPG}}$ as a general topological source for production-line synthetic data, independent of any one reference line. The sampler is parameterised by the node count $n$, an optional station count $K$, and a single density level $\eta\in[0,1]$ that sets the within-station and the cross-station edge rates, with unspecified parameters drawn per seed from configured ranges. The functional layer, that is, the structural equation attached to each node, is described in Section~\ref{sec:synth-pipeline}, so that the topological sampler remains separable from the choice of mechanism family. The full construction, the parameter ranges, and the sampling algorithm are documented in the open-source implementation\footnote{\url{https://github.com/FelixSaretzky/ProductionLineSimulator}}.

The sampler is designed around six structural principles that are consistent with published reference graphs of production lines~\cite{causalAssembly,saretzkyNetworks} and with the way manufacturing process data are described in the engineering literature~\cite{neuer2024maschinelles,finkPHMPartI}. First, material flows forward through the line, so every cause precedes its effects along the operational timeline. Second, the line is modular: the variables group into an ordered sequence of stations whose material flow forms either a single chain or parallel branches that merge for final assembly. Third, station sizes are heterogeneous, so small buffer stations and large processing stations coexist on one line. Fourth, within-station structure is sparse: a station contributes few internal forward edges relative to its pair count. Fifth, cross-station coupling dominates the edge mass and may skip intermediate stations, which reflects the long-range dependencies of the published reference line. Sixth, measurement and quality variables aggregate many upstream influences, so nodes with a high in-degree, comparable to quality gates, can arise without being imposed. The first three principles act as hard structural constraints, while the last three enter through the edge probabilities and the target weights. The list is not claimed to be exhaustive. It is sufficient to separate \acs{CPG} samples from a structureless random-graph null, as established at the end of this section.

In physical terms these principles make a \acs{CPG} anisotropic: the graph carries a distinguished input-to-output axis along which the stations are ordered, and this axis is fixed by the physics of the production process. The Erd\H{o}s--R\'enyi distribution is isotropic in the corresponding sense, as its skeleton distribution is exchangeable under permutation of node labels and carries no privileged axis. It is therefore adopted as the structureless null against which the structural plausibility of $G_{\mathrm{CPG}}$ is tested. The scale-free, small-world, and stochastic-block families that also recur in graph-prior benchmarks are not used as competing baselines here. The structural characteristics they would contribute arise in the sampler as emergent properties instead. Hub structure follows from the heavy-tailed in-attractiveness weights of the construction, and community structure follows from the station partition.

The construction realises the six principles in five steps, summarised here and specified in the open-source implementation. The density level $\eta$ is drawn if unspecified and mapped to a within-station edge rate $r_{\mathrm{w}}$ and a cross-station edge rate $r_{\mathrm{c}}$. The $n$ nodes are partitioned into $K$ ordered stations of heterogeneous size that form an in-line or branched-merge backbone. Every node carries a heavy-tailed in-attractiveness weight, denoted $w_b$ for a node $V_b$, drawn lognormally with unit mean. Edges are placed forward within stations and between all backbone-reachable station pairs, with probability $\min\{1,\,r_{\mathrm{w}}\,w_b\}$ within a station and $\min\{1,\,r_{\mathrm{c}}\,w_b\}$ across stations, independent of the hop distance. A final step adds one edge along every backbone arc that would otherwise carry none and one edge for every otherwise unconnected node. Node identifiers follow the station order, so every edge runs from a lower to a higher identifier and $G_{\mathrm{CPG}}$ is acyclic by construction. Structural roles are not imposed by the construction. Source, internal, and aggregating nodes emerge from the realised degrees, and these emergent positions are distinct from the task-specific variable-role map $r_{\mathcal{K}}$ of Chapter~\ref{chap:prima-causa}, which assigns intervention-target, context, and outcome labels per decision case. The procedure is summarised in Algorithm~\ref{alg:cpg-sample}.

\begin{algorithm}[!htbp]
\DontPrintSemicolon
\KwIn{node count $n$, a seed, optional density level $\eta\in[0,1]$, optional station count $K$.}
\KwOut{a \acs{CPG} instance $G_{\mathrm{CPG}}$, acyclic by construction.}
Draw the density level $\eta$ if unspecified and map it to a within-station rate $r_{\mathrm{w}}$ and a cross-station rate $r_{\mathrm{c}}$\;
Partition the $n$ nodes into $K$ ordered stations of heterogeneous size\;
Draw a station backbone, either an in-line chain or parallel branches merging at the sink\;
Draw an in-attractiveness weight $w_b$ for every node $V_b$\;
\ForEach{station}{add forward within-station edges, each with probability $\min\{1,\,r_{\mathrm{w}}\,w_b\}$\;}
\ForEach{backbone-reachable station pair}{add forward cross-station edges, each with probability $\min\{1,\,r_{\mathrm{c}}\,w_b\}$, independent of the hop distance\;}
Add one edge along every backbone arc that would otherwise carry none, and one edge for every otherwise unconnected node\;
\Return $G_{\mathrm{CPG}}$\;
\caption{\texttt{sample\_dag}$(n,\eta,\mathrm{seed},K)$}
\label{alg:cpg-sample}
\end{algorithm}

\begin{figure}[!htbp]
    \centering
    \includegraphics[width=\textwidth]{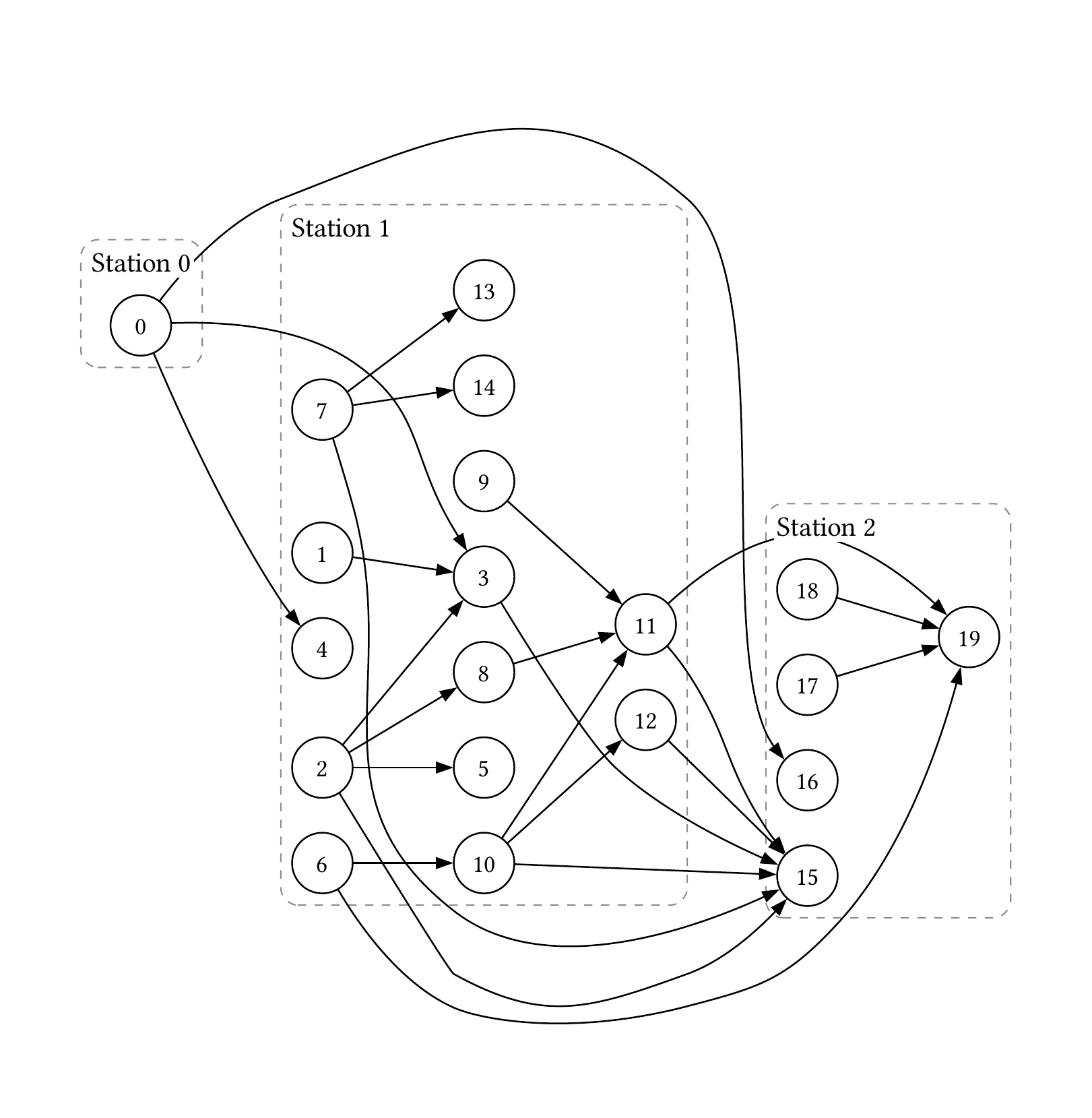}\\[8pt]
    \includegraphics[width=\textwidth]{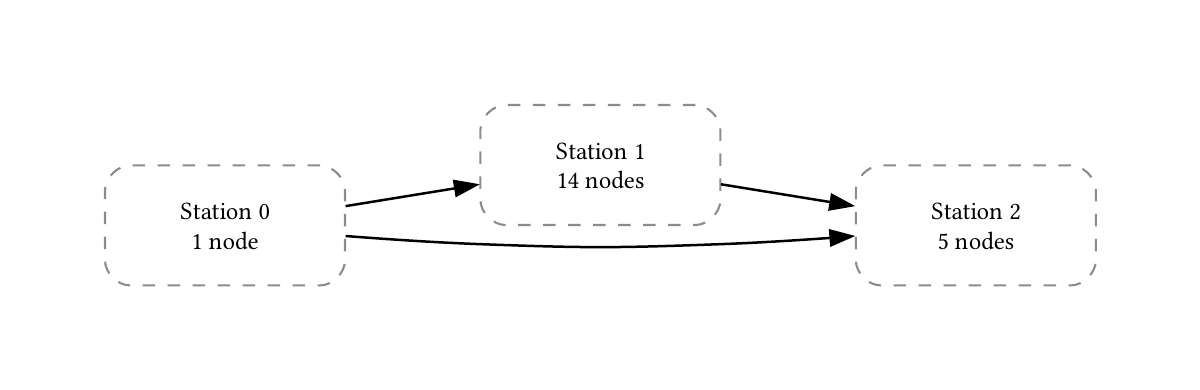}
    \caption[Example CPG instance with station backbone.]{A sampled \acs{CPG} instance with $n=20$ and $K=3$ (top) and its station backbone (bottom).}
    \label{fig:cpg-example}
\end{figure}

Figure~\ref{fig:cpg-example} shows a sampled instance with $n=20$ nodes and $K=3$ stations, together with its station backbone. The nodes are partitioned into three stations, drawn as dashed boxes ordered left to right, and the same stations reappear in the backbone below. Forward edges wire the nodes within a station and across stations along the backbone-reachable pairs, so the instance carries within-station and cross-station structure on top of the station ordering.

The structural plausibility of $G_{\mathrm{CPG}}$ is assessed against the Erd\H{o}s--R\'enyi \acs{DAG} sampler $G_{\mathrm{ER}}$, a standard null in synthetic causal benchmarks~\cite{reisachBewareSimulatedDAG} that also underlies the graph prior of parametric benchmark generators~\cite{poinsot2026causalProfiler}. The assessment follows the distance-contrast protocol of Wills and Meyer~\cite{wills2020metrics} for comparing two graph populations through a spectral distance. Because the adjacency matrix of a \acs{DAG} is nilpotent and carries an all-zero spectrum, the comparison is conducted on the undirected skeleton of each graph. It therefore probes the skeletal connectivity of $G_{\mathrm{CPG}}$ and not the edge orientation imposed by the construction.

For the undirected skeleton of a graph, the adjacency matrix $\mathbf{A}$, the combinatorial Laplacian $\mathbf{L}$, and the normalized Laplacian $\boldsymbol{\mathcal{L}}$, not to be confused with the hierarchy layers $\mathcal{L}_1$ to $\mathcal{L}_3$ of Chapter~\ref{chap: Background}, are defined as~\cite{wills2020metrics}
\begin{equation}
\mathbf{A}_{ij}=
\begin{cases}1 & \text{if } V_i \text{ and } V_j \text{ are adjacent},\\ 0 & \text{otherwise},\end{cases}
\qquad \mathbf{L}=\mathbf{D}-\mathbf{A},\qquad \boldsymbol{\mathcal{L}}=\mathbf{D}^{-1/2}\,\mathbf{L}\,\mathbf{D}^{-1/2},
\label{eq:cpg-matrices}
\end{equation}
where $d_i=\sum_j \mathbf{A}_{ij}$, $\mathbf{D}=\operatorname{diag}(d_1,\dots,d_n)$, and $\mathbf{D}^{-1/2}_{ii}=d_i^{-1/2}$ for $d_i>0$ and $0$ otherwise. The eigenvalues of each matrix are sorted, the adjacency spectrum in descending order and the two Laplacian spectra in non-decreasing order from zero~\cite{wills2020metrics}:
\begin{equation}
\begin{aligned}
\text{(adjacency)} &\quad \lambda_1^{\mathbf{A}}\geq \lambda_2^{\mathbf{A}}\geq\dots\geq\lambda_n^{\mathbf{A}},\\
\text{(combinatorial Laplacian)} &\quad 0=\lambda_1^{\mathbf{L}}\leq \lambda_2^{\mathbf{L}}\leq\dots\leq\lambda_n^{\mathbf{L}},\\
\text{(normalized Laplacian)} &\quad 0=\lambda_1^{\boldsymbol{\mathcal{L}}}\leq \lambda_2^{\boldsymbol{\mathcal{L}}}\leq\dots\leq\lambda_n^{\boldsymbol{\mathcal{L}}}\leq 2 .
\end{aligned}
\label{eq:cpg-spectra}
\end{equation}
For two skeletons $G,G'$ of equal size $n$, the adjacency spectral distance is the $\ell_2$ distance between their sorted adjacency spectra~\cite{wills2020metrics},
\begin{equation}
d_{\mathbf{A}}(G,G')=\sqrt{\sum_{i=1}^{n}\bigl(\lambda_i^{\mathbf{A}}(G)-\lambda_i^{\mathbf{A}}(G')\bigr)^{2}},
\label{eq:lambda-distance}
\end{equation}
and the combinatorial-Lapla\-cian and normalized-Lapla\-cian spectral distances $d_{\mathbf{L}}$ and $d_{\boldsymbol{\mathcal{L}}}$ are defined in the same way on the respective spectra. A spectral distance requires no correspondence between the node sets, so it compares the two populations on structure alone. The labels adjacency, combinatorial Laplacian, and normalized Laplacian therefore refer to the matrix whose spectrum enters the distance. They do not denote an entrywise comparison of the matrices themselves, which would require a node correspondence~\cite{wills2020metrics}.

Two graph populations are contrasted through the studentized distance contrast of Wills and Meyer~\cite{wills2020metrics}. Let $D_0=d(G_0,G_0')$ denote a within-null distance between two Erd\H{o}s--R\'enyi draws and $D_1=d(G_0,G_1)$ a cross distance between an Erd\H{o}s--R\'enyi draw and a \acs{CPG} draw. With $\mu_{D_0}$ and $\sigma_{D_0}$ the mean and standard deviation of the within-null distances $D_0$, the contrast of a cross distance $D_1$ is
\begin{equation}
\widehat{D}_1=\frac{D_1-\mu_{D_0}}{\sigma_{D_0}}.
\label{eq:distance-contrast}
\end{equation}
Each \acs{CPG} skeleton is compared against Erd\H{o}s--R\'enyi nulls matched to its own realised edge density, because the \acs{CPG} density varies widely across seeds and an unmatched comparison would let an edge-count gap rather than structure drive the distance. A contrast above three is read as separation and a contrast between one and three as weak separation, in both cases as an effect size relative to the within-null spread rather than as a calibrated significance test.

\begin{figure}[!htbp]
    \centering
    \includegraphics[width=\textwidth]{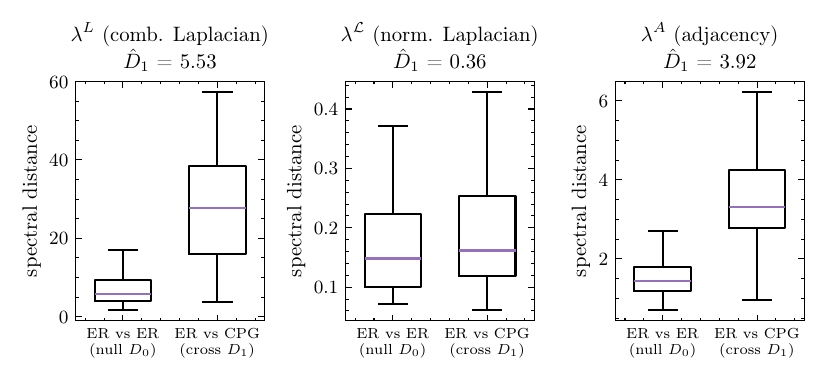}
    \caption[Spectral two-sample contrast: \acs{CPG} versus Erd\H{o}s--R\'enyi.]{Spectral-distance contrast $\widehat{D}_1$ between \acs{CPG} and density-matched Erd\H{o}s--R\'enyi skeletons. Each panel shows the distance computed from one spectrum: the combinatorial-Lapla\-cian spectrum $\lambda^{\mathbf{L}}$, the normalized-Lapla\-cian spectrum $\lambda^{\boldsymbol{\mathcal{L}}}$, and the adjacency spectrum $\lambda^{\mathbf{A}}$.}
    \label{fig:cpg-spectral-contrast}
\end{figure}

At $n=98$ with fifty samples per population and per-pair density matching, the contrast is $\widehat{D}_1\approx 3.9$ for the adjacency spectral distance, $\widehat{D}_1\approx 5.5$ for the combinatorial-Lapla\-cian spectral distance, and $\widehat{D}_1\approx 0.4$ for the normalized-Lapla\-cian spectral distance (Figure~\ref{fig:cpg-spectral-contrast}). The \acs{CPG} population is therefore separated from a same-density Erd\H{o}s--R\'enyi null on structure alone in the adjacency and the combinatorial-Lapla\-cian spectrum, whereas the normalized-Lapla\-cian contrast falls below one and indicates no separation on that representation. The combinatorial-Lapla\-cian spectral distance largely tracks the degree sequence~\cite{wills2020metrics}. Its separation is consistent with the degree heterogeneity induced by the heavy-tailed in-attractiveness weights, which departs from the binomial degree distribution of the null. The normalized Laplacian factors out the overall degree scale, so the remaining contrast is small once the degree-driven signal is removed. Because the three spectral distances occupy different absolute scales, the panels of Figure~\ref{fig:cpg-spectral-contrast} carry independent vertical axes and are compared through $\widehat{D}_1$ rather than through box height.\footnote{The whiskers extend to $1.5$ times the interquartile range, and outliers beyond this range are omitted for legibility. The contrast $\widehat{D}_1$ is computed on the full sample of distances and is therefore unaffected by this display choice.}

\begin{figure}[!htbp]
    \centering
    \includegraphics[width=0.75\textwidth]{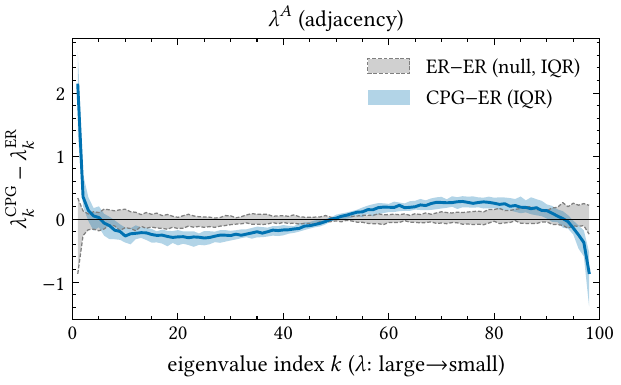}
    \caption[Per-index adjacency spectral difference.]{Per-index difference $\lambda_k^{\mathrm{CPG}}-\lambda_k^{\mathrm{ER}}$ along the sorted adjacency spectrum (blue, with median and interquartile band) against the within-null band (grey).}
    \label{fig:cpg-spectral-diff}
\end{figure}

Figure~\ref{fig:cpg-spectral-diff} localises the separation of the adjacency spectra across the eigenvalue index. The per-index difference $\lambda_k^{\mathrm{CPG}}-\lambda_k^{\mathrm{ER}}$ departs from the within-null band along large parts of the spectrum, with the strongest signal at the two spectral ends. The leading eigenvalue $\lambda_1^{\mathbf{A}}$ is markedly elevated, which is consistent with the heterogeneous, hub-weighted degree sequence of the \acs{CPG} relative to a uniform Erd\H{o}s--R\'enyi null, in line with the reading of the adjacency spectrum as a representation sensitive to community and high-connectivity structure~\cite{wills2020metrics}. The most negative eigenvalues fall below the null band. Cross-station edges dominate the skeleton and connect distinct station blocks, so the skeleton is locally close to a multipartite graph between stations, and such near-bipartite structure deepens the negative end of the adjacency spectrum.

Distinguishability from a structureless null establishes that the sampler produces structure. It does not establish that real lines lie inside the family the sampler produces. A complementary coverage check is therefore conducted against the \texttt{causalAssembly} ground-truth graph, the only published ground-truth structure derived from measured data of a real production line~\cite{causalAssembly}. Over $300$ draws at the reference node and station count, the real line lies inside the generated range on every headline statistic considered, including edge density, cross-station edge share, root-node share, and in-degree dispersion. The generated family therefore covers the one available real line without imitating it, which is the behaviour required of a topological prior rather than of a replica of one dataset.

The \acs{CPG} sampler defined here is the topological foundation for the synthetic data pipeline of Section~\ref{sec:synth-pipeline}, which composes $G_{\mathrm{CPG}}$ instances with one or more mechanism families to produce \acs{SCM} instances from which observational and interventional data are drawn. The pre-training corpus of Chapter~\ref{chap:prima-causa-model} additionally mixes $G_{\mathrm{CPG}}$ draws with Erd\H{o}s--R\'enyi draws in a ratio of three to one, so the learned estimator is not conditioned exclusively on production-line topology.

\section{Synthetic Data}
\label{sec:synth-pipeline}

The generation of synthetic data under a known causal mechanism is well established in causal machine learning~\cite{poinsot2025synthetic}, yet remains uncommon in manufacturing and industrial engineering, where empirical work has relied on recorded process data. This section turns the \acs{CPG} instance $G_{\mathrm{CPG}}$ of Section~\ref{sec:synth-cpg} into a structural causal model from which observational and interventional data are drawn.\footnote{The \acs{SCM}s are constructed and sampled with open-source code from~\cite{sauter2024causalplayground,qu2025tabicl,pfefferle2025tfmplayground}.}

Manufacturing data are mixed. Continuous sensor readings coexist with discrete and categorical signals such as actuator states, OK/NOK quality flags, and fault classes~\cite{causalMan2025}. To reproduce both signal types under a known mechanism, the \acs{CPG} $G=(\mathbf{V},\mathbf{E})$ is equipped with a \acs{SCM} $\mathcal{M}=(\mathbf{U},\mathbf{V},\mathcal{F},P(\mathbf{U}))$~\cite{bareinboim2025causal}, where $\mathcal{M}$ is the structural causal model, distinct from the decision mode $M$ of Chapter~\ref{chap:prima-causa}. Each process variable $V_i$ is assigned a value by a structural function $f_i\in\mathcal{F}$ of its parents $\mathrm{pa}(V_i)$ in $G$ and an independent exogenous noise variable $U_i$, following Equation~\eqref{eq:bg-scm-assignment}.

The form of $f_i$ is chosen by the data type of the variable, so that one \acs{SCM} carries the mixed signals of a real line. A continuous sensor variable is the sum of a deterministic mechanism $m_i$ of its parents and additive noise,
\begin{equation}
V_i := m_i\bigl(\mathrm{pa}(V_i)\bigr) + U_i.
\label{eq:scm-additive}
\end{equation}
A binary variable, an actuator state or an OK/NOK quality flag, thresholds the continuous latent $z_i = m_i(\mathrm{pa}(V_i)) + U_i$, as a single-sided switch or a two-sided in-spec flag,
\begin{equation}
V_i := \mathbf{1}\{z_i > \theta_i\}\quad\text{(switch)},\qquad
V_i := \mathbf{1}\{\mathrm{LTL}_i \le z_i \le \mathrm{UTL}_i\}\quad\text{(in-spec flag)},
\label{eq:scm-binary}
\end{equation}
with $\mathbf{1}\{\cdot\}$ the indicator function, $\theta_i$ a threshold, and $\mathrm{LTL}_i,\mathrm{UTL}_i$ tolerance limits. A categorical variable, such as an unordered fault class, draws one latent per category and takes the largest,
\begin{equation}
V_i := \operatorname*{arg\,max}_{k\in\{1,\dots,C\}} \bigl( m_{i,k}(\mathrm{pa}(V_i)) + U_{i,k}\bigr),
\label{eq:scm-argmax}
\end{equation}
with $C$ the number of categories. The deterministic mechanism $m_i$ is a randomly initialised multilayer perceptron, which makes the parent-to-child maps nonlinear, or, for a small fraction of nodes, a gradient-boosted tree, and the noise $U_i$ is drawn from a mixture family. The mechanism family is independent of the topology, so a single $G_{\mathrm{CPG}}$ instance yields many \acs{SCM} instances.

\begin{figure}[!htbp]
    \centering
    \includegraphics[width=\textwidth]{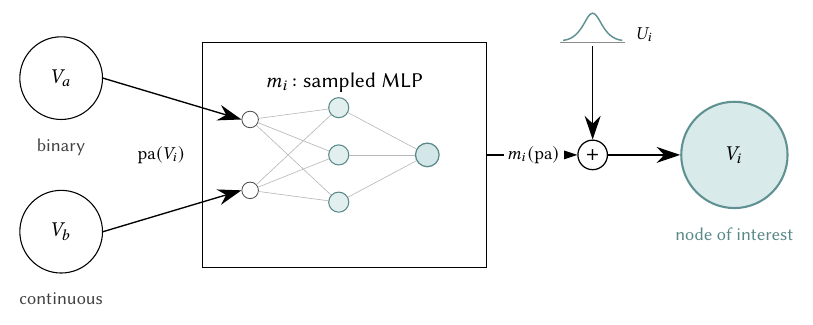}
    \caption[Structural mechanism at one CPG node.]{Data generation at one \acs{CPG} node: the mixed-type parents feed the sampled mechanism $m_i$, and zero-mean mixture noise $U_i$ is added.}
    \label{fig:synth-node-mechanism}
\end{figure}

Figure~\ref{fig:synth-node-mechanism} illustrates the mechanism for a single node with two parents of different data type. The parent values are assembled into one input vector $\mathrm{pa}(V_i)$, in which a binary parent enters as a zero-one column and a continuous parent as a real column, so a node that mixes signal types is admitted without special handling. The deterministic part $m_i$ maps this vector through the sampled mechanism, a multilayer perceptron whose nonlinearity is drawn per node or per layer from a diverse pool adapted from~\cite{qu2025tabicl}, for example the identity, $\tanh$, $\sin$, the rectified linear unit, or a Gaussian bump $e^{-x^2}$. The parent-to-child response is therefore nonlinear and varies from node to node. The exogenous term $U_i$ is drawn from a zero-mean mixture over Gaussian, Laplace, Student-$t$, and Gumbel components, so the noise is in general heavy-tailed rather than Gaussian. The activation pool, the layer dimensions, and the mixture weights are reported in the open-source implementation.

Observational data are drawn by an ancestral forward pass: the exogenous noise is sampled, and each $V_i$ is evaluated in topological order, which exists because $G_{\mathrm{CPG}}$ is acyclic. Interventional data follow Pearl's $\dop$-operator. An intervention replaces the mechanism of the target $T$ and cuts off its incoming edges, the submodel of Equation~\eqref{eq:bg-submodel}~\cite{bareinboim2025causal}, while every other mechanism is retained. An atomic intervention $\dop(T{=}t)$ fixes $T$ at the level $t$, giving the truncated factorisation~\cite{pearl2009}
\begin{equation}
P\bigl(\mathbf{V}\mid \dop(T{=}t)\bigr)=\delta_t(T)\prod_{i:\,V_i\neq T} P\bigl(V_i\mid \mathrm{pa}(V_i)\bigr),
\label{eq:scm-truncated}
\end{equation}
with $\delta_t(T)$ the point mass at $t$. Because the target $T$ is itself a process variable, the intervention can be continuous or binary: a continuous target admits a dose-response intervention $\dop(T{=}t)$ at any level, and a binary target the contrast between $\dop(T{=}0)$ and $\dop(T{=}1)$. The default sampler instead draws $T$ from its observational marginal, which cuts its dependence on its parents while preserving its marginal. The generated interventional data satisfy the defining invariances of the $\dop$-operator, so the estimand $P(Y\mid \dop(T{=}t),X{=}x)$ that the \acs{CRL} evaluates follows by conditioning, subject to interventional overlap at $x$. This is the quantity the observational records lack.

\begin{figure}[!htbp]
    \centering
    \includegraphics[width=0.75\textwidth]{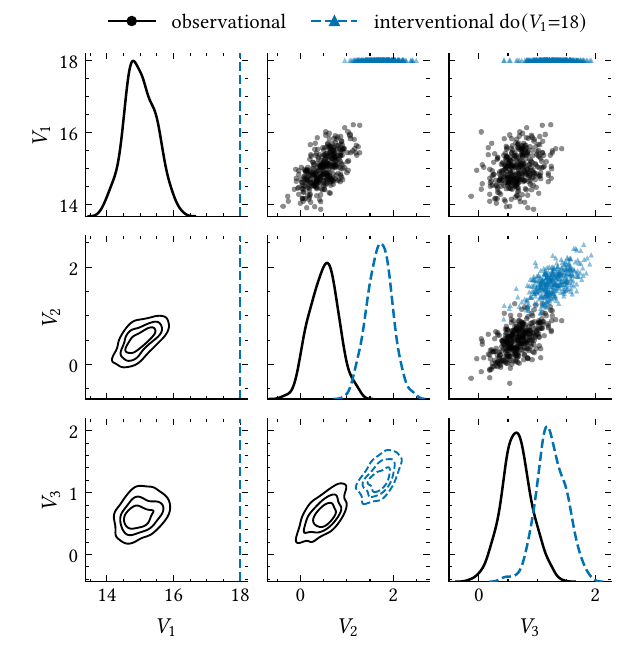}
    \caption[Observational versus interventional data.]{Observational and interventional samples of three process variables from an illustrative linear \acs{SCM}.}
    \label{fig:obs-int-image}
\end{figure}

Figure~\ref{fig:obs-int-image} makes the contrast between the two regimes visible at the level of the drawn data for an illustrative linear \acs{SCM}. For a chain of three process variables, the marginal densities, the pairwise scatter, and the joint density contours are compared between an observational and an interventional sample. The intervention is atomic, $\dop(T{=}t)$, on the upstream process variable $V_1$. The marginal of the intervened variable degenerates to the intervention level, as the point mass of Equation~\eqref{eq:scm-truncated} prescribes. The downstream variables are displaced through the retained mechanisms, and the displacement weakens along the chain. The difference between $P(\mathbf{V})$ and $P(\mathbf{V}\mid \dop(T{=}t))$ is thus directly observable in the generated samples.

The spectral contrast of Section~\ref{sec:synth-cpg} certifies the topology alone. A joint check across graph, mechanisms, and data follows the profiling methodology of CausalProfiler~\cite{poinsot2026causalProfiler}. Each sampled instance is summarised by a profile vector $\varphi\in\mathbb{R}^{35}$, adapted from the metric families of that framework, that collects thirteen graph statistics (degrees, ancestors, descendants, causal paths), six confounding statistics (c-components, bidirected siblings), six mechanism statistics (stochasticity, linearity, monotonicity), and ten statistics of the drawn observational data (positivity, marginal and joint imbalance, entropy).

\begin{figure}[!htbp]
    \centering
    \includegraphics[width=0.8\textwidth]{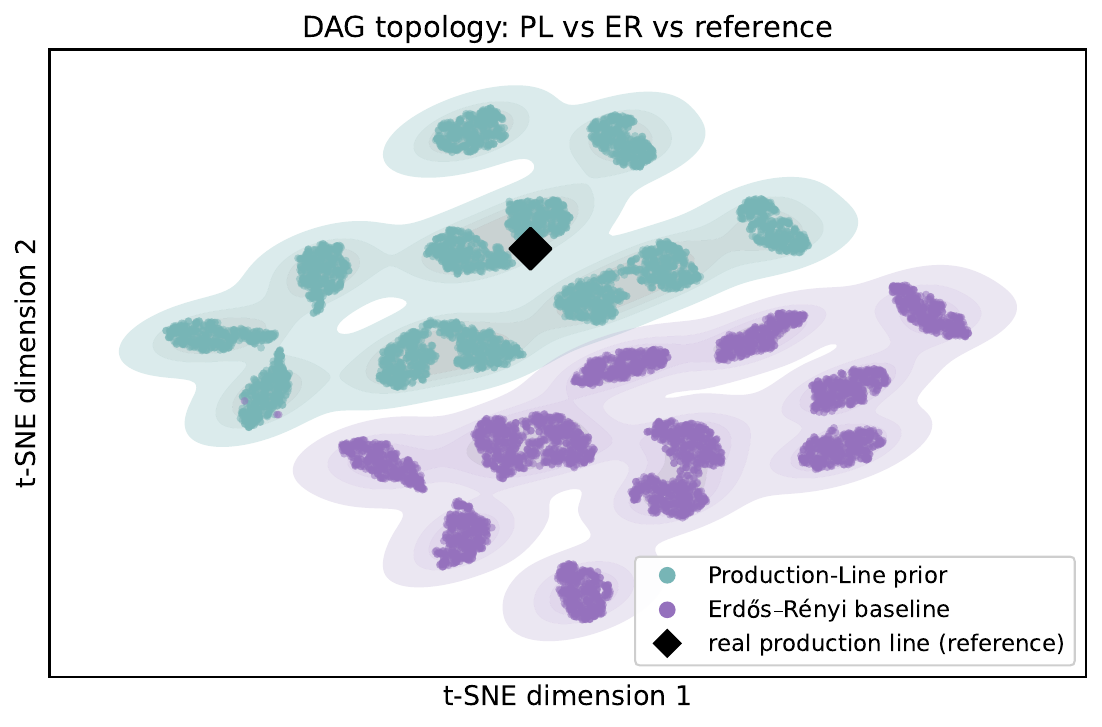}
    \caption[t-SNE embedding of generated and real instances.]{t-SNE embedding of the profile vectors $\varphi$ of \acs{CPG} and Erd\H{o}s--R\'enyi instances and of the real production-line reference.}
    \label{fig:tsne-plot}
\end{figure}

\begin{figure}[!htbp]
    \centering
    \includegraphics[width=\textwidth]{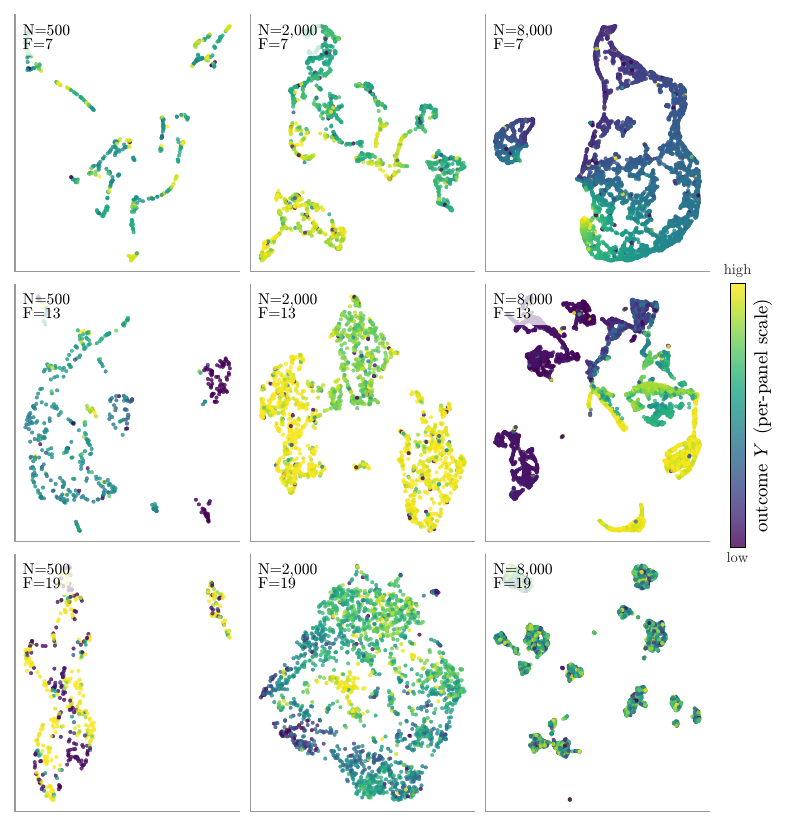}
    \caption{UMAP embeddings of the sample rows of nine independent generator draws.}
    \label{fig:umap-diversity}
\end{figure}

Figure~\ref{fig:tsne-plot} embeds the profile vectors of \acs{CPG} and Erd\H{o}s--R\'enyi instances in two dimensions with t-SNE~\cite{maaten2008tsne}, together with the profile computed from recorded reference data of a real production line. Since t-SNE primarily preserves local neighbourhood structure, the embedding is read descriptively. The two generated families occupy separated regions, and the real reference falls inside the \acs{CPG} region. On the profile statistics considered, the generated family is therefore closer to the real production line than an Erd\H{o}s--R\'enyi benchmark family, which complements the coverage check of Section~\ref{sec:synth-cpg}. Figure~\ref{fig:tsne-plot} embeds one profile vector per \acs{SCM} instance to illustrate the \emph{fidelity} relative to real data, while Figure~\ref{fig:umap-diversity} presents the generated datasets themselves to demonstrate their \emph{diversity}. Each panel represents an independent draw from the generator, a new \acs{CPG} instance and a \acs{SCM} of varying size, whose sample rows were embedded in two dimensions using UMAP~\cite{mcinnes2018umap} based on the drawn covariates and labelled according to the result $Y$. Each panel states the number of sample rows $N$ and covariates $F$ of its draw. The nine panels differ clearly in the geometry of their point clouds, ranging from a few compact clusters to filaments and diffuse manifolds, and the outcome varies continuously within each panel. This heterogeneity is intentional, since the generator serves as a source of pre-training data for a causal foundation model, and such a model must be exposed to a wide range of data geometries rather than to a single recurring form.

\begin{figure}[!htbp]
    \centering
    \includegraphics[width=0.62\textwidth]{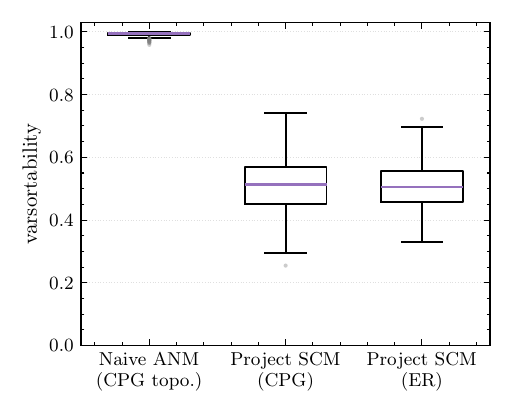}
    \caption[Varsortability of the generated data.]{Varsortability over $200$ sampled \acs{SCM} instances at $n=98$: naive additive-noise mechanism (left), proposed mechanisms (centre), and Erd\H{o}s--R\'enyi reference (right).}
    \label{fig:synth-varsortability}
\end{figure}

The generated observational data are, in addition, not varsortable~\cite{reisachBewareSimulatedDAG}. Across $200$ sampled \acs{SCM} instances at $n=98$, the varsortability is $0.51$, close to the chance level of $0.5$ and far below the value of $0.99$ attained by a naive additive-noise mechanism on the same \acs{CPG} topologies. It matches the level attained by the same mechanisms on an Erd\H{o}s--R\'enyi reference (Figure~\ref{fig:synth-varsortability}). The marginal-variance order therefore carries no causal-order information, and a benchmark drawn from these data cannot be solved by ranking the variables on their variance.

\section{Semi-Synthetic Data}
\label{sec:synth-semi-fmcg}

The synthetic \acs{SCM} data of Section~\ref{sec:synth-pipeline} establish a controlled ground truth, but their covariates are drawn from the sampled \acs{CPG} topologies and mechanism library of that section rather than from an existing dataset. The semi-synthetic step closes this gap from the other direction: covariates are taken unchanged from an existing dataset (Table~\ref{tab:synth-semi-datasets}) and only the treatment and the outcome are simulated on top of them. The causal ground truth remains known by construction, while the covariate distribution, its marginal shapes, its correlation structure, and its measurement scales are those of the host dataset. This gives the frozen model of Chapter~\ref{chap:prima-causa-model} its first evaluation on covariates it was not trained to represent (out-of-distribution). The resulting evaluation is reported in Section~\ref{sec:eval-transfer}.

\paragraph{Datasets.} Five public industrial and manufacturing datasets supply the covariates (Table~\ref{tab:synth-semi-datasets}). Each contributes only its pre-treatment sensor and process readings, retaining $N_{\mathrm{host}}$ units and $F$ covariates after cleaning. Original outcome, failure, or quality labels are discarded before simulation, and near-constant columns are removed since they carry no signal for the treatment or response mechanism. The five datasets differ in how their covariates were obtained. This distinction is preserved explicitly in the table's provenance column rather than being collapsed into a uniform category, separating covariates measured on a real system, produced by a calibrated numerical simulator, or synthetically generated to resemble real data. SECOM, Steel, and Hydraulic are measurements from a real production line, a real fault inspection, and a real test rig, respectively. CBM is the output of a physics-based numerical simulator whose component models were tuned against real propulsion plants, not a source of directly measured sensor data. AI4I is, by its own documentation, a synthetic dataset designed to resemble real predictive-maintenance data rather than measurements from an operating machine. AI4I is included as the weakest and most conservative anchor of the five.

\begin{table}[!htbp]
    \centering
    \caption{The five host datasets and their covariate provenance.}
    \label{tab:synth-semi-datasets}
    \footnotesize
    \setlength{\tabcolsep}{6pt}
    \renewcommand{\arraystretch}{1.15}
    \begin{tabularx}{\textwidth}{@{}lYrrl@{}}
        \toprule
        Dataset & Domain & $N_{\mathrm{host}}$ & $F$ & Provenance \\
        \midrule
        AI4I~\cite{matzka2020ai4i}     & discrete manufacturing, predictive maintenance & 10{,}000 & 6  & synthetic \\
        CBM~\cite{coraddu2014cbm}      & naval gas-turbine propulsion, condition monitoring & 11{,}934 & 14 & calibrated simulator \\
        Hydraulic~\cite{helwig2015hydraulic} & hydraulic test rig, condition monitoring & 2{,}205 & 17 & real (test rig) \\
        SECOM~\cite{mccann2008secom}   & semiconductor manufacturing, process control & 1{,}567 & 30 & real (production line) \\
        Steel~\cite{buscema2010steel}  & steel-plate fault inspection & 1{,}941 & 25 & real (inspection) \\
        \bottomrule
    \end{tabularx}
\end{table}

\paragraph{Simulated treatment and outcome.} Given the covariates $X^{\mathrm{host}}\in\mathbb{R}^{N_{\mathrm{host}}\times F}$ of one host dataset, a treatment indicator $T$ and an outcome $Y$ are drawn from the data-generating process of the 2016 Atlantic Causal Inference Competition~\cite{dorie2019acic}.\footnote{Implementation: \url{https://github.com/vdorie/aciccomp}~\cite{dorie2016aciccomp} (R package \texttt{aciccomp2016}).} A treatment-assignment function $f_{\mathrm{trt}}$ and a response-surface function $f_{\mathrm{rsp}}$ are each drawn as a generalised additive function of $X^{\mathrm{host}}$, composed of randomly selected univariate and pairwise-interaction terms from a fixed library of nonlinear shapes (linear, polynomial, step, exponential). Their complexity is set by \texttt{model.trt} and \texttt{model.rsp}. The propensity score and treatment assignment follow
\begin{equation}
e(x^{\mathrm{host}}) = \operatorname{expit}\bigl(f_{\mathrm{trt}}(x^{\mathrm{host}})\bigr), \qquad T_i \sim \mathrm{Bernoulli}\bigl(e(x^{\mathrm{host}}_i)\bigr).
\label{eq:acic-propensity}
\end{equation}
The outcomes under the two treatment levels are evaluations of the same response function at the treatment indicator held fixed,
\begin{equation}
\mu_0(x^{\mathrm{host}}) = f_{\mathrm{rsp}}(x^{\mathrm{host}}, 0), \qquad \mu_1(x^{\mathrm{host}}) = f_{\mathrm{rsp}}(x^{\mathrm{host}}, 1),
\label{eq:acic-surfaces}
\end{equation}
with observed outcome $Y_i = \mu_{T_i}(x^{\mathrm{host}}_i) + \varepsilon_i$, $\varepsilon_i\sim\mathcal{N}(0,1)$, and true per-unit effect $\tau_i=\mu_1(x^{\mathrm{host}}_i)-\mu_0(x^{\mathrm{host}}_i)$. The sample average effect on the treated is calibrated toward a target drawn once per run, so the oracle effect is known exactly for every unit under a controlled realised average.

Five parameters set the estimation difficulty. \texttt{model.trt} and \texttt{model.rsp} fix the nonlinearity of $f_{\mathrm{trt}}$ and $f_{\mathrm{rsp}}$. \texttt{overlap.trt} optionally biases $f_{\mathrm{trt}}$ to force $e(x^{\mathrm{host}})\to 0$ on part of the covariate space, an engineered positivity violation. \texttt{alignment}$\,\in[0,1]$ sets the fraction of $f_{\mathrm{trt}}$'s terms reused inside $f_{\mathrm{rsp}}$ and thereby the strength of confounding. \texttt{te.hetero} determines whether $\tau(x^{\mathrm{host}})$ is constant across units (\emph{none}) or a genuine function of the covariates through added treatment-by-covariate interaction terms (\emph{med}, \emph{high}). For each (dataset, configuration) cell the generator is redrawn under a resample-until-valid rule, discarding draws that the original competition protocol would reject because they fail to instantiate the target overlap, alignment, and heterogeneity at the configured strength, until a valid draw is obtained.

\paragraph{Difficulty ladder.} Rather than varying one generator parameter in isolation, four rungs of escalating difficulty are defined by moving five parameters together, so that each rung is a qualitatively distinct estimation problem rather than a level on a single continuous difficulty scale (Table~\ref{tab:synth-semi-ladder}). \texttt{R1\_easy} combines full overlap, a linear treatment and response mechanism, and no heterogeneity. \texttt{R2\_mild} adds nonlinearity to the treatment mechanism. \texttt{R3\_hard} adds an engineered positivity violation, an exponential response surface, and reduces the alignment between assignment and response from $0.75$ to $0.25$. \texttt{R4\_severe} additionally switches on high treatment-effect heterogeneity. \texttt{te.hetero} is \emph{none} on R1--R3, so $\tau(x)$ is a single constant on those three rungs. Only R4 exercises genuine heterogeneous-effect estimation, a caveat stated explicitly wherever this ladder is used for evaluation. The baseline treatment probability \texttt{root.trt}$=0.5$ is held fixed throughout.

\begin{table}[!htbp]
    \centering
    \caption[The difficulty ladder.]{The difficulty ladder. Each rung fixes a joint setting of the generator parameters of Equations~\eqref{eq:acic-propensity} and~\eqref{eq:acic-surfaces}.}
    \label{tab:synth-semi-ladder}
    \footnotesize
    \setlength{\tabcolsep}{6pt}
    \renewcommand{\arraystretch}{1.15}
    \begin{tabular}{@{}lccccc@{}}
        \toprule
        Rung & \texttt{model.trt} & \texttt{overlap.trt} & \texttt{model.rsp} & \texttt{alignment} & \texttt{te.hetero} \\
        \midrule
        R1\_easy   & linear     & full     & linear      & 0.75 & none \\
        R2\_mild   & polynomial & full     & linear      & 0.75 & none \\
        R3\_hard   & polynomial & one-term & exponential & 0.25 & none \\
        R4\_severe & polynomial & one-term & exponential & 0.25 & high \\
        \bottomrule
    \end{tabular}
\end{table}

\chapter{PriMa-Causa Model}
\label{chap:prima-causa-model}

\begin{jointwork}
This chapter incorporates my prior peer-reviewed publication~\cite{saretzky2026integratingcausalfoundationmodel}. Figure~\ref{fig:ch6_structure} outlines the structure of this chapter, which develops the PriMa-Causa model, the pre-trained causal foundation model that fills the estimation block of the Causal Reasoning Layer. The chapter objective and the model interface are stated first (Section~\ref{sec:ch6-objective}), followed by the derivation of the model class from the meta-requirements (Section~\ref{sec:ch6-model-class}) and by the causal inductive bias and its attention backbone (Section~\ref{sec:ch6-backbone}). The graph-conditioned interventional estimator is developed in Section~\ref{sec:ch6-arch-effect-estimator}, its decision-mode coverage is collected in Section~\ref{sec:ch6-mode-coverage}, and the scope of the instantiation is discussed in Section~\ref{sec:ch6-discussion}. The chapter yields one artefact: the pre-trained model weights.
\end{jointwork}

\vspace{\baselineskip}

\begin{center}
    \includegraphics[width=\textwidth]{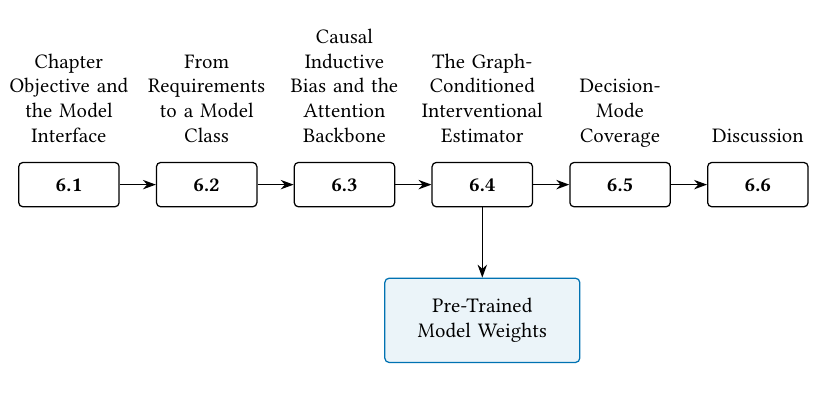}%
    \captionof{figure}{Structure of Chapter~\ref{chap:prima-causa-model}.}
    \label{fig:ch6_structure}
\end{center}

\newpage

\section{Chapter Objective and the Model Interface}
\label{sec:ch6-objective}

The causal reasoning layer of Chapter~\ref{chap:prima-causa} leaves its estimation block as a defined slot. For each selected cell of the digital-intervention tensor, the block returns an estimated interventional distribution $\widehat{P}(Y\mid \dop(T{=}t),\,X{=}x)$ under the role map $r_{\mathcal{K}}$ and the identification constraints $\mathcal{I}_{\mathcal{K}}$ (Figure~\ref{fig:ch4-estimation}, Table~\ref{tab:ch4-notation}). This chapter fills the slot. It specifies the model class of the estimator, derives the class from the meta-requirements of Chapter~\ref{chap:prima-causa}, and develops the architecture that instantiates the class. The interventional estimand and the $\dop$-operator are not re-derived here. They are established in Chapter~\ref{chap: Background} (Section~\ref{sec:bg-causal}). Identifiability likewise lies outside the scope of the network. It is an upstream gate enforced by design principle DP8 (Subsection~\ref{subsec:ch4-crl-assumptions}). Every architectural choice in the remainder of this chapter is traced to a meta-requirement or a design principle of Chapter~\ref{chap:prima-causa}, so that the resulting model is the operationalisation of those requirements at the level of a trainable estimator. In answering research question RQ2 introduced in Chapter~\ref{chap:introduction}, this chapter, together with the data foundation of Chapter~\ref{chap:synthetic-data}, constitutes contribution~C2, addressing research gaps RG2 and RG3 through design objectives DO2 and DO3.

\section{From Requirements to a Model Class}
\label{sec:ch6-model-class}

\begin{figure}[!htbp]
    \centering
    \includegraphics[width=\textwidth]{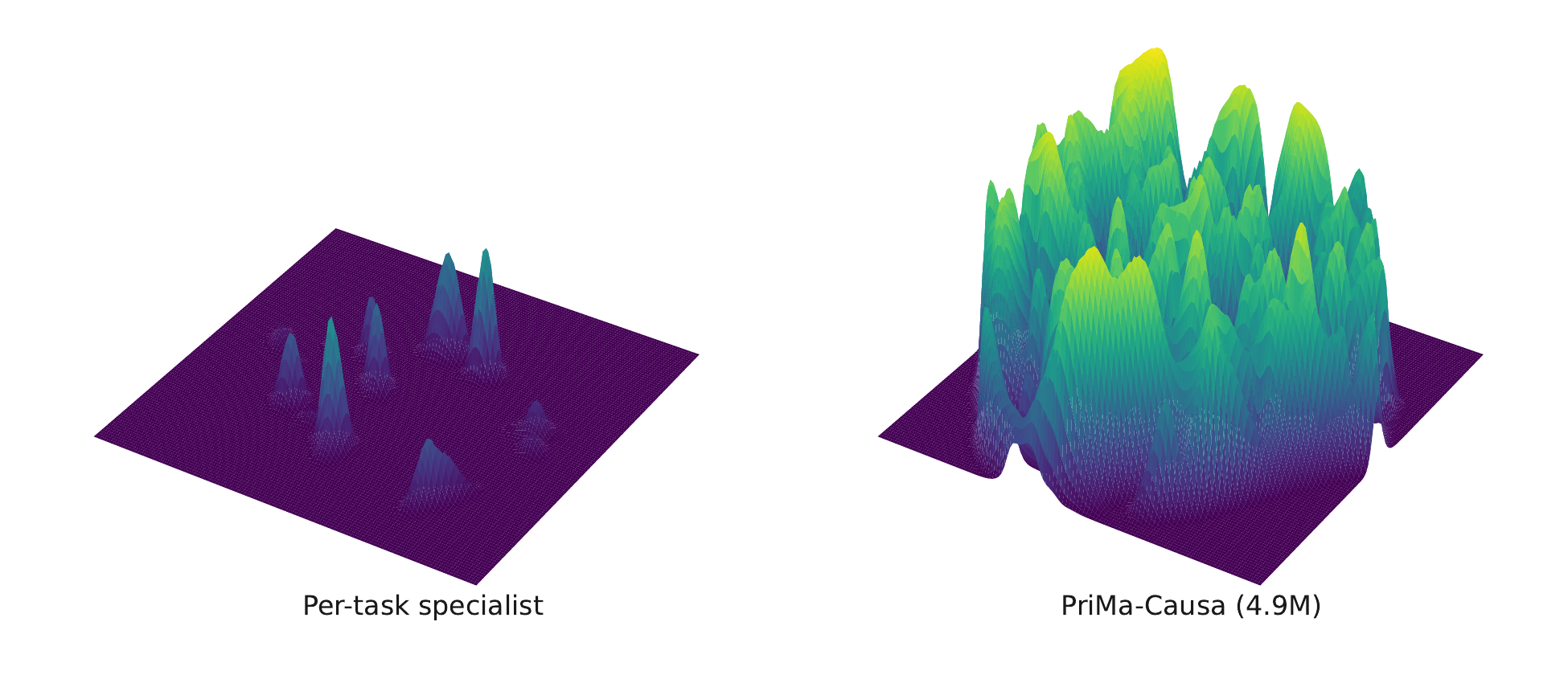}
    \caption[Amortised competence across the task prior.]{Illustration of per-task specialist competence (left) and amortised competence across the task prior (right).}
    \label{fig:ch6-amortisation}
\end{figure}

\paragraph{An amortised, pre-trained causal foundation model.}
The operating regime targeted by the framework is the data-sparse case, in which the relevant action-outcome relation is not encoded at decision time (Subsection~\ref{subsec:decision-time-knowledge-horizon}). Two consequences follow. First, no estimator can be fitted and validated per case within the production-stop window, so the model is prepared before the incident and answers each case without per-dataset retraining, as required by meta-requirement MR6 (Table~\ref{tab:ch4-meta-requirements}). Second, the interventional target is never recorded in plant logs, because historical data document the regime under which the line operated rather than the outcome of an intervention, so a supervised interventional estimator cannot be trained on field data. Both consequences are met by amortisation. The contrast between an estimator fitted per task and a single amortised estimator is illustrated in Figure~\ref{fig:ch6-amortisation}. A single model is trained once over a prior of tasks and returns a predictive distribution for each query in one forward pass with fixed weights, the amortised in-context paradigm of Section~\ref{sec:bg-pfn}. This paradigm is not specific to one architecture. A pre-trained causal model in this sense subsumes both the effect estimators that predict an interventional outcome directly and the structural-causal-model learners that reconstruct the mechanisms and answer queries from them. Of these, the direct effect estimator is the realisation instantiated in this dissertation. The interventional supervision that production line data cannot provide is generated synthetically, which is the purpose of the generator of Chapter~\ref{chap:synthetic-data}, where the structural causal models are known by construction and interventional labels are drawn under the truncated factorisation induced by the $\dop$-operator (Equation~\ref{eq:scm-truncated}).

\section{Causal Inductive Bias and the Attention Backbone}
\label{sec:ch6-backbone}

\paragraph{Causal inductive bias.}
An estimator trained on associations alone returns $P(Y\mid X)$. This is not the quantity that a prescription requires, and it loses validity once an action is taken, because an intervention changes the distribution under which the system operates (Section~\ref{sec:ch4-bridge-problem}). The interventional answer is obtained only if the model is biased toward it, and this causal inductive bias is the realisation, at the level of the analytical layer, of the research gap RG1 (Chapter~\ref{chap:sota}). In an amortised, pre-trained model the bias has two carriers. The first is the pre-training prior, which is itself a primary determinant of the inductive bias of an amortised estimator (Section~\ref{sec:bg-pfn}), so that a prior of structural causal models with interventional labels biases the learned map toward the interventional rather than the observational regime. The second carrier is the prior knowledge supplied with a case.

\paragraph{Prior knowledge as conditioning channels.}
Prior knowledge here denotes the information that conditions the estimate beyond the raw query and the observational data. The observational data are not restricted to tabular records and may also take the form of time series or images. Prior knowledge comprises the structural prior given by the causal graph in full or partial form, and representations derived from it, for example spectral descriptors of the graph such as the eigenvalues of its Laplacian. Each form is a channel through which structural information enters the model, and the architecture determines how it is rendered machine-usable, in line with the treatment of prior knowledge as machine-usable constraints across the structural-completeness levels of Subsection~\ref{subsec:knowledge-availability}. The constraint map $\Phi_{\mathcal{K}}(x)$ of Chapter~\ref{chap:prima-causa} (Subsection~\ref{subsec:knowledge-availability}) is the object that delivers prior knowledge to the estimator as inductive bias, realising meta-requirement MR5 and design principle DP4 (Table~\ref{tab:ch4-meta-requirements}, Table~\ref{tab:ch4-design-principles}). The architecture developed below consumes prior knowledge in this sense and not only the observational context. Its specific channel, the graph-conditioned attention, is described in Section~\ref{sec:ch6-arch-effect-estimator}.

\paragraph{The attention backbone.}
The backbone is a transformer rather than a recurrent or state-space network, for two structural reasons. A production-line dataset is a set of exchangeable observations over a set of variables. Attention is permutation-equivariant over a set, whereas a recurrent backbone imposes an order on the observations that carries no meaning. More decisively, the causal graph is a relation between variables, so it maps directly onto the attention matrix. Structural knowledge is injected by masking or biasing the attention along the variable axis, which is the single point at which prior structure becomes a usable inductive bias for the estimator. The transformer is also the established backbone for prior-data fitted networks (Section~\ref{sec:bg-pfn}) and for the related amortised causal estimators discussed below.

\paragraph{Related amortised causal estimators.}
The model class is shared by a small set of concurrent estimators. Do-PFN estimates interventional distributions in context from observational data alone, without consuming a causal graph~\cite{robertson2025dopfn}. CausalFM pre-trains prior-data fitted networks for treatment-effect estimation on a prior over structural causal models whose mechanisms are randomly parameterised multilayer perceptrons, likewise without graph conditioning~\cite{ma2025causalfm}. The first instantiation of the estimator of this chapter also operated without graph input. It was pre-trained on the production-line prior of Chapter~\ref{chap:synthetic-data} and published in~\cite{saretzky2026integratingcausalfoundationmodel}, where the use of partially known causal structure was identified as future work. Conditioning an amortised effect estimator on partially known graphs is proposed in the concurrent preprint of Reuter et al.~\cite{reuter2026usewhatyouknow}. The instantiation of this chapter adopts this conditioning mechanism, a three-valued ancestor encoding that enters the variable-axis attention as a learned soft bias (Section~\ref{sec:ch6-arch-effect-estimator}). It differs from the estimators above in two respects, the production-line prior and the constraint map of Chapter~\ref{chap:prima-causa} that this channel realises. For optimisation, the Muon optimiser is applied to the two-dimensional weight matrices of the transformer blocks and a schedule-free AdamW optimiser to the remaining parameters.

\paragraph{One class, one learning target.}
The architecture developed in the following section realises this pattern, a dual-axis attention that alternates over the observation axis and the variable axis. Its learning target is the total interventional effect, since every query that the Causal Reasoning Layer submits to the estimation block concerns one treatment-outcome pair per tensor cell (Section~\ref{sec:ch4-causal-reasoning-layer}). The mode coverage that follows is collected in Section~\ref{sec:ch6-mode-coverage}.

\section{The Graph-Conditioned Interventional Estimator}
\label{sec:ch6-arch-effect-estimator}

The PriMa-Causa model instantiates the model class of Section~\ref{sec:ch6-model-class} as a direct estimator of the interventional effect. Its learning target is fixed per case: for a specified treatment-outcome pair $(T,Y)$ it maps the observational context, a query value $\dop(T{=}t)$, and the available graph onto the interventional distribution $\widehat{P}(Y\mid \dop(T{=}t),\,X{=}x)$ in a single forward pass. It does not reconstruct the structural causal model and carries no mechanism for a counterfactual or a Mode~4 structural query. Its scope is therefore the modes that ask for a total effect, namely Mode~1, Mode~2, and the forward-query route of Mode~3. The architecture is a prior-data fitted network in the sense of Section~\ref{sec:bg-pfn}, realised as a graph-conditioned interventional estimator. Its two points of departure from the tabular prior-data fitted network of Section~\ref{sec:bg-pfn} are the prior with interventional labels, supplied by the generator of Chapter~\ref{chap:synthetic-data}, and the graph-conditioned attention derived below.

\subsection{Architecture internals}
\label{sec:ch6-eff-internals}

One case reaches the estimator as a single table together with a graph code and a query block. The observational context forms $N$ rows over the intervention target $T$, the outcome $Y$, and the covariates $X_1,\dots,X_F$. Each query forms one additional row. Its treatment column carries the intervention level of $\dop(T{=}t)$, its covariate columns carry the decision-time context, and its outcome entry is unobserved and is replaced by a neutral placeholder derived from the context. All columns are standardised by the statistics of the context rows, so the network operates on a unit-free representation of the case and no measurement scale of the line enters the weights. The input layout has a fixed width of $35$ variables, namely the intervention target, the outcome, and up to $33$ covariates, and unused columns are zero-padded. Graphs wider than this layout cannot be evaluated by the trained instantiation. The graph code is the three-valued ancestor encoding described below.

The backbone is a two-way attention over the case table, following the cell-wise attention of the second-generation tabular prior-data fitted network~\cite{hollmann2025tabpfn2}. Each block alternates between an attention across the variable axis within an observation and an attention across the observation axis within a variable. On the observation axis the attention is asymmetric, following the prior-data fitted network convention~\cite{muller2022transformers,hollmann2023tabpfn}: the observational records self-attend, and each interventional query cross-attends to the observational set rather than to the other queries. This asymmetry keeps the answers to different queries independent of one another.

\begin{figure}[!htbp]
    \centering
    \includegraphics[width=\textwidth]{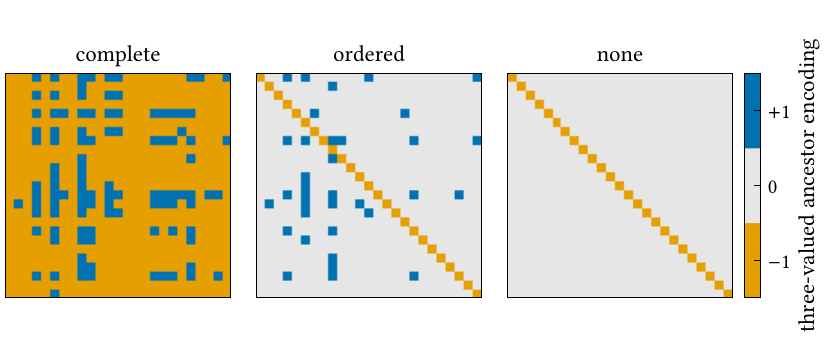}
    \caption[The three-valued ancestor encoding at three structural-completeness levels.]{The three-valued ancestor encoding of one sampled \acs{CPG} at the three structural-completeness levels.}
    \label{fig:ch6-ancestor-encoding}
\end{figure}

The causal graph is injected on the variable axis, because the graph is a statement about variables rather than about observations. The ancestor relation between each ordered pair of variables, the existence of a directed path, is encoded with the three values $\{-1,0,1\}$, in which $1$ denotes a known ancestor relation, $-1$ a known absence of such a relation, and $0$ an unknown relation. The three-valued encoding spans the three structural-completeness levels of Subsection~\ref{subsec:knowledge-availability} without a level-specific branch: the \emph{none} level is all-zero off the diagonal, the \emph{ordered} level is a sparse, order-consistent fill, and the \emph{complete} level is a fully specified graph (Figure~\ref{fig:ch6-ancestor-encoding}). In that figure each panel is a $25\times25$ matrix with one row and column per variable of the sampled graph, drawn in an arbitrary variable order. The encoded graph enters the attention computation as a learned per-head additive bias on the pre-softmax logits, following~\cite{reuter2026usewhatyouknow}. One bias parameter is applied where a relation is known, a second where a relation is known to be absent, and none where it is unknown. A partially trusted graph is thereby treated as a prior over the attention pattern rather than as ground truth, in line with the data-sparse operating regime targeted by meta-requirement MR6. The consumption of the constraint map required by meta-requirement MR5 and design principle DP4 is realised at this single point of the architecture. A dedicated treatment token carries the value of $T$ together with a role embedding that distinguishes the observational regime, in which $T$ is recorded under its own mechanism, from the interventional regime, in which $T$ is set by the $\dop$-operator. The role embedding binds each forward pass to the queried treatment-outcome pair and respects meta-requirement MR3. The output head emits a full predictive density rather than a point estimate, because decision-relevant uncertainty is required downstream by meta-requirement MR11 and design principle DP5. The head is a bar distribution, a fixed categorical distribution over a set of discretised outcome intervals with Gaussian half-tails~\cite{muller2022transformers}, which admits heavy-tailed and multimodal interventional outcomes without a parametric assumption on the outcome family and returns the density in one forward pass. The interval boundaries are fixed quantiles of the standardised outcome, fitted once over the pre-training corpus and stored with the model. The returned density and its mean are mapped back to the outcome scale of the current case through the context statistics.

\subsection{Pre-training and in-context learning}
\label{sec:ch6-eff-pretraining}

\begin{figure}[p]
    \centering
    \includegraphics[angle=90, height=0.9\textheight]{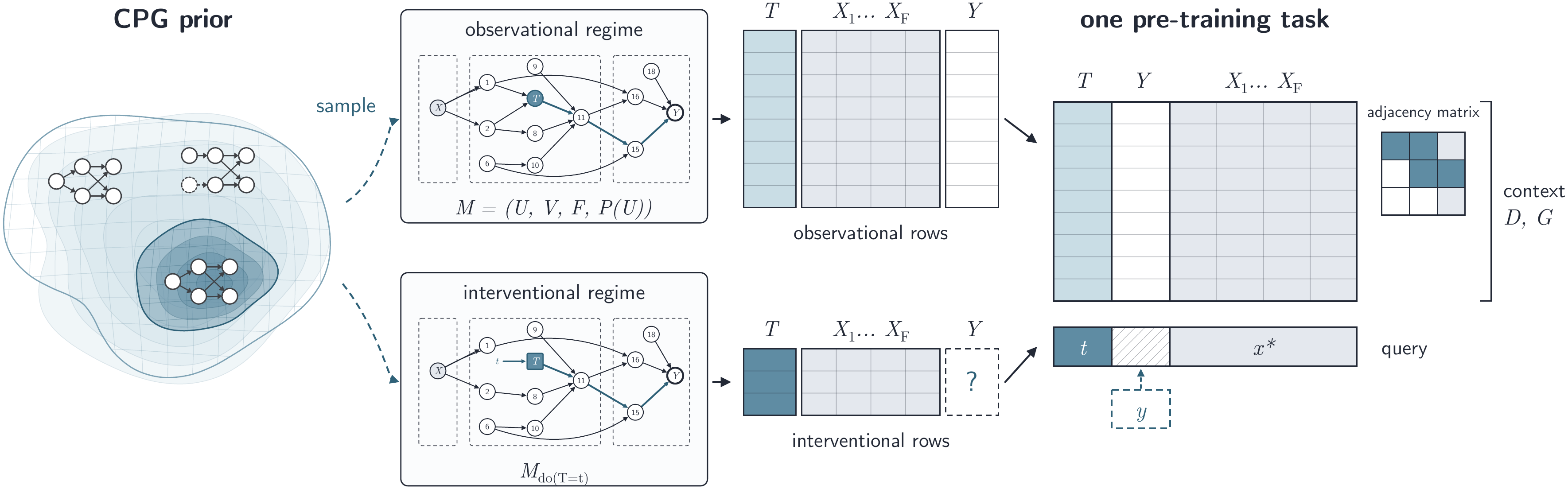}
    \caption[Assembly of one pre-training task.]{Assembly of one pre-training task: a structural causal model sampled from the \acs{CPG} prior supplies the observational context and, under $\dop(T{=}t)$, the interventional label $y$ of the query.}
    \label{fig:ch6-pretraining-task}
\end{figure}

The network is trained once, over many synthetic datasets drawn from the causal prior, and it thereafter answers each query with its weights frozen. The prior is the Causal Process Graph generator of Chapter~\ref{chap:synthetic-data} (Section~\ref{sec:synth-cpg}, Section~\ref{sec:synth-pipeline}). A structural causal model is sampled together with its graph, observational records are drawn by an ancestral forward pass, and the interventional label is obtained under the truncated factorisation induced by the $\dop$-operator (Equation~\ref{eq:scm-truncated}). The $\dop$-semantics are established in Chapter~\ref{chap: Background} (Equation~\ref{eq:bg-submodel}) and are not re-derived here. The essential property is that the supervision is interventional by construction, since the label is drawn under graph surgery rather than under the observational regime. This is the first carrier of the causal inductive bias identified in Section~\ref{sec:ch6-backbone}, specialised to the total-effect target.
The interventional query admits binary and continuous intervention targets. A binary target is queried as the contrast between $\dop(T{=}0)$ and $\dop(T{=}1)$, and a continuous target admits dose-response queries at admissible levels (Section~\ref{sec:synth-pipeline}). In the pre-training corpus of this instantiation the treatment is a binary action drawn under a context-dependent propensity, and each query row carries one of the two levels together with the corresponding interventional outcome. The assembly of one pre-training task from these components is shown in Figure~\ref{fig:ch6-pretraining-task}.

The graph-conditioned attention is trained against the full spectrum of partial-knowledge conditions rather than against a single level of graph disclosure. The prior loads each sampled directed acyclic graph at a disclosure level drawn uniformly at random per task: a random subset of its known ancestor relations and known non-ancestor relations is replaced by the unknown code, and the relations still implied by the disclosed subset are restored by transitive closure before the graph reaches the network. Each pre-training task therefore exposes the model to a different point on the range from the complete graph down to the empty one, and this range is spanned over the course of training rather than within a single task. This is the mechanism by which one artefact consumes the second carrier of Section~\ref{sec:ch6-backbone}, the prior knowledge supplied with a case, across all three structural-completeness levels of Subsection~\ref{subsec:knowledge-availability}.

Each training task supplies the observational context, the interventional query, and the interventional label. With $q_\theta$ the density returned by the bar head, the objective is the prior-data negative log-likelihood of Equation~\ref{eq:bg-pfn-loss}, specialised to the held-out interventional outcome under the bar distribution. At its optimum the network output approximates the posterior predictive distribution induced by the causal prior, so that a forward pass on a previously unseen case performs amortised inference rather than per-case optimisation. This is the in-context property demanded by meta-requirement MR6, and it is the reason no estimator is fitted within the production-stop window (Subsection~\ref{subsec:decision-time-knowledge-horizon}). The degree to which a model trained on the generic prior transfers to a concrete line is tested empirically in Chapter~\ref{chap:empirical-evaluation}, against the interventional-effect metric defined in that chapter.

The estimator is instantiated once. The network comprises approximately $4.9$~million parameters, structured into five dual-axis blocks with an embedding width of $256$, four attention heads, and a $1000$-interval bar-distribution head. Pre-training is performed in a single pass over one million tasks drawn from the synthetic prior. Each task contains between $3$ and $35$ variables with $500$ sample rows, half of which serve as the pre-intervention observational context. The corpus mixes Causal Process Graph and Erd\H{o}s--R\'enyi topologies in a ratio of three to one (Section~\ref{sec:synth-cpg}), so the estimator is not conditioned exclusively on production-line topology. After this run the weights are frozen. The same checkpoint is used unchanged in every experiment of Chapter~\ref{chap:empirical-evaluation} and at every structural-knowledge level. The implementation, the training configuration, and the pre-trained weights are released as open-source software.\footnote{\url{https://github.com/FelixSaretzky/PriMaCausa}}

\subsection{Inference}
\label{sec:ch6-eff-inference}

At decision time a forward pass consumes the observational support set, the pre-intervention context $X{=}x$, the query value $\dop(T{=}t)$, and the available graph, and it returns the full bar-distribution density over the interventional outcome. No fitting or retraining takes place at decision time, consistent with the decision-time horizon of Subsection~\ref{subsec:decision-time-knowledge-horizon}. The estimand is the conditional interventional distribution $P(Y\mid \dop(T{=}t),\,X{=}x)$, where $X$ is the pre-intervention decision-time context of Chapter~\ref{chap:prima-causa} (Table~\ref{tab:ch4-notation}). Conditioning on the pre-intervention context rather than on post-intervention observations preserves the estimate as a total effect on $Y$. Under meta-requirement MR7 and design principle DP8 no descendant of $T$ enters the conditioning set, so the effect mediated through such descendants is not conditioned away. The marginal estimand $P(Y\mid \dop(T{=}t))$ is recovered as the special case in which the query covariates are not supplied. The mode-by-mode coverage that follows from this forward pass is collected in Section~\ref{sec:ch6-mode-coverage}. The estimates are governed by the upstream identifiability gate of design principle DP8 and meta-requirement MR7 rather than by the network itself. A tensor cell that fails the admissibility conditions of Subsection~\ref{subsec:ch4-crl-assumptions} is not assigned a point estimate. The gate prescribes bounds, a sensitivity interval, or an escalation instead. The conditions are not re-enumerated here, since they are governed by Chapter~\ref{chap:prima-causa}. For admissible cells, the output fills the corresponding entry $\widehat{\mathcal{D}}_c^{(M)}[\ell,j,k]$ of the estimated digital-intervention tensor and is passed to the robust recommendation synthesis block (Figure~\ref{fig:ch4-robust-recommendation-synthesis}).

\section{Decision-Mode Coverage}
\label{sec:ch6-mode-coverage}

\begin{figure}[!htbp]
    \centering
    \includegraphics[width=\textwidth]{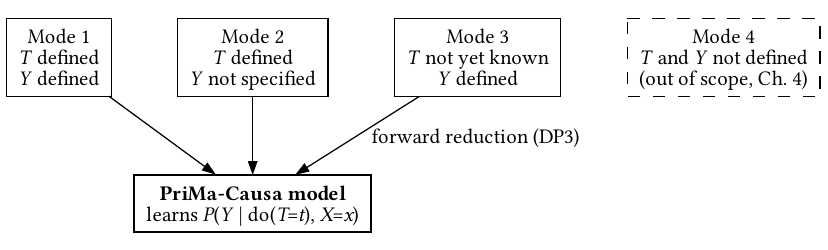}
    \caption{Decision-mode coverage of the PriMa-Causa model.}
    \label{fig:ch6-mode-coverage}
\end{figure}

The coverage of the decision modes of Chapter~\ref{chap:prima-causa} follows directly from the forward pass (Figure~\ref{fig:ch6-mode-coverage}). Mode~1 is served by the conditional interventional distribution, from which the contrast $\tau(x)$ of Equation~\ref{eq:cate} is read off as a difference of two interventional means. Mode~2 is served by repeating the forward pass per candidate outcome. Mode~3 is served through the forward reduction of design principle DP3, as a set of forward interventional queries over the diagnosis-restricted candidate-action set $\mathcal{A}_{\mathcal{K}}(x)$ scored by $S_j(x)$. Mode~4 is not served, since the graph is an input of the estimator and not an object of estimation, consistent with the scoping of Chapter~\ref{chap:prima-causa}. Along the knowledge axis, the three-valued ancestor encoding serves the none, ordered, and complete levels of Subsection~\ref{subsec:knowledge-availability} without a level-specific model-selection step. The configuration without a supplied graph and the configuration with the complete graph are compared in Chapter~\ref{chap:empirical-evaluation}. The evaluation of Chapter~\ref{chap:empirical-evaluation} exercises Mode~1 throughout and Mode~3 in the root-cause-analysis experiment. Mode~2 is specified but not empirically tested.

\section{Discussion}
\label{sec:ch6-discussion}

Table~\ref{tab:ch6-traceability} summarises how the architectural elements of this chapter realise the meta-requirements and design principles of Chapter~\ref{chap:prima-causa}.

\begin{table}[!htbp]
    \centering
    \caption[Traceability of the model architecture.]{Traceability of the model architecture to the meta-requirements and design principles of Chapter~\ref{chap:prima-causa}.}
    \label{tab:ch6-traceability}
    \footnotesize
    \setlength{\tabcolsep}{4pt}
    \renewcommand{\arraystretch}{1.15}
    \begin{tabularx}{\textwidth}{@{}p{0.18\textwidth}Y@{}}
        \toprule
        \textbf{Source} & \textbf{Architectural realisation} \\
        \midrule
        MR3 & The treatment token and its role embedding bind each forward pass to the queried treatment-outcome pair $(T,Y)$ and separate the observational from the interventional regime. \\
        MR4 & The supervision is drawn under the truncated factorisation, so the output is an interventional distribution rather than an associational score. \\
        MR5, DP4 & The three-valued ancestor encoding is consumed as a learned attention bias, the single entry point of the constraint map $\Phi_{\mathcal{K}}(x)$. \\
        MR6 & Amortised pre-training and frozen weights answer a case in one forward pass without per-case fitting. \\
        MR7, DP8 & Identifiability and support are gated upstream per tensor cell, and only admissible cells are estimated. \\
        MR7, DP8 & Conditioning is restricted to the pre-intervention context, so no descendant of $T$ enters. \\
        MR11, DP5 & The bar-distribution head returns a full predictive density. \\
        MR13 & Estimates fill the auditable digital-intervention tensor, validated in stages beginning with the data foundation of Chapter~\ref{chap:synthetic-data}. \\
        DP1, DP3 & Diagnostic evidence enters only through the candidate-action set, and reverse questions are answered as forward interventional queries. \\
        \bottomrule
    \end{tabularx}
\end{table}

The scope of the instantiation is bounded by four properties. The input layout caps a case at $35$ variables, so wider lines must be reduced to a decision-relevant variable subset before the estimator is queried. The pre-training corpus draws binary treatment contrasts over continuous process signals, so continuous dose-response queries and mixed-type signals belong to the model class and the generator, and are not validated for this checkpoint. The graph is an input and not an output, so structural discovery remains upstream of the estimator. Identifiability and empirical support are decided upstream by the gate of design principle DP8, which the estimator informs through its predictive density but does not replace. Whether the frozen estimator retains its accuracy under the data conditions of operating production lines, including weak overlap, sparse context, and covariates outside the pre-training distribution, is an empirical question. It is answered against controllable ground truth in Chapter~\ref{chap:empirical-evaluation}.

		    \chapter{Industrial Application}
\label{chap:usecase}

\begin{jointwork}
The demonstrations of this chapter were developed within the industrial research project PANIC on the production line of an industrial partner. The troubleshooting application of Section~\ref{sec:usecase-app} was co-developed with Dropslab Technologies. The integration architecture, the causal reasoning workflow, and the operator console are the author's contribution. Figure~\ref{fig:ch7_structure} outlines the structure of this chapter, which demonstrates the PriMa-Causa framework in its industrial target environment and thereby realises the production-line application announced in Chapter~\ref{chap:introduction}. Section~\ref{sec:usecase-architecture} describes the deployment environment and the integration architecture on the production line. Section~\ref{sec:usecase-console} traces a single troubleshooting case through the operator console, and Section~\ref{sec:usecase-app} presents the operator-facing troubleshooting application on mobile and head-mounted devices. The chapter yields one artefact: the line-deployed software demonstrator. The controlled evaluation of the estimator under known ground truth is the subject of Chapter~\ref{chap:empirical-evaluation}.
\end{jointwork}

\vspace{\baselineskip}

\begin{center}
    \includegraphics[width=\textwidth]{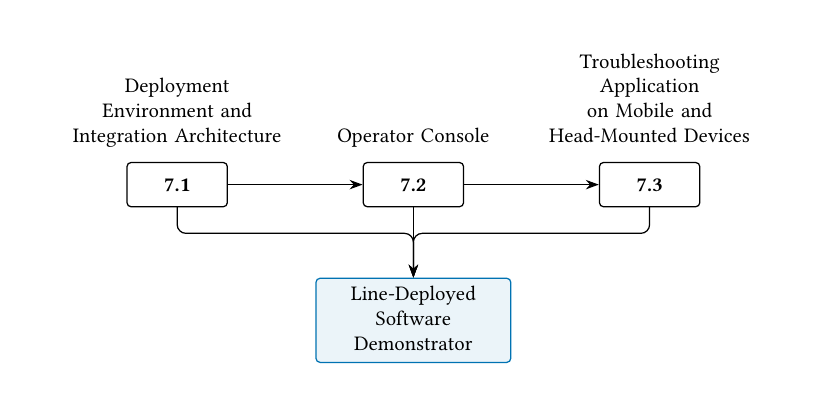}%
    \captionof{figure}{Structure of Chapter~\ref{chap:usecase}.}
    \label{fig:ch7_structure}
\end{center}

\newpage

\paragraph{Technology readiness.} The maturity of the resulting artefacts is assessed against the Technology Readiness Level (TRL) scale \cite{EuropeanCommission2014TRL}. The causal-prescriptive core reaches approximately TRL~4. It is validated under laboratory conditions on synthetic and semi-synthetic data with real production covariates (Chapters~\ref{chap:synthetic-data} and~\ref{chap:empirical-evaluation}). The operator-facing troubleshooting application reaches TRL~5 to~6, as it is demonstrated on an operating production line within the industrial project PANIC (Section~\ref{sec:usecase-app}). The validation of the closed causal recommendation loop in live production operation remains open. This validation is deferred to future work in Chapter~\ref{chap:conclusions}.

\section{Deployment Environment and Integration Architecture}
\label{sec:usecase-architecture}

\begin{figure}[!htbp]
    \centering
    \includegraphics{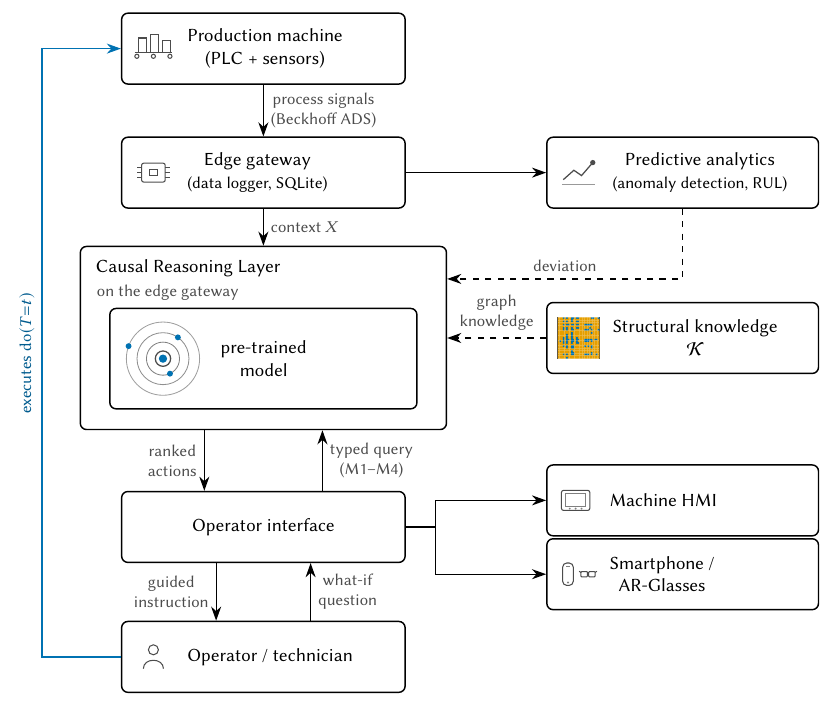}
    \caption{Integration architecture of the PriMa-Causa framework.}
    \label{fig:usecase-architecture}
\end{figure}

The demonstration environment is a high-speed production line for fast-moving consumer goods, operated by an industrial partner within the research project PANIC. The demonstrated use cases originate from an adhesive-application and sealing station of this line and thereby revisit the station family from which the motivating example of Section~\ref{sec:intro-gentle-intro} was drawn. Figure~\ref{fig:usecase-architecture} shows the integration of the framework and its pre-trained model into this environment. The architecture comprises three tiers, namely the machine, the edge gateway, and the operator interfaces. Two paths connect these tiers. A data path carries the machine data to the causal reasoning layer, and an interaction path connects the operator to the framework. The intervention selected by the operator returns to the machine and closes this loop.

The line is connected through the Beckhoff ADS protocol, over which the process signals are sampled from the programmable logic controller and logged locally. The gateway is realised on an NVIDIA Jetson platform and hosts the causal reasoning layer, including the local inference of the pre-trained model. Since no estimator can be fitted and validated per case within the operational window of a production stop (Subsection~\ref{subsec:decision-time-knowledge-horizon}), recommendations are computed at the line by the amortised pre-trained model without a per-case fit, consistent with the data-sparse operating conditions of meta-requirement MR6. The local inference on the gateway avoids any dependency on external infrastructure. The operator interfaces issue the typed queries of the four decision modes and receive their results (M1--M4 in Figure~\ref{fig:usecase-architecture}). Their output is presented on the end devices, namely the machine HMI and the mobile and head-mounted devices of Section~\ref{sec:usecase-app}.

\begin{figure}[!htbp]
    \centering
    \includegraphics{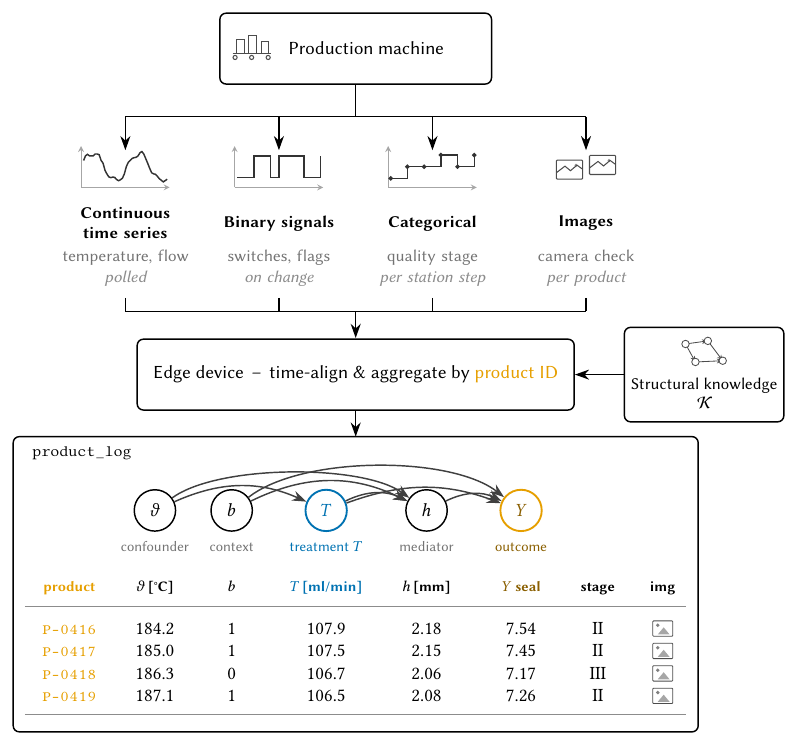}
    \caption{Data input of the PriMa-Causa framework.}
    \label{fig:usecase-data-types}
\end{figure}

Figure~\ref{fig:usecase-data-types} details the data path. The station emits heterogeneous data, namely continuous time series, binary and categorical signals, and images of the product. On the edge device, these streams are time-aligned and aggregated by a common product identifier into one record per produced unit. The operationalised prior knowledge $\mathcal{K}$ enters as an external input alongside the recorded data (Subsection~\ref{subsec:knowledge-availability}). The supplied graph identifies the structural position of each measured column for the motivating station (Figure~\ref{fig:usecase-data-types}). In this setup, the process temperature serves as a confounder, while the machine status defines the background context. The adhesive-flow setpoint represents the treatment $T$, the adhesive height acts as a mediator, and the seal quality constitutes the primary outcome $Y$. The role map $r_{\mathcal{K}}$ translates these positions into the variable roles of Chapter~\ref{chap:prima-causa}: the setpoint is the admissible intervention target, the temperature and the machine status enter the pre-intervention context, and the adhesive height is a post-intervention descendant that is not an admissible adjustment variable. The quality stage and the product image are labelled as excluded for the current case.

\begin{figure}[!htbp]
    \centering
    \includegraphics{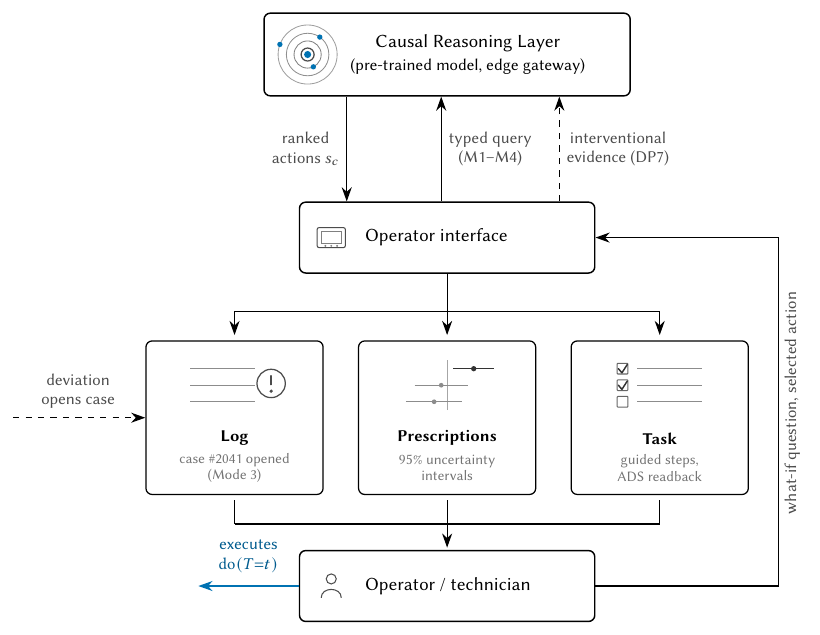}
    \caption{Interaction between the operator and the framework at the operator interface.}
    \label{fig:usecase-interaction}
\end{figure}

The interaction path is shown in Figure~\ref{fig:usecase-interaction}. A diagnostic case is entered in one of two ways. A deviation reported by the predictive-analytics layer opens a case automatically, or the operator issues a what-if question that is translated into a typed query of the four decision modes (M1--M4). In both paths the causal reasoning layer evaluates the candidate actions with the pre-trained model and returns the ranked actions $s_c$ together with their uncertainty intervals. The operator interface presents this exchange in three views. The log view collects the predictive triggers and the opened cases, the prescription view presents the ranked actions with their $95\%$ intervals, and the task view guides the execution of the selected intervention and records the observed outcome as interventional evidence (DP7). The executed intervention closes the loop of Figure~\ref{fig:usecase-architecture}. The realisation of this interface, the operator console, is traced through a complete troubleshooting case in the following section.

\section{Operator Console}
\label{sec:usecase-console}

\begin{figure}[!htbp]
    \centering
    \includegraphics[width=\textwidth]{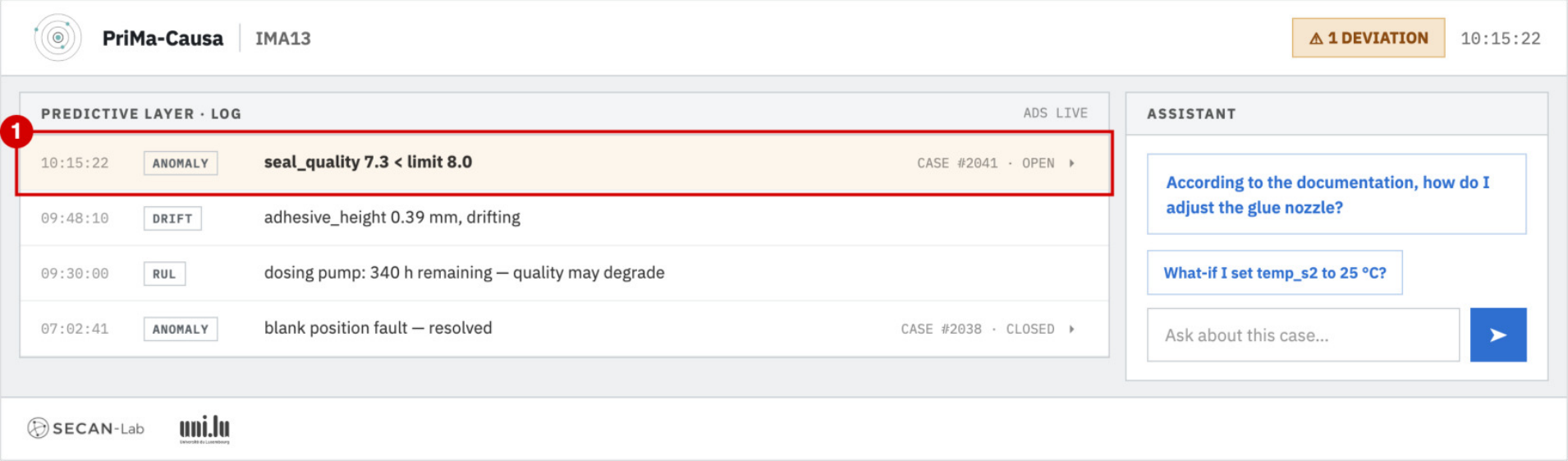}
    \caption{Log view of the operator console.}
    \label{fig:usecase-console-log}
\end{figure}

\begin{figure}[!htbp]
    \centering
    \includegraphics[angle=90, width=\textwidth, height=0.92\textheight, keepaspectratio]{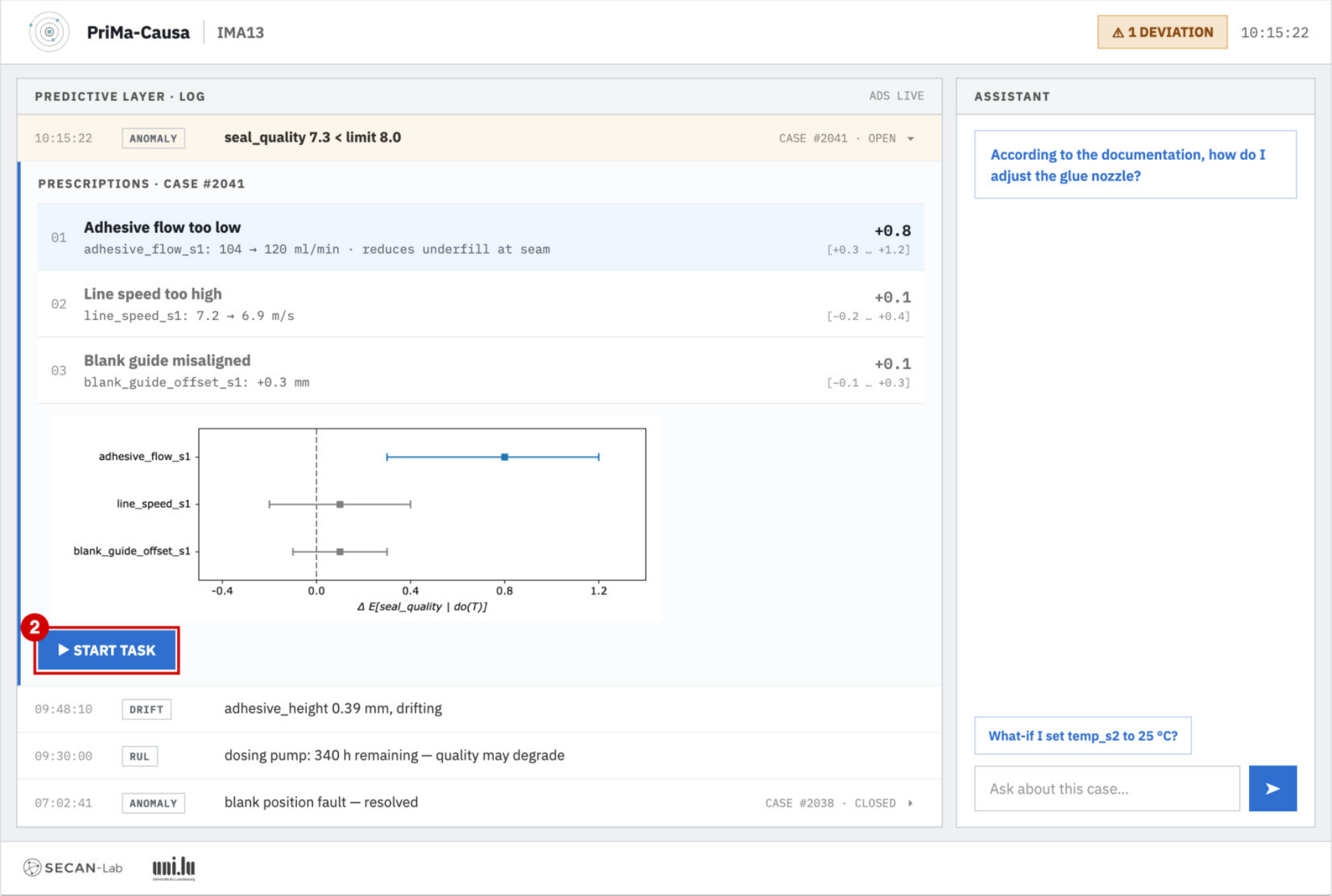}
    \caption{Prescription view of the operator console.}
    \label{fig:usecase-console-prescriptions}
\end{figure}

\begin{figure}[!htbp]
    \centering
    \includegraphics[width=\textwidth]{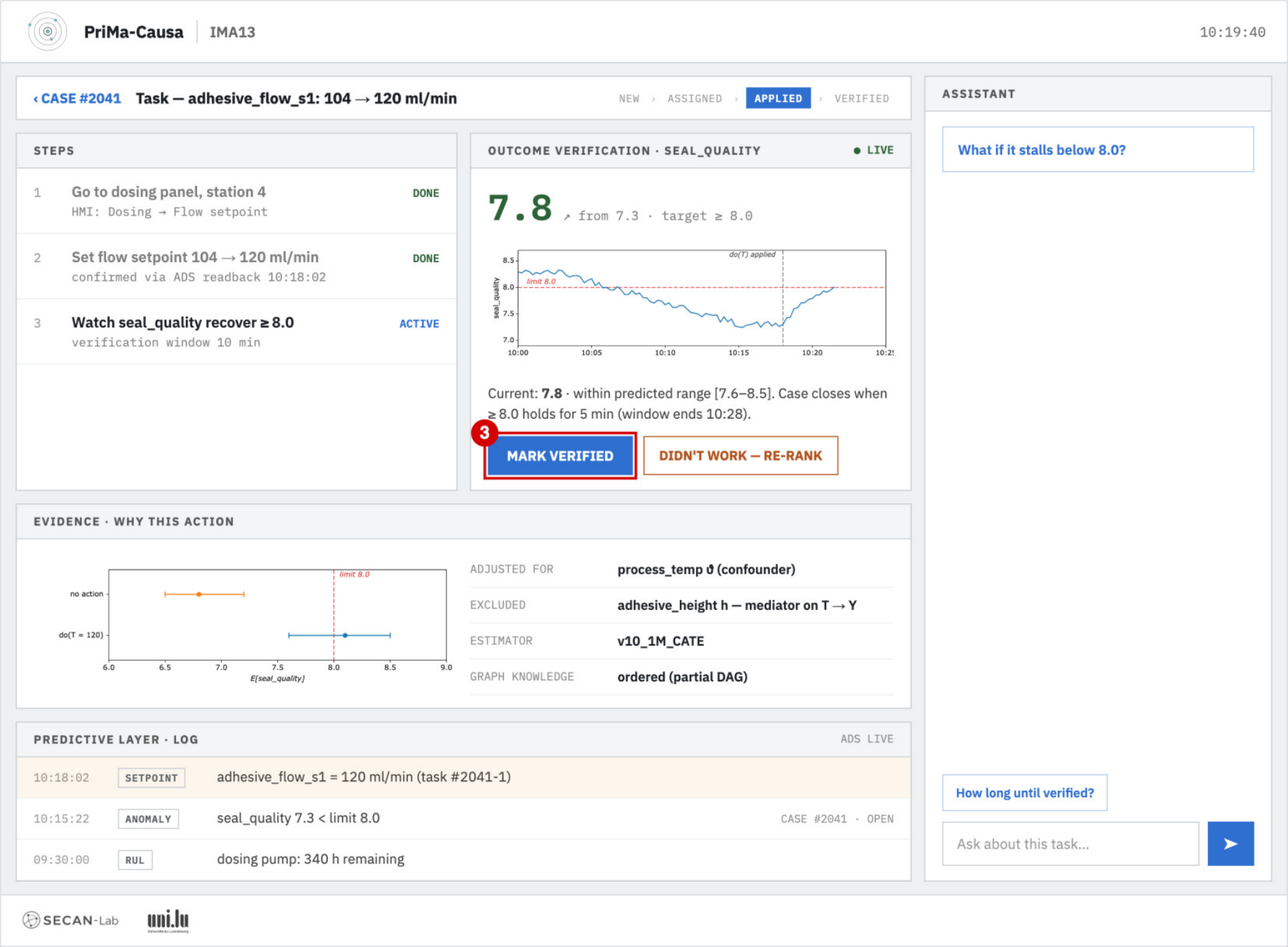}
    \caption{Task view of the operator console.}
    \label{fig:usecase-console-task}
\end{figure}

The first proof of concept is the operator console of PriMa-Causa, which realises the operator interface of the integration architecture (Figure~\ref{fig:usecase-architecture}). Figures~\ref{fig:usecase-console-log} to~\ref{fig:usecase-console-task} trace a single troubleshooting case through the console. The depicted case follows the Mode~3 pattern of Subsection~\ref{subsec: Four quadrants}: a deviation in a quality characteristic is observed, the intervention target is initially unspecified, and candidate interventions are ranked through forward interventional queries (DP3). The affected station is an adhesive-application and sealing station, so the case mirrors the confounded station structure used to motivate the dissertation in Section~\ref{sec:intro-gentle-intro}.

The entry point is the log view shown in Figure~\ref{fig:usecase-console-log}. The log aggregates the artefacts of the predictive-analytics layer, namely anomaly detections, drift warnings, and remaining-useful-life forecasts, received from the machine over the Beckhoff ADS connection of the edge gateway (Figure~\ref{fig:usecase-architecture}). Each entry is time-stamped and typed, and entries that open a diagnostic case carry a case reference. In the depicted situation, the anomaly monitor has flagged a seal-quality value of $7.3$, below the specification limit of $8.0$, and case~\#2041 has been opened automatically (marker~1). This trigger corresponds to the prescriptive case framing block of the \acs{CRL} (Section~\ref{sec:ch4-causal-reasoning-layer}). The outcome is fixed by the observed deviation while the intervention target remains unspecified, which classifies the case into Mode~3. An assistant panel on the right accepts case-contextual natural-language requests, for example documentation lookups or what-if questions, and translates them into the typed queries of the operator interface (M1--M4 in Figure~\ref{fig:usecase-architecture}). Selecting the triggering entry expands the associated prescriptions.

Figure~\ref{fig:usecase-console-prescriptions} shows the expanded prescription view for case~\#2041. The view carries the Digital Intervention Experiments, estimation, and robust recommendation synthesis blocks of the \acs{CRL} to the operator interface. The candidate actions $a_j=(T_j,t_j)\in\mathcal{A}_{\mathcal{K}}(x)$ are evaluated through forward interventional queries $\dop(T_j{=}t_j)$ with the pre-trained PriMa-Causa model of Chapter~\ref{chap:prima-causa-model} and are ranked by their estimated effect on the deviating quality characteristic. Each row names the candidate root cause, the concrete setpoint intervention, and the estimated change in the expected seal quality under the intervention, reported with a $95\%$ uncertainty interval derived from the spread of the model's estimates over the recent context rows. For the top-ranked candidate, the intervention raises the adhesive-flow setpoint from $104$ to $120\,\mathrm{ml/min}$. The forest plot visualises these qualified scores $s_c$. Only the interval of the top-ranked intervention excludes zero. The remaining candidates show no resolvable effect under the current line state and are therefore deprioritised as root-cause candidates. The ranking is thereby coupled to the uncertainty of the evidence rather than to the point estimates alone (DP5). The technician initiates the recommended intervention through the task workflow (marker~2).

The task view of Figure~\ref{fig:usecase-console-task} guides the technician through the execution of the intervention and links the prescription to the observed outcome. The left panel lists the execution steps, and the setpoint change is confirmed automatically through the ADS readback. The right panel monitors the outcome variable. The time series shows the degradation, the intervention time, and the subsequent recovery, with the current value of $7.8$ lying inside the predicted range of $[7.6,\,8.5]$. The case is closed once the quality characteristic holds above the specification limit for a verification window (marker~3). If the estimated effect fails to materialise, the observed outcome is recorded as interventional evidence and the remaining candidates are re-ranked within the robust recommendation synthesis block. The recorded evidence additionally updates the encoded process knowledge for subsequent cases, which implements the feedback mechanism of design principle DP7. The evidence panel below documents the identification basis of the estimate, namely the adjusted confounder (the process temperature) and the excluded mediator (the adhesive height). The panel further records the identifier of the pre-trained model checkpoint and the structural-knowledge level $\mathcal{K}_{\mathrm{op}}$ assumed for the case, here the ordered level of Subsection~\ref{subsec:knowledge-availability}. The recommendation thereby remains traceable to its interventional queries and identification assumptions, which constitutes the per-case audit trail required by meta-requirement MR13. The adjusted confounder and the excluded mediator correspond to the temperature $\vartheta$ and the adhesive height $h$ of the motivating example (Figure~\ref{fig:intro-dag}). The conflation of these two variables by a purely predictive model was demonstrated in Section~\ref{sec:intro-gentle-intro}.

The depicted case demonstrates the console workflow on the connected line, from the predictive trigger to the verified recovery. It does not constitute a validation of the closed causal recommendation loop in live production operation, which remains open as stated at the beginning of this chapter and in Chapter~\ref{chap:conclusions}.

\section{Troubleshooting Application on Mobile and Head-Mounted Devices}
\label{sec:usecase-app}

\begin{figure}[!htbp]
    \centering
    \setlength{\fboxsep}{0pt}%
    \setlength{\fboxrule}{0.4pt}%
    \begin{subfigure}[t]{0.31\textwidth}
        \centering
        \fbox{\includegraphics[width=\linewidth]{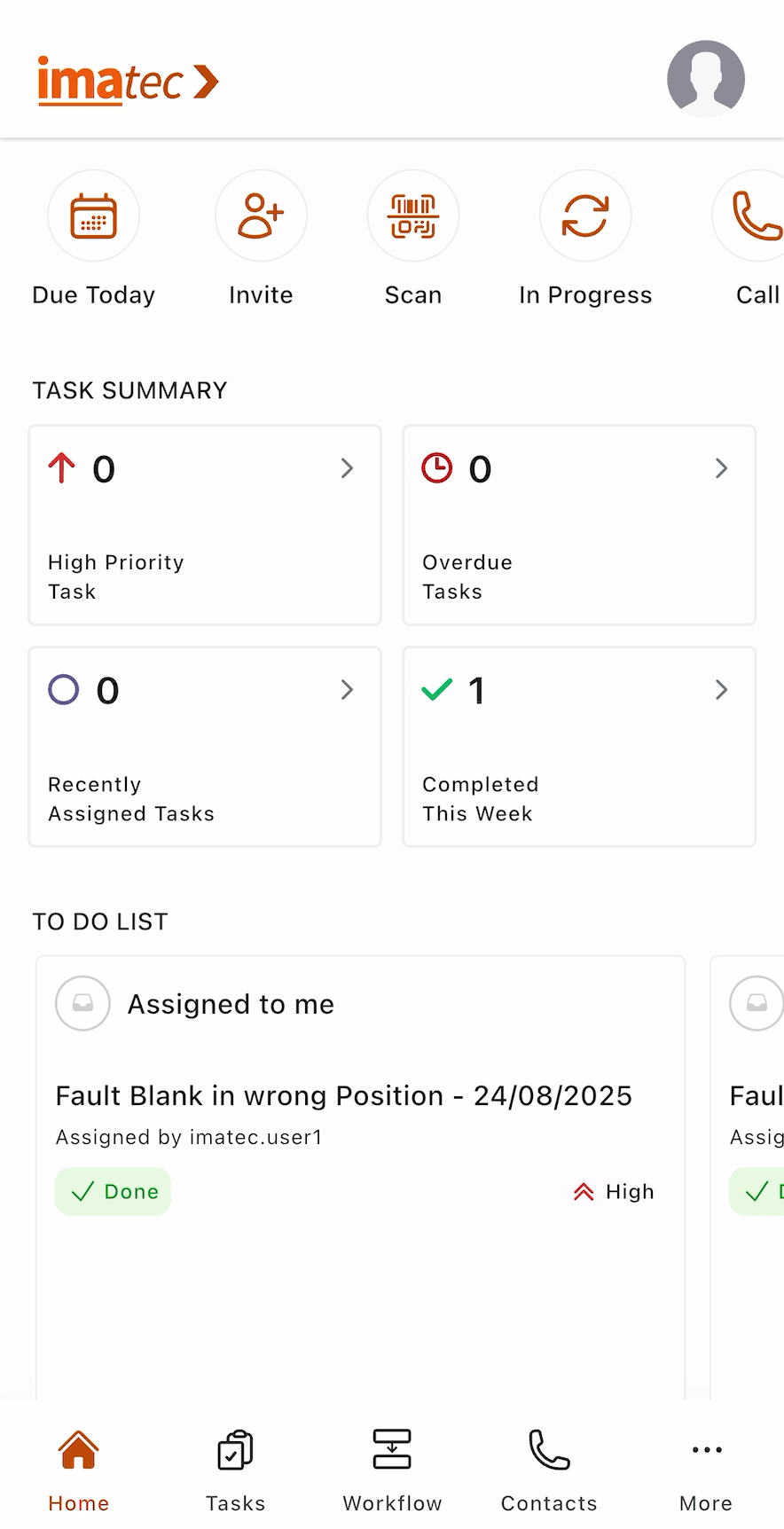}}
        \caption{Task overview.}
        \label{fig:usecase-app-tasks}
    \end{subfigure}
    \hfill
    \begin{subfigure}[t]{0.31\textwidth}
        \centering
        \fbox{\includegraphics[width=\linewidth]{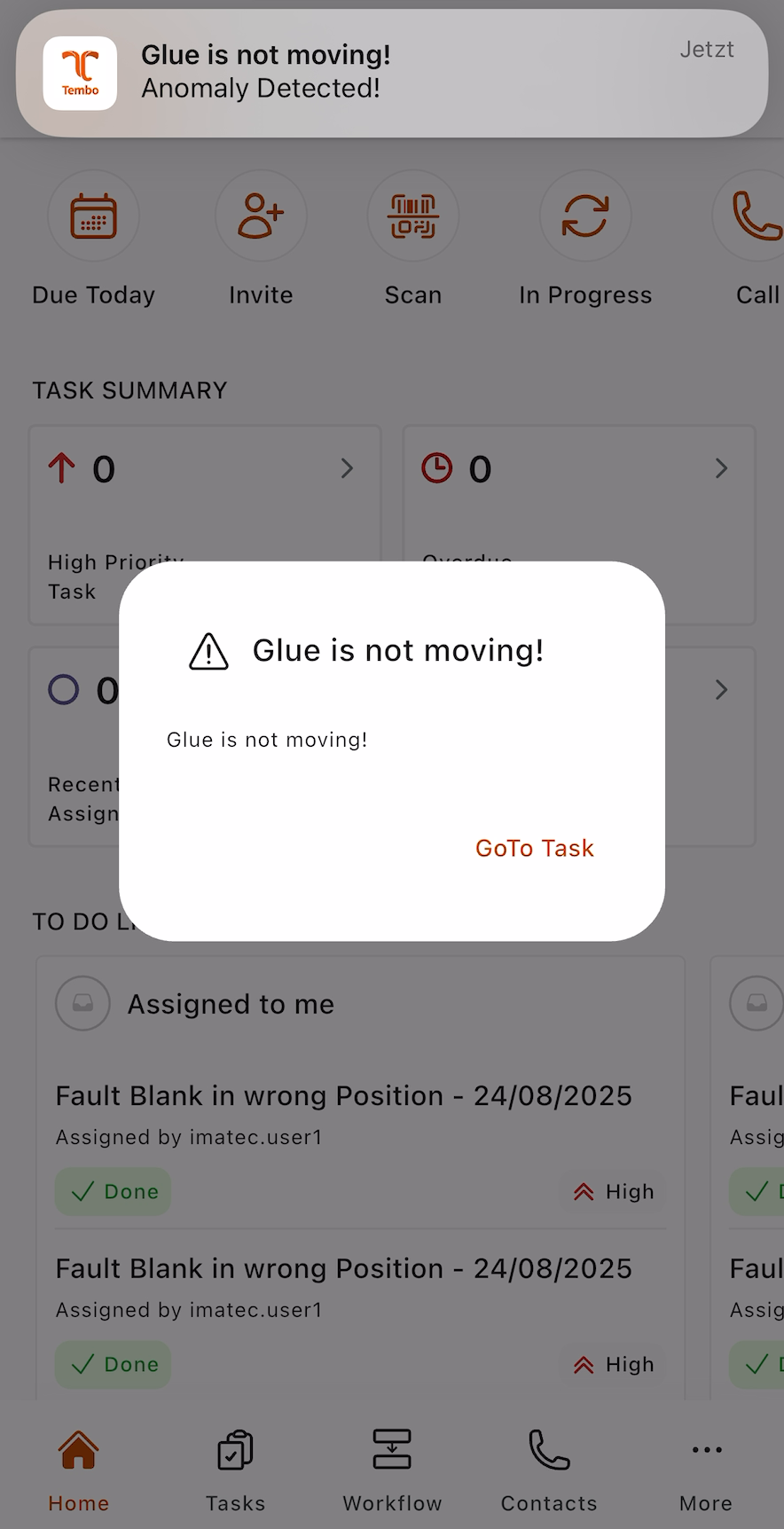}}
        \caption{Anomaly notification.}
        \label{fig:usecase-app-alert}
    \end{subfigure}
    \hfill
    \begin{subfigure}[t]{0.31\textwidth}
        \centering
        \fbox{\includegraphics[width=\linewidth]{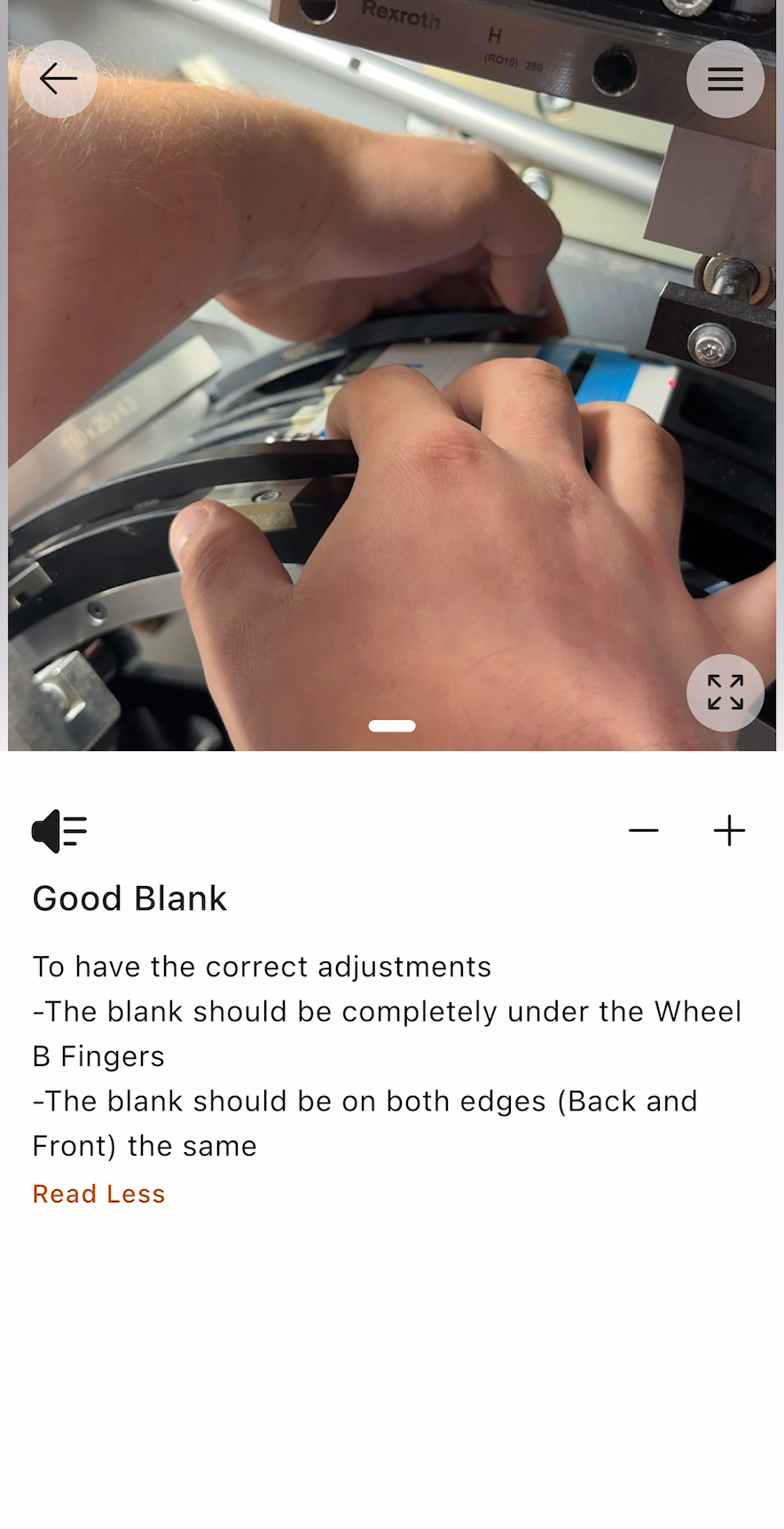}}
        \caption{Guided repair instruction.}
        \label{fig:usecase-app-guide}
    \end{subfigure}
    \caption{Operator-facing troubleshooting application.}
    \label{fig:usecase-app}
\end{figure}

Figure~\ref{fig:usecase-app} shows three views of the operator-facing troubleshooting application, which was co-developed with Dropslab Technologies. Open and completed maintenance tasks are aggregated in a task overview (Figure~\ref{fig:usecase-app-tasks}). When an anomaly is detected, a notification is pushed to the operator and linked to a corresponding task (Figure~\ref{fig:usecase-app-alert}). Each task is resolved through guided repair instructions with reference imagery of the correct machine state (Figure~\ref{fig:usecase-app-guide}). The application realises the operator interface of the integration architecture (Figure~\ref{fig:usecase-architecture}) on a mobile device. In the Mode~3 flow of the preceding section, the pushed notification corresponds to the predictive trigger of the case framing, and the guided instruction carries the recommended intervention to its execution at the machine.

\begin{figure}[!htbp]
    \centering
    \setlength{\fboxsep}{0pt}%
    \setlength{\fboxrule}{0.4pt}%
    \fbox{\includegraphics[width=0.85\textwidth]{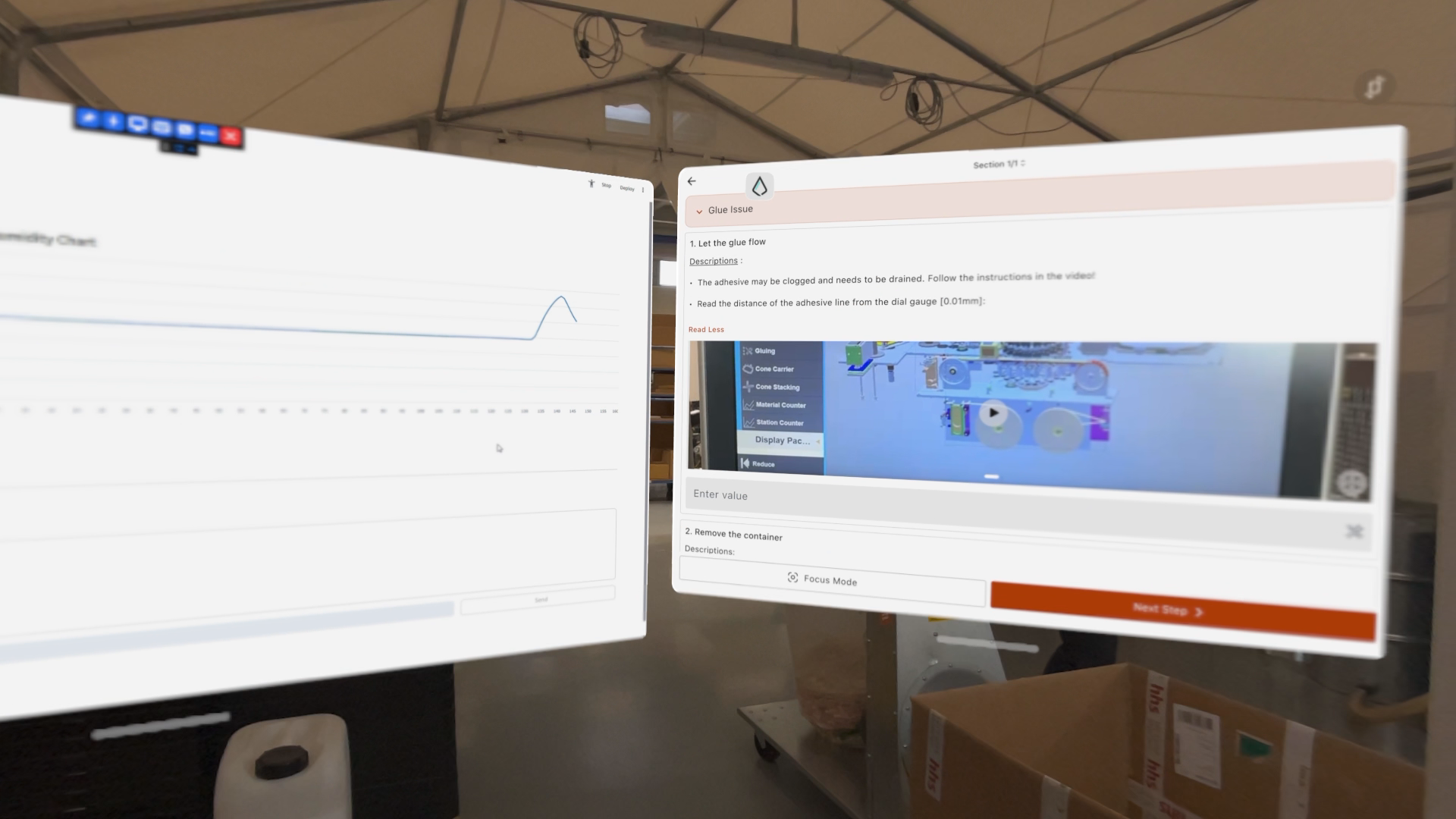}}
    \caption{Head-mounted variant of the troubleshooting application on Apple Vision Pro.}
    \label{fig:usecase-avp}
\end{figure}

The application was additionally demonstrated on a head-mounted device (Figure~\ref{fig:usecase-avp}). On Apple Vision Pro the task views of the application, connected to the same task backend as the mobile variant, are rendered in the operator's field of view together with the live process trends. The execution at the machine thereby remains hands-free. The depicted task addresses an adhesive fault of the same station family as the console case of Section~\ref{sec:usecase-console}.

\begin{table}[!htbp]
    \centering
    \caption[Design principles in the demonstrator.]{Realisation of the design principles and meta-requirements of Chapter~\ref{chap:prima-causa} in the demonstrated artefacts.}
    \label{tab:usecase-dp}
    \footnotesize
    \renewcommand{\arraystretch}{1.2}
    \begin{tabularx}{\textwidth}{@{}lXl@{}}
        \toprule
        Principle & Realisation in the demonstrator & Reference \\
        \midrule
        MR2  & Automatic framing of the observed deviation into Mode~3 when a case is opened & Fig.~\ref{fig:usecase-console-log} \\
        DP3  & Candidate interventions ranked through forward interventional queries of the pre-trained model & Fig.~\ref{fig:usecase-console-prescriptions} \\
        DP5, MR11 & Ranking coupled to the predictive uncertainty through $95\%$ intervals in the forest plot & Fig.~\ref{fig:usecase-console-prescriptions} \\
        DP7  & Observed outcomes recorded as interventional evidence that informs subsequent cases & Fig.~\ref{fig:usecase-console-task} \\
        DP8, MR7 & Evidence panel with the adjusted confounder and the excluded mediator of the estimate & Fig.~\ref{fig:usecase-console-task} \\
        MR13 & Per-case audit trail with model checkpoint, knowledge level $\mathcal{K}_{\mathrm{op}}$, and identification basis & Fig.~\ref{fig:usecase-console-task} \\
        MR6  & Amortised recommendation from the pre-trained model without a per-case fit at decision time & Sec.~\ref{sec:usecase-architecture} \\
        DP1  & Diagnostic notification separated from the prescriptive guided instruction & Fig.~\ref{fig:usecase-app} \\
        \bottomrule
    \end{tabularx}
\end{table}

Table~\ref{tab:usecase-dp} summarises how the design principles and meta-requirements of Chapter~\ref{chap:prima-causa} surface in the demonstrated artefacts. The demonstrations cover the path from the predictive trigger to the guided execution on the operating line and thereby establish the feasibility of the integration. The controlled evaluation of the estimator under known ground truth follows in Chapter~\ref{chap:empirical-evaluation}.

		    \chapter{Empirical Evaluation}
\label{chap:empirical-evaluation}

\begin{jointwork}
This chapter constitutes contribution~C3 and answers research question RQ3 of Chapter~\ref{chap:introduction}. This chapter incorporates my prior peer-reviewed publication~\cite{saretzky2026integratingcausalfoundationmodel}. Figure~\ref{fig:ch8_structure} outlines the structure of this chapter. Sections~\ref{sec:eval-setup} and~\ref{sec:eval-hypotheses} establish the experimental setup and the evaluation hypotheses. The estimation experiments of Sections~\ref{sec:eval-gap} to~\ref{sec:eval-conditions} test interventional accuracy, structural knowledge, scalability, transferability, and the manufacturing data conditions. The operational experiments of Sections~\ref{sec:eval-downstream} and~\ref{sec:eval-rca} test downstream impact and causal root-cause analysis. Section~\ref{sec:eval-discussion} synthesises the results into the empirical validation of the central hypothesis.
\end{jointwork}

\vspace{\baselineskip}

\begin{center}
    \includegraphics[width=\textwidth]{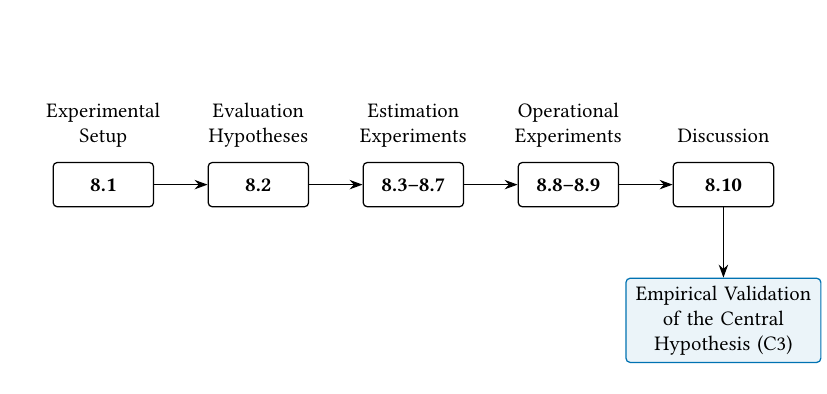}%
    \captionof{figure}{Structure of Chapter~\ref{chap:empirical-evaluation}.}
    \label{fig:ch8_structure}
\end{center}

\newpage

This chapter evaluates the PriMa-Causa model instantiation against the central hypothesis of Chapter~\ref{chap:introduction}: under the data conditions of \acs{FMCG} production lines, an interventional estimator of $P(Y\mid \dop(T{=}t),X{=}x)$ equipped with causal inductive bias yields more accurate intervention-effect estimates and a higher expected maintenance efficiency than an associational estimator of $P(Y\mid X{=}x)$. The evaluation is organised as a design-of-experiments study in which two axes of variation are controlled independently:

\begin{enumerate}
    \item the \emph{identification problem} (confounder, mediator, collider-induced selection bias), and
    \item the \emph{statistical problem} (data quality: overlap, noise level, sample size, distribution shift).
\end{enumerate}

\section{Experimental Setup}
\label{sec:eval-setup}

\paragraph{Models under comparison.} The PriMa-Causa model is instantiated as the graph-conditioned interventional estimator of Chapter~\ref{chap:prima-causa-model}. It is evaluated across the structural-knowledge levels of $\mathcal{K}_{\mathrm{op}}$: \emph{PriMa-Causa (no DAG)}, abbreviated PC-ng, at the \emph{none} level with no graph, the same model supplied with a partially specified graph at the \emph{ordered} level, abbreviated PC-pg, and \emph{PriMa-Causa (DAG)}, abbreviated PC-wg, at the \emph{complete} level with the true causal graph. The graph-consuming baselines are a back-door-adjustment estimator with heterogeneous effect modifiers (DoWhy~\cite{sharma2020dowhyendtoendlibrarycausal}), which derives an adjustment set from the supplied graph, and a discovery-then-adjust estimator (DiscoverAdjust), which learns the graph from the context by constraint-based search, the PC algorithm~\cite{10.7551/mitpress/1754.001.0001} in the pgmpy implementation~\cite{JMLR:v25:23-0487}, before adjusting as DoWhy does. The baselines that do not consume a graph are an S-learner, a T-learner, and an X-learner~\cite{doi:10.1073/pnas.1804597116} on random forests (ML-S, ML-T, ML-X), a causal forest~\cite{Wager03072018} in the CausalML implementation~\cite{chen2020causalmlpythonpackagecausal}, and the graph-free causal foundation model CausalFM of Section~\ref{sec:ch6-backbone}. The meta-learners and the causal forest are refitted on the context of every sampled instance, whereas CausalFM is queried in a single forward pass. The baselines that use neither the causal graph nor interventional pre-training, namely ML-S, ML-T, ML-X, and the causal forest, are referred to as associational baselines.

\paragraph{Data source and conditions.} Each sampled SCM is one experimental replication, drawn from the Causal Process Graph generator of Chapter~\ref{chap:synthetic-data}. These instances are drawn independently of the pre-training corpus of Chapter~\ref{chap:prima-causa-model}, so the frozen model is evaluated only on tasks unseen during pre-training. The matched per-unit interventional outcomes are obtained from the truncated factorisation induced by the $\dop$-operator (Equation~\eqref{eq:scm-truncated}), so the oracle effect is known by construction. Three parameters govern the data conditions: an overlap parameter $\theta$, the strength of the direct treatment effect, and the marginal treatment prevalence. Increasing $\theta$ sharpens the treatment-assignment propensity as a function of the covariates and thereby weakens the overlap between the treated and control covariate distributions.

\paragraph{Metrics.} The primary metric is the precision in estimation of heterogeneous effects,
\begin{equation}
\begin{gathered}
    \mathrm{PEHE} = \sqrt{\frac{1}{N}\sum_{i=1}^{N}\bigl(\hat\tau_i - \tau_i\bigr)^{2}},\\
    \tau_i = \mathbb{E}\bigl[Y \mid \dop(T{=}1),\,X{=}x_i\bigr] - \mathbb{E}\bigl[Y \mid \dop(T{=}0),\,X{=}x_i\bigr],
\end{gathered}
    \label{eq:pehe}
\end{equation}
where the sum runs over the $N$ evaluation units, $\hat\tau_i$ is the model's estimate of the conditional total effect $\tau(x_i)$ of Equation~\eqref{eq:cate} and $\tau_i$ is the oracle per-unit effect obtained from the structural causal model. The metric retains its established name from the treatment-effect literature~\cite{Hill01012011}. The estimand it scores is the do-calculus conditional total effect of Equation~\eqref{eq:cate}, and the potential-outcomes lineage of the name is adopted as nomenclature only. The secondary metric is the per-query inference latency. 

\paragraph{Comparison granularity.} For every (model, condition) cell, the per-SCM aggregate metric is reported as a boxplot over $N_{\mathrm{SCM}}\geq 1000$ instances. The paired design allows a Wilcoxon signed-rank test per model pair. Boxplots follow the convention of Chapter~\ref{chap:synthetic-data}.\footnote{The whiskers extend from the 5th to the 95th percentile, and outliers beyond this range are omitted for legibility.}

\section{Evaluation Hypotheses}
\label{sec:eval-hypotheses}

This chapter answers research question RQ3 introduced in Chapter~\ref{chap:introduction} and constitutes contribution~C3, addressing research gaps RG2 and RG3 through design objectives DO2 and DO3. The root-cause hypothesis EH11 additionally addresses research gap RG4. The central hypothesis of Chapter~\ref{chap:introduction} operationalises RQ3 as a falsifiable claim and is decomposed into ten evaluation hypotheses, each traced to the meta-requirements of Chapter~\ref{chap:prima-causa} and, where applicable, to a design principle of Subsection~\ref{subsec:ch4-design-principles-sub}, and tested by one experiment of this chapter. Each hypothesis is stated as a falsifiable claim that predicts a measurable outcome under a named condition, so that a single experiment can refute it. The traceability of each hypothesis to the research gaps, meta-requirements, and design principles is collected in Table~\ref{tab:eval-traceability}. In the table, the column Hyp.\ assigns a hypothesis to the accuracy half (i) or the maintenance-efficiency half (ii) of the central hypothesis where applicable. Research gap RG1, meta-requirements MR1, MR3, and MR10, and design principles DP2 and DP9 are design-stage constructs established in Chapter~\ref{chap:prima-causa} and are satisfied by the architecture of the artefact rather than by a separate experiment, so no evaluation hypothesis is mapped to them. Decision mode M4 (structural discovery) is outside the scope of this dissertation. The numbering skips EH8, a regime-dependence hypothesis retired during evaluation, and EH10, a Mode-2 side-effect-ranking hypothesis dropped because no multi-outcome ranking harness was built. %

\begin{description}
  \item[EH1 (interventional accuracy).] Under weak overlap and sparse data with confounded treatment assignment, the interventional estimator attains a lower PEHE than every associational baseline. Under strong overlap and abundant data it does not.
  \item[EH2 (structural knowledge).] Under weak overlap and sparse data, PriMa-Causa attains a lower PEHE than the graph-consuming back-door and discovery estimators at every level of $\mathcal{K}_{\mathrm{op}}$ at which each baseline is defined. The reduction in its own PEHE from supplying the complete graph is larger under linear mechanisms, where the structure is not identified from the observational distribution, than under nonlinear mechanisms.
  \item[EH3 (sparse data and scalability).] Across the evaluated context sizes the amortised estimator attains a lower PEHE than the fitted baselines, whose error plateaus at a positive floor as rows are added, and its PEHE grows more slowly than theirs over the evaluated graph sizes $n$.
  \item[EH4 (inference latency).] The amortised estimator's median per-query latency is higher than every baseline's. The magnitude of this latency determines whether a recommendation is available within the production-stop window.
  \item[EH5 (transfer).] Evaluated without refitting on covariates drawn from public industrial datasets rather than from the pre-training mechanism, the amortised estimator keeps its normalised PEHE below the no-effect reference across the difficulty ladder, and a model pre-trained on the structured \acs{CPG} prior retains a larger margin than one pre-trained on density-matched Erd\H{o}s--R\'enyi graphs.
  \item[EH6 (confounding and overlap).] As the overlap parameter $\theta$ increases, the associational baselines' PEHE increases faster than the interventional estimator's.
  \item[EH7 (causal inductive bias).] Supplying the causal graph reduces the amortised estimator's PEHE on mediated and selection-biased structural causal models.
  \item[EH9 (impact).] Under a bounded adjustment budget, ranking production states by the estimated interventional effect and adjusting the top-ranked fraction yields a positive expected net \acs{OEE} gain that exceeds the gain of the treatment-unaware predictive ranking at every budget level, and the gain saturates as the budget grows.
  \item[EH11 (root-cause analysis, Mode~3).] For an observed outcome deviation with the intervention target unspecified, restricting the candidates through the causal graph identifies the injected root cause within the top-$k$ variables more often than an unrestricted correlation-based attribution. For root causes that remain within their normal operating band while the outcome deviates, the forward interventional queries recover the injected cause, whereas anomaly-based and correlation-based attribution cannot.
  \item[EH12 (graph density).] The reduction in PEHE that a partially disclosed graph grants over the no-graph variant of the same estimator grows with the realised edge density over the low-to-medium range, and the partial graph recovers the complete-graph accuracy.
\end{description}

\begin{table}[!htbp]
  \centering
  \caption[Traceability of the evaluation hypotheses.]{Traceability of the evaluation hypotheses to the central hypothesis halves, research gaps, meta-requirements, design principles, decision modes, and the floats that test them.}
  \label{tab:eval-traceability}
  \footnotesize
  \setlength{\tabcolsep}{4pt}
  \renewcommand{\arraystretch}{1.15}
  \begin{tabular}{@{}llllll@{}}
    \toprule
    \textbf{EH} & \textbf{Hyp.} & \textbf{RG} & \textbf{MR} & \textbf{DP} & \textbf{Mode / float} \\
    \midrule
    EH1  & (i)  & RG2 & MR4, MR8   & --        & M1 / Fig.~\ref{fig:eval-gap}, Fig.~\ref{fig:eval-gap-scatter} \\
    EH2  & --   & RG3 & MR5        & DP4       & M1 / Fig.~\ref{fig:eval-kop}, Tab.~\ref{tab:eval-kop} \\
    EH3  & (i)  & RG3 & MR6        & --        & M1 / Fig.~\ref{fig:eval-scarce-context}, Fig.~\ref{fig:eval-graph-size} \\
    EH4  & --   & RG3 & MR12       & DP6       & M1 / Fig.~\ref{fig:eval-latency} \\
    EH5  & --   & RG3 & MR6, MR13  & DP7       & M1 / Fig.~\ref{fig:eval-semisynth-cpg}, Fig.~\ref{fig:eval-semisynth-er} \\
    EH6  & (i)  & RG3 & MR7        & DP8       & M1 / Fig.~\ref{fig:eval-overlap} \\
    EH7  & (i)  & RG3 & MR5        & DP4       & M1 / Fig.~\ref{fig:eval-inductive-bias} \\
    EH9  & (ii) & RG2 & MR11, MR12 & DP5, DP6  & M1 / Fig.~\ref{fig:eval-oee-budget} \\
    EH11 & --   & RG4 & MR2, MR9   & DP1, DP3  & M3 / Sec.~\ref{sec:eval-rca} \\
    EH12 & --   & RG3 & MR5, MR6   & DP4       & M1 / Fig.~\ref{fig:eval-density} \\
    \bottomrule
  \end{tabular}
\end{table}

\section{Interventional Accuracy and the Prediction--Prescription Gap}
\label{sec:eval-gap}

\begin{figure}[!htbp]
    \centering
    \includegraphics[width=0.95\textwidth]{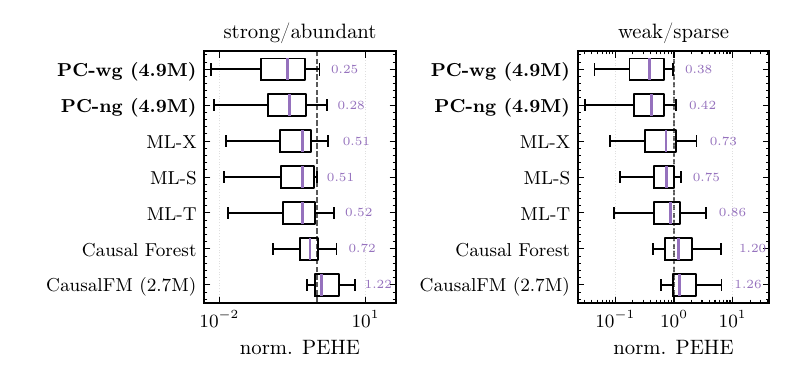}
    \caption[Prescription error by model across data conditions.]{Normalised prescription error (PEHE) by model under strong overlap and abundant data (left) and under weak overlap and sparse data (right), medians annotated.}
    \label{fig:eval-gap}
\end{figure}

This experiment tests EH1. Each model is evaluated on structural causal models drawn from the Causal Process Graph generator with confounded treatment assignment, the failure mode introduced with the single-station example of Figure~\ref{fig:intro-dag}. Two data conditions are contrasted, shown in the two panels of Figure~\ref{fig:eval-gap}: strong overlap with abundant data ($\theta=1$, $400$ context rows) and weak overlap with sparse data ($\theta=6$, $100$ context rows), both on $25$-node graphs. For each condition the normalised prescription error, the PEHE divided by the RMS of the oracle effect per SCM, is reported for every model, so that the dependence of the relative accuracy on the data condition is made explicit.

Figure~\ref{fig:eval-gap} compares the normalised prescription error of the models under the two data conditions. Under weak overlap and sparse data the two PriMa-Causa variants attain the lowest median error, and every associational baseline incurs a higher median error. Under strong overlap and abundant data the ordering is unchanged and the margins to the fitted baselines narrow. A paired Wilcoxon signed-rank test on the per-SCM error confirms a significant separation between PriMa-Causa and each baseline in both conditions ($p<0.05$ for every model pair). These results support the first clause of EH1. The advantage persists under strong overlap and abundant data, a more favourable outcome than the second clause of EH1 predicted. The second clause is therefore refuted in the model's favour at the current scale.

\begin{figure}[!htbp]
    \centering
    \includegraphics[width=0.6\textwidth]{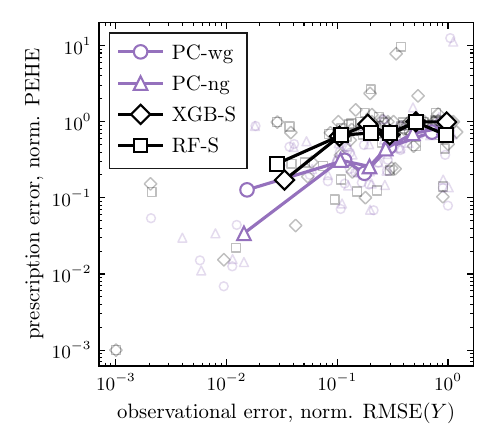}
    \caption[Decoupling of observational and interventional accuracy.]{Observational error against prescription error, one point per sampled SCM and model, at weak overlap and sparse data.}
    \label{fig:eval-gap-scatter}
\end{figure}

The prediction-prescription gap is the divergence between observational fit and interventional accuracy. Figure~\ref{fig:eval-gap-scatter} plots, for each sampled SCM under the weak-overlap and sparse-data condition, the observational error (normalised RMSE on the factual $Y$) against the prescription error (normalised PEHE) on logarithmic axes. A model with a low observational error but a high prescription error fits the observed outcome well yet estimates the interventional effect poorly, the failure mode of Figure~\ref{fig:intro-predict-prescribe}. Both axes are computed from the identical fitted estimator per SCM, isolating the model rather than the CATE construction. PriMa-Causa (PC-wg and PC-ng, purple) and two dedicated S-learner variants, on a random forest (RF-S) and on gradient-boosted trees (XGB-S, black), each predict $Y$ directly for the observational error and combine the two interventional predictions under $\dop(T{=}1)$ and $\dop(T{=}0)$ for the prescription error. Faded markers are the raw per-SCM points. The connected markers trace the median prescription error within six quantiles of the pooled observational error, isolating the model-specific gap from cross-SCM difficulty. The two errors are positively correlated (pooled Spearman $\rho=0.51$) rather than unrelated, and the gap persists conditionally: at matched observational error, the PriMa-Causa variants attain a lower median prescription error than both S-learner variants in five of the six quantiles. In the highest-error quantile the random-forest variant attains the lowest median.

\section{Structural-Knowledge Regime}
\label{sec:eval-kop}

This experiment tests EH2. The structural knowledge supplied to each model is varied along the regime axis $\mathcal{K}_{\mathrm{op}}\in\{\text{none},\text{ordered},\text{complete}\}$ of Subsection~\ref{subsec:knowledge-availability}. At the \emph{none} level no graph is provided. At the \emph{ordered} level a partially specified graph is supplied, operationalised as the disclosure of one half of the true edges together with their transitive closure, an order-consistent partial structure in which the remaining relations, including absent ones, are left unknown. At the \emph{complete} level the full causal graph is available. The data conditions are held at weak overlap and sparse context ($\theta=6$, $100$ context rows, the condition of Section~\ref{sec:eval-gap}), the regime in which the structural-knowledge axis is most informative.

Each sampled instance belongs to one of three structural families that isolate distinct identification settings. In the mediator family (T1) an observed variable lies on a directed path from $T$ to $Y$, so that adjustment for the full observed context violates the back-door criterion and biases the estimate. In the collider family (T2) a pre-intervention variable is a common child of a cause of $T$ and a cause of $Y$ that are both withheld from the observed covariates, so that conditioning on it opens a spurious association (M-bias). The control family (T0) contains neither construct, so adjustment for the observed context is admissible. The results are reported on the mediator family, with the control family retained as a reference under linear mechanisms. The collider family is not part of this comparison. The value of structural knowledge on the collider structure is reported in Section~\ref{sec:eval-mediator}.

PriMa-Causa is compared against the two graph-consuming baselines of Section~\ref{sec:eval-setup}. PriMa-Causa consumes the graph as a soft prior on its attention weights and operates at every level of $\mathcal{K}_{\mathrm{op}}$. The back-door estimator (DoWhy) derives an adjustment set from the supplied graph and fits a linear regression on it. The discovery estimator (DiscoverAdjust) learns a graph from the context by constraint-based search, seeded by the disclosed edges at the \emph{ordered} level, and then adjusts as the back-door estimator does.

\begin{figure}[!htbp]
    \centering
    \includegraphics[width=0.95\textwidth]{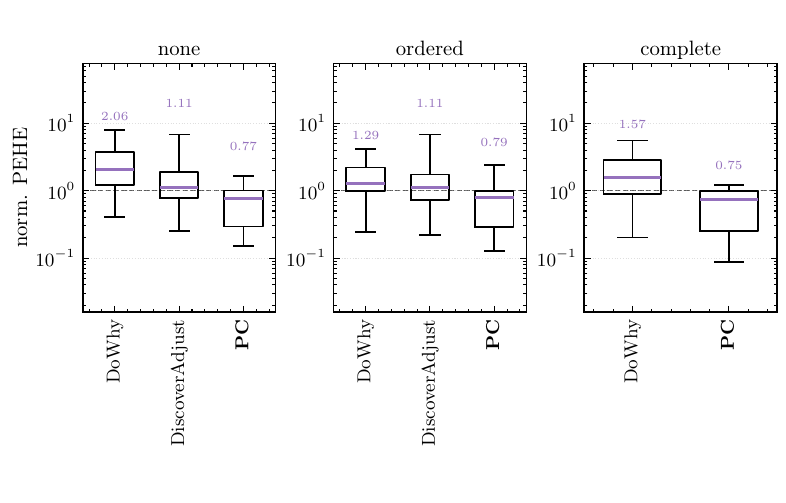}
    \caption[Structural-knowledge regime.]{Normalised PEHE across the structural-knowledge regime $\mathcal{K}_{\mathrm{op}}$ on the mediator family, at weak overlap and sparse context.}
    \label{fig:eval-kop}
\end{figure}

Figure~\ref{fig:eval-kop} reports the normalised PEHE across the three regimes on the mediator family. PriMa-Causa attains the lowest error at every level and stays below the predict-no-effect reference of unity. The back-door estimator exceeds unity at the \emph{none} level, where all observed covariates are declared as adjustment variables and the mediator is therefore adjusted for. It improves once disclosed structure restricts the adjustment set, with its best result at the \emph{ordered} level rather than at the \emph{complete} level. The discovery estimator remains near the predict-no-effect reference at both levels, essentially unchanged by the disclosed edges. The estimators are therefore distinguished less by the gain from added structure than by their robustness to its absence. These results support the first clause of EH2.

The dependence of PriMa-Causa on the graph itself is isolated in Table~\ref{tab:eval-kop}, which reports the median normalised PEHE without and with the complete graph under two mechanism classes. The reduction is computed per instance and reported as the median of the fifty paired relative reductions. Under nonlinear mechanisms the reduction is about two percent (Wilcoxon $p=0.022$), since the estimator already recovers most of the relevant structure from the observational context. Under linear mechanisms, in which the mediator and confounder roles are not distinguished by the observational distribution, the reduction rises to about seven percent, although this difference is not statistically significant at fifty instances ($p=0.16$), and it stays negligible on the control family. The ordering of the reductions is consistent with the second clause of EH2. The value of structural knowledge is thus conditional on the extent to which the observational data already carry it.

\begin{table}[!htbp]
    \centering
    \caption[Value of structural knowledge.]{Median normalised PEHE of PriMa-Causa without and with the complete graph, by mechanism class and structural family, at weak overlap and sparse context. Fifty instances per cell.}
    \label{tab:eval-kop}
    \footnotesize
    \renewcommand{\arraystretch}{1.2}
    \begin{tabularx}{\textwidth}{@{}lXrrrr@{}}
        \toprule
        Mechanism & Structural family & no graph & full graph & reduction & instances improved \\
        \midrule
        nonlinear & mediator (T1)      & $0.77$ & $0.75$ & $2.0\%$  & $31/50$ \\
        linear    & control (T0)       & $0.20$ & $0.20$ & $-0.3\%$ & $25/50$ \\
        linear    & mediator (T1)      & $0.42$ & $0.32$ & $7.0\%$  & $28/50$ \\
        \bottomrule
    \end{tabularx}
    \par\smallskip
    {\footnotesize Note: the reduction column reports the median of the fifty paired per-instance relative reductions, not the relative change of the two median columns.}
\end{table}

\section{Scalability and the Small-Data Regime}
\label{sec:eval-scalability}

\paragraph{Sparse context.} This experiment tests the sparse-data half of EH3. The number of observational rows $N_c$ is varied over $\{25,50,100,200,250,400\}$, identical across the foundation model's context and the fitted baselines' training set, at weak overlap with confounding ($\theta = 6$, the same overlap setting as the weak/sparse condition of Figure~\ref{fig:eval-gap}, so that the variation isolates the effect of $N_c$ from the overlap condition). The expected pattern is that additional rows reduce the variance of every estimator but not the confounding bias of the associational baselines, so their prescription error declines and then plateaus at a positive floor above PriMa-Causa rather than crossing below it.

\begin{figure}[!htbp]
    \centering
    \includegraphics[width=0.9\textwidth]{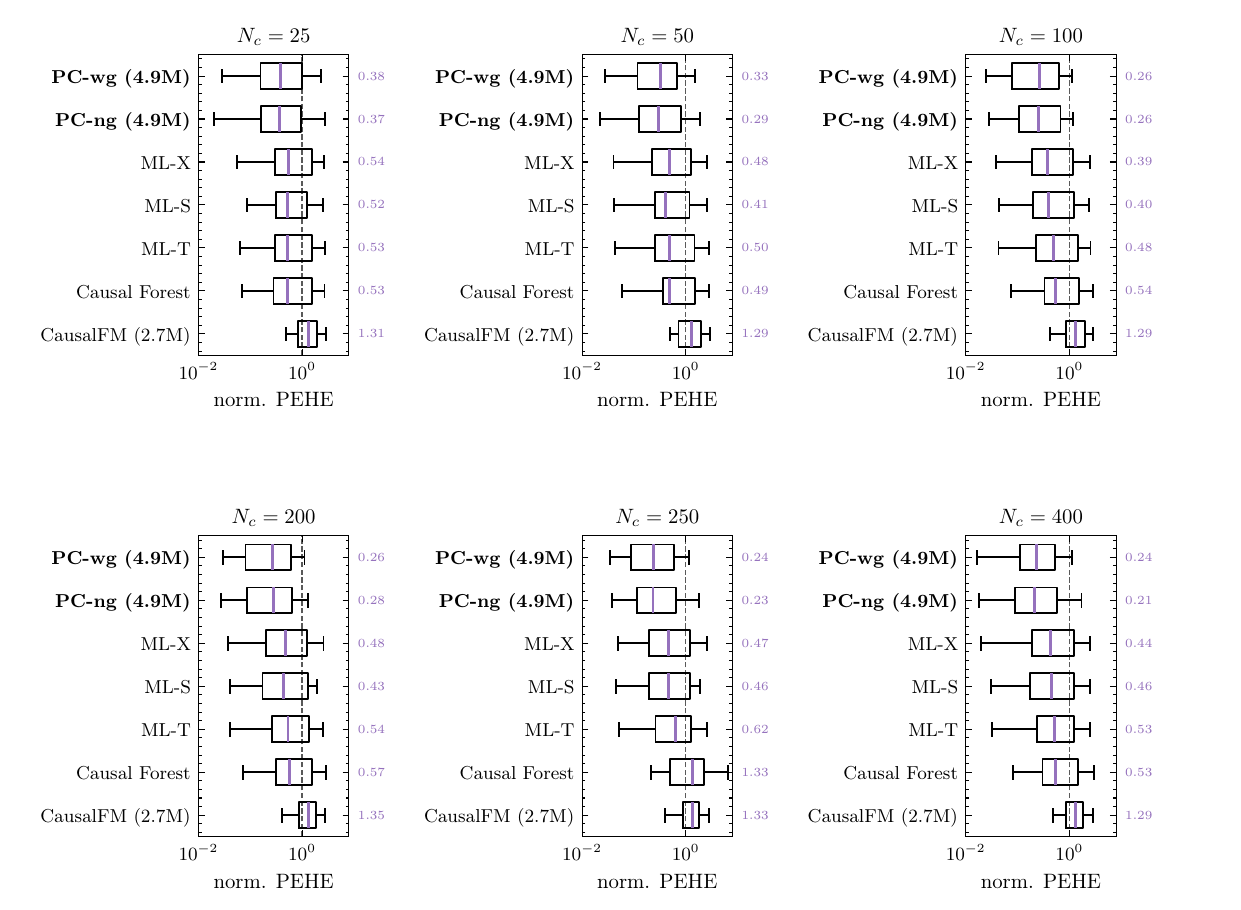}
    \caption[Prescription error against context size.]{Normalised PEHE against the number of context rows $N_c$, at weak overlap with confounding ($\theta = 6$).}
    \label{fig:eval-scarce-context}
\end{figure}

Figure~\ref{fig:eval-scarce-context} shows this pattern with two exceptions. Most baselines' errors decline with additional rows and plateau at a positive floor above both PriMa-Causa variants. The two exceptions are the causal forest, whose error spikes at $N_c{=}250$ before falling back, an instability consistent with run-to-run variance rather than a systematic trend, and ML-T, whose error rises beyond one hundred rows. No baseline crosses below either variant at any context size. PriMa-Causa's own error flattens from about $100$ rows onward. The largest context size lies beyond the pre-training context range of the checkpoint and is retained as a stress point. These results partially support the sparse-data clause of EH3.

\paragraph{Graph size.} This experiment tests the scalability half of EH3. PEHE is reported against the number of nodes $n$ over $\{10,15,20,25\}$, the range accessible to all models of the panel. The expected pattern is that the two PriMa-Causa variants degrade only mildly as the graph grows, whereas the fitted baselines lose accuracy on larger graphs.

\begin{figure}[!htbp]
    \centering
    \includegraphics[width=0.9\textwidth]{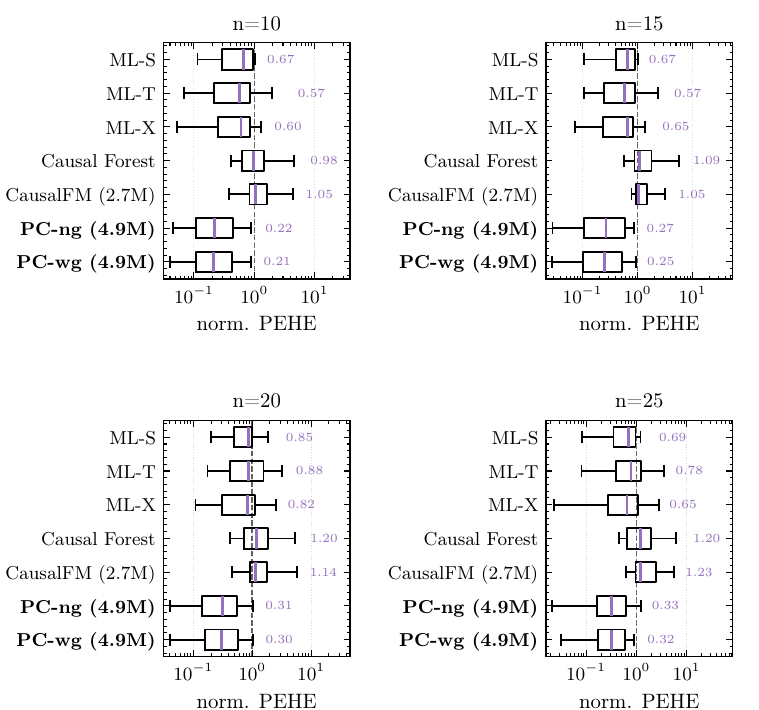}
    \caption[Prescription error against graph size.]{Normalised PEHE against the number of nodes $n$.}
    \label{fig:eval-graph-size}
\end{figure}

Figure~\ref{fig:eval-graph-size} shows that both PriMa-Causa variants attain the lowest error at every graph size and degrade only mildly and monotonically over the evaluated range, while the causal forest and CausalFM degrade more strongly and the random-forest meta-learners vary without a monotone trend. These results partially support the scalability clause of EH3. The decisive contrast is one of level rather than slope. Both variants stay between $0.21$ and $0.33$ across the evaluated sizes, while every baseline remains substantially higher ($0.57$ to $1.23$), with the causal forest and CausalFM at or above the no-effect reference from $n=15$ onwards, and with the random-forest meta-learners non-monotone. The level advantage of both variants therefore holds against all baselines over the evaluated graph sizes.

\paragraph{Inference latency and fit time.} This experiment tests EH4. Per-query latency is a hard operational constraint, because the inference window of a production stop is bounded. Per-query latency is measured after a one-off checkpoint load that is excluded from the timing, so that only the forward pass enters the measurement. Fit time is the cost of preparing a model for new context data, which per-query latency alone omits.

\begin{figure}[!htbp]
    \centering
    \includegraphics[width=0.95\textwidth]{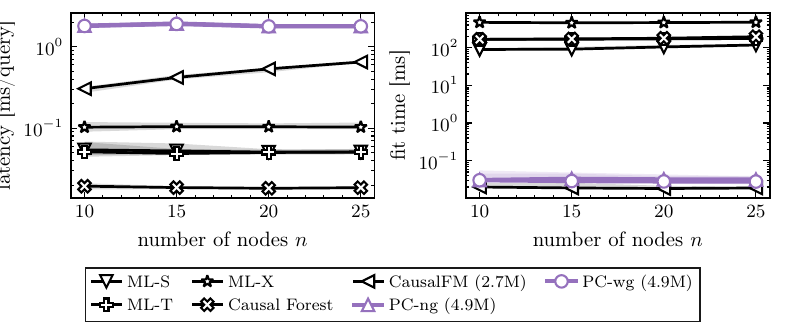}
    \caption[Inference latency and fit time.]{Median per-query latency (left) and per-context fit time (right) by model against the number of nodes $n$, on logarithmic axes. Shaded bands show the interquartile range.}
    \label{fig:eval-latency}
\end{figure}

Figure~\ref{fig:eval-latency} (left) shows that PriMa-Causa's per-query latency (around 1.8~ms, median) is one to two orders of magnitude higher than the tree-based baselines (roughly 0.02--0.10~ms) and about $3$--$4\times$ higher than CausalFM (around 0.5~ms). This relative latency gap is a genuine cost, yet in absolute wall-clock terms it remains small. Figure~\ref{fig:eval-latency} (right) shows the inverse: PriMa-Causa's amortised architecture treats the context as a conditioning set rather than a training set, so its fit time is negligible and near-constant across graph sizes, while the tree-based baselines instead retrain from scratch on every new context, an operational cost that predict-only latency does not capture, and that grows with the number of nodes. The two panels together show that PriMa-Causa trades a higher, roughly constant per-query cost for a near-zero data-refresh cost, whereas the fitted, tree-based baselines trade a lower per-query cost for a data-refresh cost that recurs on every new context. At the measured median of $1.8$~ms per query, a digital-intervention tensor of $3\times10^{4}$ cells is evaluated in under one minute and thus within the bounded inference window of a production stop. These results support EH4.

\section{Transferability and Distribution Shift}
\label{sec:eval-transfer}

This experiment tests EH5 through the semi-synthetic construction of Section~\ref{sec:synth-semi-fmcg}. The covariates are taken from five public industrial datasets, while the treatment assignment and the outcome are simulated on top of them, so that the causal ground truth is known by construction and the covariate distribution is the one the frozen model was never trained to represent. The pre-trained estimator is scored without refitting, since a real production line provides no labelled interventional outcomes at decision time and adaptation is available only by conditioning on the observational context. Estimation difficulty is varied along the ladder of Table~\ref{tab:synth-semi-ladder}, whose rungs jointly hardens the assignment mechanism, the response surface, the overlap, and the effect heterogeneity.

Normalised PEHE is reported, defined as the PEHE divided by the root-mean-square of the true conditional effect. This places the five host datasets, whose effect scales differ by up to an order of magnitude across datasets and rungs, on a common scale. On this scale the value $1$ is the error of a predictor that assigns no treatment effect, so that a value below $1$ indicates recovery of the true effect beyond that null reference.

\begin{figure}[!htbp]
    \centering
    \includegraphics[width=0.95\textwidth]{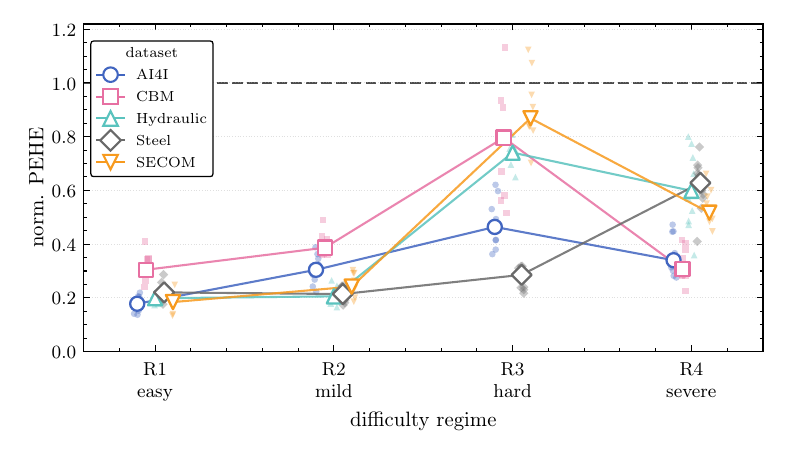}
    \caption[Semi-synthetic difficulty ladder, \acs{CPG} pre-training.]{Normalised PEHE across the difficulty ladder for the \acs{CPG}-pre-trained model, one marker per host dataset. The dashed line marks the no-effect reference at $1$.}
    \label{fig:eval-semisynth-cpg}
\end{figure}

Figure~\ref{fig:eval-semisynth-cpg} reports the model pre-trained on the \acs{CPG} prior. The normalised error remains below the no-effect reference on every rung for all five datasets, so the frozen estimator retains useful accuracy on covariates outside its pre-training distribution. The error increases from the linear regime to the hard rung, where the engineered positivity violation and the nonlinear response surface take effect. At the severe rung the normalised error eases for four of the five datasets, with steel the exception. This observation does not indicate an easier problem, since the introduced effect heterogeneity enlarges the true-effect scale in the denominator faster than the absolute error grows. These results support the first clause of EH5.

\begin{figure}[!htbp]
    \centering
    \includegraphics[width=0.95\textwidth]{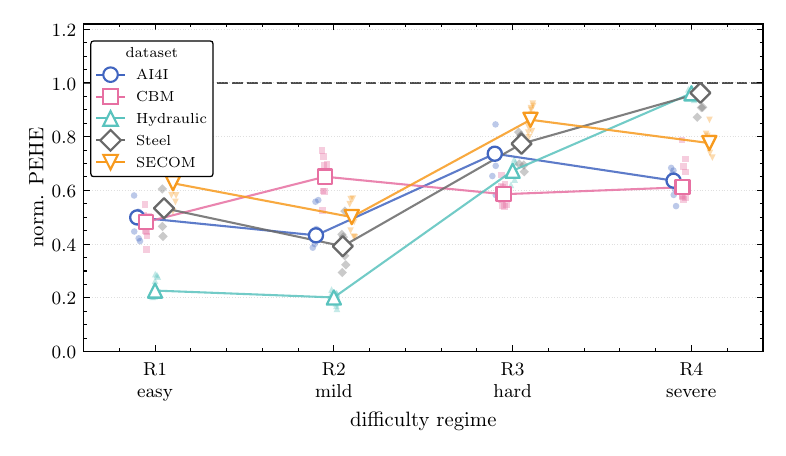}
    \caption[Semi-synthetic difficulty ladder, Erd\H{o}s--R\'enyi pre-training.]{Normalised PEHE across the difficulty ladder for the model pre-trained on density-matched Erd\H{o}s--R\'enyi graphs, otherwise identical to Figure~\ref{fig:eval-semisynth-cpg}.}
    \label{fig:eval-semisynth-er}
\end{figure}

Figure~\ref{fig:eval-semisynth-er} repeats the evaluation for a model pre-trained instead on density-matched Erd\H{o}s--R\'enyi graphs, with the architecture and the training budget held fixed, scored on the identical context-query splits, so the comparison is paired per tile. The \acs{CPG}-pre-trained model attains the lower normalised error on sixteen of the twenty dataset-rung cells, and at the severe rung it retains a margin below the Erd\H{o}s--R\'enyi variant on all five datasets. At the hard rung the picture is mixed, with three datasets favouring the Erd\H{o}s--R\'enyi variant. Because the two models differ only in the graph family seen during pre-training, the contrast is attributable to the structured pre-training prior rather than to architecture or capacity. The comparison rests on a pilot in which each dataset and rung is a single mechanism draw, so the five datasets form the independent unit. These results support the second clause of EH5.

\section{Manufacturing Data Conditions}
\label{sec:eval-conditions}

\subsection{Confounding and Overlap}
\label{sec:eval-overlap}

This experiment tests EH6. The confounder strength is varied via the overlap knob $\theta$ over $\{0.5,2,4,6,8\}$. Larger $\theta$ separates the treated and control covariate clouds, which strengthens the $T$--$X$ confounding and weakens overlap. The weak-overlap tail ($\theta\geq6$) is the regime for which design principle DP8 prescribes escalation instead of a point estimate. This provision is architectural and is not evaluated empirically here. The qualitative pattern follows the caussim benchmark, in which the relative error of an associational model selection grows steeply as overlap fails~\cite{DoutreligneHowToSelect}.

\begin{figure}[!htbp]
    \centering
    \includegraphics[width=0.78\textwidth]{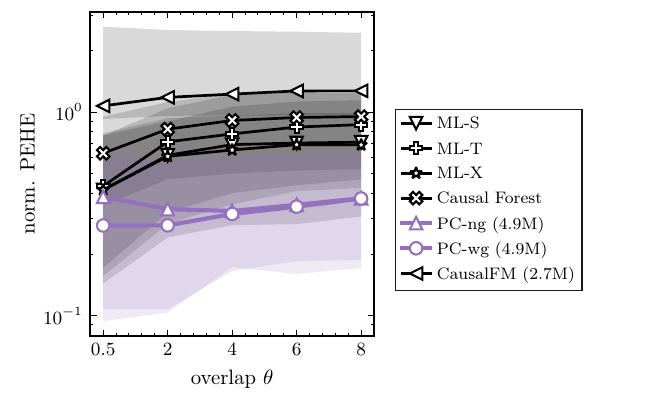}
    \caption[Confounding-strength variation.]{Normalised PEHE against the overlap knob $\theta$. Shaded bands span the per-model interquartile range.}
    \label{fig:eval-overlap}
\end{figure}

Figure~\ref{fig:eval-overlap} shows that at strong overlap the strongest associational baselines approach the no-graph variant, and that as overlap weakens their PEHE grows faster than that of either PriMa-Causa variant. The faster degradation is consistent with the collapse of the propensity and the shrinking common support under weak overlap. These results support EH6.

\subsection{Value of the Causal Inductive Bias}
\label{sec:eval-mediator}

This experiment tests EH7. The contribution of the causal inductive bias is isolated by comparing PriMa-Causa without a supplied graph (PC-ng) against the graph-conditioned variant (PC-wg) on the same data, so that the difference is attributable to the graph alone. Two structural threats are examined separately, each in its own family of structural causal models: a post-treatment mediator on the $T\!\to\!Y$ path, and a selection collider. By the back-door criterion~\cite{pearl2009}, no descendant of $T$ is an admissible adjustment variable for the total effect, so the supplied graph carries the information that the mediator and the collider are to be excluded from adjustment. The graph enters the estimator through the graph-conditioned attention of Chapter~\ref{chap:prima-causa-model}, realising meta-requirement MR5 and design principle DP4. The confounder is not repeated here, as the dependence of the graph-conditioned estimator on the supplied structure is already reported in Section~\ref{sec:eval-kop}. Both structures are evaluated at a CPG-scale, weak-overlap regime, matching the harder condition of Section~\ref{sec:eval-gap}. A design-stage regime-dependence hypothesis (originally EH8) that would have contrasted this regime against a small-graph, strong-overlap regime for the same two structures was retired, as no experiment in this chapter isolates the graph's benefit on mediated and selection-biased structural causal models specifically at the easier regime (see Section~\ref{sec:eval-hypotheses}).

\begin{figure}[!htbp]
    \centering
    \includegraphics[width=0.9\textwidth]{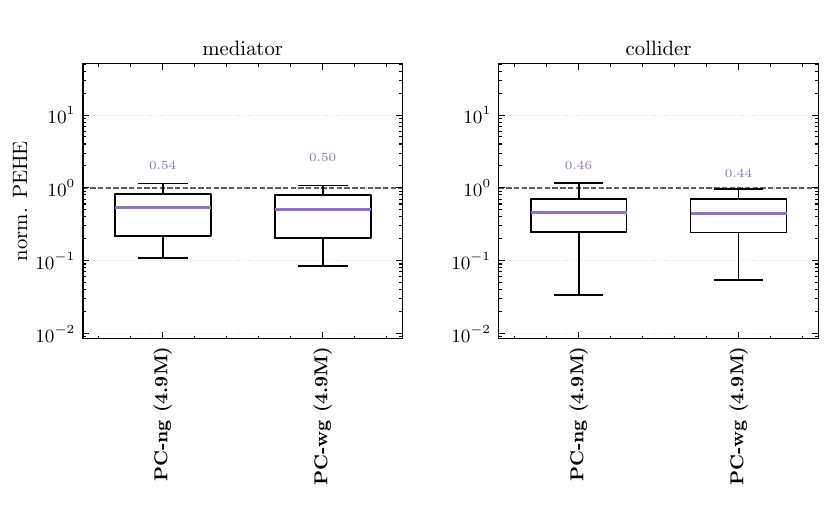}
    \caption[Value of the causal inductive bias.]{Normalised PEHE of PC-ng and PC-wg on a mediator (a) and a selection-collider (b) structure, both at a CPG-scale, weak-overlap regime. The dashed line marks the predict-zero-effect reference.}
    \label{fig:eval-inductive-bias}
\end{figure}

Figure~\ref{fig:eval-inductive-bias} shows that PC-wg attains a lower median normalised PEHE than PC-ng on both the mediator structure (0.50 against 0.54) and the selection-collider structure (0.44 against 0.46), and a paired Wilcoxon signed-rank test rejects equality for both (mediator: $p=0.021$; collider: $p=0.007$; $N_{\mathrm{SCM}}{=}50$ structural causal models per structure). The supplied graph therefore reduces the estimator's error specifically on the two structures that the back-door criterion marks as needing an adjustment set, consistent with EH7. A per-structure decomposition of this benefit is not reported by the graph-conditioned benchmark of Section~\ref{sec:eval-kop}, which does not include the collider family.

\subsection{Graph Density under Partial Structural Knowledge}
\label{sec:eval-density}

This experiment tests EH12. The structural-knowledge experiment of Section~\ref{sec:eval-kop} holds the graph fixed and varies how much of it is disclosed. This experiment instead holds the estimator fixed and varies the density of the structure itself. Three variants of the same amortised estimator are compared: the no-graph variant, the variant supplied with a partially disclosed graph in which half of the ancestor relations are masked to unknown, and the variant supplied with the complete graph as an upper reference. Holding the estimator fixed isolates the graph as the sole varying factor. A contrast against associational baselines would conflate the graph with the estimator architecture and is therefore not used here.

\begin{figure}[!htbp]
    \centering
    \includegraphics[width=0.78\textwidth]{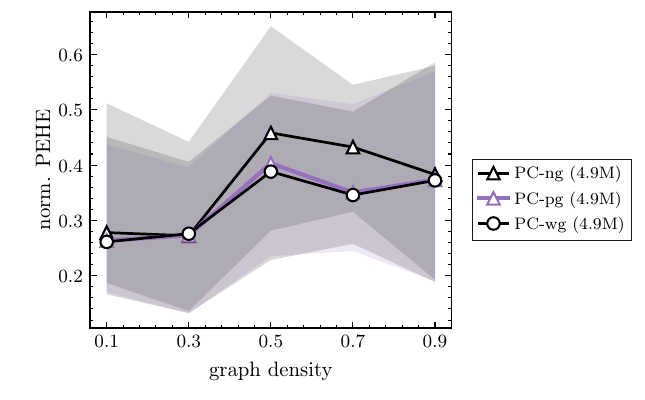}
    \caption[Graph density under partial structural knowledge.]{Median normalised PEHE against the graph density dial for the three structural-knowledge variants of PriMa-Causa: no graph, partially disclosed, and complete. Weak overlap and sparse context. Bands span the $95\%$ bootstrap CI of the median over $N_{\mathrm{SCM}}{=}50$ structural causal models per level.}
    \label{fig:eval-density}
\end{figure}

The density is varied through the density dial of the Causal Process Graph generator (Section~\ref{sec:synth-cpg}) over five levels, so that the realised edge density of the sampled line increases monotonically in expectation. Within each seed the station count and the backbone are held fixed across the levels, so the edge density is the sole varying structural factor, while the accepted instances vary with the density owing to the validity screening. A denser graph admits more back-door paths between the intervention target and the outcome and enlarges the admissible adjustment set, so identification is expected to become harder up to the medium-density range, where the confounding burden is largest relative to the available signal. At still higher densities the abundant connectivity supplies redundant signal and the difficulty eases again. The dial spans the full range that the amortised estimator observed during pre-training, so the evaluated range is in-distribution throughout. The experiment therefore measures the density-scaling of the structural advantage rather than an out-of-distribution boundary.

Figure~\ref{fig:eval-density} shows that the three variants are practically indistinguishable at the two lowest densities, where their median normalised PEHE values differ by less than two hundredths and the sparse graph leaves little structure for the disclosure to resolve. The separation opens in the middle of the range: at the medium-high density the no-graph variant reaches a median of $0.43$ whereas the partial-graph variant reaches $0.35$, a $19\%$ reduction that a paired Wilcoxon signed-rank test rates significant ($p{<}0.001$, $N_{\mathrm{SCM}}{=}50$). The partial and the complete variants track each other within four percent across the whole evaluated range, so the partial disclosure recovers essentially all of the complete-graph accuracy at every density. The advantage attenuates at the densest level, where the three variants converge again. Since the level remains inside the pre-training prior, this is read as an intrinsic difficulty effect rather than a distribution shift. The structural advantage is therefore density-dependent rather than monotone: negligible on sparse graphs, largest at medium-high density, and recovered in full by a partial disclosure. These results support EH12.

\section{Operational Impact}
\label{sec:eval-downstream}

\subsection{Budget-Constrained Intervention Prioritisation}
\label{sec:eval-mttr}

This experiment tests EH9, the efficiency half of the central hypothesis. It reports the policy evaluation of~\cite{saretzky2026integratingcausalfoundationmodel}, redrawn from the original run data. The data source differs from the Causal Process Graph generator of Section~\ref{sec:eval-setup}: real \acs{FMCG} production-state covariates are retained, while the treatment assignment and heterogeneous interventional outcomes are simulated on top of them, so the effect of every candidate adjustment is known by construction. Each production state defines one decision point at which a parameter adjustment is either applied for the next production cycle or withheld. Since adjustment capacity is limited in practice, the task is formulated as a budget-constrained prioritisation. Each policy ranks the production states by its estimated conditional effect, the adjustment is applied to the top-ranked fraction permitted by the budget, and the remaining states are left unchanged. The resulting policy is evaluated against a no-adjustment reference through the known interventional outcomes, which yields the expected net \acs{OEE} gain per budget level. Five ranking policies are compared: the oracle ranking on the true effects, PriMa-Causa, a causal forest, an S-learner on random forests (ML-S), and a treatment-unaware predictive ranking on a random forest (Predictive RF). The predictive ranking orders states by the predicted outcome alone and instantiates the predictive-maintenance-only reference of EH9.

\begin{figure}[!htbp]
    \centering
    \includegraphics[width=0.75\textwidth]{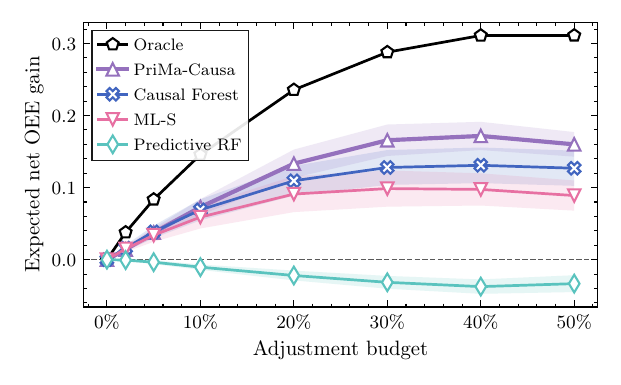}
    \caption[Budget-constrained intervention prioritisation.]{Expected net \acs{OEE} gain against the adjustment budget per ranking policy. Bands span the $95\%$ confidence interval over $20$ repeated runs.}
    \label{fig:eval-oee-budget}
\end{figure}

Figure~\ref{fig:eval-oee-budget} reports the expected net \acs{OEE} gain against the adjustment budget. The treatment-unaware predictive ranking remains at or below zero at every budget level, so a ranking by predicted outcome alone does not translate into an operational gain. Every effect-based ranking attains a positive gain that grows with the budget and saturates beyond a budget of roughly $30\%$, as the highest-impact adjustments are selected first. At the smallest budgets ($2$ to $5\%$) the fitted effect-based baselines are competitive with PriMa-Causa. From a budget of $10\%$ onward PriMa-Causa attains the largest gain among the non-oracle policies, reaching a net gain of $0.17$ at a budget of $40\%$ against $0.13$ for the causal forest and $0.10$ for the S-learner, and it remains bounded by the oracle ranking throughout. The advantage over the effect-based baselines is therefore confined to the mid-to-high budget range, whereas the advantage over the predictive ranking holds at every non-zero budget level. These results support EH9.

The gain is expressed in units of the \acs{OEE}, the product of an availability factor, a performance factor, and a quality factor~\cite{iso22400},
\begin{equation}
\label{eq:oee}
\mathrm{OEE} = A \cdot P \cdot Q ,
\end{equation}
where \(A\) is the fraction of planned production time during which the line operates, \(P\) the ratio of actual to nominal throughput, and \(Q\) the fraction of conforming output. The reported gain acts on this product directly, since every applied adjustment moves the state-level \acs{OEE} outcome. A second channel concerns the availability factor alone. For failure-induced downtime it is governed in steady state by
\begin{equation}
\label{eq:mttr}
A = \frac{\mathrm{MTBF}}{\mathrm{MTBF} + \mathrm{MTTR}} ,
\qquad
\mathrm{MTTR} = \frac{1}{n_r}\sum_{i=1}^{n_r} d_i ,
\end{equation}
with \(n_r\) the number of repairs and \(d_i\) the duration of repair~\(i\). A ranking that selects effective adjustments earlier shortens the troubleshooting and intervention-selection part of the MTTR (Figure~\ref{fig:operational-knowledge-mttr}) and raises \(A\) through Equation~\eqref{eq:mttr}. This second channel is the mechanism behind the outcome horizon of Chapter~\ref{chap:prima-causa} and is not quantified separately here.

\section{Causal Root-Cause Analysis}
\label{sec:eval-rca}

The decision-mode taxonomy of Subsection~\ref{subsec: Four quadrants} is covered in two modes by this chapter. Mode~1 underlies every preceding experiment, since each queries the interventional outcome of a specified treatment--outcome pair. This section evaluates Mode~3. Mode~2 is specified but not empirically tested and Mode~4 is outside the scope of this dissertation (Section~\ref{sec:eval-hypotheses}).

This experiment tests EH11. In Mode~3 the outcome presents an observed deviation while the intervention target is unspecified, so the analysis is a reverse causal query that attributes the deviation to its upstream cause. Following design principle DP3, the reverse question is answered as a set of forward interventional queries over the diagnosis-restricted candidates. The protocol adapts the in-distribution intervention scheme of Nagalapatti et al.~\cite{nagalapatti2025idi}. A root cause must satisfy an anomaly condition, being anomalous while its direct causes are not, and a fix condition, namely that setting it to its usual value restores the outcome. The fix condition is evaluated interventionally rather than through counterfactual anomaly attribution~\cite{budhathoki22a}, since the counterfactual route requires abduction of the exogenous variables at inputs outside the training distribution and is unreliable there~\cite{nagalapatti2025idi}. Counterfactual attribution is nevertheless retained as a baseline of its own. The restriction of the candidates to graph ancestors of the outcome follows the network-based root-cause identification introduced in~\cite{saretzkyNetworks}.

Ground truth is generated with the Causal Process Graph sampler of Chapter~\ref{chap:synthetic-data} at $25$ nodes. The outcome node is the node with the largest ancestor set, the analogue of a quality gate. A fault is injected at a uniformly drawn ancestor by perturbing its exogenous noise. Three fault classes are evaluated separately. In the level-shift and dispersion classes the fault is escalated until the outcome deviates by at least five standard deviations of normal operation and the fault node by at least three. In roughly half of these instances a competing anomalous variable is injected on a causally disjoint branch, escalated until it satisfies the anomaly condition itself while its oracle-verified effect on the outcome remains below half a standard deviation. In the masked class the fault is a shift of at most two standard deviations at an ancestor with a high gain onto the outcome, accepted only when the outcome deviates by at least five standard deviations while the fault node itself remains inside its normal band. Test cases are the units on which the acceptance criteria materialise, yielding $325$ level-shift and $355$ dispersion cases over $50$ structural causal models each, and $180$ masked cases over $25$ structural causal models. The masked class is acceptance-limited, as only half of the screened structural causal models admit a masked case within the screening budget.

Candidates are the known ancestors of the outcome that satisfy the anomaly condition at $|z|\geq 2$ under normal-operation statistics. In the masked class the fix-condition methods rank all known ancestors instead, since the anomaly condition cannot contain the cause there by construction. Eight attribution methods are compared, covering each baseline class of~\cite{nagalapatti2025idi}. PC-wg and PC-ng evaluate the fix condition through the pre-trained estimator, as the forward interventional query of setting the candidate to its normal level in the observed context, with and without the causal graph. The fitted counterpart, denoted GCM-IDI, evaluates the same fix condition through a graphical causal model with automatically assigned mechanisms (DoWhy-GCM~\cite{JMLR:v25:22-1258}), fitted on the identical context rows~\cite{nagalapatti2025idi}. Counterfactual attribution ranks the ancestors by their noise contribution to the observed deviation~\cite{budhathoki22a} and is evaluated on one repetition per structural causal model at $N_c{=}500$ for runtime reasons. The anomaly ranking ablates the fix condition and orders the same candidates by their own anomaly score. The traversal baseline follows the anomalous-path traversal of the root-cause-diagnosis literature, as implemented as a baseline in~\cite{nagalapatti2025idi}: it climbs from the outcome through anomalous variables only and returns those variables on such a path whose direct causes are not anomalous. The correlation baseline ranks all process variables by their absolute correlation with the outcome on the normal-operation rows and consumes no graph, and a restricted variant applies the same score to the candidate set. The context size is varied over $N_c\in\{50,500\}$ rows. Reported is the recall of the injected root cause within the top-$k$ ranked variables at $k=1$ and $k=3$~\cite{nagalapatti2025idi}, with method contrasts accompanied by bootstrap confidence intervals clustered at the structural-causal-model level.

\begin{table}[!htbp]
    \centering
    \caption[Mode-3 root-cause recall.]{Recall of the injected root cause (Recall@1/Recall@3) at the complete knowledge level, by fault class and context size. Best value per column in bold.}
    \label{tab:eval-rca-recall}
    \footnotesize
    \renewcommand{\arraystretch}{1.2}
    \begin{tabularx}{\textwidth}{@{}Xcccccc@{}}
        \toprule
        & \multicolumn{2}{c}{Level shift} & \multicolumn{2}{c}{Dispersion} & \multicolumn{2}{c}{Masked} \\
        \cmidrule(lr){2-3}\cmidrule(lr){4-5}\cmidrule(lr){6-7}
        Method & $N_c{=}50$ & $N_c{=}500$ & $N_c{=}50$ & $N_c{=}500$ & $N_c{=}50$ & $N_c{=}500$ \\
        \midrule
        PC-wg                     & $0.70$/$\mathbf{0.98}$ & $0.75$/$0.99$ & $0.72$/$\mathbf{0.98}$ & $0.78$/$0.99$ & $0.16$/$0.49$ & $0.22$/$0.51$ \\
        PC-ng                     & $0.59$/$\mathbf{0.98}$ & $0.68$/$0.98$ & $0.61$/$0.97$ & $0.73$/$0.99$ & $0.21$/$\mathbf{0.53}$ & $0.19$/$0.45$ \\
        GCM-IDI                   & $0.67$/$0.97$ & $0.75$/$0.99$ & $0.62$/$\mathbf{0.98}$ & $0.76$/$0.99$ & $0.16$/$0.42$ & $\mathbf{0.24}$/$0.56$ \\
        CF attribution            & --            & $0.65$/$0.84$ & --            & $0.64$/$0.78$ & --            & $0.23$/$\mathbf{0.62}$ \\
        Anomaly rank              & $0.73$/$\mathbf{0.98}$ & $0.78$/$0.99$ & $0.75$/$\mathbf{0.98}$ & $0.86$/$0.99$ & $0.15$/$0.49$ & $0.16$/$0.47$ \\
        Traversal                 & $\mathbf{0.80}$/$\mathbf{0.98}$ & $\mathbf{0.84}$/$\mathbf{1.00}$ & $\mathbf{0.78}$/$\mathbf{0.98}$ & $\mathbf{0.88}$/$\mathbf{1.00}$ & $0.14$/$0.49$ & $0.16$/$0.47$ \\
        Correlation               & $0.22$/$0.31$ & $0.29$/$0.55$ & $0.17$/$0.27$ & $0.18$/$0.35$ & $\mathbf{0.28}$/$0.42$ & $0.14$/$0.58$ \\
        Correlation (candidates)  & $0.60$/$0.96$ & $0.71$/$0.99$ & $0.64$/$0.97$ & $0.75$/$0.99$ & $0.14$/$0.49$ & $0.16$/$0.47$ \\
        \bottomrule
    \end{tabularx}
    \par\smallskip
    {\footnotesize Note: counterfactual attribution is evaluated on one repetition per structural causal model at $N_c{=}500$ ($250$ cases per observable fault class and $125$ masked cases).}
\end{table}

Table~\ref{tab:eval-rca-recall} reports the recall at the complete knowledge level. In the two observable fault classes every method restricted to the graph-derived candidate set attains a Recall@3 of at least $0.96$, whereas counterfactual attribution over the full ancestor set reaches $0.78$ to $0.84$ and the unrestricted correlation attribution remains between $0.27$ and $0.55$. The restricted correlation control isolates this effect, as the identical correlation score gains between $0.38$ and $0.57$ in Recall@1 once it is confined to the graph-derived candidate set. These results support the first clause of EH11. The contribution of the interventional fix condition on these faults is bounded by the contrasts on the identical candidate sets. PC-wg does not exceed the anomaly ranking in any observable condition, with clustered Recall@1 differences between $-0.14$ ($95\%$ CI $[-0.26,-0.02]$) and $-0.06$ ($[-0.19,+0.07]$). The two fix-condition engines operate at a comparable accuracy, as the clustered difference between PC-wg and GCM-IDI includes zero in three of the four observable conditions and reaches $+0.15$ ($[+0.02,+0.28]$) under the dispersion fault at $N_c{=}50$. The traversal attains the highest Recall@1 ($0.78$ to $0.88$), consistent with the traversal results on the PetShop benchmark~\cite{nagalapatti2025idi}, since under a unique root cause with a propagating symptom chain the anomalous path is informative. The pre-trained estimator reaches this accuracy region without a per-case fit, as a single amortised model whose queries additionally return the interventional fix scores consumed by the recommendation stage. Supplying the graph raises the pre-trained estimator's Recall@1 in every observable condition.

In the masked class no method identifies the injected cause reliably. The Recall@1 lies between $0.14$ and $0.28$ across all methods and both context sizes, close to the level expected from ranking the ancestor set without information. The anomaly condition contains the cause in $9\%$ of the masked cases, confirming the construction. Neither the interventional nor the counterfactual fix condition recovers the cause from its small deviation. The second clause of EH11 is therefore not supported. Root causes that remain inside their normal operating band are an open problem for every evaluated attribution class.

\begin{table}[!htbp]
    \centering
    \caption[Mode-3 recall across the structural-knowledge levels.]{Recall@1 of the injected root cause across the three structural-knowledge levels at $N_c{=}500$, observable fault classes. Methods that require structural knowledge are undefined at the none level.}
    \label{tab:eval-rca-kop}
    \footnotesize
    \renewcommand{\arraystretch}{1.2}
    \begin{tabularx}{\textwidth}{@{}Xcccccc@{}}
        \toprule
        & \multicolumn{3}{c}{Level shift} & \multicolumn{3}{c}{Dispersion} \\
        \cmidrule(lr){2-4}\cmidrule(lr){5-7}
        Method & none & ordered & complete & none & ordered & complete \\
        \midrule
        PC-wg                    & --     & $0.51$ & $0.75$ & --     & $0.37$ & $0.78$ \\
        PC-ng                    & $0.56$ & $0.51$ & $0.68$ & $0.60$ & $0.36$ & $0.73$ \\
        GCM-IDI                  & --     & $0.54$ & $0.75$ & --     & $0.36$ & $0.76$ \\
        Anomaly rank             & $0.66$ & $0.54$ & $0.78$ & $0.72$ & $0.41$ & $0.86$ \\
        Traversal                & --     & $0.55$ & $0.84$ & --     & $0.39$ & $0.88$ \\
        Correlation              & $0.29$ & $0.29$ & $0.29$ & $0.18$ & $0.18$ & $0.18$ \\
        Correlation (candidates) & $0.55$ & $0.52$ & $0.71$ & $0.50$ & $0.35$ & $0.75$ \\
        \midrule
        Cause inside the known ancestor set & -- & $0.58$ & $1.00$ & -- & $0.42$ & $1.00$ \\
        \bottomrule
    \end{tabularx}
\end{table}

Table~\ref{tab:eval-rca-kop} reports the sensitivity to the structural-knowledge level over the three regimes of Subsection~\ref{subsec:knowledge-availability}, with the ordered level operationalised as in Section~\ref{sec:eval-kop} and the diagnosis and every engine restricted consistently to the disclosed structure. Two effects separate. At the ordered level the recall of every graph-consuming method is governed by whether the injected cause lies inside the known ancestor set, which holds for $58\%$ of the level-shift and $42\%$ of the dispersion cases. Within this subset all graph-based methods retain or exceed their complete-knowledge accuracy, with Recall@1 between $0.84$ and $0.97$ at $N_c{=}500$, and outside it they cannot rank the cause by construction. At the none level no restriction applies and the injected cause remains in the candidate pool, where the anomaly-magnitude ranking attains a Recall@1 of $0.66$ to $0.72$ and the graph-free estimator variant $0.56$ to $0.60$ at $N_c{=}500$. The recall is therefore not monotone in the knowledge level: a partially known structure applied as a hard candidate restriction performs worse than no restriction, since the excluded share outweighs the gained precision. The effect is most pronounced in the dispersion class, where the $42\%$ coverage caps the attainable recall of the candidate-restricted methods below their none-level values, and the complete structure performs best in both classes. The unrestricted correlation attribution is unaffected by the knowledge level and remains the weakest method in every defined comparison. The benefit of the causal graph in Mode~3 therefore scales with the coverage of the disclosed structure rather than with the ranking engine, and partially known structure degrades the attribution when it is enforced as a hard candidate restriction. The pre-trained estimator is the only fix-condition engine that remains defined at every knowledge level, since the three-valued ancestor encoding of Chapter~\ref{chap:prima-causa-model} accepts the complete, the partially disclosed, and the empty graph without a level-specific branch. Its attribution accuracy at the none level remains below the anomaly-magnitude ranking, so this coverage of the knowledge axis, together with the interventional fix scores returned by the same queries, constitutes an operational property rather than a ranking advantage.

Two boundaries remain. The outcome is always the deepest node of the sampled graph, so the protocol covers the single-quality-gate case and not shallow or multi-outcome targets. The generator's mechanisms are additive in the noise, which favours the fitted additive-noise baseline and is stated as a prior property of the evaluation rather than corrected.

\section{Discussion}

\label{sec:eval-discussion}
\begin{figure}[!htbp]
    \centering
    \includegraphics[width=\textwidth]{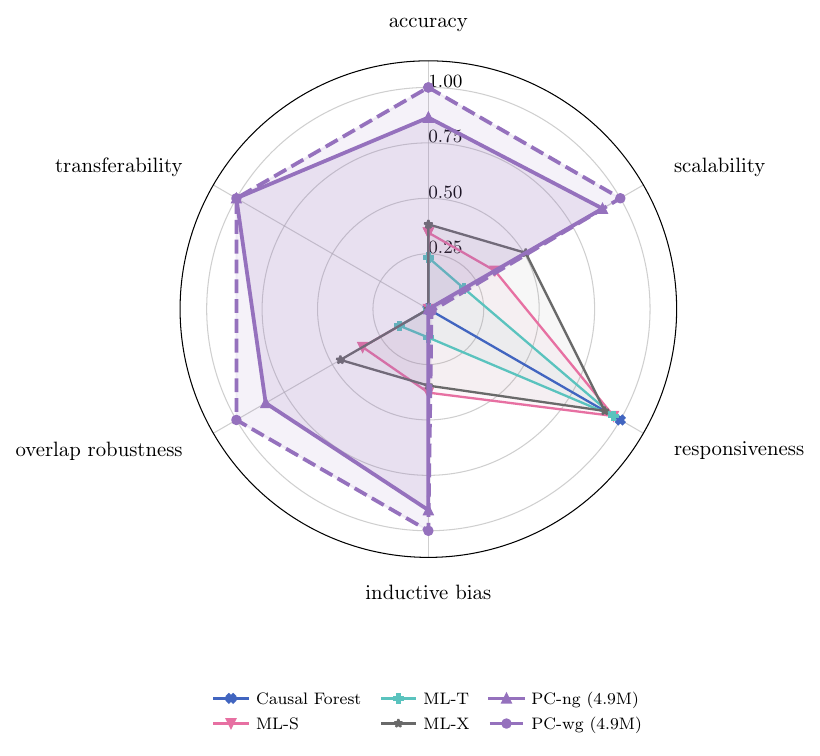}
    \caption{Normalised per-model profiles across six evaluation axes.}
    \label{fig:eval-profiles}
\end{figure}
Figure~\ref{fig:eval-profiles} aggregates six axes into a single per-model profile: interventional accuracy (EH1, weak-overlap and sparse-data condition), scalability (EH3, $n=25$ nodes), inference responsiveness (EH4, per-query latency inverted so that a larger value denotes a faster estimator), causal inductive bias (EH7, mediator and selection-collider structures pooled), overlap robustness (EH6, the most adverse evaluated overlap setting, $\theta=8$), and transfer accuracy (EH5, the hardest rung of the semi-synthetic difficulty ladder of Table~\ref{tab:synth-semi-ladder}). Each axis value is the median of the corresponding per-structural-causal-model metric under one fixed data condition, chosen per axis rather than averaged across the overlap and difficulty regimes the hypotheses contrast, since several hypotheses of this chapter state a direction that depends on the regime. Every axis is rescaled to the $[0,1]$ interval across the models that attain a value on that axis, with $1$ denoting the best-performing model observed on that axis and $0$ the worst. On the transfer axis the schema-agnostic PriMa-Causa is scored zero-shot on each host line, whereas the classical baselines are schema-bound and cannot be deployed across host lines of differing sensor schema without a target-side refit. Deployed without that refit they reduce to the no-effect predictor, so they are placed at the no-effect reference (normalised PEHE $=1$), which maps to the axis minimum in the figure, rather than at a re-fit in-distribution score that would not reflect transfer.

The two PriMa-Causa variants (PC-ng, PC-wg), an S-learner and a T-learner on random forests (ML-S, ML-T), an X-learner on random forests (ML-X), and a causal forest each attain a value on all six axes. On the transfer axis PC-wg reuses PC-ng's value, since no evaluated host line carries a known causal graph and the graph-conditioned variant therefore reduces to the graph-free forward pass. CausalFM is omitted from this summary, as it was not scored on the transfer axis and a profile with a missing axis cannot be compared by enclosed area.

To provide a single-number summary across the six evaluation axes, the polygon area enclosed by each profile is computed. This area equals the sum of the triangular areas formed between the origin and adjacent normalized axis values. Results are expressed as a fraction of the maximum theoretical area attained when a model achieves the best-observed score across all six axes simultaneously. Table~\ref{tab:eval-profile-area} reports this fraction per model. PC-wg attains the largest area, followed by PC-ng, then ML-X, ML-S, ML-T, and the causal forest. The two PriMa-Causa variants lead every axis except responsiveness, with PC-wg improving on PC-ng on accuracy, scalability, causal inductive bias, and overlap robustness and the two tying at the top of the transfer axis. The causal forest encloses a negligible area, since among the compared models it attains the worst value on five of the six axes and is competitive only on responsiveness. CausalFM is not summarised here, as it lacks a transfer measurement and the enclosed area is undefined when an axis value is missing.

\begin{table}[!htbp]
    \centering
    \caption[Profile-area summary.]{Polygon area of each model's normalised six-axis profile in Figure~\ref{fig:eval-profiles}, as a fraction of the area attained by a model at the best-observed value on every axis. Computed over all six models of Figure~\ref{fig:eval-profiles}.}
    \label{tab:eval-profile-area}
    \footnotesize
    \renewcommand{\arraystretch}{1.2}
    \begin{tabularx}{\textwidth}{@{}Xr@{}}
        \toprule
        Model & Profile area (\% of max) \\
        \midrule
        PC-wg             & $67\%$ \\
        PC-ng             & $54\%$ \\
        ML-X            & $19\%$ \\
        ML-S            & $16\%$ \\
        ML-T            & $6\%$ \\
        Causal Forest   & $0\%$ \\
        \bottomrule
    \end{tabularx}
\end{table}

PriMa-Causa trades inference latency for accuracy and transferability. Figure~\ref{fig:eval-gap} shows that under strong overlap and abundant data the margins to the fitted baselines narrow, so the size of the advantage depends on the data conditions, while the ordering of the models is unchanged. The largest differences are observed under weak overlap with sparse context, where the fitted baselines plateau at a biased estimate of the interventional effect while the pre-trained PriMa-Causa retains its accuracy.

Taken together, these results provide the empirical assessment of the central hypothesis. The accuracy half is supported under the evaluated conditions: the interventional estimator attains a lower prescription error than every associational baseline under weak overlap and sparse data (EH1), the advantage widens as overlap weakens (EH6), it persists on real industrial covariates (EH5), and it is reinforced by structural knowledge (EH2, EH7, EH12). Contrary to the second clause of EH1, the advantage does not vanish under strong overlap and abundant data, so the evaluated conditions bound where the advantage is largest rather than where it exists. The efficiency half is supported by the budget-constrained prioritisation (EH9), in which the effect-based ranking attains a positive net \acs{OEE} gain that the treatment-unaware ranking does not. The root-cause mode (EH11) is supported in its first clause, as the candidate-restricted attribution identifies the injected root cause within the top three candidates in at least $96\%$ of the observable-fault cases given the complete graph, against at most $55\%$ for the unrestricted correlation attribution. Its second clause is not supported, since no evaluated method recovers masked root causes above chance-level ranking of the ancestor set. Within the graph-restricted regime the pre-trained estimator attains this attribution accuracy as the same amortised model that serves Mode~1, without a per-case fit at decision time. It is also the only fix-condition engine that remains defined across all three structural-knowledge levels.

    \chapter{Conclusions and Outlook}
    \label{chap:conclusions}

\begin{jointwork}
This chapter draws together the strands of the preceding chapters and brings the dissertation to a close. Section~\ref{sec:conclusions-rq-answers} answers research questions RQ1 to RQ3. Section~\ref{sec:conclusions-contribution} consolidates the contribution to knowledge across artefacts C1, C2, and C3. Section~\ref{sec:conclusions-limitations} outlines the limitations of the evidence, and Section~\ref{sec:conclusions-outlook} presents model-level and programme-level directions for future work. Figure~\ref{fig:ch9_structure} outlines the structure of this chapter.
\end{jointwork}

\vspace{\baselineskip}

\begin{center}
    \includegraphics[width=\textwidth]{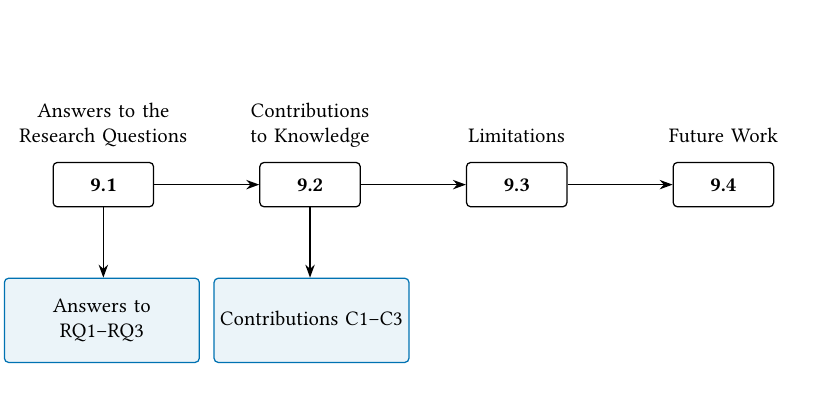}%
    \captionof{figure}{Structure of Chapter~\ref{chap:conclusions}.}
    \label{fig:ch9_structure}
\end{center}

\newpage
\section{Answers to the Research Questions}
\label{sec:conclusions-rq-answers}

\noindent\textbf{RQ1 (design).}\enspace Causal inductive bias can be embedded into a prescriptive-main\-te\-nance reference model under three structural conditions, established in Chapter~\ref{chap:prima-causa}. First, an explicit causal reasoning layer is inserted between the predictive-analytics step and the recommendation step, so that recommendations are justified by the interventional quantity $P(Y\mid \dop(T{=}t),X{=}x)$ rather than by the associational quantity $P(Y\mid X{=}x)$. Second, each prescriptive situation is framed as a typed query and organised into one of four decision modes, which connects causal diagnosis to action selection and separates the two analytically. Third, the available structural knowledge is rendered machine-usable across the none, ordered, and complete levels, so the layer operates under the full range of knowledge conditions rather than relying on a curated graph. The design knowledge is consolidated from meta-requirements into design principles, which makes the answer transferable beyond the single artefact of this dissertation.

\medskip
\noindent\textbf{RQ2 (instantiation).}\enspace The design knowledge is formalised as a trainable model by amortisation. The PriMa-Causa model (Chapter~\ref{chap:prima-causa-model}) is a graph-conditioned interventional estimator. It is pre-trained once on a synthetic prior of structural causal models. At inference time, each case is evaluated in a single forward pass by conditioning on the observational context and a ternary encoding of the disclosed graph structure. Rare fault events and the absence of interventional records at decision time are thereby circumvented rather than modelled, since no per-case fitting is required. The model is made analysable through the data foundation of Chapter~\ref{chap:synthetic-data}: the Causal Process Graph generator supplies production-line-shaped structural causal models whose interventional ground truth is available by construction, and the semi-synthetic procedure anchors the evaluation on real industrial covariates. Validation, identified in Chapter~\ref{chap:introduction} as the binding constraint, is thus provided by the data foundation rather than assumed.

\medskip
\noindent\textbf{RQ3 (evaluation).}\enspace The causal approach improves both target quantities under the evaluated conditions, to a degree that depends on the data regime, as quantified in Chapter~\ref{chap:empirical-evaluation}. On the accuracy side, the interventional estimator attains a lower prescription error than every associational baseline, with the largest margins under weak overlap and sparse context. The advantage narrows but does not vanish under strong overlap and abundant data, so the evaluated conditions delimit where the advantage is largest rather than whether it is present at all. Supplied structural knowledge yields a further reduction whose size grows where the observational data carry less of the structure. The accuracy transfers to real industrial covariates without refitting. On the efficiency side, ranking production states by the estimated interventional effect converts into a positive expected net \acs{OEE} gain at every non-zero adjustment budget, which the treatment-unaware predictive ranking does not attain. The central hypothesis of Chapter~\ref{chap:introduction} is thereby supported in both of its halves under the evaluated conditions.

\section{Contributions to Knowledge}
\label{sec:conclusions-contribution}

Following the design-science methodology of Chapter~\ref{chap:introduction}, the dissertation returns three artefacts to the knowledge base. C1 contributes reference-level design knowledge: the taxonomy of causal-prescriptive decision situations, the causal reasoning layer, and the design principles consolidated from meta-requirements (Chapter~\ref{chap:prima-causa}). C2 contributes the instantiation: the PriMa-Causa model and its open-source data foundation, which jointly demonstrate that interventional estimation is feasible under the data conditions of \acs{FMCG} production lines (Chapters~\ref{chap:synthetic-data} and~\ref{chap:prima-causa-model}). C3 contributes the intervention-oriented evaluation methodology and its results, in which accuracy and an \acs{OEE}-based efficiency criterion are measured against controllable ground truth (Chapter~\ref{chap:empirical-evaluation}). In a design-of-experiments study that varies the identification problem and the statistical problem independently, the instantiated model attains a lower prescription error than every associational baseline, with the largest margins under weak overlap and sparse context. The advantage is retained over the evaluated graph sizes and transfers to real industrial covariates without refitting.

Taken together, the contributions operationalise the research gaps of Chapter~\ref{chap: literature review} as a validated design. The causal inductive bias called for by RG1 is realised in a trainable architecture. The end-to-end methodology with intervention-oriented evaluation demanded by RG2 is instantiated and tested. The feasibility under the data conditions of \acs{FMCG} production lines required by RG3 is demonstrated across the synthetic, semi-synthetic, and real-covariate regimes. The integration of root-cause analysis with prescriptive action called for by RG4 is realised in the decision-mode taxonomy, although its empirical evaluation remains confined to Mode~3 and to synthetic experiments. The peer-reviewed publications underlying these contributions return the design knowledge to the scientific community, closing the rigor cycle.

The knowledge contributions are carried by three software artefacts, summarised in Table~\ref{tab:conclusions-artefacts}. The data generator and the pre-trained model are released as open-source software, and the maturity of the demonstrator is assessed in Chapter~\ref{chap:usecase}.

\begin{table}[!htbp]
    \centering
    \caption{Software artefacts of the dissertation.}
    \label{tab:conclusions-artefacts}
    \footnotesize
    \renewcommand{\arraystretch}{1.2}
    \begin{tabularx}{\textwidth}{@{}lXl@{}}
        \toprule
        Software artefact & Realisation & Source \\
        \midrule
        Production-line data generator & Causal Process Graph sampler, synthetic \acs{SCM} pipeline, and semi-synthetic anchoring on real industrial covariates & Ch.~\ref{chap:synthetic-data} \\
        Pre-trained PriMa-Causa foundation model & Graph-conditioned interventional estimator, released with implementation, training configuration, and weights & Ch.~\ref{chap:prima-causa-model} \\
        Line-deployed software demonstrator & Causal reasoning runtime and operator console on the edge gateway, with the troubleshooting application on mobile and head-mounted devices (co-developed with Dropslab Technologies) & Ch.~\ref{chap:usecase} \\
        \bottomrule
    \end{tabularx}
\end{table}

\newpage
\section{Limitations}
\label{sec:conclusions-limitations}

The evidence of this dissertation is bounded in five respects, summarised in \mbox{Table~\ref{tab:conclusions-limitations}}. First, reasoning is performed at the sensor and process layer on the observable variable set, so the framework narrows the search space for actionable levers and does not recover latent physical mechanisms. Second, the treatment is confined to the interventional layer of the causal hierarchy, and the structural assumptions are static, so counterfactual reasoning, time-varying structures, and latent-confounder identification are outside the validated scope. Third, the empirical evidence rests on synthetic and semi-synthetic ground truth with real covariates rather than on closed-loop field data. This boundary is not specific to the present artefact but the general evaluation condition of causal inference: the per-case interventional effect is never observed in recorded data, so estimation accuracy is verifiable only where the generating mechanism is controlled~\cite{DoutreligneHowToSelect,poinsot2025synthetic}. A closed-loop field trial would measure the realised gain of the recommendation policy rather than the accuracy of the estimator. Within this evidence, the decision-mode coverage is partial: side-effect analysis (Mode~2) is specified but not empirically tested, and the evaluation of causal root-cause analysis (Mode~3) is done with synthetic experiments. Fourth, the match between the generic pre-training prior and a concrete line is established offline, on the topology statistics of the only published real-line ground-truth graph (Section~\ref{sec:synth-cpg}) and on real industrial covariates (Section~\ref{sec:eval-transfer}), and not at query time. For tabular foundation models, the measured distance between a synthetic prior and real data has been found to explain downstream performance only partially~\cite{turkmen2026evaluating} or not detectably~\cite{davies2026mind}, so such a distance is not an established indicator of whether a prior mismatch degrades the estimate for a given case. Fifth, the input layout of the trained instantiation has a fixed width of $35$ variables, so wider lines require a reduction to a decision-relevant variable subset or retraining (Subsection~\ref{sec:ch6-eff-internals}). These boundaries follow from the operating conditions targeted by the dissertation and mark the transition points for future work.

\begin{table}[!htbp]
    \centering
    \caption{Boundaries of the evidence.}
    \label{tab:conclusions-limitations}
    \footnotesize
    \renewcommand{\arraystretch}{1.2}
    \begin{tabularx}{\textwidth}{@{}lY>{\raggedright\arraybackslash}p{27mm}@{}}
        \toprule
        Boundary & What lies outside the validated scope & Source \\
        \midrule
        Sensor and process layer & Recovery of latent physical mechanisms (the framework narrows the search space for actionable levers). & Subsec.~\ref{subsec:ch4-crl-abgrenzung} \\
        Static interventional scope & Counterfactual reasoning, time-varying structures, and latent-confounder identification. & Sec.~\ref{sec:bg-causal} \\
        Controlled ground truth & Realised gain of the recommendation policy in closed-loop field operation (accuracy is verified on synthetic and semi-synthetic data with real covariates, Mode~2 is untested, Mode~3 is tested synthetically). & \cite{DoutreligneHowToSelect,poinsot2025synthetic}, Sec.~\ref{sec:ch6-mode-coverage}, \ref{sec:eval-rca} \\
        Offline prior match & Query-time check of whether a case lies within the pre-training prior (the match is established offline). & Sec.~\ref{sec:synth-cpg}, \ref{sec:eval-transfer}, \mbox{\cite{turkmen2026evaluating,davies2026mind}} \\
        Fixed architecture & Lines wider than the $35$-variable input layout without reduction or retraining. & Subsec.~\ref{sec:ch6-eff-internals} \\
        \bottomrule
    \end{tabularx}
\end{table}

\section{Future Work}
\label{sec:conclusions-outlook}

Four model-level directions follow. First, the foundation model is to be retrained and optimised. A larger training budget, refined priors, and a density-matched Erd\H{o}s--R\'enyi control sharpen the attribution of the structured-prior advantage. Second, the evaluation is to be scaled to larger graphs, and thus to larger numbers of process variables, than the current pre-training ceiling admits. Extending the evaluation in this way requires retraining the model. Third, the backbone itself sets a computational ceiling. The transformer is adopted in Chapter~\ref{chap:prima-causa-model} because the causal graph maps directly onto the attention matrix, which ties the cost of inference to attention over the full observational context. Sequence and state-space architectures offer linear-time alternatives for long contexts. The open question is how structural knowledge can be injected into an architecture that provides no attention matrix to carry the graph bias. Fourth, the framework is to be tested in live production operation, where non-stationarity, closed-loop control, and the integration with existing manufacturing systems decide the practical value of the recommendations.

Beyond these model-level directions, five programme-level directions follow. First, the extension to the counterfactual layer and to time-varying causal structures would widen the class of maintenance questions the framework can answer, including retrospective queries over closed-loop control. Second, the treatment of latent confounding, through sensitivity analysis or partial-identification bounds, would relax the structural assumptions under which the recommendations are valid. Third, the data foundation can be developed into an automated pre-deployment evaluation suite, in which a candidate model is screened across generated data conditions before it is admitted to production use, turning the evaluation design of this dissertation into a reusable test procedure. Fourth, the integration with operator-facing workflows warrants investigation, including interfaces that communicate the causal evidence behind a recommendation and conversational access to the reasoning layer. Fifth, the extension to multimodal data would broaden the observable sensor layer, with image-valued outcomes as estimation targets and free-text workflow records from operators as additional context.

\begin{figure}[!htbp]
  \centering
  \begin{subfigure}{0.48\textwidth}
    \centering
    \includegraphics[width=0.65\linewidth]{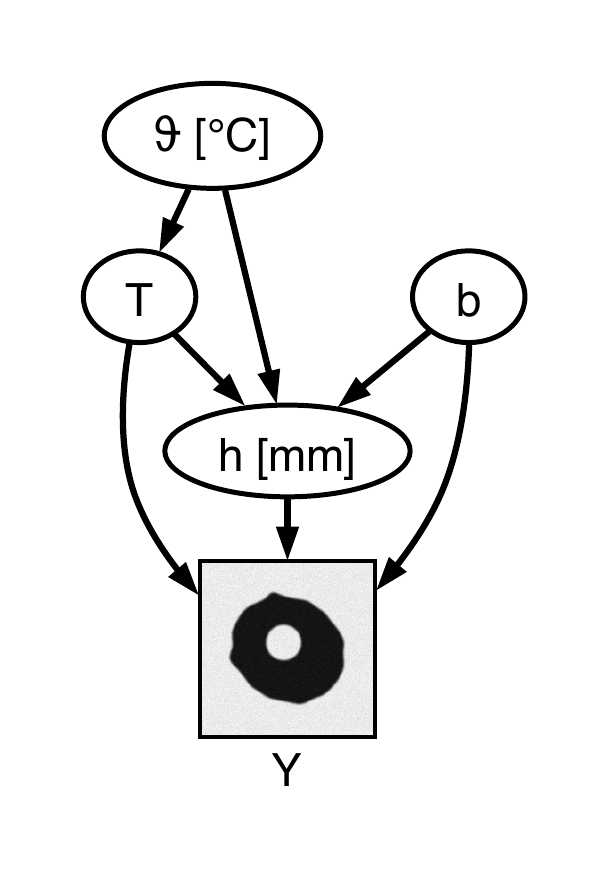}
    \caption{Observational regime.}
    \label{fig:conclusions-image-dag}
  \end{subfigure}
  \hfill
  \begin{subfigure}{0.48\textwidth}
    \centering
    \includegraphics[width=0.65\linewidth]{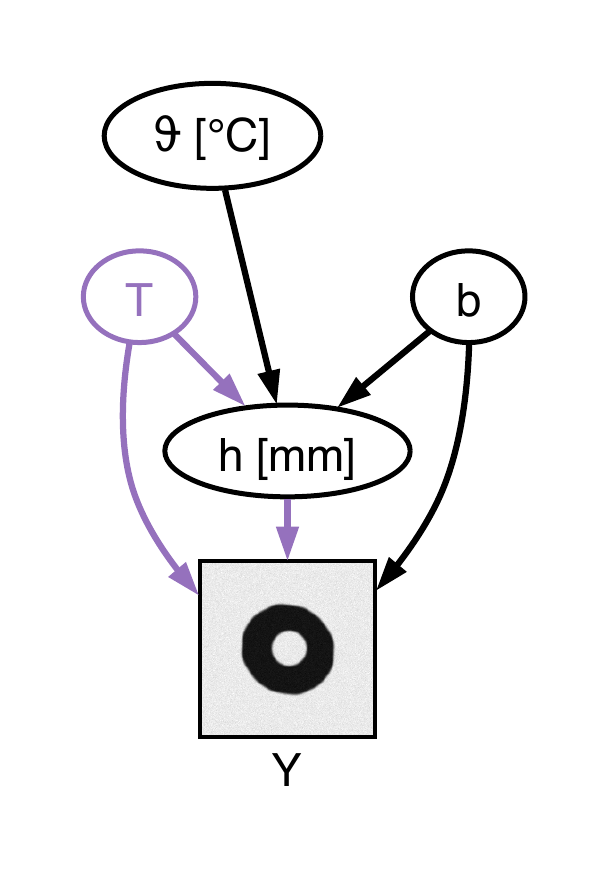}
    \caption{Under an intervention $\dop(T)$.}
    \label{fig:conclusions-image-dag-do}
  \end{subfigure}
  \caption[Single-station causal graph with an image-valued outcome.]{Single-station causal graph of Figure~\ref{fig:intro-dag} with an image-valued outcome. The intervention $\dop(T)$ on the corrective action removes the incoming edge $\vartheta \to T$ by graph surgery.}
  \label{fig:conclusions-image-dag-pair}
\end{figure}

\begin{figure}[!htbp]
  \centering
  \includegraphics[width=\textwidth]{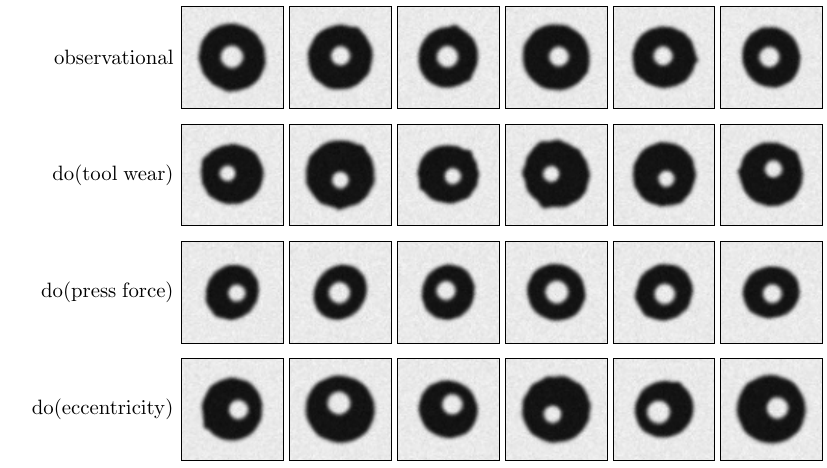}
  \caption{Rendered inspection images under observation and under three interventions.}
  \label{fig:conclusions-image-grid}
\end{figure}

The extension to image-valued outcomes can be stated precisely. In end-of-line quality inspection, the outcome is a camera image rather than a scalar measurement, and its structural equation factorises into two stages. The process parameters determine a low-dimensional vector of product attributes, such as diameters, eccentricity, and burr. A specified rendering mechanism then maps this attribute vector, together with exogenous imaging noise, onto the inspection image~\cite{pawlowski2020deep}. Figure~\ref{fig:conclusions-image-dag-pair} shows the single-station graph of Chapter~\ref{chap:introduction} under this extension. The intervention $\dop(T)$ on the corrective action removes the incoming edge $\vartheta \to T$ by graph surgery, shown in Figure~\ref{fig:conclusions-image-dag-do}. It propagates through the process mechanisms into the attributes and through the renderer into the image, so the observational and the interventional image distributions differ exactly as the structural causal model prescribes. Figure~\ref{fig:conclusions-image-grid} illustrates this with samples from a prototype rendering mechanism, in which each intervention leaves a distinct signature in the inspection image. \textit{What would the inspection image look like if the corrective action were taken?} Counterfactual image distributions are not identifiable from observational images in general, even when the graph is known~\cite{pan2024counterfactual}. The specification of the rendering mechanism therefore preserves the ground-truth property of the data foundation. The estimand remains well defined when it targets the attributes that are observable in the image, since the image constitutes a high-dimensional and partial measurement of the attribute vector. Pre-training on such a prior would extend the PriMa-Causa model from tabular outcomes to visual interventional queries, a combination that existing causal foundation models do not yet offer. The required extension is architectural as well as statistical. The image enters the estimator as a single node whose representation is an embedding rather than a scalar, so the backbone requires an image head. The graph conditioning of Chapter~\ref{chap:prima-causa-model} is unaffected, since the graph is a relation between variables and the image occupies one node of it. Real parameter-labelled camera data from manufacturing processes provide transfer anchors for such a prior~\cite{brion2022generalisable}.

The live-production validation prepared by the software artefact of Chapter~\ref{chap:usecase} is the step on which the model-level and the programme-level directions converge.

    \clearpage
    \appendix

    \chapter{Use of Software and AI Tools}
    \label{app:software-ai-tools}
    This appendix documents the software and artificial-intelligence-based tools used during the preparation of this dissertation. The disclosure is intended to make the role of these tools transparent. The tools used are listed in Table~\ref{tab:app-tools}.

\begin{table}[htbp]
    \centering
    \caption{Software and AI-based tools used in the dissertation.}
    \label{tab:app-tools}
    \footnotesize
    \setlength{\tabcolsep}{6pt}
    \renewcommand{\arraystretch}{1.15}
    \begin{tabularx}{\textwidth}{@{}l l Y@{}}
        \toprule
        Tool or tool class & Provider / Reference & Use in the dissertation \\
        \midrule
        DeepL & \url{https://www.deepl.com} & Translation and writing support. \\
        Claude Code & \url{https://claude.ai} & Optimisation and debugging of code. \\
        Positron IDE & \url{https://positron.posit.co} & Integrated development environment for Python and R code development and analysis. \\
        Python & \url{https://www.python.org} & Data generation, model architecture implementation, and empirical evaluation. \\
        R Project & \url{https://www.r-project.org} & Statistical computing, benchmark analysis, and synthetic data generation. \\
        Beckhoff TwinCAT & \url{https://www.beckhoff.com} & Programmable logic controller (PLC) programming, machine telemetry acquisition, and ADS protocol integration. \\
        TikZ / PGF & \url{https://ctan.org/pkg/pgf} & Manual authoring and typesetting of vector diagrams, flowcharts, and structural causal graphs. \\
        Affinity Suite & \url{https://www.affinity.studio} & Creation, vector illustration, and graphical layout of figures. \\
        \bottomrule
    \end{tabularx}
\end{table}

    \makeatletter
        \def\toclevel@chapter{-1}
        \def\toclevel@section{0}
    \makeatother

    \pagestyle{plain}
    
    \setuptoc{lof}{totoc}
    \listoffigures

    \setuptoc{lot}{totoc}
    \listoftables

    \renewcommand*{\bibfont}{\small}
    \printbibheading[title={References}]
    \addtocontents{toc}{\protect\enlargethispage{2\baselineskip}}
    \addcontentsline{toc}{chapter}{References}
    \printbibliography[heading = none]
    \renewcommand*{\bibfont}{\normalsize}

\end{document}